\documentclass[10pt,twocolumn,letterpaper]{article}

 \usepackage{cvpr}              

\usepackage{tabularx, makecell, tabularx}
\usepackage{graphicx}
\usepackage{mathtools}
\usepackage{amssymb}
\usepackage{booktabs}
\usepackage{multirow}
\usepackage{diagbox}
\usepackage[toc,page]{appendix}
\usepackage{pifont}
\usepackage{ifthen}
\usepackage{microtype}
\usepackage{color,soul}
\usepackage{adjustbox}
\usepackage{overpic}
\usepackage{xcolor}
\usepackage{xr}
\usepackage{tablefootnote}

\newcommand{\sota}{state-of-the-art\xspace}

\newcolumntype{Y}{>{\centering\arraybackslash}X}

\definecolor{mygray}{gray}{0.6}

\newboolean{arxiv}
\setboolean{arxiv}{true}

\makeatletter
\newcommand*{\addFileDependency}[1]{
  \typeout{(#1)}
  \@addtofilelist{#1}
  \IfFileExists{#1}{}{\typeout{No file #1.}}
}
\makeatother

\usepackage[pagebackref,breaklinks,colorlinks]{hyperref}

\usepackage[capitalize]{cleveref}
\crefname{section}{Sec.}{Secs.}
\Crefname{section}{Section}{Sections}
\Crefname{table}{Table}{Tables}
\crefname{table}{Tab.}{Tabs.}

\def\cvprPaperID{413} 
\def\confName{CVPR}
\def\confYear{2022}

\begin{document}
\title{Multimodal Conditional Image Synthesis with Product-of-Experts GANs}

\makeatletter
\apptocmd\@maketitle{{\myfigure{}\par}}{}{}
\makeatother

\author{Xun Huang\qquad Arun Mallya\qquad Ting-Chun Wang\qquad Ming-Yu Liu\\NVIDIA}

\newcommand\myfigure{%
	\centering
	\vspace{-6mm}
	\includegraphics[width=\linewidth,trim={0.5in 0.3in 0.3in 0}]{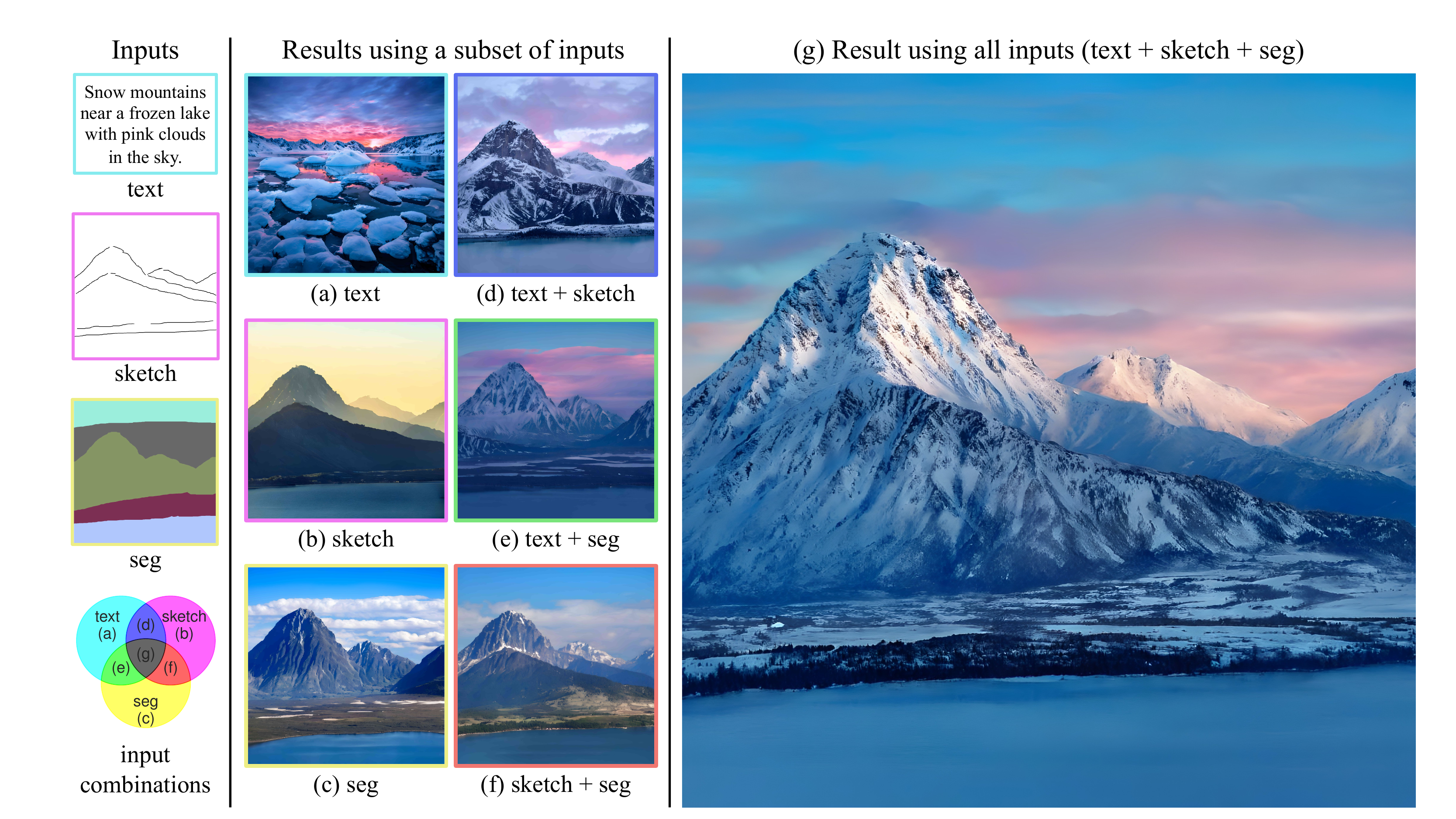}
	\vspace{1mm}
	\captionof{figure}{Given conditional inputs in multiple modalities~(the left column), our approach can synthesize images that satisfy all input conditions~(the right column, (g)) or any arbitrary subset of the input conditions~(the middle column, (a)--(f)) with \textit{a single model}.}
	\label{fig:teaser}
	\vspace{6mm}
}

\maketitle

\begin{abstract}
   Existing conditional image synthesis frameworks generate images based on user inputs in a single modality, such as text, segmentation, sketch, or style reference. They are often unable to leverage multimodal user inputs when available, which reduces their practicality. To address this limitation, we propose the Product-of-Experts Generative Adversarial Networks (PoE-GAN) framework, which can synthesize images conditioned on multiple input modalities or any subset of them, even the empty set. PoE-GAN consists of a product-of-experts generator and a multimodal multiscale projection discriminator. Through our carefully designed training scheme, PoE-GAN learns to synthesize images with high quality and diversity. Besides advancing the state of the art in multimodal conditional image synthesis, PoE-GAN also outperforms the best existing unimodal conditional image synthesis approaches when tested in the unimodal setting. \ifthenelse{\boolean{arxiv}}{The project website is available at \href{https://deepimagination.github.io/PoE-GAN/}{this link}.}{Code and pretrained models will be released.}  
\end{abstract}

\section{Introduction}
\label{sec:intro}

Conditional image synthesis allows users to use their creative inputs to control the output of image synthesis methods. It has found applications in many content creation tools.
Over the years, a variety of input modalities have been studied, mostly based on conditional GANs~\cite{goodfellow2014generative,isola2017image,odena2017conditional,liu2020generative}. To this end, we have various single modality-to-image models. When the input modality is text, we have the \textit{text-to-image} model~\cite{reed2016generative,zhang2017stackgan,xu2018attngan,zhu2019dm,ming2020DFGAN,ye2021improving,ramesh2021zero}. When the input modality is a segmentation mask, we have the \textit{segmentation-to-image} model~\cite{isola2017image,wang2018high,park2019semantic,liu2019learning,schonfeld2020you,esser2021taming}. When the input modality is a sketch, we have the \textit{sketch-to-image} model~\cite{sangkloy2017scribbler,chen2018sketchygan,ghosh2019interactive,chen2020deepfacedrawing}.


However, different input modalities are best suited for conveying different types of conditioning information. 
For example, as seen in the first column of \cref{fig:teaser}, segmentation makes it easy to define the coarse layout of semantic classes in an image---the relative locations and sizes of sky, cloud, mountain, and water regions. Sketch allows us to specify the structure and details within the same semantic region, such as individual mountain ridges. 
On the other hand, text is well-suited for modifying and describing objects or regions in the image, which cannot be achieved by using segmentation or sketch, \eg `\emph{frozen} lake' and `\emph{pink} clouds' in \cref{fig:teaser}.  
Nevertheless, despite this synergy among modalities, prior work has considered image generation conditioned on each modality as a distinct task and studied it in isolation. Existing models thus fail to utilize complementary information available in different modalities. Clearly, a conditional generative model that can combine input information from all available modalities would be of immense value.

Even though the benefits are enormous, the task of conditional image synthesis with multiple input modalities poses several challenges. First, it is unclear how to combine multiple modalities with different dimensions and structures in a single framework. Second, from a practical standpoint, the generator needs to handle missing modalities since it is cumbersome to ask users to provide every single modality all the time. This means that the generator should work well even when only a subset of modalities are provided. Lastly, conditional GANs are known to be susceptible to mode collapse~\cite{isola2017image,odena2017conditional}, wherein the generator produces identical images when conditioned on the same inputs. This makes it difficult for the generator to produce diverse output images that capture the full conditional distribution when conditioned on an arbitrary set of modalities.

We present Product-of-Experts Generative Adversarial Networks~(PoE-GAN), a framework that can generate images conditioned on any subset of the input modalities presented during training, as illustrated in Fig.~\ref{fig:teaser} (a)-(g). 
This framework provides users unprecedented control, allowing them to specify exactly what they want using multiple complementary input modalities.
When users provide no inputs, it falls back to an unconditional GAN model~\cite{goodfellow2014generative,radford2016unsupervised,miyato2018spectral,karras2018progressive,brock2019large,karras2019style,karras2020analyzing,karras2021alias}. One key ingredient of our framework is a novel product-of-experts generator that can effectively fuse multimodal user inputs and handle missing modalities~(Sec.~\ref{sec:gen}). A novel hierarchical and multiscale latent representation leads to better usage of the structure in spatial modalities, such as segmentation and sketch (Sec.~\ref{sec:latent}). Our model is trained with a multimodal projection discriminator (Sec.~\ref{sec:discriminator}) together with contrastive losses for better input-output alignment. In addition, we adopt modality dropout for additional robustness to missing inputs (Sec.~\ref{sec:training}). Extensive experiment results show that PoE-GAN outperforms prior work in both multimodal and unimodal settings (Sec.~\ref{sec:experiments}), including \sota approaches specifically designed for a single modality. 
We also show that PoE-GAN can generate diverse images when conditioned on the same inputs.

\section{Related work}

\noindent\textbf{Image Synthesis.}
Our network architecture design is inspired by previous work in unconditional image synthesis. Our decoder employs some techniques proposed in StyleGAN~\cite{karras2019style} such as global modulation. Our latent space is constructed in a way similar to hierarchical variational auto-encoders (VAEs)~\cite{sonderby2016ladder,maaloe2019biva,vahdat2020NVAE,child2021very}. While hierarchical VAEs encode the image itself to the latent space, our network encodes conditional information from different modalities into a unified latent space. Our discriminator design is inspired by the projection discriminator~\cite{miyato2018cgans} and multiscale discriminators~\cite{wang2018high,liu2019learning}, which we extend to our multimodal setting.

\medskip
\noindent\textbf{Multimodal Image Synthesis.}
Prior work~\cite{suzuki2016joint,vedantam2018generative,wu2018multimodal,shi2019variational,sutter2020generalized,kutuzova2021multimodal} has used VAEs~\cite{vae,rezende2014stochastic} to learn the joint distribution of multiple modalities. Some of them~\cite{vedantam2018generative,wu2018multimodal,kutuzova2021multimodal} use a product-of-experts inference network to approximate the posterior distribution. This is conceptually similar to how our generator combines information from multiple modalities. While their goal is to estimate the complete joint distribution, we focus on learning the image distribution conditioned on other modalities. Besides, our framework is based on GANs rather than VAEs and we perform experiments on high-resolution and large-scale datasets, unlike the above work. 
Recently, Xia~\etal~\cite{xia2021tedigan} propose a GAN-based multimodal image synthesis method named TediGAN. Their method relies on a pretrained unconditional generator. However, such a generator is difficult to train on a complex dataset such as MS-COCO~\cite{lin2014microsoft}. 
Concurrently, Zhang~\etal~\cite{zhang2021m6ufc} propose a multimodal image synthesis method based on VQGAN. The way they combine different modalities is similar to our baseline using concatenation and modality dropout~(\cref{sec:ablation}). We will show that our product-of-experts generator design significantly improves upon this baseline.

\section{Product-of-experts GANs}

Given a dataset of images $x$ paired with $M$ different input modalities $(y_{1}, y_{2}, ..., y_{M})$, our goal is to train a single generative model that learns to capture the image distribution conditioned on an arbitrary subset of possible modalities $p(x|\mathcal{Y}), \forall\mathcal{Y}\subseteq \{y_{1}, y_{2}, ..., y_{M}\}$. 
In this paper, we consider four different modalities including text, semantic segmentation, sketch, and style reference. Note that our framework is general and can easily incorporate additional modalities.

Learning image distributions conditioned on \textit{any subset} of $M$ modalities is challenging because it requires a single generator to simultaneously model $2^M$ distributions. Of particular note, the generator needs to capture the unconditional image distribution $p(x)$ when $\mathcal{Y}$ is an empty set, and the unimodal conditional distributions $p(x|y_{i}), \forall i\in\{1, 2, ..., M\}$, such as the image distribution conditioned on text alone. These settings have been popular and widely studied in isolation, and we aim to bring them all under a unified framework.

\begin{figure}[!t]
	\centering
	\subcaptionbox{Intersection of sets}
	{\includegraphics[height=0.26\linewidth]{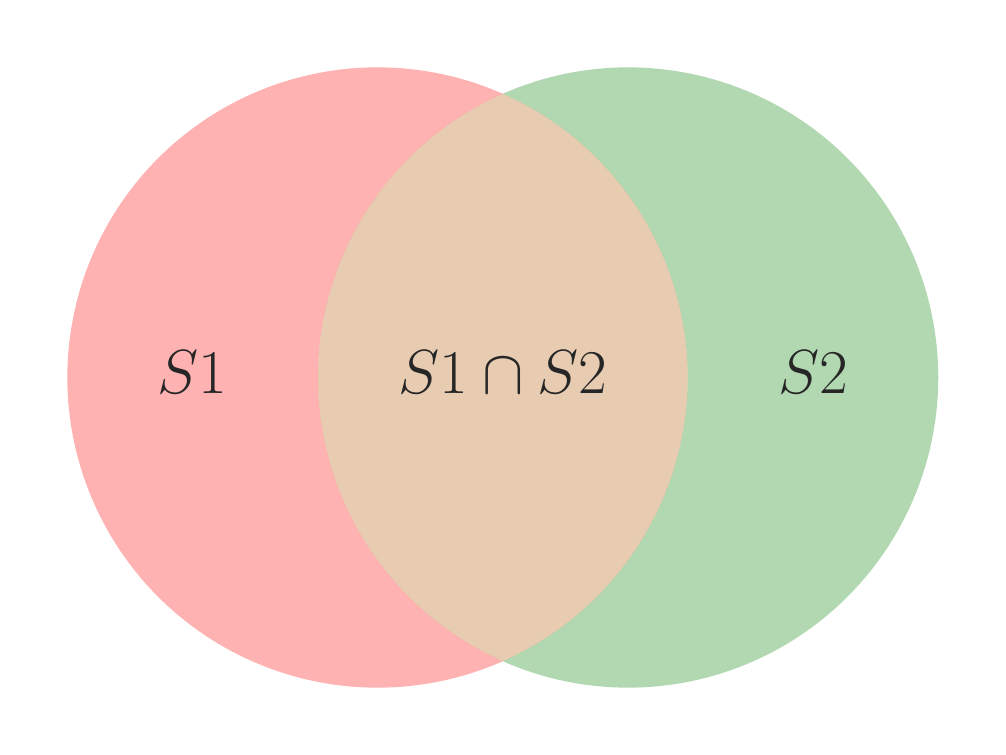}}
	\subcaptionbox{Product of distributions}
	{\includegraphics[height=0.26\linewidth]{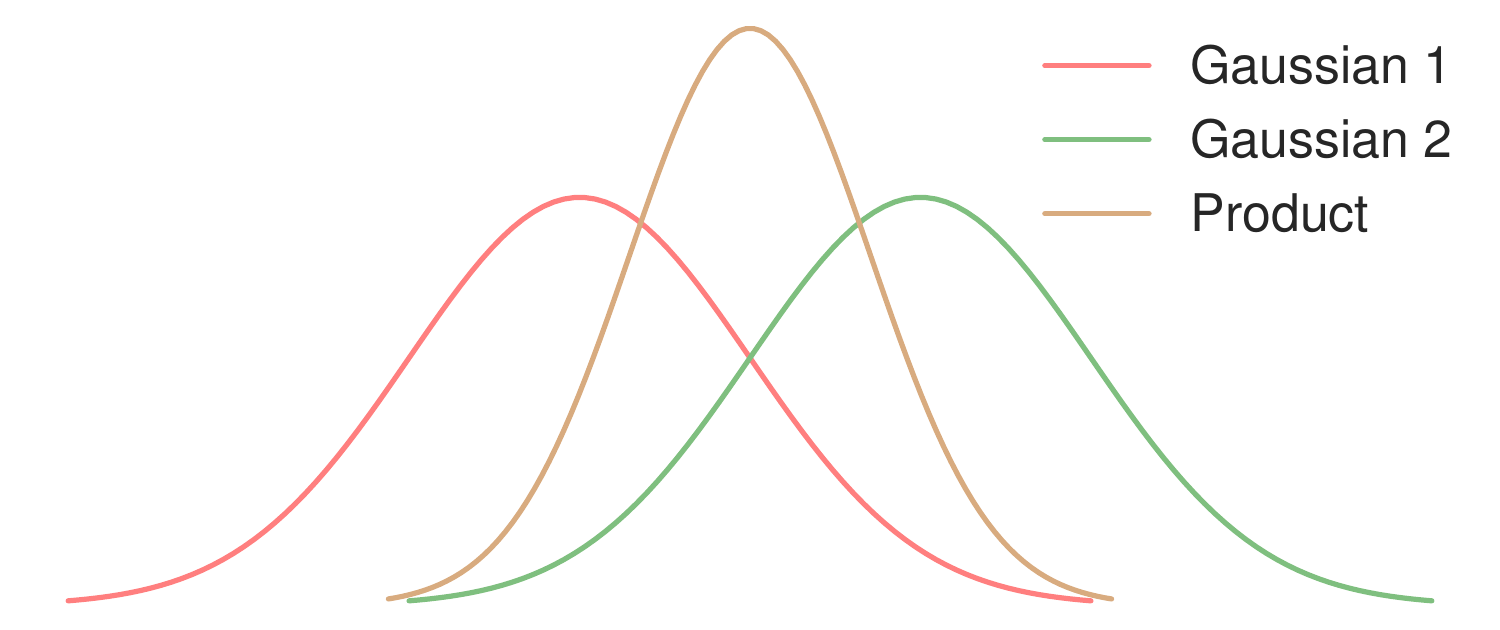}}
	\vspace{-2mm}
	\caption{The product of distributions is analogous to the intersection of sets. The product distribution has a high density where both distributions have relatively high densities.}
	\label{fig:setpoe}
	\vspace{-2mm}
\end{figure}

\subsection{Product-of-experts modeling}
\label{sec:gen}
Intuitively, each input modality adds a constraint the synthesized image must satisfy. The set of images that satisfies all constraints is the intersection of the sets each of which satisfies one individual constraint. As shown in \cref{fig:setpoe}, we model this by making a strong assumption that the jointly conditioned probability distribution $p(x|y_{i}, y_{j})$ is proportional to the \textit{product} of singly conditioned probability distributions $p(x|y_{i})$ and $p(x|y_{j})$. Under this setting, for the product distribution to have a high density in a region, each of the individual distributions needs to have a high density in that region, thereby satisfying each constraint. This idea of combining several distributions (``\emph{experts}'') by multiplying them has been previously referred to as product-of-experts~\cite{hinton2002training}.

Our generator is trained to map a latent code $z$ to an image $x$.  Since the output image is uniquely determined by the latent code, the problem of estimating $p(x|\mathcal{Y})$ is equivalent to that of estimating $p(z|\mathcal{Y})$.
We use product-of-experts to model the conditional latent distribution
\begin{equation}
	p(z|\mathcal{Y})\propto p^{\prime}(z)\prod_{y_{i}\in\mathcal{Y}}q(z|y_{i}) \,,
	\label{eq:poe}
\end{equation}
where $p^{\prime}(z)$ is the prior distribution and
each expert $q(z|y_{i})$ is a distribution predicted by the encoder of that single modality. 
When no modalities are given, the latent distribution is simply the prior. As more modalities are provided, the number of constraints increases, the space of possible output images shrinks, and the latent distribution becomes narrower.

\subsection{Multiscale and hierarchical latent space}
\label{sec:latent}

Some of the modalities we consider~(\eg, sketch, segmentation) are two-dimensional and naturally contain information at multiple scales. Therefore, we devise a hierarchical latent space with latent variables at different resolutions. This allows us to directly pass information from each resolution of the spatial encoder to the corresponding resolution of the latent space, so that the high-resolution control signals can be better preserved. {Mathematically, our latent code is partitioned into groups $z=(z^0, z^1, ..., z^N)$ where $z^0\in\mathbb{R}^{c_0}$ is a feature vector and $z^k\in\mathbb{R}^{c_k\times r_k\times r_k}, 1\leq k\leq N$ are feature maps of increasing resolutions~($r_{k+1}=2r_{k}, r_{1}=4$, $r_{N}$ is the image resolution). We can therefore decompose the prior $p^{\prime}(z)$ into $\prod_{k=0}^{N}p^{\prime}(z^k|z^{<k})$ and the experts $q(z|y_{i})$ into $\prod_{k=0}^{N}q(z^k|z^{<k}, y_i)$.
This design is similar to the prior and posterior networks in hierarchical VAEs~\cite{sonderby2016ladder,maaloe2019biva,vahdat2020NVAE,child2021very}. The difference is that our encoder encodes the conditional information in the input modalities rather than the image itself.
Following \cref{eq:poe}, we assume the conditional latent distribution at each resolution is a product-of-experts
\begin{equation}
	p(z^k|z^{<k},\mathcal{Y})\propto p^{\prime}(z^k|z^{<k})\prod_{y_{i}\in\mathcal{Y}}q(z^{k}|z^{<k},y_{i})\,,
	\label{eq:poe_hier}
\end{equation}
where $p^{\prime}(z^k|z^{<k})=\mathcal{N}\left(\mu^{k}_{0}, \sigma^{k}_{0}\right)$ and  $q(z^{k}|z^{<k},y_{i})=\mathcal{N}\left(\mu^{k}_{i}, \sigma^{k}_{i}\right)$ are independent Gaussian distributions with mean and standard deviation parameterized by a neural network.\footnote{Except for $p^{\prime}(z^0)$, which is simply a standard Gaussian distribution.} It can be shown~\cite{williams2001products} that the product of Gaussian experts is also a Gaussian $p(z^k|z^{<k},\mathcal{Y})=\mathcal{N}(\mu^{k}, \sigma^{k})$, with
\begin{align}
	& \mu^{k}=\left(\frac{\mu^{k}_{0}}{(\sigma^{k}_{0})^{2}}+\sum_{i}\frac{\mu^{k}_{i}}{(\sigma^{k}_{i})^{2}}\right)\left(\frac{1}{(\sigma^{k}_{0})^{2}}+\sum_{i}\frac{1}{(\sigma^{k}_{i})^{2}}\right)^{-1},\nonumber\\ 
	& \sigma^{k}=\left(\frac{1}{(\sigma^{k}_{0})^{2}}+\sum_{i}\frac{1}{(\sigma^{k}_{i})^{2}}\right)^{-1}\,.
	\label{eq:poe_gaussian}
\end{align}
\subsection{Generator architecture}
\begin{figure}[!t]
	\centering
	{\includegraphics[width=\linewidth]{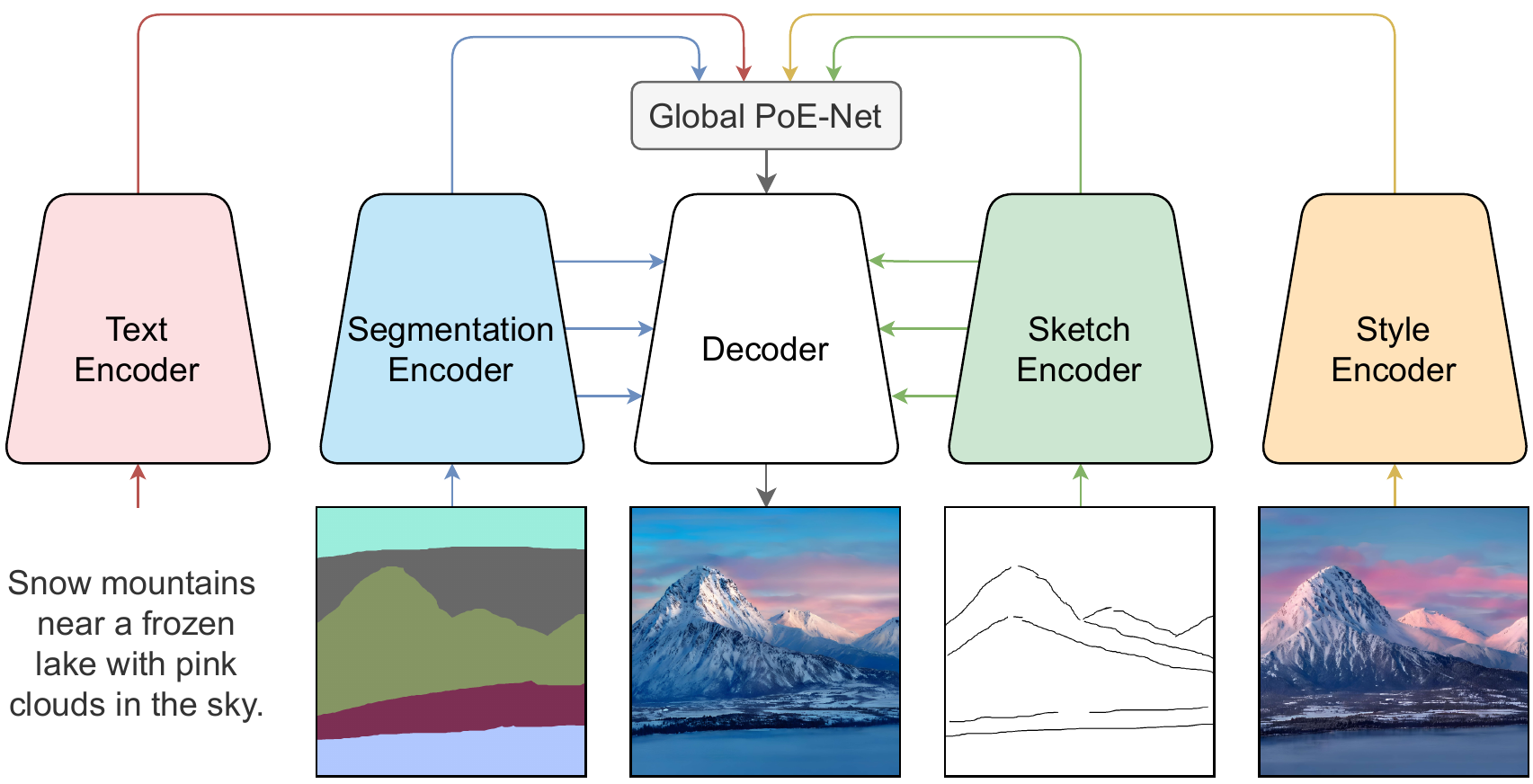}}
	\vspace{-6mm}
	\DeclareRobustCommand{\Switch}[1] { \ifthenelse{\boolean{arxiv}}{\cref{app:hyperparameters}}{Supplementary Material \ref*{app:hyperparameters}}}
	\caption{An overview of our generator. The detailed architectures of Global PoE-Net and the decoder are shown in \cref{fig:global_penet} and \cref{fig:decoder} respectively. The architectures of all modality encoders are provided in\Switch{1}.}
	\vspace{-3mm}
	\label{fig:overview_gen}
\end{figure}

\begin{figure}[!t]
	\centering
	{\includegraphics[width=0.85\linewidth]{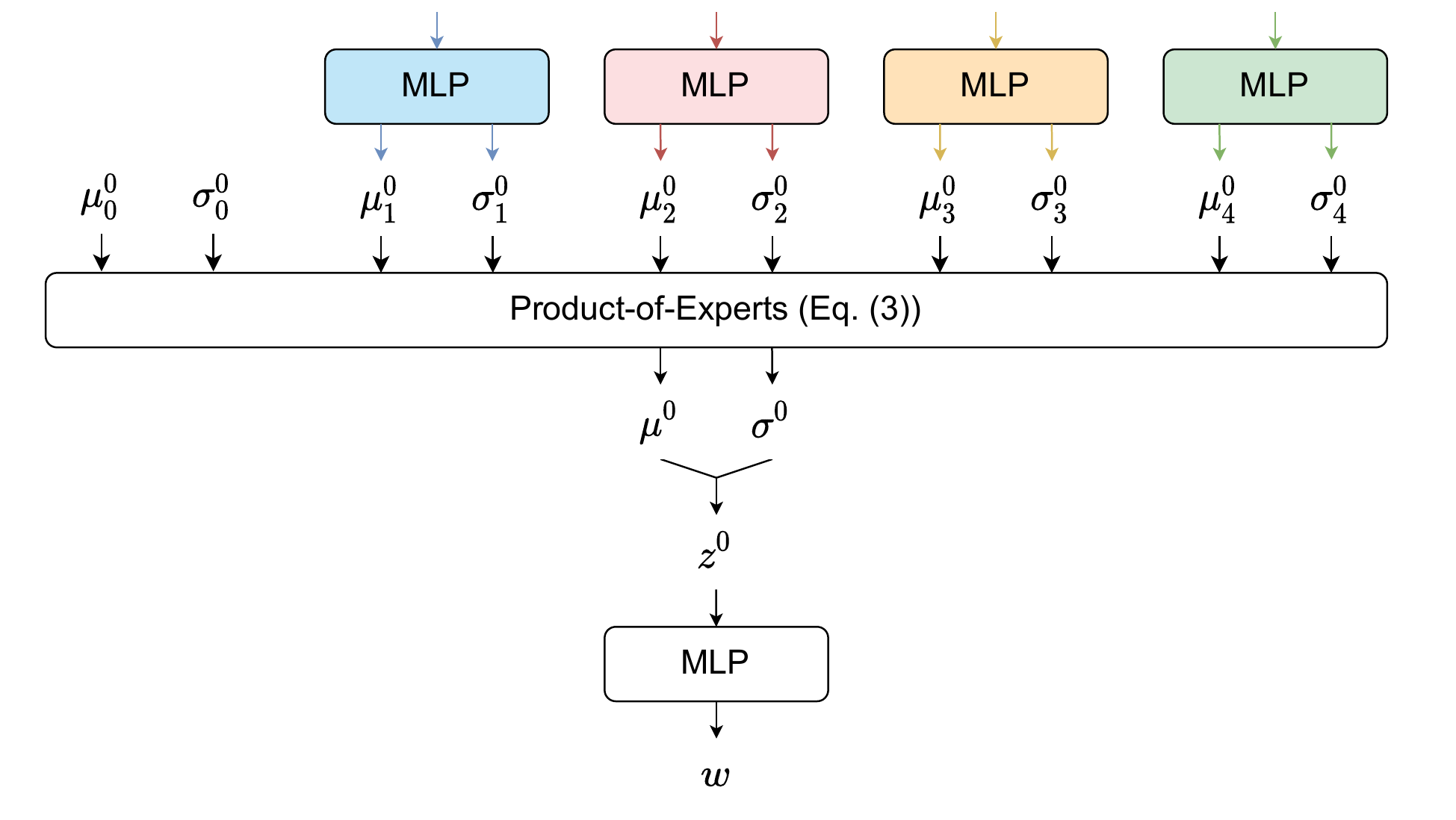}}
	\vspace{-2mm}
	\caption{Global PoE-Net. We sample a latent feature vector $z^{0}$ using product-of-experts~(\cref{eq:poe_gaussian} in \cref{sec:latent}), which is then processed by an MLP to output a feature vector $w$.}
	\vspace{-3mm}
	\label{fig:global_penet}
\end{figure}
\begin{figure}[t]
\centering
{\includegraphics[width=\linewidth]{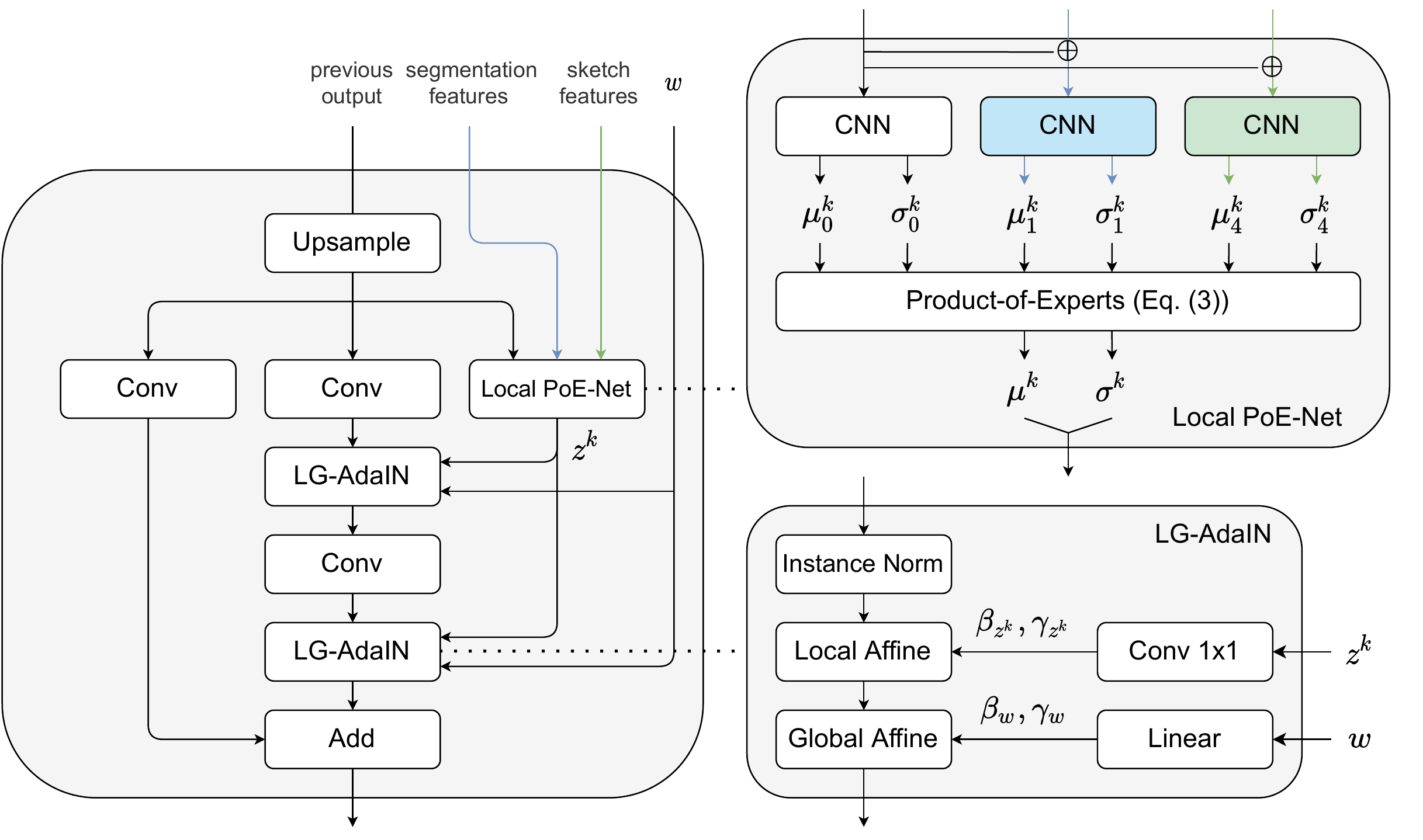}}
\vspace{-4mm}
\caption{A residual block in our decoder. Local PoE-Net samples a latent feature map $z^k$ using product-of-experts~(\cref{eq:poe_gaussian} in \cref{sec:latent}). Here $\oplus$ denotes concatenation. LG-AdaIN uses $w$ and $z^{k}$ to modulate the feature activations in the residual branch.}
\label{fig:decoder}
\vspace{-3mm}
\end{figure}
\Cref{fig:overview_gen} shows an overview of our generator architecture. We encode each modality into a feature vector which is then aggregated in \emph{Global PoE-Net}. We use convolutional networks with input skip connections to encode segmentation and sketch maps, a residual network to encode style images, and CLIP~\cite{radford2021learning} to encode text. Details of all modality encoders are given in \ifthenelse{\boolean{arxiv}}{\cref{app:hyperparameters}}{Supplementary Material \ref*{app:hyperparameters}}. The decoder generates the image using the output of Global PoE-Net and skip connections from the segmentation and sketch encoders. 

In Global PoE-Net~(\cref{fig:global_penet}), we predict a Gaussian $q(z^{0}|y_{i})=\mathcal{N}\left(\mu^{0}_{i}, \sigma^{0}_{i}\right)$ from the feature vector of each modality using an MLP. We then compute the product of Gaussians including the prior $p^{\prime}(z^{0})=\mathcal{N}(\mu^0_0, \sigma^0_0)=\mathcal{N}(0, I)$ and sample $z^{0}$ from the product distribution. We use another MLP to convert $z^0$ to another feature vector $w$.

The decoder mainly consists of a stack of residual blocks\footnote{Except for the first layer that convolves a constant feature map and the last layer that convolves the previous output to synthesize the output.}~\cite{he2016deep}, each of which is shown in \cref{fig:decoder}. \emph{Local PoE-Net} samples the latent feature map $z^{k}$ at the current resolution from the product of $p^{\prime}(z^{k}|z^{<k})=\mathcal{N}\left(\mu^{k}_{0}, \sigma^{k}_{0}\right)$ and $q(z^{k}|z^{<k}, y_{i})=\mathcal{N}\left(\mu^{k}_{i}, \sigma^{k}_{i}\right), \forall y_{i}\in\mathcal{Y}$, where $\left(\mu^{k}_{0}, \sigma^{k}_{0}\right)$ is computed from the output of the last layer and $\left(\mu^{k}_{i}, \sigma^{k}_{i}\right)$ is computed by concatenating the output of the last layer and the skip connection from the corresponding modality. Note that only modalities that have skip connections~(segmentation and sketch, i.e.\,$i=1,4$) contribute to the computation. Other modalities~(text and style reference) only provide global information but not local details. The latent feature map $z^k$ produced by \emph{Local PoE-Net} and the feature vector $w$ produced by \emph{Global PoE-Net} are fed to our  
local-global adaptive instance normalization~(LG-AdaIN) layer,
\begin{equation}
\resizebox{0.9\columnwidth}{!}{%
	$\text{LG-AdaIN}(h^{k}, z^{k}, w) = \gamma_{w}\left(\gamma_{z^{k}}\frac{h^{k}-\mu(h^{k})}{\sigma(h^{k})}+\beta_{z^{k}}\right)+\beta_{w}$\,,}
\end{equation}       
where $h^k$ is a feature map in the residual branch after convolution, $\mu(h^{k})$ and $\sigma(h^{k})$ are channel-wise mean and standard deviation. $\beta_{w}$, $\gamma_{w}$ are feature vectors computed from $w$, while $\beta_{z^{k}}$, $\gamma_{z^{k}}$ are feature maps computed from $z^{k}$. 
The LG-AdaIN layer can be viewed as a combination of AdaIN~\cite{huang2017arbitrary} and SPADE~\cite{park2019semantic} that takes both a global feature vector and a spatially-varying feature map to modulate the activations.

\begin{figure}[t]
	\centering
	\subcaptionbox{Projection discriminator}[2.9cm]
	{\includegraphics[height=0.18\textheight]{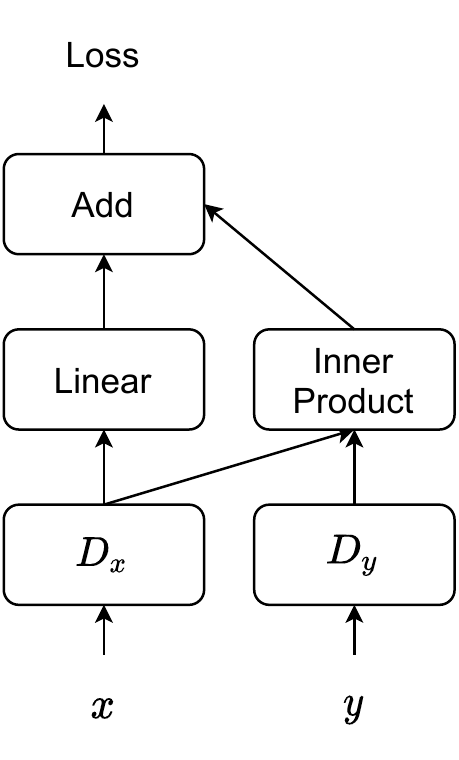}}
	\hfill
	\subcaptionbox{Multimodal projection discriminator\label{fig:mpd}}[5.2cm]
	{\includegraphics[height=0.18\textheight]{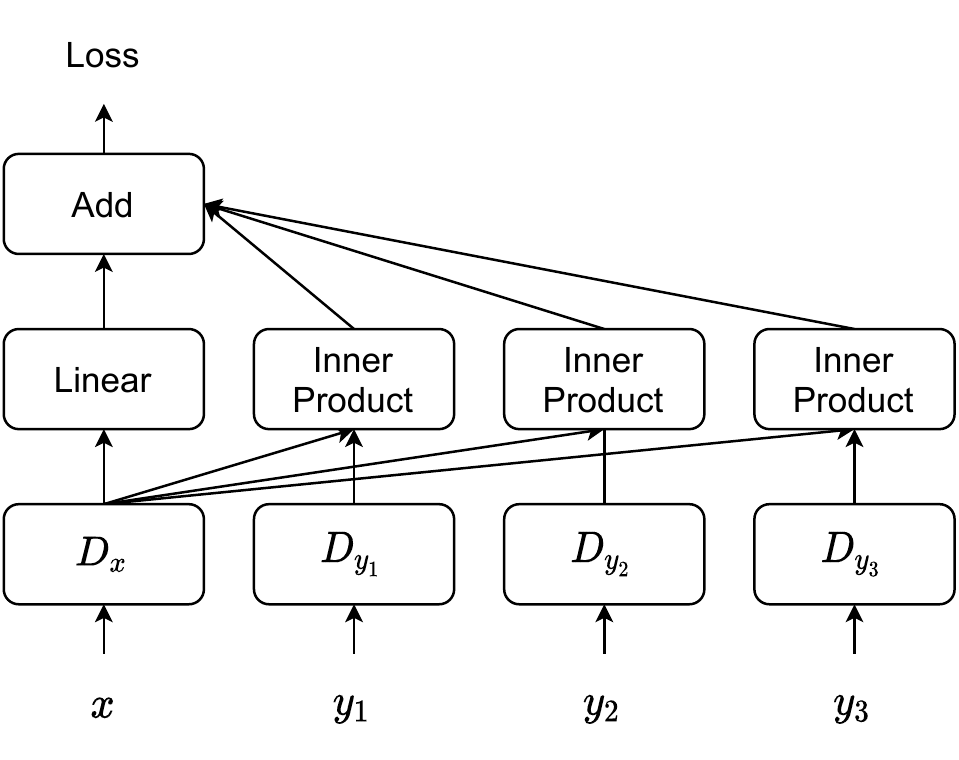}}
	\vspace{-2mm}
	\caption{Comparison between the standard projection discriminator and our multimodal projection discriminator~(MPD).}
	\label{fig:pdmpd}
    \vspace{2mm}
	\centering
	{\includegraphics[width=0.65\linewidth]{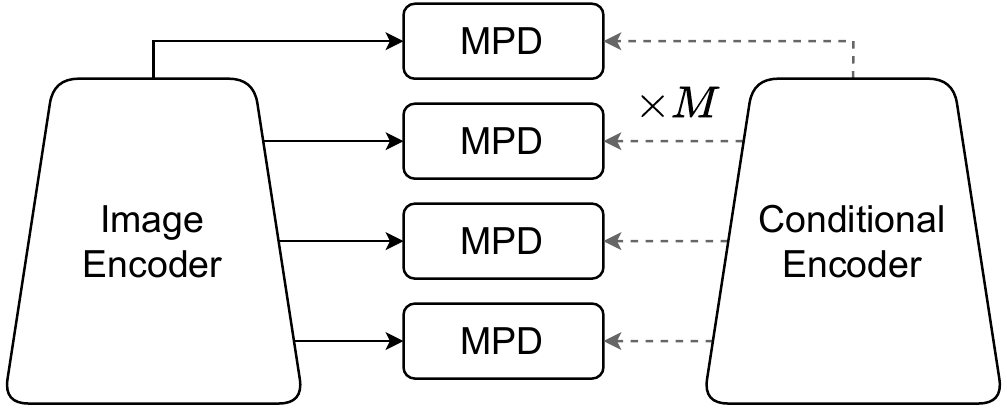}}
	\vspace{-2mm}
	\caption{Our multiscale multimodal projection discriminator. The image and other modalities are encoded into feature maps of multiple resolutions and we compute the MPD loss at each resolution.}
	\label{fig:mmpd}
	\vspace{-3mm}
\end{figure}

\subsection{\mbox{Multiscale multimodal projection discriminator}}
\label{sec:discriminator}
Our discriminator receives the image $x$ and a set of conditions $\mathcal{Y}$ as inputs and produces a score $D(x, \mathcal{Y})=\text{sigmoid}(f(x, \mathcal{Y}))$ indicating the realness of $x$ given $\mathcal{Y}$ . Under the GAN objective~\cite{goodfellow2014generative}, the optimal solution of $f$ is
\begin{equation}\textsc{}
	f^*(x, \mathcal{Y}) = \underbrace{\log\frac{q(x)}{p(x)}}_{\text{unconditional term}}+\sum_{y_{i}\in\mathcal{Y}}\underbrace{\log\frac{q(y_{i}|x)}{p(y_{i}|x)}}_{\text{conditional term}}\,,
\end{equation}
if we assume conditional independence of different modalities given $x$. 
The projection discriminator~(PD)~\cite{miyato2018cgans} proposes to use the inner product to estimate the conditional term. This implementation restricts the conditional term to be relatively simple, which imposes a good inductive bias that leads to strong empirical results. We propose a multimodal projection discriminator~(MPD) that generalizes PD to our multimodal setting. As shown in \cref{fig:pdmpd}, the original PD first encodes both the image and the conditional input into a shared latent space. It then uses a linear layer to estimate the unconditional term from the image embedding and uses the inner product between the image embedding and the conditional embedding to estimate the conditional term. The unconditional term and the conditional term are summed to obtain the final discriminator logits. In our multimodal scenario, we simply encode each observed modality and add its inner product with the image embedding to the final loss
\begin{equation}
	f(x, \mathcal{Y}) = \text{Linear}(D_{x}(x)) + \sum_{y_{i}\in\mathcal{Y}}D^{T}_{y_{i}}(y_i)D_{x}(x)\,.
	\label{eq:projd}
\end{equation}

For spatial modalities such as segmentation and sketch, it is more effective to enforce their alignment with the image in multiple scales~\cite{liu2019learning}. As shown in \cref{fig:mmpd}, we encode the image and spatial modalities into feature maps of different resolutions and compute the MPD loss at each resolution.
We compute a loss value at each location and resolution, and obtain the final loss by averaging first across locations then across resolutions.
We name the resulting discriminator as the multiscale multimodal projection discriminator~(MMPD) and describe its details in \ifthenelse{\boolean{arxiv}}{\cref{app:hyperparameters}}{Supplementary Material \ref*{app:hyperparameters}}.


\subsection{Losses and training procedure}
\label{sec:training}

\noindent{\bf Latent regularization.} Under the product-of-experts assumption~(\cref{eq:poe}), the marginalized conditional latent distribution should match the unconditional prior distribution:
\begin{equation}
	\int p(z|y_{i})p(y_{i})dy_{i}=p(z|\varnothing)=p^{\prime}(z)\,.
\end{equation}
To this end, we minimize the Kullback-Leibler~(KL) divergence from the prior distribution $p^{\prime}(z)$ to the conditional latent distribution $p(z|y_{i})$ at every resolution
\begin{align}
\mathcal{L}_{\text{KL}} = 	&\sum_{y_{i}\in\mathcal{Y}}\omega_{i}\sum_{k}\omega^{k}\mathbb{E}_{p(z^{<k}|y_{i})}\big{[}\nonumber\\
&D_{\text{KL}}(p(z^{k}|z^{<k}, y_{i})||p^{\prime}(z^{k}|z^{<k}))\big{]}\,,\label{eq:kl}
\end{align}
where $\omega^{k}$ is a resolution-dependent rebalancing weight and $\omega_{i}$ is a modality-specific loss weight. We describe both weights in detail in \ifthenelse{\boolean{arxiv}}{\cref{app:hyperparameters}}{Supplementary Material \ref*{app:hyperparameters}}. 

The KL loss also reduces conditional mode collapse since it encourages the conditional latent distribution to be close to the prior and therefore have high entropy. From the perspective of information bottleneck\mbox{~\cite{alemi2017deep}}, the KL loss encourages each modality to only provide the minimum information necessary to specify the conditional image distribution.

\begin{figure*}[!t]
	\centering
	\newcommand{\sizea}{0.2\linewidth}
	\setlength{\tabcolsep}{0.5pt}
	\renewcommand{\arraystretch}{0.3}
	\begin{tabular}{ccccc}
		Segmentation & Ground truth & SPADE & OASIS & PoE-GAN~(Ours) \\
		\includegraphics[width=\sizea]{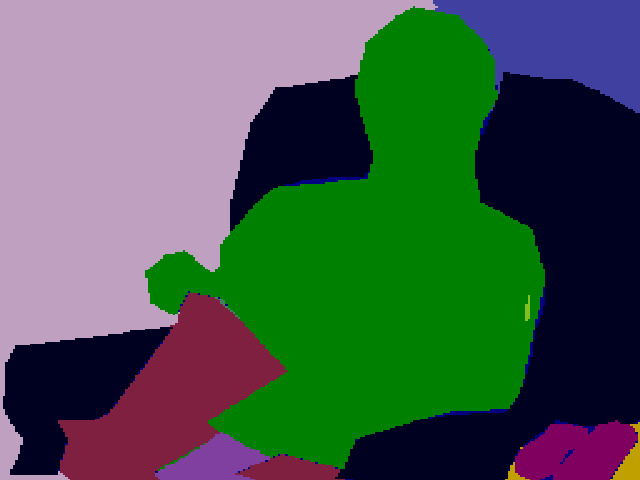}     & \includegraphics[width=\sizea]{figures/coco_seg/gt_000000055528.jpg}    & \includegraphics[width=\sizea]{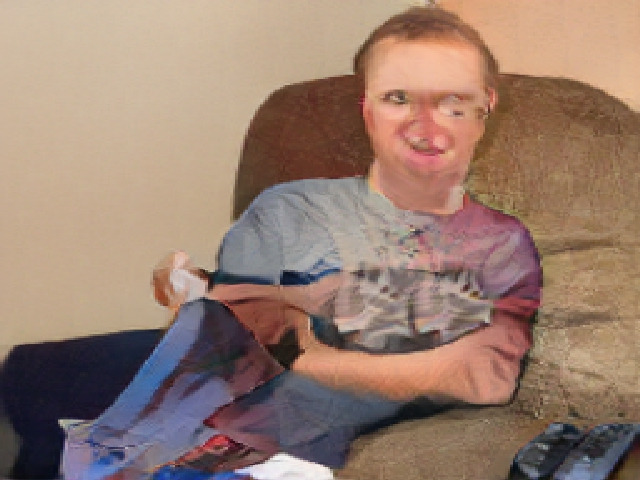}  & \includegraphics[width=\sizea]{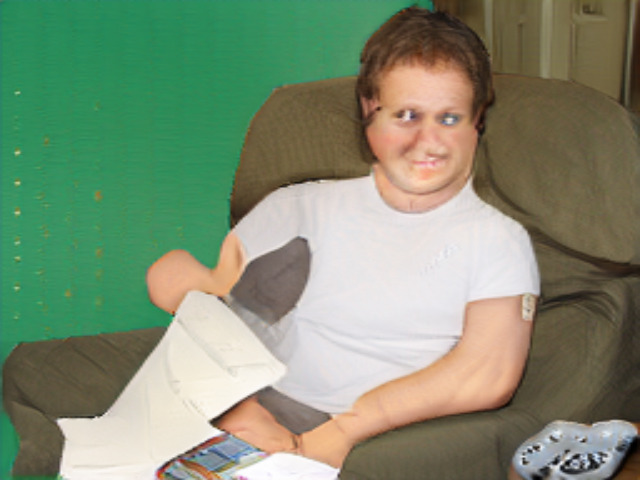}  & \includegraphics[width=\sizea]{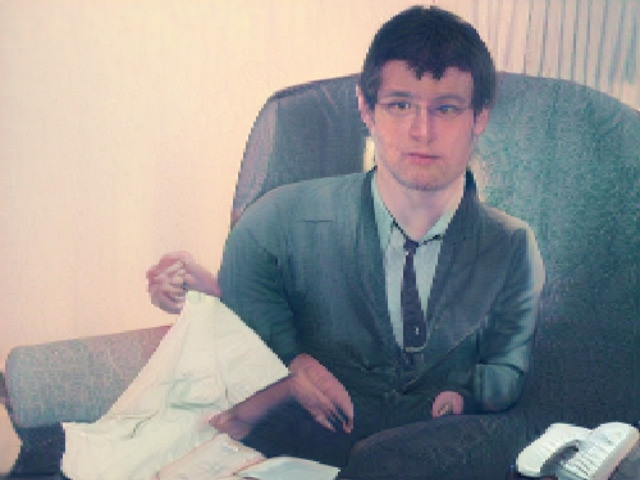} \\
		\includegraphics[width=\sizea]{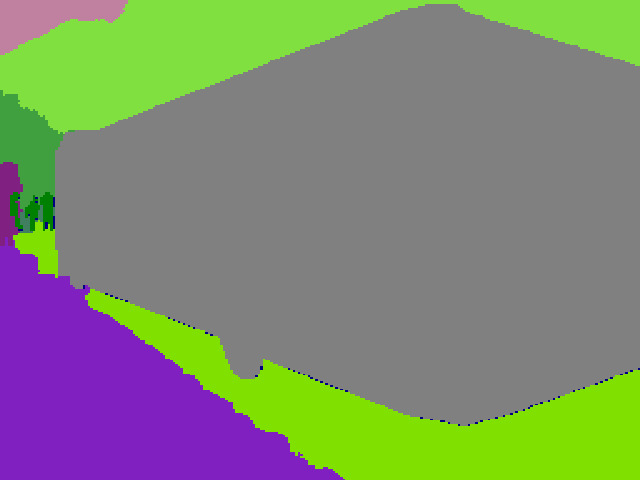}     & \includegraphics[width=\sizea]{figures/coco_seg/gt_000000357748.jpg}    & \includegraphics[width=\sizea]{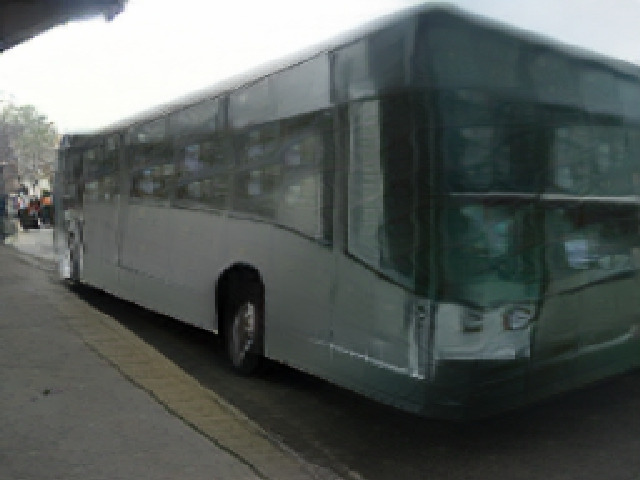}  & \includegraphics[width=\sizea]{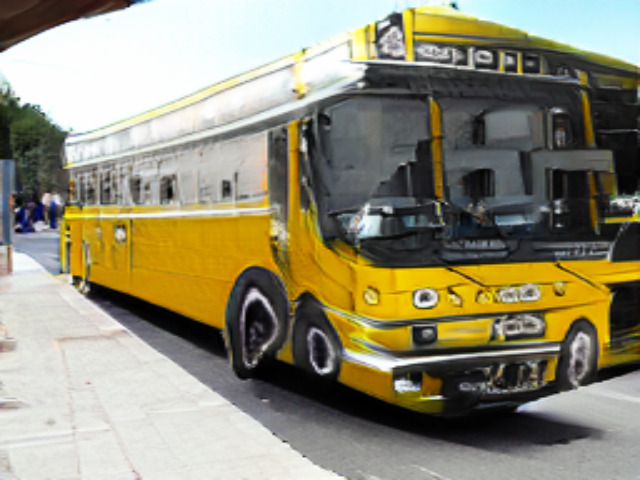}  & \includegraphics[width=\sizea]{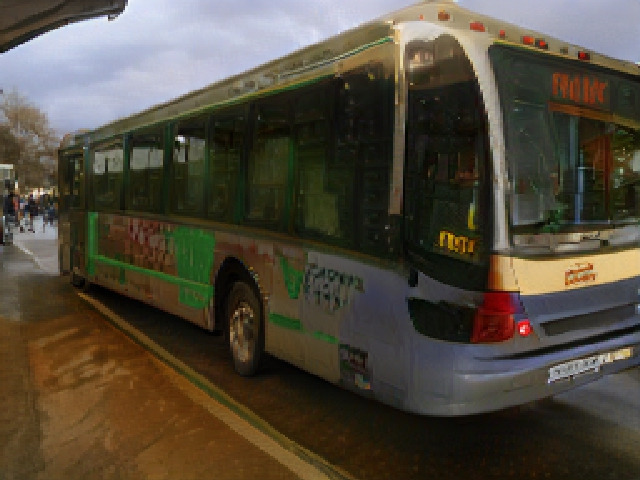} \\
	\end{tabular}
	\vspace{-3mm}
	\caption{Qualitative comparison of segmentation-to-image synthesis results on MS-COCO 2017.}
	\label{fig:coco_seg}
    \vspace{-3mm}	
\end{figure*}
\begin{figure*}[!t]
	\centering
	\newcommand{\sizea}{0.2\linewidth}
	\setlength{\tabcolsep}{0.5pt}
	\renewcommand{\arraystretch}{0.3}
	\begin{tabular}{ccccc}
		Text & Ground truth & DF-GAN & DM-GAN + CL & PoE-GAN~(Ours) \\
	    \begin{minipage}[b]{\sizea}
	    	\centering A man riding a snowboard down a \\ snow covered slope.\vspace{20pt} 
	    \end{minipage}     
	    & \includegraphics[width=\sizea]{figures/coco_text/gt_000000117645.jpg}    & \includegraphics[width=\sizea]{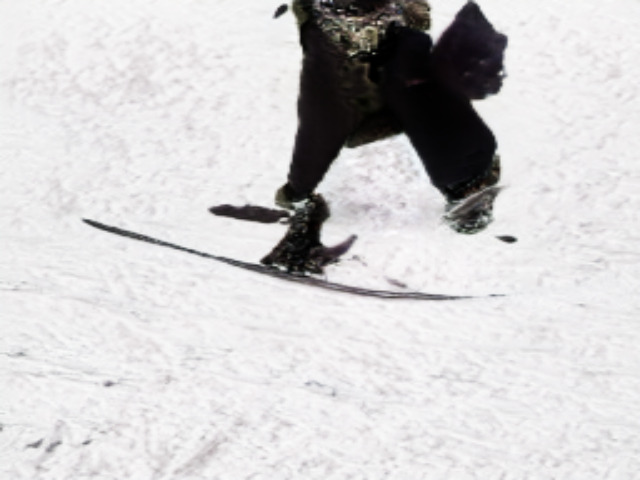}  & \includegraphics[width=\sizea]{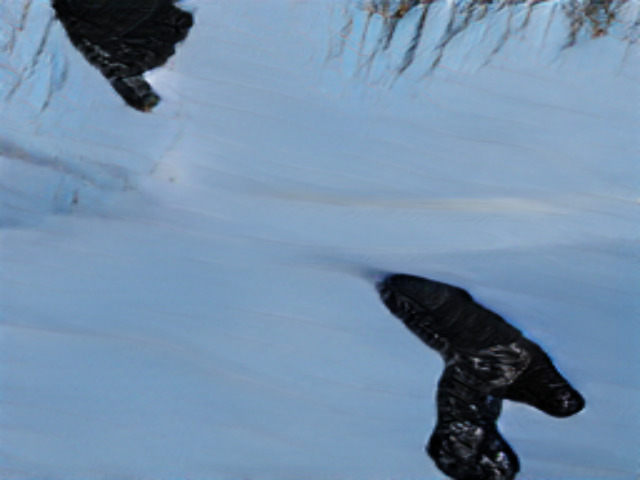}  & \includegraphics[width=\sizea]{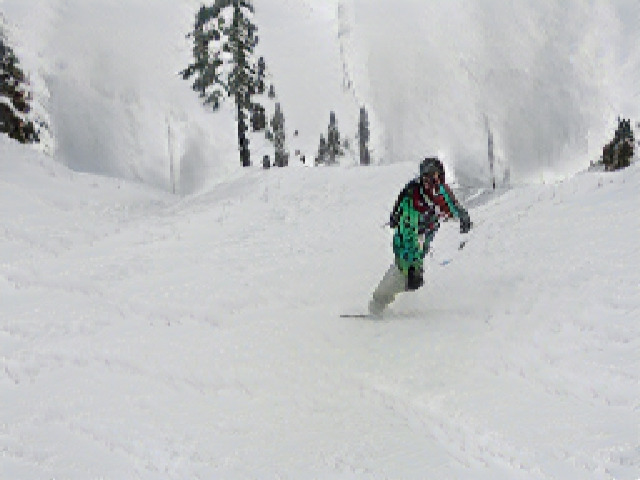} \\
	    \begin{minipage}[b]{\sizea}
	    	\centering A red stop sign sitting \\ on the side of a road.\vspace{26pt} 
	    \end{minipage}     
	    & \includegraphics[width=\sizea]{figures/coco_text/gt_000000193884.jpg}    & \includegraphics[width=\sizea]{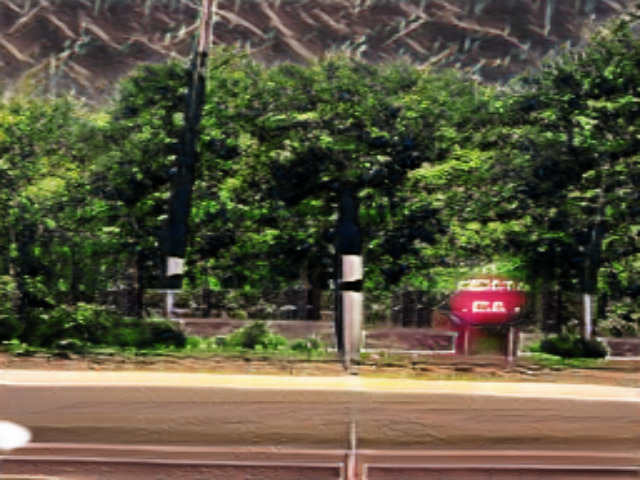}  & \includegraphics[width=\sizea]{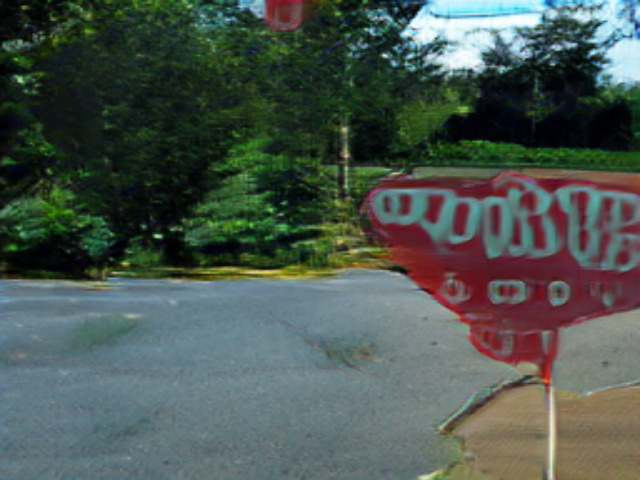}  & \includegraphics[width=\sizea]{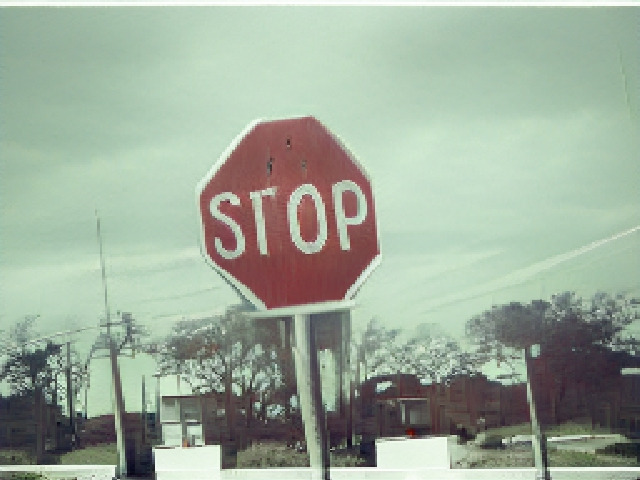}
	\end{tabular}
	\vspace{-3mm}
	\caption{Qualitative comparison of text-to-image synthesis results on MS-COCO 2017.}
	\label{fig:coco_text}
	\vspace{-5mm}
\end{figure*}
\medskip
\noindent{\bf Contrastive losses.} The contrastive loss has been widely adopted in representation learning~\cite{chen2020simple,he2020momentum} and more recently in image synthesis~\cite{park2020contrastive,zhang2021cross,liu2021divco,han2021dual}. Given a batch of paired vectors $(\mathbf{u}, \mathbf{v})=\{(u_{i}, v_{i}), i=1, 2, ..., N\}$, the symmetric cross-entropy loss~\cite{zhang2020contrastive,radford2021learning} maximizes the similarity of the vectors in a pair while keeping non-paired vectors apart
\begin{align}
	\mathcal{L}^{\text{ce}}(&\mathbf{u}, \mathbf{v}) =  -\frac{1}{2N}\sum_{i=1}^{N}\log\frac{\exp(\cos(u_{i}, v_{i})/\tau)}{\sum_{j=1}^{N}\exp(\cos(u_{i}, v_{j})/\tau)}  \nonumber\\ &
	 -\frac{1}{2N}\sum_{i=1}^{N}\log\frac{\exp(\cos(u_{i}, v_{i})/\tau)}{\sum_{j=1}^{N}\exp(\cos(u_{j}, v_{i})/\tau)}
	\label{eq:ce2}\,,
\end{align}
where $\tau$ is a temperature hyper-parameter. We use two kinds of pairs to construct two contrastive loss terms: the image contrastive loss and the conditional contrastive loss. 

The \textit{image contrastive loss} maximizes the similarity between a real image $x$ and a random fake image $\tilde{x}$ synthesized based on the corresponding conditional inputs:
\begin{equation}
	\mathcal{L}_{cx} = \mathcal{L}^{\text{ce}}(E_{\text{vgg}}(\mathbf{x}), E_{\text{vgg}}(\tilde{\mathbf{x}}))\,,
\end{equation}
where $E_{\text{vgg}}$ is a pretrained VGG~\cite{simonyan2015very} encoder. This loss is similar to the perceptual loss widely used in conditional synthesis but has been found to perform better~\cite{park2020contrastive,zhang2021cross}.

The \textit{conditional contrastive loss} aims to better align images with the corresponding conditions. Specifically, the discriminator is trained to maximize the similarity between its embedding of a real image $\mathbf{x}$ and \mbox{the conditional input $\mathbf{y}_{i}$.}
\begin{equation}
\mathcal{L}_{cy}^{D} = \mathcal{L}^{\text{ce}}(D_{x}(\mathbf{x}), D_{y_{i}}(\mathbf{y}_{i}))\,,
\end{equation}
where $D_x$ and $D_{y_{i}}$ are two modules in the discriminator that extract features from $x$ and $y_i$, respectively, as shown in \cref{eq:projd} and  \cref{fig:mpd}. The generator is trained with the same loss, but using the generated image $\tilde{\mathbf{x}}$ instead of the real image to compute the discriminator embedding,
\begin{equation}
\mathcal{L}_{cy}^{G} = \mathcal{L}^{\text{ce}}(D_{x}(\tilde{\mathbf{x}}), D_{y_{i}}(\mathbf{y}_{i}))\,.
\end{equation}
In practice, we only use the conditional contrastive loss for the text modality since it consumes too much GPU memory to use the conditional contrastive loss for the other modalities, especially when the image resolution and batch size are large. A similar image-text contrastive loss is used in XMC-GAN~\cite{zhang2021cross}, where they use a non-symmetric cross-entropy loss that only includes the first term in \cref{eq:ce2}.

\medskip
\noindent{\bf Full training objective.} 
In summary, the generator loss $\mathcal{L}^{G}$ and the discriminator loss $\mathcal{L}^{D}$ can be written as
\begin{align}
	& \mathcal{L}^{G}=\mathcal{L}^{G}_{\text{GAN}} + \mathcal{L}_{\text{KL}} + \lambda_{1}\mathcal{L}_{cx} + \lambda_{2}\mathcal{L}_{cy}^{G}\,, \\
	& \mathcal{L}^{D}=\mathcal{L}^{D}_{\text{GAN}} + \lambda_{2}\mathcal{L}_{cy}^{D} + \lambda_{3}\mathcal{L}_{\text{GP}}\,,
\end{align}
where $\mathcal{L}^{G}_{\text{GAN}}$ and $\mathcal{L}^{D}_{\text{GAN}}$ are non-saturated GAN losses~\cite{goodfellow2014generative}, $\mathcal{L}_{\text{GP}}$ is the $R_1$ gradient penalty loss~\cite{mescheder2018training}, and $\lambda_{1}, \lambda_{2}, \lambda_{3}$ are weights associated with the loss terms.

\medskip
\noindent{\bf Modality dropout.} 
By design, our generator, discriminator, and loss terms are able to handle missing modalities. We also find that randomly dropping out some input modalities before each training iteration further improves the robustness of the generator towards missing modalities at test time.

	\begin{table}[t]
	\centering
	\setlength{\tabcolsep}{1pt}
	\begin{small}
		\begin{tabularx}{\linewidth}{l*{5}{Y}}
			\toprule
			& Uncond & Text & Seg & Sketch & All \\
			StyleGAN2~\cite{karras2020analyzing}    & 11.7 & ---  & ---  & ---  & ---  \\
			SPADE-Seg~\cite{park2019semantic}       & ---  & ---  & 48.6 & ---  & ---  \\
			pSp-Seg~\cite{richardson2021encoding}   & ---  & ---  & 44.1 & ---  & ---  \\
			SPADE-Sketch~\cite{park2019semantic}    & ---  & ---  & ---  & 33.0 & ---  \\
			pSp-Sketch~\cite{richardson2021encoding}& ---  & ---  & ---  & 45.8 & ---  \\ \midrule
			TediGAN~\cite{xia2021tedigan}           & ---  & 38.4 & 45.1 & 45.1 &  45.1  \\
			PoE-GAN~(Ours)              & \textbf{10.5} & \textbf{10.1} & \textbf{9.9} & \textbf{9.9} & \textbf{8.3} \\
			\bottomrule
		\end{tabularx}
	\end{small}
	\vspace{-2mm}
	\caption{Comparison on MM-CelebA-HQ (1024$\times$1024) using FID. We evaluate models 	conditioned on different modalities~(from left to right: no conditions, text, segmentation, sketch, and all three modalities). The best scores are highlighted in bold.}
	\label{tab:main_celeba}
	\vspace{-3mm}
\end{table}

\begin{table}[t]
	\centering
	\setlength{\tabcolsep}{1pt}
	\begin{small}
		\begin{tabularx}{\linewidth}{l*{5}{Y}}
			\toprule
			& Uncond & Text & Seg & Sketch & All \\
			StyleGAN2~\cite{karras2020analyzing}            & 43.6 & --- & --- & ---  & --- \\
			DF-GAN~\cite{ming2020DFGAN}            & --- & 45.2 & --- & ---  & --- \\
			DM-GAN + CL~\cite{ye2021improving}          & --- & 29.9 & --- & ---  & --- \\
			SPADE-Seg~\cite{park2019semantic}        & ---  & ---  & 22.1 & --- & --- \\
			VQGAN~\cite{esser2021taming}            & ---  & ---  & 21.6 & ---  & --- \\
			OASIS~\cite{schonfeld2020you}            & ---  & ---  & 19.2 & ---  & --- \\
			SPADE-Sketch~\cite{park2019semantic}       & ---  & ---  & ---  & 63.7 & --- \\ \midrule
			PoE-GAN~(Ours)              & \textbf{43.4} & \textbf{20.5} & \textbf{15.8} & \textbf{25.5}& \textbf{13.6} \\
			\bottomrule
		\end{tabularx}
	\end{small}
	\vspace{-2mm}
	\caption{Comparison on MS-COCO 2017~(256$\times$256) using FID. We evaluate models conditioned on different modalities~(from left to right: no conditions, text, segmentation, sketch, and all three modalities). The best scores are highlighted in bold.}
	\label{tab:main_coco}
	\vspace{-3mm}
\end{table}

\section{Experiments}
\label{sec:experiments}
We evaluate the proposed approach on several datasets, including MM-CelebA-HQ~\cite{xia2021tedigan}, MS-COCO 2017~\cite{lin2014microsoft} with COCO-Stuff annotations~\cite{caesar2018cvpr}, and a proprietary dataset of landscape images. Images are labeled with all input modalities obtained from either manual annotation or pseudo-labeling methods. More details about the datasets and the pseudo-labeling procedure are in \ifthenelse{\boolean{arxiv}}{\cref{app:datasets}}{Supplementary Material \ref*{app:datasets}}.

\begin{figure*}
	\centering
	\begin{overpic}[width=0.32\textwidth, height=0.32\textwidth]{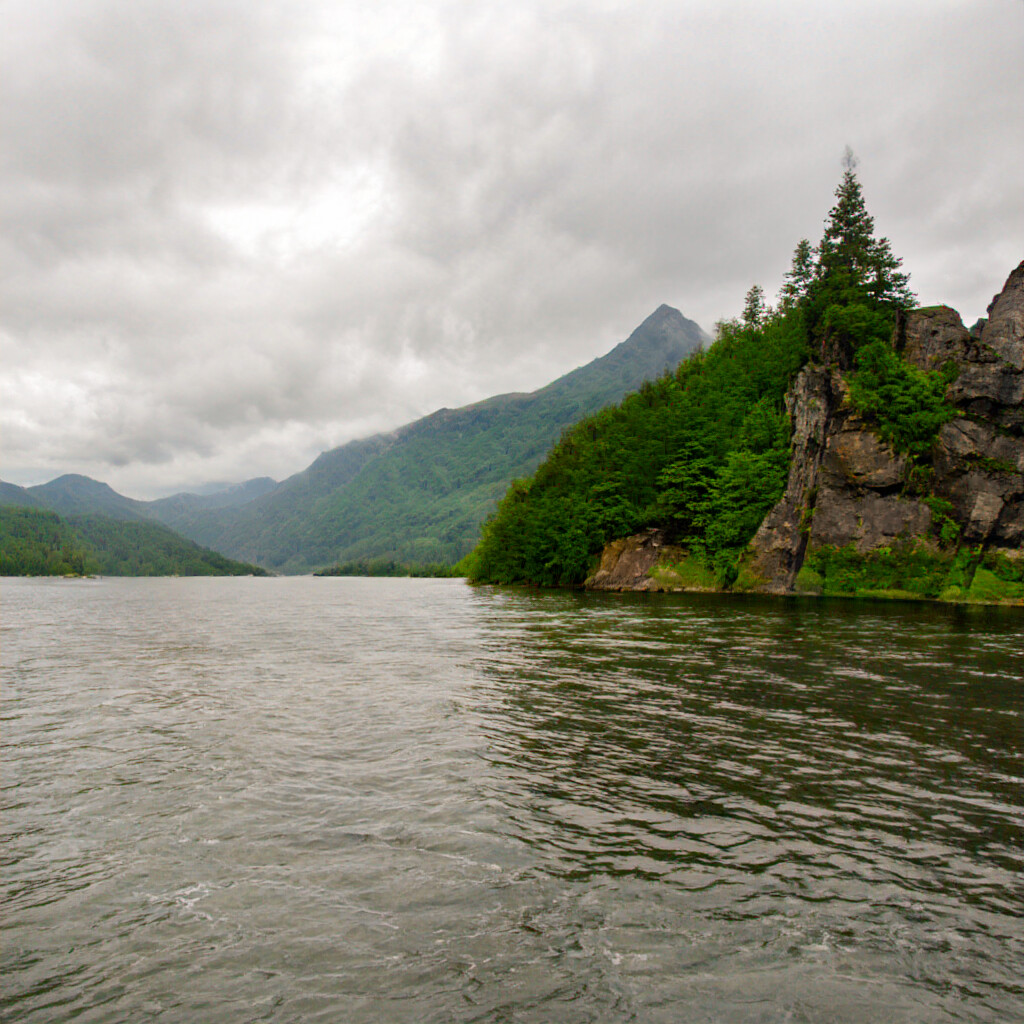}
	\put(79.5, 0.2){
		\frame{\includegraphics[width=0.06\textwidth, height=0.06\textwidth]{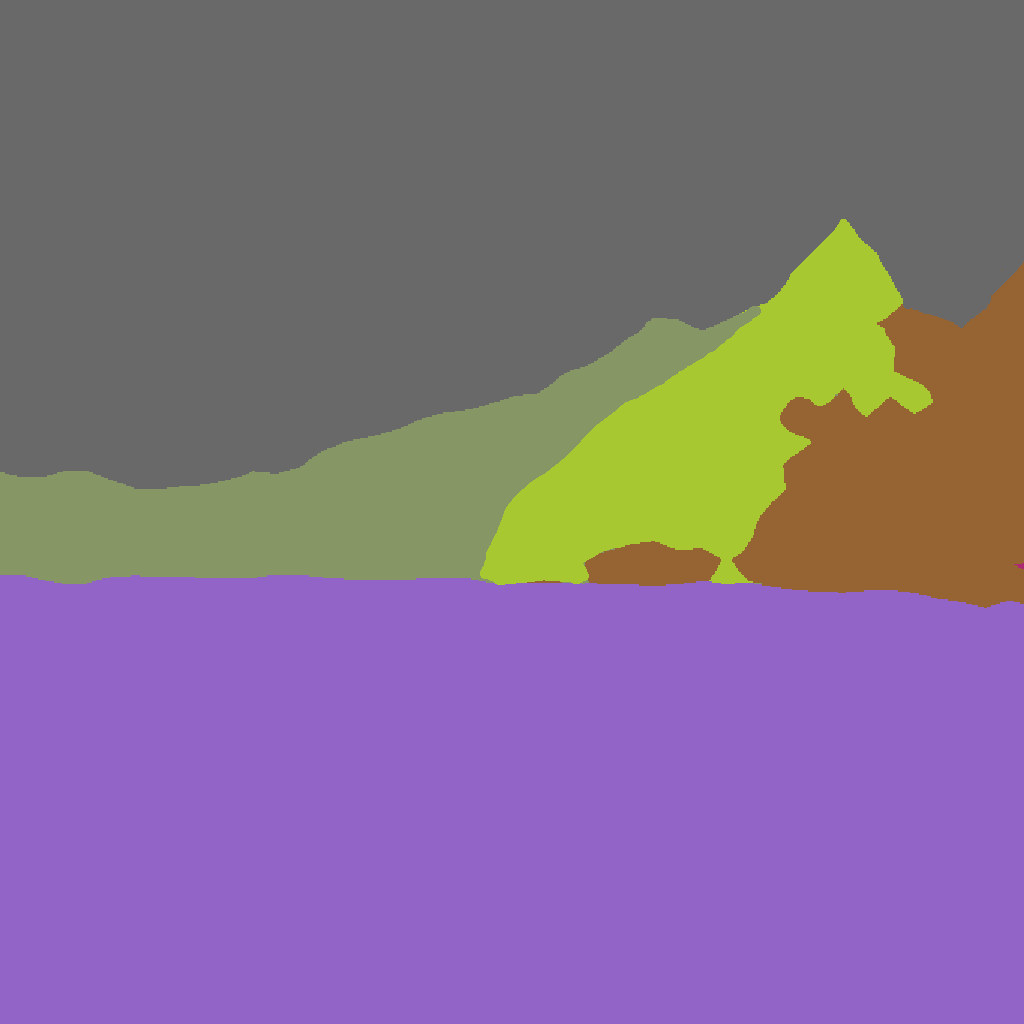}}}
	\end{overpic}
	\begin{overpic}[width=0.32\textwidth, height=0.32\textwidth]{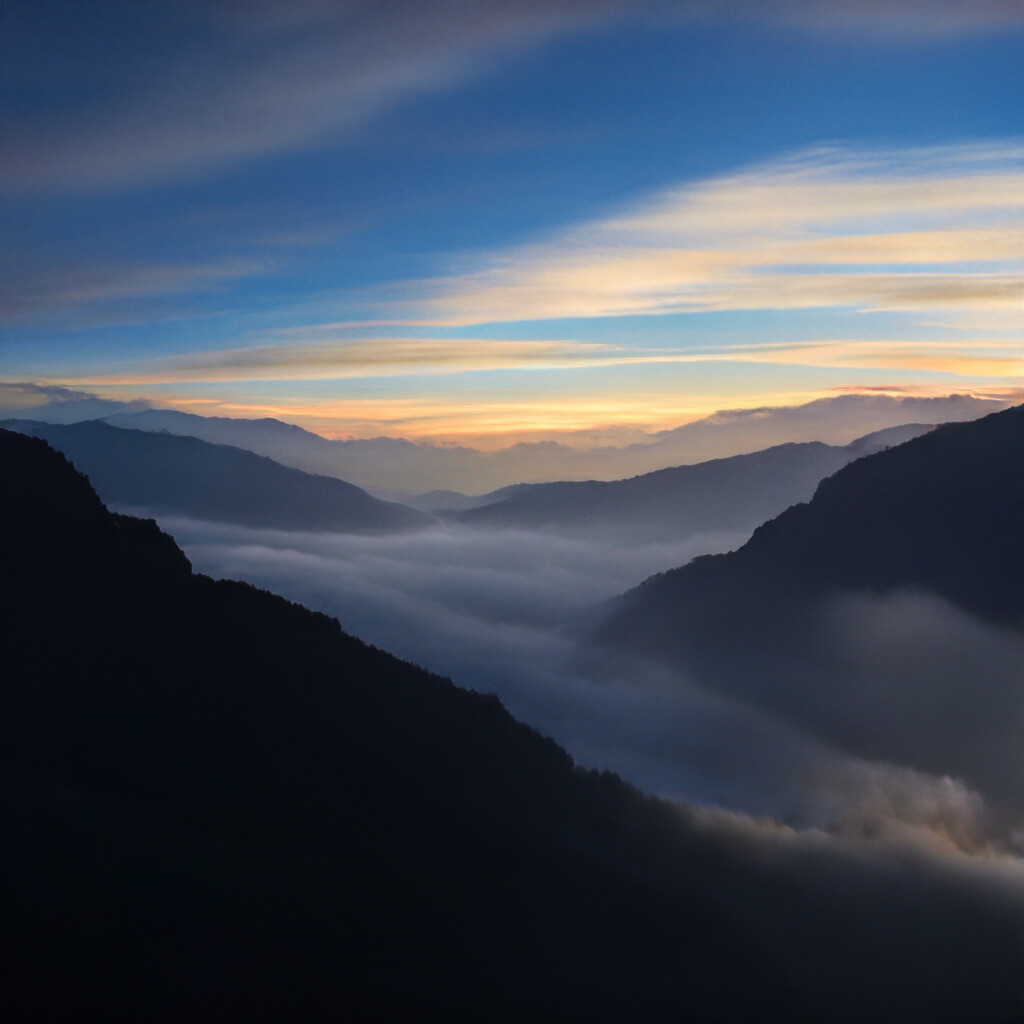}
	\put(79.5, 0.2){
		\frame{\includegraphics[width=0.06\textwidth, height=0.06\textwidth]{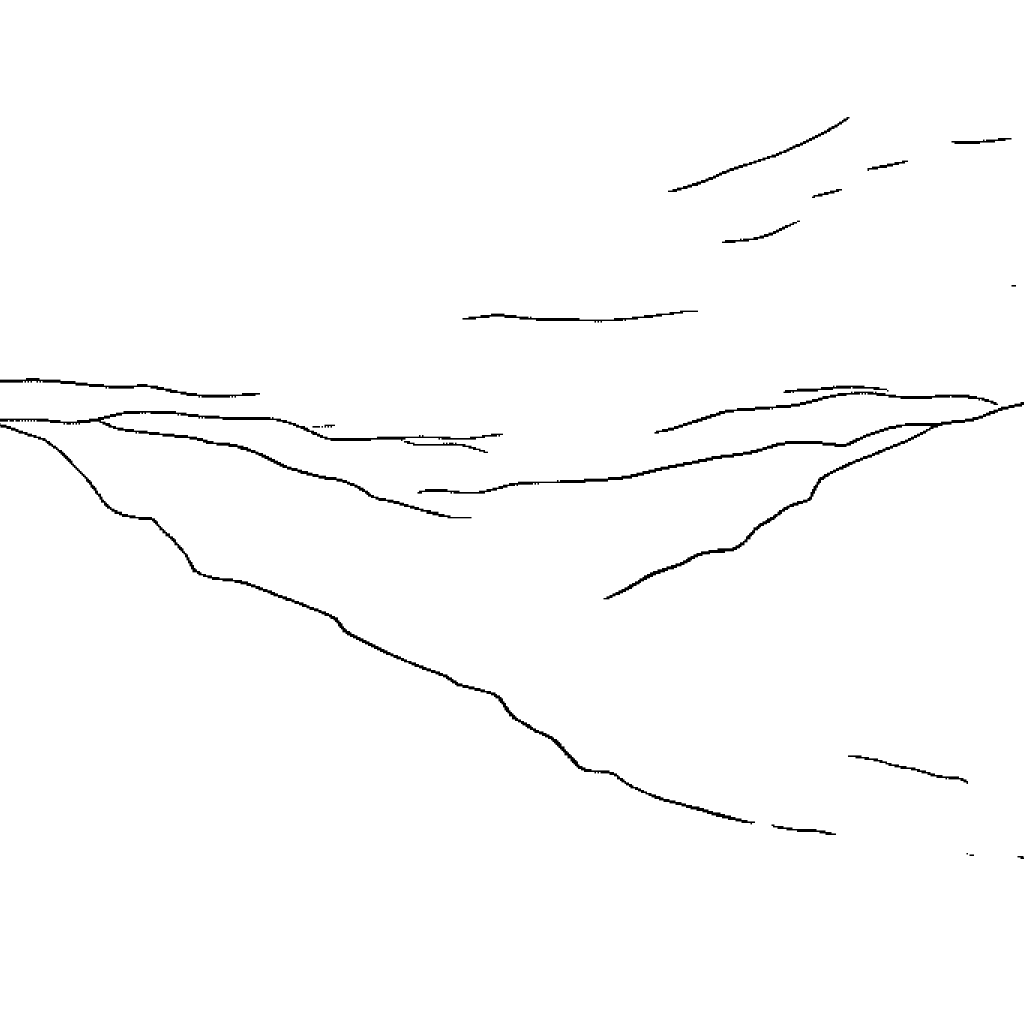}}}
	\end{overpic}
	\begin{overpic}[width=0.32\textwidth, height=0.32\textwidth]{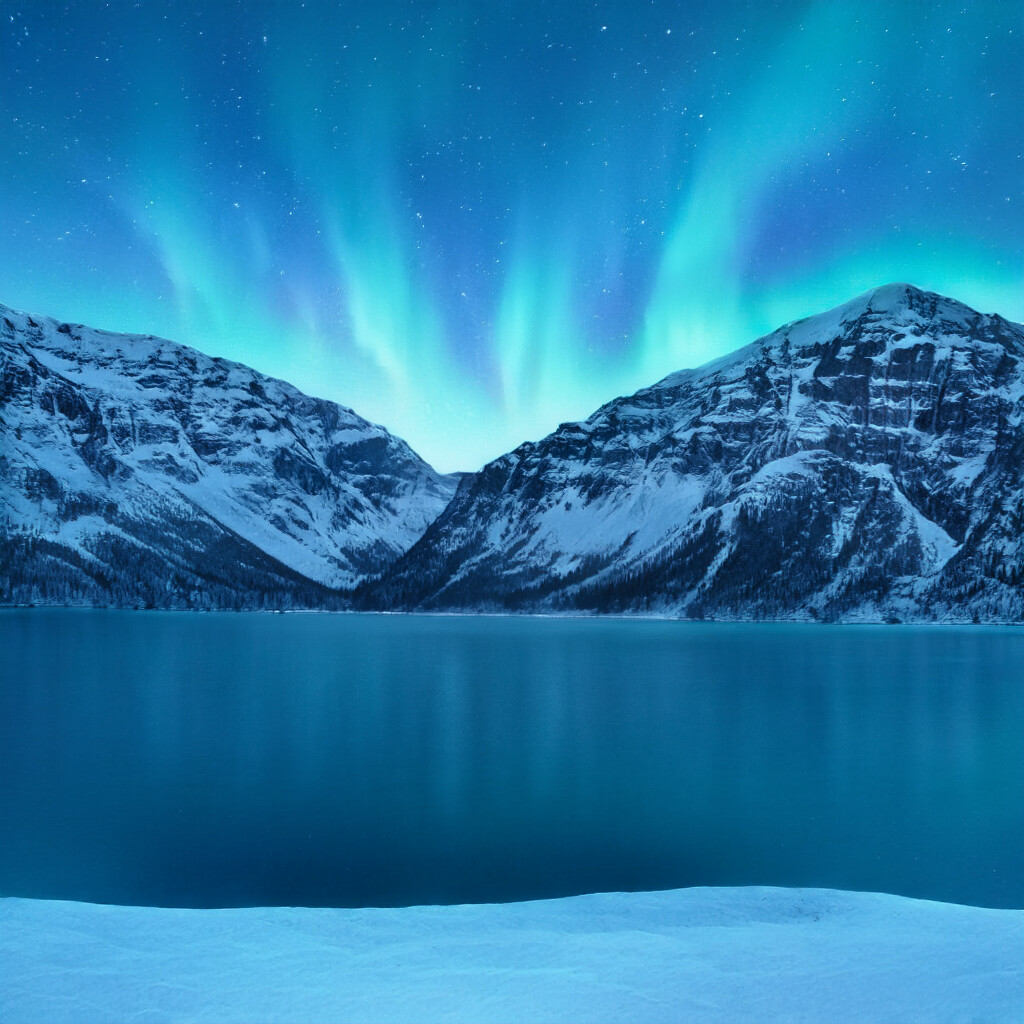}
		\put(60.5, 0.2){
			\frame{\includegraphics[width=0.06\textwidth, height=0.06\textwidth]{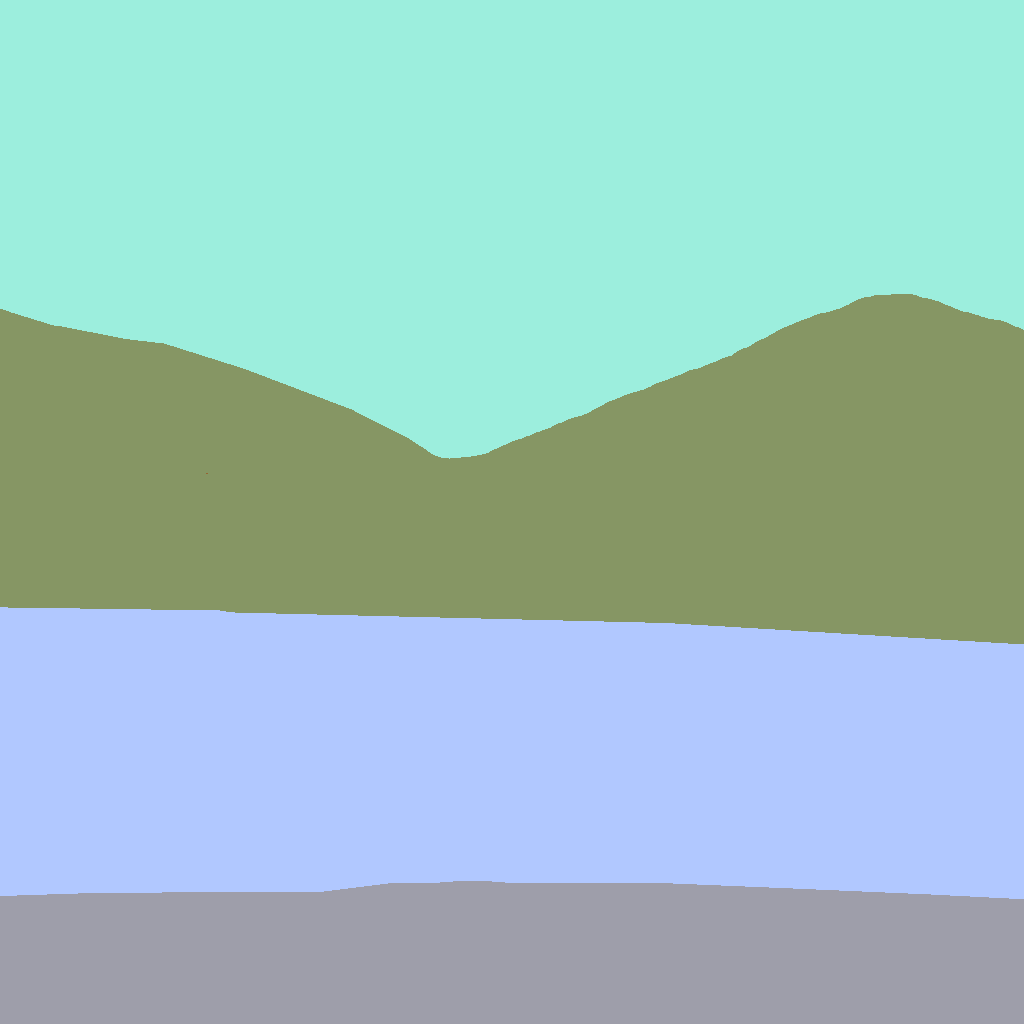}}}
		\put(79.5, 0.2){
			\frame{\includegraphics[width=0.06\textwidth, height=0.06\textwidth]{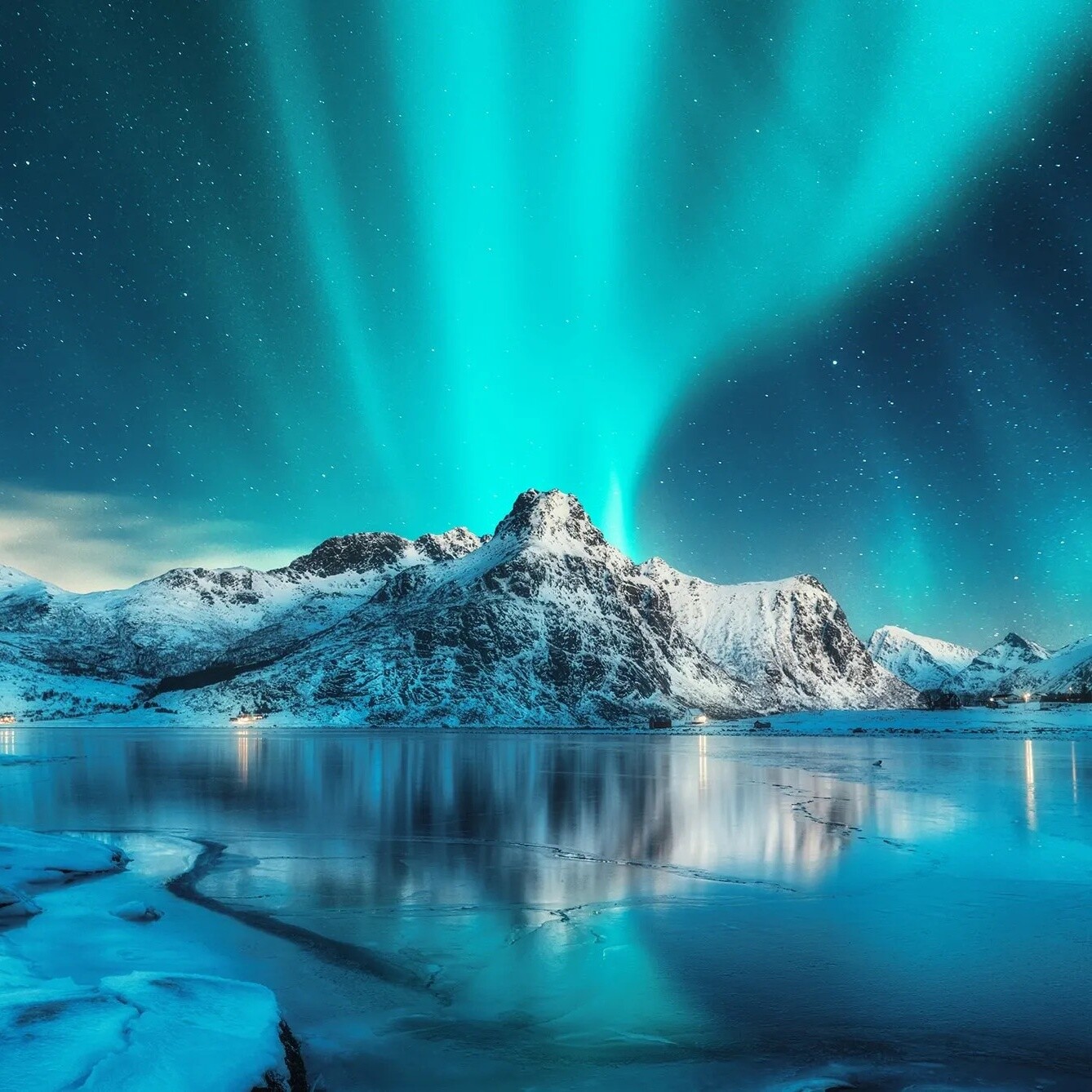}}}
	\end{overpic}
	\begin{overpic}[width=0.32\textwidth, height=0.32\textwidth]{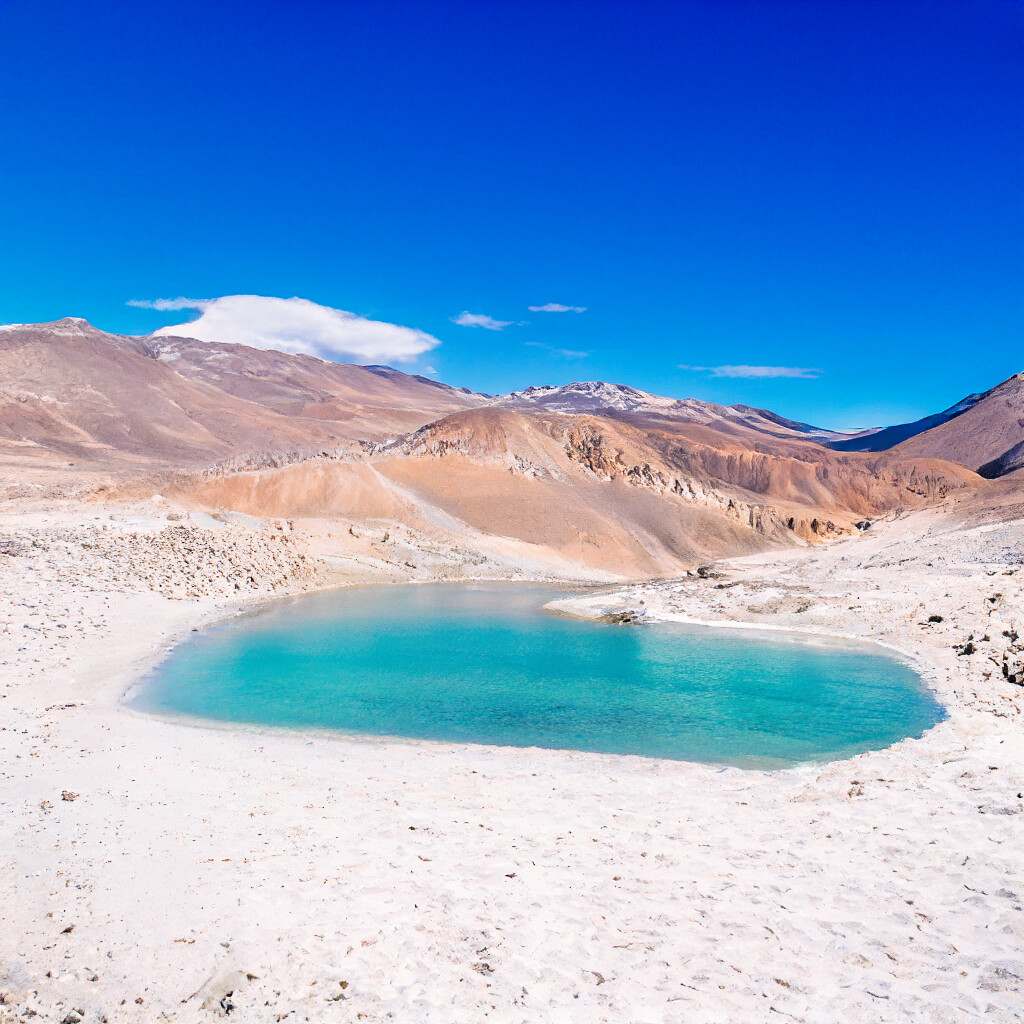}
	\end{overpic}
	\begin{overpic}[width=0.32\textwidth, height=0.32\textwidth]{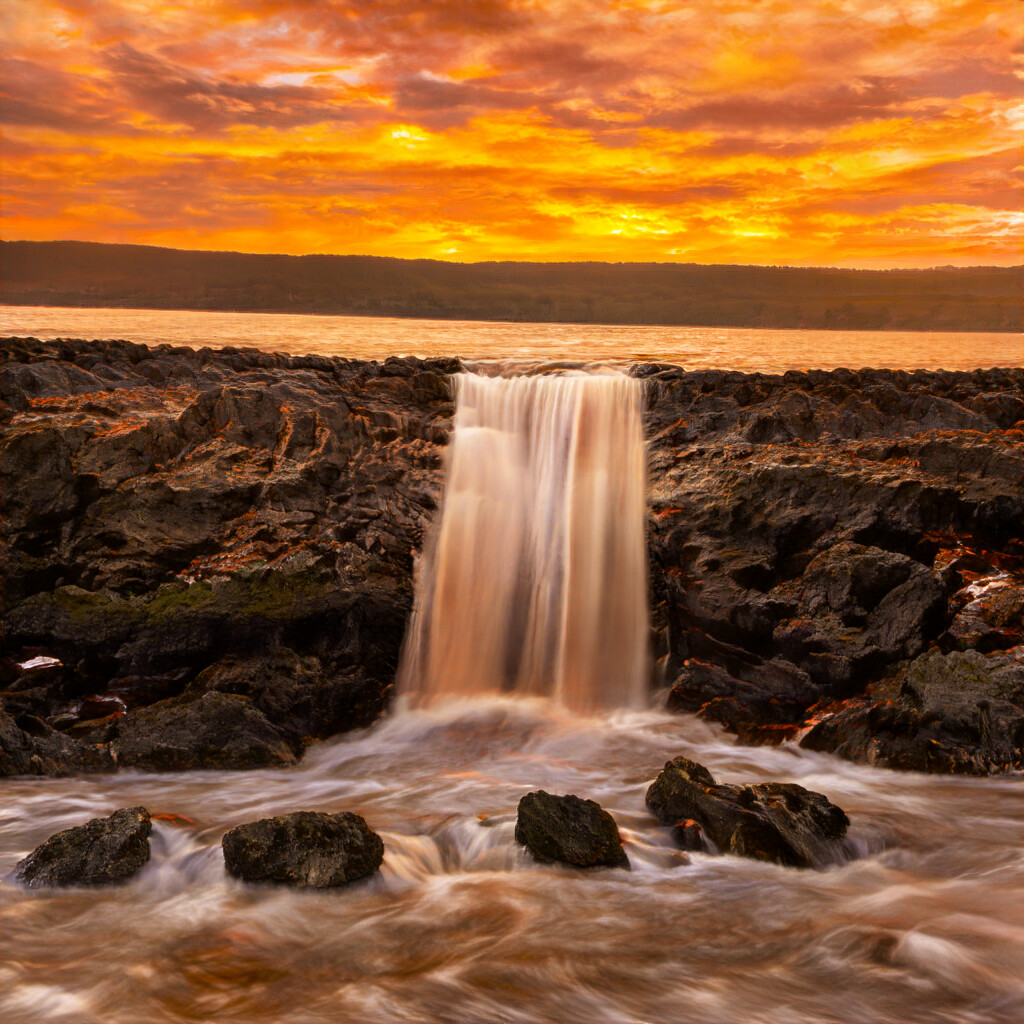}
		\put(79.5, 0.2){
		\frame{\includegraphics[width=0.06\textwidth, height=0.06\textwidth]{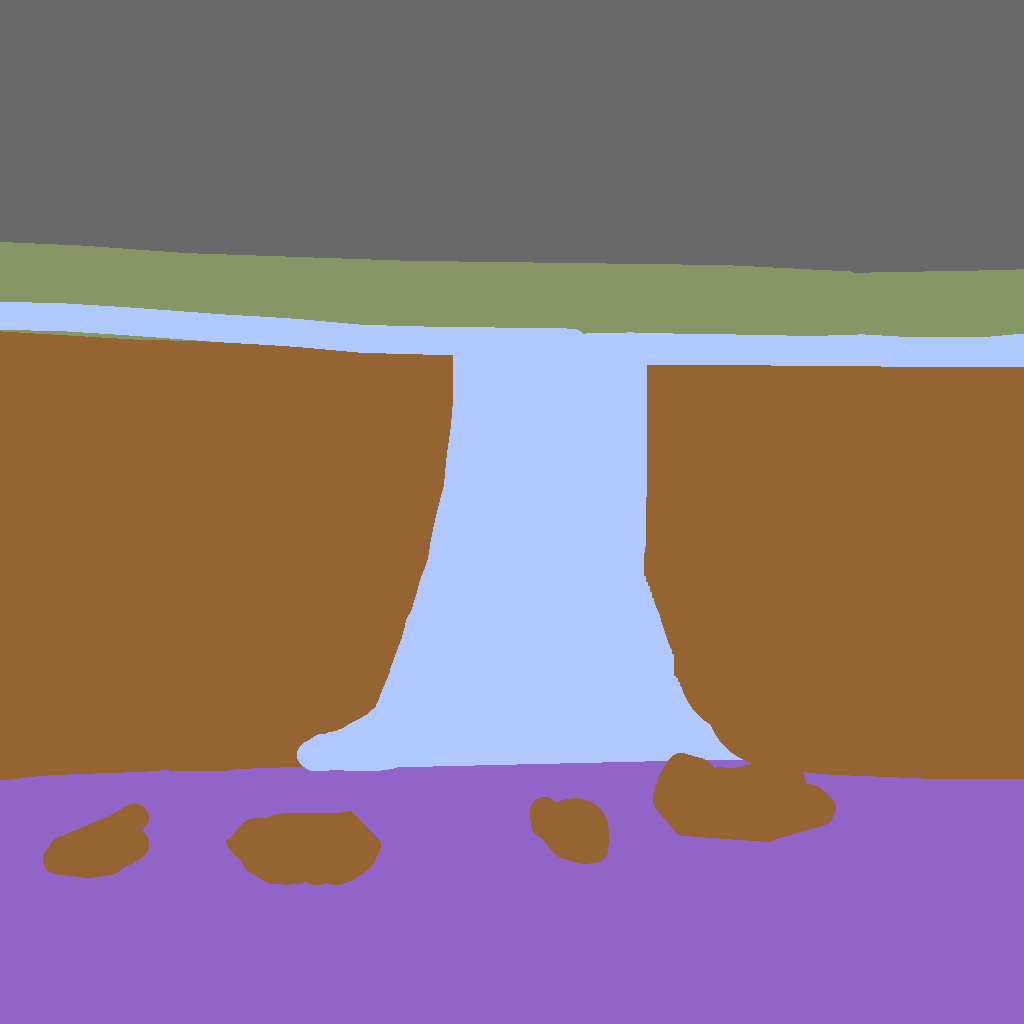}}}
	\end{overpic}
	\begin{overpic}[width=0.32\textwidth, height=0.32\textwidth]{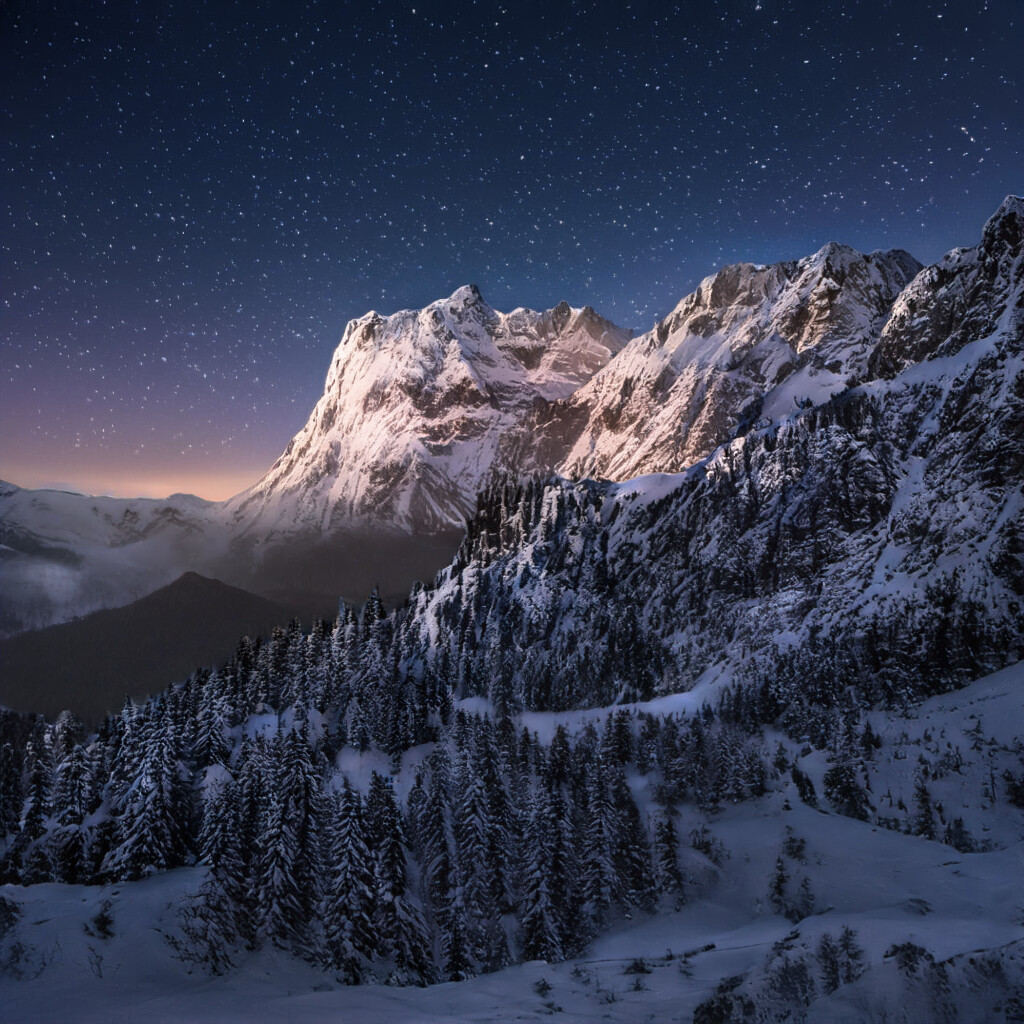}
		\put(79.5, 0.2){
		\frame{\includegraphics[width=0.06\textwidth, height=0.06\textwidth]{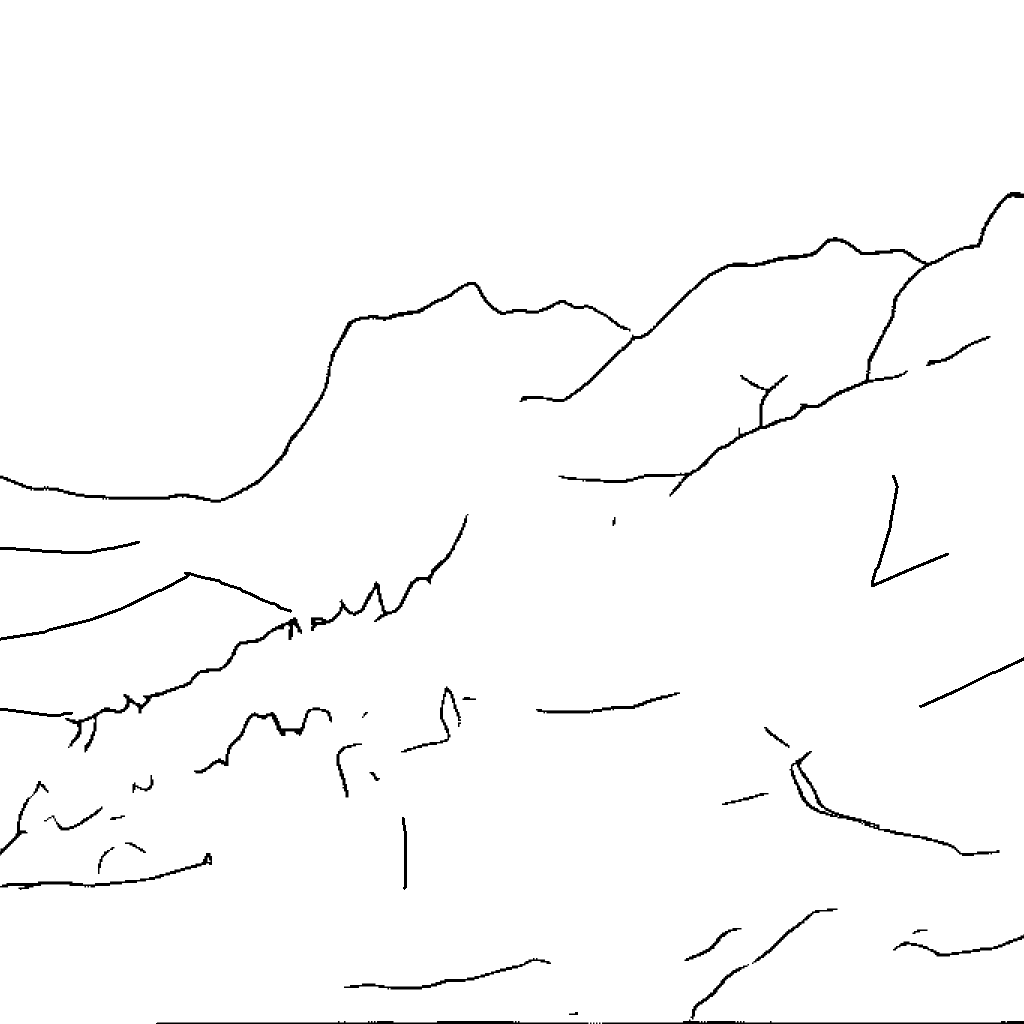}}}
	\end{overpic}
	\begin{minipage}[t]{0.32\textwidth}
		\centering
 		\footnotesize{\textit{A lake in the desert with mountains at a distance.}}
	\end{minipage}
	\begin{minipage}[t]{0.32\textwidth}
		\centering
 		\footnotesize{\textit{A waterfall in sunset.}}
	\end{minipage}
	\begin{minipage}[t]{0.32\textwidth}
		\centering
 		\footnotesize{\textit{A mountain with pine trees in a starry winter night.}}
	\end{minipage}
	\vspace{-2mm}
   	\caption{Examples of multimodal conditional image synthesis results produced by PoE-GAN trained on the 1024$\times$1024 landscape dataset. We show the segmentation/sketch/style inputs on the bottom right of the generated images for the results in the first row. The results in the second row additionally leverage text inputs, which are shown below the corresponding generated images. Please zoom in for details.}
	\label{fig:getty_mm_main}
	\vspace{-5mm}
\end{figure*}
\subsection{Main results}
We compare PoE-GAN with a recent multimodal image synthesis method named TediGAN~\cite{xia2021tedigan} and also with state-of-the-art approaches specifically designed for each modality. For text-to-image, we compare with DF-GAN~\cite{ming2020DFGAN} and DM-GAN + CL~\cite{ye2021improving} on MS-COCO. Since the original models are trained on the 2014 split, we retrain their models on the 2017 split using the official code. For segmentation-to-image synthesis, we compare with SPADE~\cite{park2019semantic}, VQGAN~\cite{esser2021taming}, OASIS~\cite{schonfeld2020you}, and pSp~\cite{richardson2021encoding}. For sketch-to-image synthesis, we compare with SPADE~\cite{park2019semantic} and pSp~\cite{richardson2021encoding}. We additionally compare with StyleGAN2~\cite{karras2020analyzing} in the unconditional setting. We use Clean-FID~\cite{parmar2021cleanfid} for benchmarking due to its reported benefits over previous implementations of FID~\cite{heusel2017gans}.\footnote{The baseline scores differ slightly from those in the original papers.}

Results on MM-CelebA-HQ and MS-COCO are summarized in \cref{tab:main_celeba} and \cref{tab:main_coco}, respectively. As shown in \cref{tab:main_celeba}, PoE-GAN obtains a much lower FID than TediGAN in all settings. In \ifthenelse{\boolean{arxiv}}{\cref{app:tedigan}}{Supplementary Material \ref*{app:tedigan}}, we compare PoE-GAN with TediGAN in more detail and show that PoE-GAN is faster and more general than TediGAN. When conditioned on a single modality, PoE-GAN \textit{surprisingly} outperforms the state-of-the-art method designed specifically for that modality on both datasets, although PoE-GAN is trained for a more general purpose. We believe our proposed architecture and training scheme are responsible for the improved performance. In \cref{fig:coco_seg} and \cref{fig:coco_text}, we qualitatively compare PoE-GAN with previous segmentation-to-image and text-to-image methods on MS-COCO. We find that PoE-GAN produces images of much better quality and can synthesize realistic objects with complex structures, such as human faces and stop signs.
More qualitative comparisons are included in \ifthenelse{\boolean{arxiv}}{\cref{app:example}}{Supplementary Material \ref*{app:example}}. 

\begin{table*}[t]
	\centering
	\begin{small}
		\begin{tabularx}{\textwidth}{p{0.1cm}l*{9}{@{}Y}}
			\toprule
			& & Uncond & \multicolumn{2}{c}{Text} & \multicolumn{2}{c}{Segmentation} & \multicolumn{2}{c}{Sketch} & \multicolumn{2}{c}{All} \\ 
			\cmidrule(lr){4-5} \cmidrule(lr){6-7} \cmidrule(lr){8-9} \cmidrule(lr){10-11}
			& Methods & FID $\downarrow$ & FID $\downarrow$ & LPIPS $\uparrow$ & FID $\downarrow$ & LPIPS $\uparrow$ & FID $\downarrow$ & LPIPS $\uparrow$ & FID $\downarrow$ & LPIPS $\uparrow$ \\
			(a) & Concatenation + modality dropout 	& 29.3             & 26.4 			  & 0.33 			 & 19.4 			& 0.22    		   & \underline{9.2} 			  & 0.11 			 & \textbf{7.7}  & 0.11  \\
			(b) & Ours w/o KL loss 				    & 29.1 			   & 26.6 	 		  & 0.35 			 & 18.1 			& 0.21 			   & \textbf{9.1} 			  & 0.10 			 & \textbf{7.7}  & 0.12 \\
			(c) & Ours w/o modality dropout 		    & 30.8 		       & 31.6 			  & 0.50 			 & 21.0 			& 0.39 			   & 28.0 			  & 0.34 			 & 9.5 			 & 0.30  \\
			(d) & Ours w/o MMPD						& 21.5             & 20.8             & 0.48 			 & 18.3 			& 0.40 			   & 16.4 			  & 0.36 			 & 16.2 		 & 0.34 \\
			(e) & Ours w/o image contrastive loss 	& \underline{15.4} & \underline{14.5}             & 0.55             & 13.5 			& \textbf{0.46}    & 10.2             & \textbf{0.44}    & 9.5 			 & \textbf{0.42} \\
			(f) & Ours w/o text contrastive loss 	    & 15.8 			   & 15.0             & \underline{0.56}             & \underline{13.1} & 0.40 			   & 10.0 & \underline{0.39} & 8.9 			 & \underline{0.38} \\
			(g) & Ours 				                & \textbf{14.9} & \textbf{13.7} & \textbf{0.58} & \textbf{12.9}    & \underline{0.43} & 9.9    & 0.37 & \underline{8.5} 			 & 0.35 \\
			\bottomrule
		\end{tabularx}
	\end{small}
	\vspace{-3mm}
	\caption{Ablation study on MM-CelebA-HQ (256$\times$256).
	The best scores are highlighted in bold and the second best ones are underlined.}
	\label{tab:ablation_celeba}
	\vspace{-3mm}
\end{table*}
\begin{table*}[t]
	\centering
	\begin{small}
		\begin{tabularx}{\textwidth}{p{0.1cm}l*{9}{@{}Y}}
			\toprule
			& & Uncond & \multicolumn{2}{c}{Text} & \multicolumn{2}{c}{Segmentation} & \multicolumn{2}{c}{Sketch} & \multicolumn{2}{c}{All} \\ 
			\cmidrule(lr){4-5} \cmidrule(lr){6-7} \cmidrule(lr){8-9} \cmidrule(lr){10-11}
			& Methods & FID $\downarrow$ & FID $\downarrow$ & LPIPS $\uparrow$ & FID $\downarrow$ & LPIPS $\uparrow$ & FID $\downarrow$ & LPIPS $\uparrow$ & FID $\downarrow$ & LPIPS $\uparrow$ \\
			(a) & Concatenation + modality dropout 	& 59.1             & 30.7 			  & 0.40 			 & 20.4 			& 0.16    		   & 36.1 			  & 0.27 			 & \textbf{16.6} & 0.12  \\
			(b) & Ours w/o KL loss 				    & 59.3 			   & 30.5 	 		  & 0.39 			 & 21.5 			& 0.16 			   & 33.0 			  & 0.27 			 & \underline{16.9}    & 0.12 \\
			(c) & Ours w/o modality dropout 		    & 86.2 		       & 87.8 			  & 0.58 			 & 19.9 			& 0.44 			   & 85.1 			  & 0.55 			 & 18.7 			& \underline{0.47}  \\
			(d) & Ours w/o MMPD						& 43.1             & 40.2             & 0.64 			 & 21.7 			& 0.46 			   & 45.5 			  & 0.57 			 & 21.1 			& 0.42 \\
			(e) & Ours w/o image contrastive loss 	& \textbf{25.7}    & \textbf{21.3}    &    0.64             & 18.0 			& \textbf{0.50}    & 37.9             &   \textbf{0.61}    & 18.5 			& \textbf{0.54} \\
			(f) & Ours w/o text contrastive loss 	    & 27.6 			   & 26.0             & \textbf{0.66}    & \underline{17.4} & 0.46 			   & \underline{33.5} 			  & 0.55 		     & 17.9 			 & 0.43 \\
			(g) & Ours 				                & \underline{26.6} & \underline{22.2} & \underline{0.65} & \textbf{17.1}    & \underline{0.47} & \textbf{30.2}    & \underline{0.58} & 17.1 			 & 0.44 \\
			\bottomrule
		\end{tabularx}
	\end{small}
	\vspace{-3mm}
	\caption{Ablation study on MS-COCO 2017 (64$\times$64).
    The best scores are highlighted in bold and the second best ones are underlined.}
	\label{tab:ablation_coco}
	\vspace{-3mm}
\end{table*}

In \cref{fig:getty_mm_main}, we show example images generated by our PoE-GAN using multiple input modalities on the landscape dataset. Our model is able to synthesize a wide range of landscapes in high resolution with photo-realistic details. More results are included in \ifthenelse{\boolean{arxiv}}{\cref{app:example}}{Supplementary Material \ref*{app:example}}, where we additionally show that PoE-GAN can generate diverse images when given the same conditional inputs.

\subsection{Ablation studies}
\label{sec:ablation}
In \cref{tab:ablation_celeba,tab:ablation_coco}, we analyze the importance of  different components of PoE-GAN. We use LPIPS~\cite{zhang2018unreasonable} as an additional metric to evaluate the diversity of images conditioned on the same input. Specifically, we randomly sample two output images conditioned on the same input and report the average LPIPS distance between the two outputs. A higher LPIPS score would indicate more diverse outputs.

First, we compare our product-of-experts generator, row (g) in \cref{tab:ablation_celeba,tab:ablation_coco}, with a baseline that simply concatenates the embedding of all modalities, while performing modality dropout (missing modality embeddings are set to zero). As seen in row (a), this baseline \textit{only} works well when all modalities are available. Its FID significantly drops when some modalities are missing. Further, the generated images have low diversity as indicated by the low LPIPS score. This is, however, not surprising as previous work has shown that conditional GANs are prone to mode collapse~\cite{NIPS2017_6650,huang2018multimodal,yang2019diversity}.

Row (b) of \cref{tab:ablation_celeba,tab:ablation_coco} shows that the KL loss is important for training our model. Without it, our model suffers from low sample diversity and lack of robustness towards missing modalities, similar to the concatenation baseline described above. The variances of individual experts become near zero without the KL loss. The latent code $z^k$ then becomes a deterministic weighted average of the mean of each expert, which is equivalent to concatenating all modality embeddings and projecting it with a linear layer. This explains why our model without the KL loss behaves similarly to the concatenation baseline. Row (c) shows that our modality dropout scheme is important for handling missing modalities. Without it, the model tends to overly rely on the most informative modality, such as segmentation in MS-COCO.

To evaluate the proposed multiscale multimodal discriminator architecture, we replace MMPD with a discriminator that receives concatenated images and all conditional inputs. Row (d) shows that MMPD is much more effective than such a concatenation-based discriminator in all settings. 

Finally in rows (e) and (f), we show that contrastive losses are useful but not essential. The image contrastive loss slightly improves FID in most settings, while the text contrastive loss improves FID for text-to-image synthesis.


\section{Conclusion}
We introduce a multimodal conditional image synthesis model based on product-of-experts and show its effectiveness for converting an arbitrary subset of input modalities to an image satisfying all conditions. While empirically superior than the prior multimodal synthesis work, it also outperforms state-of-the-art unimodal conditional image synthesis approaches when conditioned on a single modality. 




\medskip
\noindent{\bf Limitations.} 
If different modalities provide \textit{contradictory} information, PoE-GAN produces unsatisfactory outputs as shown in \ifthenelse{\boolean{arxiv}}{\cref{app:limitation}}{Supplementary Material \ref*{app:limitation}}. PoE-GAN trained in the unimodal setting performs better than the model trained in the multimodal setting, when tested in the unimodal setting. This indicates room for improvement
in the fusing of multiple modalities, and we leave this for future work.

\medskip
\noindent{\bf Potential negative societal impacts.} 
Image synthesis networks can help people express themselves and artists create digital content, but they can undeniably also be misused for visual misinformation~\cite{westerlund2019emergence}. Enabling users to synthesize images using multiple modalities makes it even easier to create a desired fake image.  We encourage research that helps detect or prevent these potential negative misuses.  We will provide implementations and training data to help forensics researchers detect fake images~\cite{wang2020cnn,chai2020makes} and develop effective schemes for watermarking networks and training data~\cite{yu2021artificial}. 

\clearpage

\noindent{\bf Acknowledgements.} We thank Jan Kautz, David Luebke, Tero Karras, Timo Aila, and Zinan Lin for their feedback on the manuscript. We thank Daniel Gifford and Andrea Gagliano on their help on data collection.

{\small
	\bibliographystyle{ieee_fullname}
	\bibliography{main}
}
\clearpage
	\begin{appendix}

\section{Datasets}
\label{app:datasets}
We evaluate the proposed PoE-GAN approach for multimodal conditional image synthesis on three datasets, including MM-CelebA-HQ~\cite{xia2021tedigan}, MS-COCO~\cite{lin2014microsoft}, and a proprietary dataset of landscape images using four modalities including text, segmentation, sketch, and style reference. The style is extracted from ground truth images and the other modalities are obtained from either human annotation or pseudo-labeling methods. We describe details about each dataset in the following.

\medskip
\noindent\textbf{MM-CelebA-HQ} contains 30,000 images of celebrity faces with a resolution of 1024$\times$1024, which are created from the original CelebA dataset~\cite{liu2015faceattributes} using a procedure that improves image quality and resolution by Karras~\etal~\cite{karras2018progressive}. Each image is annotated with text, segmentation, and sketch. While the segmentation maps are annotated by humans~\cite{lee2020maskgan}, the text descriptions are automatically generated from the ground truth attribute labels by Xia~\etal~\cite{xia2021tedigan}. On the other hand, sketch maps are generated by Chen~\etal~\cite{chen2020deepfacedrawing} using the Photoshop edge extractor and sketch simplification~\cite{simo2016learning}. We use the pretrained CLIP~\cite{radford2021learning} text encoder to encode the text before feeding them to the generator. We use the standard train-test split~\cite{richardson2021encoding,xia2021tedigan} which yields 24,000 images in the training set and 6,000 images in the test set. We use images in the original resolution~(1024$\times$1024) in our main experiment and use images downsampled to 256$\times$256 in ablation studies.

\medskip
\noindent\textbf{MS-COCO} contains 123,287 images of complex indoor/outdoor scenes, containing various common objects. We use the segmentation maps provided in COCO-Stuff~\cite{caesar2018cvpr} as the ground truth segmentation maps for the images. In MS-COCO, each image has up to 5 text descriptions. We use the pretrained CLIP text encoder to extract a feature vector per description. We additionally annotate each image with a sketch map produced by running HED~\cite{xie2015holistically} edge detector followed by a sketch simplification process~\cite{simo2016learning}. We use the 2017 split, which leads to 118,287 training images and 5,000 test images. We use images resized to 256$\times$256 in our main experiment and to 64$\times$64 in ablation studies.

\medskip
\noindent\textbf{Landscape} is a proprietary dataset containing around 10 million landscape images of resolution higher than 1024$\times$1024. It does not come with any manual annotation and we use DeepLab-v2~\cite{chen2017deeplab} to produce pseudo segmentation annotation and HED~\cite{xie2015holistically} with sketch simplification~\cite{simo2016learning} to produce pseudo sketch annotation. For the text annotation, we use the CLIP image embedding as the pseudo text embedding to train our model. We randomly choose 50,000 images as the test set and use the rest of the images as the training set. Images are randomly cropped to 1024$\times$1024 during training.

\section{Implementation details}
\label{app:hyperparameters}

\begin{table*}[t]
	\centering
	\renewcommand{\arraystretch}{1.46}
	\begin{small}
		\begin{tabularx}{\textwidth}{l*{5}{Y}}
			\toprule
			\multirow{2}{*}{Hyper-parameters} & MM-CelebAHQ & MM-CelebAHQ & MS-COCO & MS-COCO & Landscape \\
			& ($1024\times 1024$) & ($256\times 256$) & ($256\times 256$) & ($64\times 64$) & ($1024\times 1024$) \\ 
			\midrule
			Learning rate                 & 0.003  & 0.003  & 0.004 & 0.004 & 0.004 \\
			Batch size                    &  64    & 64     & 256   & 128   & 768   \\
			Text KL weight                & 0.01   & 0.1    & 0.01  & 0.1   & 0.05  \\
			Segmentation KL weight        & 0.01   & 0.1    & 0.01  & 0.1   & 0.1   \\
			Sketch KL weight              & 0.1    & 1      & 0.1   & 1     & 1     \\
			Style KL weight               & 0.0005 & 0.005  & 0.001 & 0.01  & 0.01  \\
			Image contrastive loss weight & 0.3    & 0.3    & 3     & 3     & 3     \\
			Text contrastive loss weight  & 0.3    & 0.3    & 0.3   & 0.3   & 0.3   \\
			Base \# channels for Dec/Dis    & 16     & 64     & 128   & 256   & 32    \\
			Maximum \# channels for Dec/Dis & 512    & 512    & 1024  & 512   & 1024  \\
			Base \# channels for Latent     & 2      & 16     & 16    & 64    & 2     \\
			Maximum \# channels for Latent  & 32     & 64     & 64    & 64    & 32    \\
			\bottomrule
		\end{tabularx}
	\end{small}
	\caption{Hyper-parameters on different datasets.}
	\label{tab:hyperparams}
\end{table*}

\begin{table*}[t]
	\centering
	\renewcommand{\arraystretch}{1.46}
	\begin{small}
		\begin{tabularx}{\textwidth}{l*{5}{Y}}
			\toprule
			& MM-CelebAHQ & MM-CelebAHQ & MS-COCO & MS-COCO & Landscape \\
			& ($1024\times 1024$) & ($256\times 256$) & ($256\times 256$) & ($64\times 64$) & ($1024\times 1024$) \\ 
			\midrule
			Number of GPUs\hspace{18.5mm} & 16      & 8      & 32    & 8       & 256    \\
			Training time                 & 71h     & 35h    & 85h   & 76h     & 101h   \\
			Inference time~(per image)      & 0.07s   & 0.04s  & 0.06s & 0.02s   & 0.12s  \\
			\bottomrule
		\end{tabularx}
	\end{small}
	\caption{Hardware and training/inference speed on different datasets. We train all models using NVIDIA Tesla V100 GPUs, except for the Landscape model which is trained using NVIDIA Ampere A100 GPUs with 80 GB of memory. The inference time is evaluated on a workstation with a single NVIDIA TITAN RTX GPU.}
	\label{tab:hardware}
\end{table*}

Our implementation is based on Pytorch~\cite{paszke2019pytorch}. We use the Adam optimizer~\cite{kingma2014adam} with $\beta_1=0$ and $\beta_2=0.99$ and the same constant learning rate for both the generator and the discriminator. The final model weight is given by an exponential moving average of the generator's weights during the course of training. To stabilize GAN training, we employ leaky ReLU~\cite{maas2013rectifier} with slope $0.2$, $R_1$ gradient penalty~\cite{mescheder2018training} with lazy regularization~\cite{karras2020analyzing} applied every $16$ iterations, equalized learning rate~\cite{karras2020analyzing}, anti-aliased resampling~\cite{zhang2019making}, and clip the gradient norms at 10. 
We also limit the range of the Gaussian log variance in product-of-experts layers to $(-\theta, \theta)$ by applying the activation function $\theta\tanh(\frac{.}{\theta})$. We use $\theta=1$ for the prior expert and $\theta=10$ for experts predicted from input modalities.
We use mixed-precision training in all of our experiments and clamp the output of every convolutional/fully-connected layer to $\pm 256$~\cite{karras2020training}. We use a dropout rate of $0.5$ when performing modality dropout. The contrastive loss temperature is initialized at $0.3$ and learnable during training. We use the \texttt{relu5\_1} feature in VGG-19 to compute the image contrastive loss.

Inspired by NVAE~\cite{vahdat2020NVAE}, we rebalance the KL terms of different resolutions. The rebalancing weight $\omega^k$ in \cref*{eq:kl} of the main paper is proportional to the unbalanced KL term and inversely proportional to the resolution:
\begin{align}
	\omega^k \propto &\frac{1}{w^{k}h^{k}}\mathbb{E}_{p(y_{i})}[\mathbb{E}_{p(z^{<k}|y_{i})}[\nonumber\\
	&D_{\text{KL}}(p(z^{k}|z^{<k}, y_{i})||p^{\prime}(z^{k}|z^{<k}))]]\,,
\end{align}
with the constraint that $\frac{1}{N}\sum_{k=1}^{N}\omega^{k}=1$.
The rebalancing weights encourage having the amount of information encoded in each latent variable. We use a running average of the rebalancing weights with a decay factor of $0.99$.

\begin{figure*}[t]
	\centering
	\subcaptionbox{Text encoder\label{fig:text_encoder}}
	{\includegraphics[height=0.23\textheight]{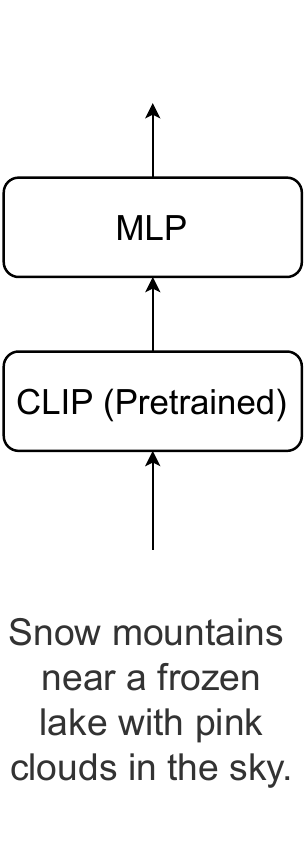}}\hspace{8mm}
	\subcaptionbox{Segmentation encoder\label{fig:seg_encoder}}
	{\includegraphics[height=0.23\textheight]{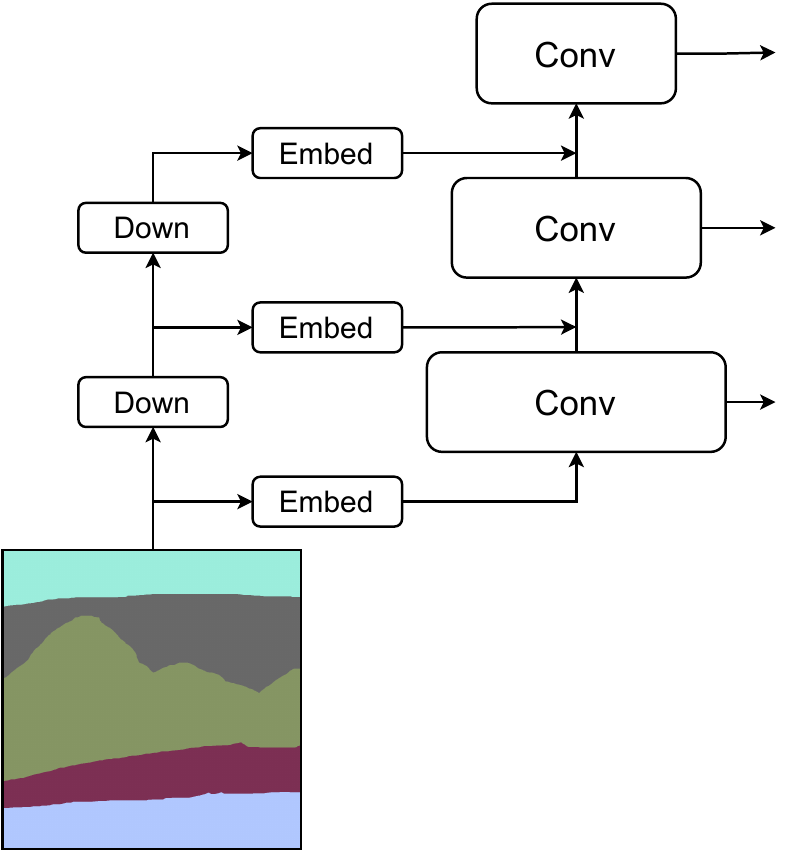}}\hspace{3mm}
	\subcaptionbox{Sketch encoder\label{fig:sketch_encoder}}
	{\includegraphics[height=0.23\textheight]{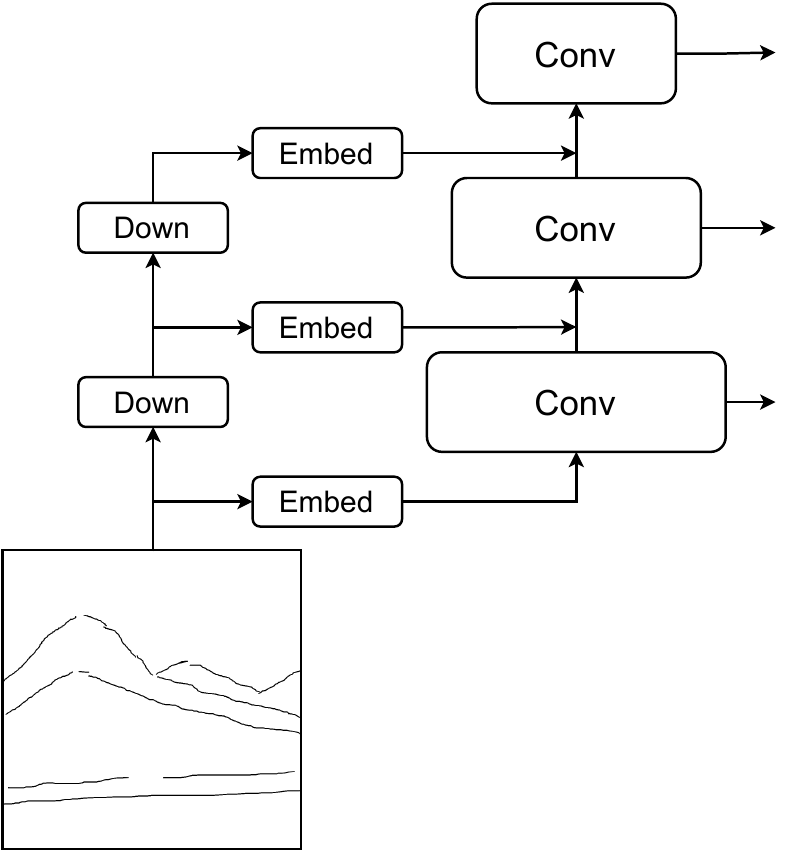}}\hspace{6mm}
	\subcaptionbox{Style encoder\label{fig:style_encoder}}
	{\includegraphics[height=0.23\textheight]{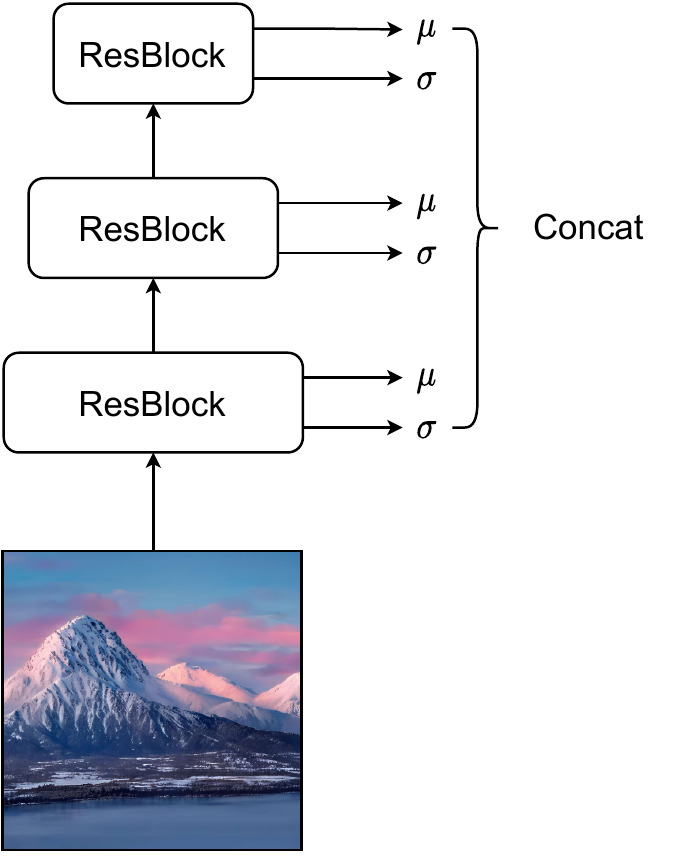}}
	\caption{Architecture of encoders. We use a pretrained CLIP~\cite{radford2021learning} and an MLP to encoder text, a convolutional network with input skip connections to encode segmentation/sketch maps, and a residual network to encoder style images.}
	\label{fig:encoders}
\end{figure*}

\subsection{Decoder}
We introduced our decoder design in \cref*{sec:gen} of the main paper. Here, we provide additional details. In Global PoE-Net (\cref*{fig:global_penet} of the main paper), each MLP consists of $4$ fully-connected layers with a hidden dimension $4$ times smaller than the input dimension. Both $z^0$ and $w$ are 512-dimensional. The output MLP has two layers with a hidden dimension of $512$. 

In Local PoE-Net (\cref*{fig:decoder} of the main paper), each CNN similarly contains $4$ convolutional layers with the number of filters $4$ times smaller than the input channel size. The first and last convolutions have $1\times 1$ kernels, and the convolutions in the middle have a kernel size of $3$. The dimension of $w$ is 512. The dimensions of $z^k$'s are described in~\cref{subsec:hyper}.

\subsection{Encoders}\label{subsec:encoders}
\cref{fig:encoders} shows the architecture of encoders used in our generator to encode each modality. The text encoder~(\cref{fig:text_encoder}) is a 4-layer MLP with dimension $512$ that processes the CLIP embedding of a caption. The segmentation and sketch encoders are CNNs with skip connections from the input, which are illustrated in \cref{fig:seg_encoder} and \cref{fig:sketch_encoder}. The segmentation or sketch map is downsampled multiple times. The embeddings of the downsampled map are added to the intermediate outputs of the corresponding convolutional layers. The intermediate outputs of convolutional layers are provided to the decoder via skip connections. The style encoder for encoding the reference style image is a residual network with instance normalization~\cite{ulyanov2017improved}. As shown in \cref{fig:style_encoder}, we obtain the style code by concatenating the mean and standard deviation of the output of every residual block.

\begin{table*}[!t]
\centering
\begin{footnotesize}
	\setlength{\tabcolsep}{11pt}
	\renewcommand{\arraystretch}{1.4}
	\begin{tabularx}{\textwidth}{*{7}{c}}
		\hline
		StyleGAN2~\cite{karras2020analyzing} & SPADE-Seg~\cite{park2019semantic} & pSp-Seg~\cite{richardson2021encoding} & SPADE-Sketch~\cite{park2019semantic} & pSp-Sketch~\cite{richardson2021encoding} & TediGAN~\cite{xia2021tedigan} & PoE-GAN \\ 
		30M & 96M & 298M & 95M & 298M & 565M & 33M  \\ 
		\hline
	\end{tabularx}
\end{footnotesize}
\caption{Number of parameter of compared models on MM-CelebA-HQ (1024$\times$1024).}
\label{tab:params_celeba}
\end{table*}
\begin{table*}[!t]
\centering
\begin{footnotesize}
	\setlength{\tabcolsep}{6.5pt}
	\renewcommand{\arraystretch}{1.4}
	\begin{tabularx}{\textwidth}{*{8}{c}}
		\toprule
		StyleGAN2~\cite{karras2020analyzing} & DF-GAN~\cite{ming2020DFGAN} & DM-GAN + CL~\cite{ye2021improving} & SPADE-Seg~\cite{richardson2021encoding} & VQGAN~\cite{esser2021taming} & OASIS~\cite{schonfeld2020you} & SPADE-Sketch~\cite{richardson2021encoding} & PoE-GAN  \\ 
		25M & 12M & 32M & 280M & 798M & 94M & 95M & 142M  \\ 
		\bottomrule
	\end{tabularx}
\end{footnotesize}
\caption{Number of parameters of compared models on MS-COCO 2017 (256$\times$256).}
\label{tab:params_coco}
\end{table*}
\subsection{Discriminator}
As shown in \cref*{fig:mmpd} of the main paper, our discriminator encodes the image and the other modalities into multiscale feature maps and computes the MPD loss at each scale. We use a residual network to encode the image. For encoding other modalities, we use the same architecture as described in the previous section (\cref{subsec:encoders} and \cref{fig:encoders}). However, the parameters are not shared with those encoders used in the generator. For encoders (the style encoder or the text encoder) that only outputs a single feature vector, we spatially replicate the feature vector to different resolutions.

\subsection{Hyper-parameters}\label{subsec:hyper}
\cref{tab:hyperparams} shows the hyper-parameters used on all the benchmark datasets, including learning rate, batch size, loss weights, and channel size. The decoder and discriminator have the same channel size, which is always $4$ times larger than the channel size of the modality encoders. We hence omit the channel size of modality encoders from the table. The layer operating at the highest resolution always has the smallest number of channels~(denoted as ``base \# channels for Dec\/Dis''). The number of channels doubles while the resolution halves until it reaches a maximum number~(denoted as ``maximum \# channels for Dec\/Dis''). The "Base \# of channels for Latent" refers to the channel size of the feature map $z_i^N$ to the decoder extracted from the spatial modalities (segmentation and sketch) in the highest resolution (the largest value of $k$ is $N$). We also note that the channel number of $\mu_i^k$ or $\sigma_i^k$ equals that of $z_i^k$. As explained in~\cref{subsec:encoders}, we double the channel size as we halve the spatial resolution. The channel number for $z_i^k$ doubles until it reaches the "Maximum \# of channels for Latent," the channel number for $z_i^0$.

\subsection{Hardware and speed}
We report the computation infrastructure used in our experiments \cref{tab:hardware}. We also report the training and inference speed of our model for different datasets in the table.
	
\section{Additional results}
We provide additional experimental results in this section, including more comparison with baselines and more analysis on the proposed approach. Please check the accompanying video for additional results on the landscape dataset.
\begin{table*}[t]
	\centering
	\begin{footnotesize}
		\begin{tabularx}{\linewidth}{l*{7}{Y}}
			\toprule
			Method & AttnGAN~\cite{xu2018attngan} & ControlGAN~\cite{li2019controllable} & DF-GAN~\cite{ming2020DFGAN} & DM-GAN~\cite{zhu2019dm} & TediGAN~\cite{xia2021tedigan} & M6-UFC~\cite{zhang2021m6ufc} & Ours \\\midrule
			FID $\downarrow$    & 125.98                       & 116.32     & 137.60 & 131.05 & 106.37  & 66.72  & \textbf{13.71} \\
			LPIPS $\uparrow$  & 0.512   					  & 0.522      & 0.581  & 0.544  & 0.456   & 0.448  & \textbf{0.583} \\
			\bottomrule
		\end{tabularx}
	\end{footnotesize}
	\caption{Comparison of text-to-image synthesis on MM-CelebA-HQ~(256$\times$256). $\uparrow$: the higher the better, $\downarrow$: the lower the better.}
	\label{tab:celeba_256}
\end{table*}

\subsection{Model size comparison}
In \cref{tab:params_celeba} and \cref{tab:params_coco}, we compare the number of parameters used in PoE-GAN and in our baselines. We show that PoE-GAN does not use significantly more parameters --- actually it uses fewer parameters than some of the single-modal baselines, although PoE-GAN is trained for a much more challenging task. This shows that our improvement does not come from using a larger model.

\subsection{Comparison on MM-CelebA-HQ~(256$\times$256)}
In \cref{tab:celeba_256}, We compare PoE-GAN trained on the 256$\times$256 MM-CelebA-HQ dataset with the text-to-image baselines reported in TediGAN~\cite{xia2021tedigan} and M6-UFC~\cite{zhang2021m6ufc}.  Our model achieves a significantly lower FID than previous methods.

\subsection{Additional comparison with TediGAN}
\label{app:tedigan}
We provide a more detailed comparison between PoE-GAN and TediGAN, the only existing multimodal image synthesis method that can produce high-resolution images. TediGAN mainly contains four steps: 1) encodes different modalities into the latent space of StyleGAN, 2) mixes the latent code according to a hand-designed rule, 3) generates the image from the latent code using StyleGAN, and 4) iteratively refines the image. It has several disadvantages compared with our PoE-GAN:
\begin{enumerate}
	\item TediGAN relies on a pretrained unconditional generator, and its image quality is upper bounded by that unconditional model. This is not ideal because unconditional models usually produce images of lower quality than conditional ones. On the other hand, PoE-GAN learns conditional and unconditional generation simultaneously, and its FID improves when more conditions are provided, as shown in \cref*{tab:main_celeba} of the main paper.
	\item TediGAN uses a handcrafted rule to combine the latent code from different modalities. Specifically, the StyleGAN latent space contains 14 layers of latent vectors. TediGAN uses the top-layer latent vectors from one modality and the bottom-layer latent vectors from another modality when combining two modalities. This combination rule cannot be generalized to other generator architectures and other modalities. In contrast, we use product-of-experts with learned parameters to combine different modalities which is more general.
	\item Sampling from TediGAN is very slow due to its instance-level optimization. It takes $51.2$ seconds to generate a 1024$\times$1024 image with TediGAN, while PoE-GAN only needs 0.07 seconds.
\end{enumerate}

\begin{table}[t!]
\centering
\setlength{\tabcolsep}{4pt}
\renewcommand{\arraystretch}{1.2}
\begin{footnotesize}
	\begin{tabularx}{\linewidth}{l*{5}{Y}}
		\toprule
		& Uncond & Text & Seg & Sketch & All \\
		Ours~(Uncond)    & \textbf{13.0} & ---  & ---  & ---  & ---  \\
		Ours~(Text)      & 13.8 & \textbf{13.4} & ---  & ---  & ---  \\
		Ours~(Seg)       & 14.2 & ---  & \textbf{10.9}  & ---  & ---  \\
		Ours~(Sketch)    & 14.2 & ---  & ---  & \textbf{9.5}  & ---  \\ \midrule
		Ours~(All)       & 14.9 & 13.7 & 12.9 & 9.9  & \textbf{8.5}  \\
		\bottomrule
	\end{tabularx}
\end{footnotesize}
\caption{Comparison on MM-CelebA-HQ (256$\times$256) using FID. We compare our PoE-GAN trained using all modalities with PoE-GAN trained using a single modality. Note that PoE-GAN trained using a single modality can also be used for unconditional synthesis and we also report the achieved FID in the table.}
\label{tab:unimodal_celeba}
\end{table}

\begin{table}[t!]
\centering
\setlength{\tabcolsep}{4pt}
\renewcommand{\arraystretch}{1.2}
\begin{footnotesize}
	\begin{tabularx}{\linewidth}{l*{5}{Y}}
		\toprule
		& Uncond & Text & Seg & Sketch & All \\
		Ours~(Uncond)    & \textbf{23.3} & ---  & ---  & ---  & ---  \\
		Ours~(Text)      & 24.0 & \textbf{21.1} & ---  & ---  & ---  \\
		Ours~(Seg)       & 25.5 & ---  & \textbf{16.3} & ---  & ---  \\
		Ours~(Sketch)    & 24.7 & ---  & ---  & \textbf{24.7} & ---  \\ \midrule
		Ours~(All)       & 26.6 & 22.2 & 17.1 & 30.2 & \textbf{17.1} \\
		\bottomrule
	\end{tabularx}
\end{footnotesize}
\caption{Comparison on MS-COCO 2017 (64$\times$64) using FID. We compare our PoE-GAN trained using all modalities with PoE-GAN trained using a single modality. Note that PoE-GAN trained using a single modality can also be used for unconditional synthesis and we also report the achieved FID in the table.}
\label{tab:unimodal_coco}
\end{table}

\subsection{Training PoE-GAN with a single modality}
\label{app:single_modality}
Although PoE-GAN is designed for the multimodal conditional image synthesis task, it can be trained in a single modality setting. That is, we can train a pure segmentation-to-image model, a pure sketch-to-image model, or a pure text-to-image model using the same PoE-GAN model. Here, we compare a PoE-GAN model trained using multiple modalities with the same PoE-GAN model but trained using one single modality. We compare their performance when applied to convert the user input in a single modality to the output image. This experiment helps understand the penalty the model pays for the multimodal synthesis capability. 

As shown in \cref{tab:unimodal_celeba} and \cref{tab:unimodal_coco}, the model trained for a specific modality always slightly outperforms the joint model when conditioned on that modality. This indicates that the increased task complexity of an additional modality outweighs the benefits of additional annotations from that modality. This result also shows that the improvement of PoE-GAN over state-of-the-art unimodal image synthesis methods comes from our architecture and training scheme rather than additional annotations.

\subsection{User study}
We conduct a user study to compare PoE-GAN with state-of-the-art text-to-image and segmentation-to-image synthesis methods on MS-COCO. We show users two images generated by different algorithms from the same conditional input and ask them which one is more realistic. As shown in \cref{tab:user_study_text} and \cref{tab:user_study_seg}, the majority of users prefer PoE-GAN over the baseline methods.

\subsection{Additional qualitative examples}
\label{app:example}
In \cref{fig:celeba_mm_2,fig:coco_mm_2,fig:coco_mm_2_style,fig:getty_mm,fig:getty_mm_style}, we show that PoE-GAN can generate diverse images when conditioned on two different input modalities. We show additional qualitative comparison of text-to-image synthesis and segmentation-to-image synthesis on MS-COCO in \cref{fig:coco_text_more,fig:coco_seg_more} respectively.
\cref{fig:celeba_uncond,fig:coco_uncond,fig:getty_uncond} show uncurated samples generated unconditionally on MM-CelebA-HQ, MS-COCO, and Landscape. 

\section{Limitation}
\label{app:limitation}
We find that the PoE-GAN model does not work well when conditioned on contradictory multimodality inputs. For example, when the segmentation and text are contradictory to each other, the text input is usually ignored, as shown in \cref{fig:limitation}. In the product-of-experts formulation, an expert with a larger variance will have a smaller influence on the product distribution, and we indeed find the variance of the text expert is usually larger than that of the segmentation expert, which explains the behavior of our model.

In addition, as discussed in~\cref{app:single_modality}, the PoE-GAN model still pays the penalty in terms of a higher FID score as achieving the multimodal conditional image synthesis capability. This indicates room for improvement in the fusing of multiple modalities, and we leave this for future work

\begin{table}[!t]
	\centering
	\renewcommand{\arraystretch}{1.2}
	\begin{tabularx}{\linewidth}{l*{2}{Y}}
		\toprule
		         & DF-GAN~\cite{ming2020DFGAN}  & DM-GAN + CL~\cite{ye2021improving} \\ 
		Ours vs. & 82.1\% & 72.9\%   \\ 
		\bottomrule
	\end{tabularx}
	\caption{User study on text-to-image synthesis. Each column shows the percentage of users that prefer the image generated by our model over that generated by the baseline method.}
	\label{tab:user_study_text}
\end{table}

\begin{table}[!t]
	\centering
	\renewcommand{\arraystretch}{1.2}
	\begin{tabularx}{\linewidth}{l*{3}{Y}}
		\toprule
		         & SPADE~\cite{park2019semantic} & VQGAN~\cite{esser2021taming}  & OASIS~\cite{schonfeld2020you}  \\ 
		Ours vs. & 69\%  & 66.7\% & 64.9\% \\ 
		\bottomrule
	\end{tabularx}
	\caption{User study on segmentation-to-image synthesis. Each column shows the percentage of users that prefer the image generated by our model over that generated by the baseline method.}
	\label{tab:user_study_seg}
\end{table}
\begin{figure}
	\centering
	\begin{overpic}[width=0.23\textwidth, height=0.23\textwidth]{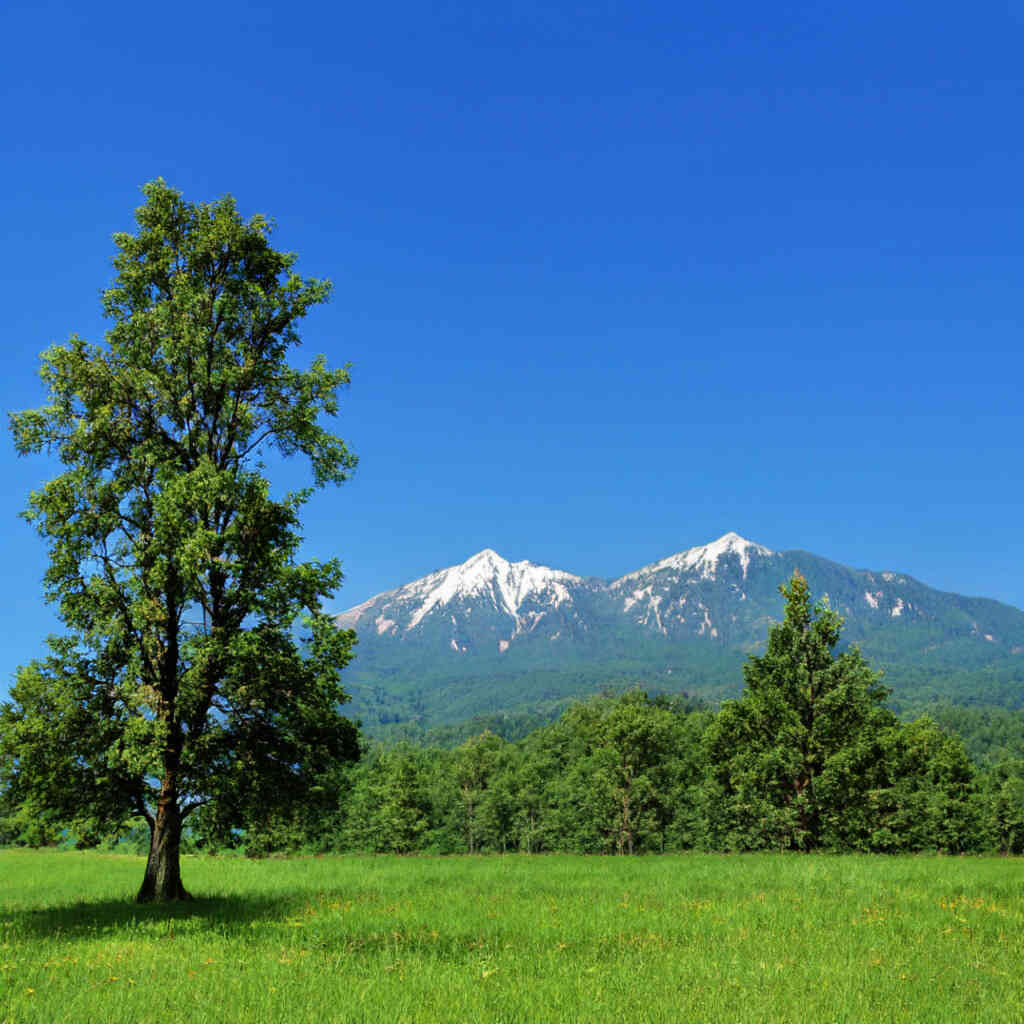}
		\put(62.8, 0.2){
		\frame{\includegraphics[width=0.08\textwidth, height=0.08\textwidth]{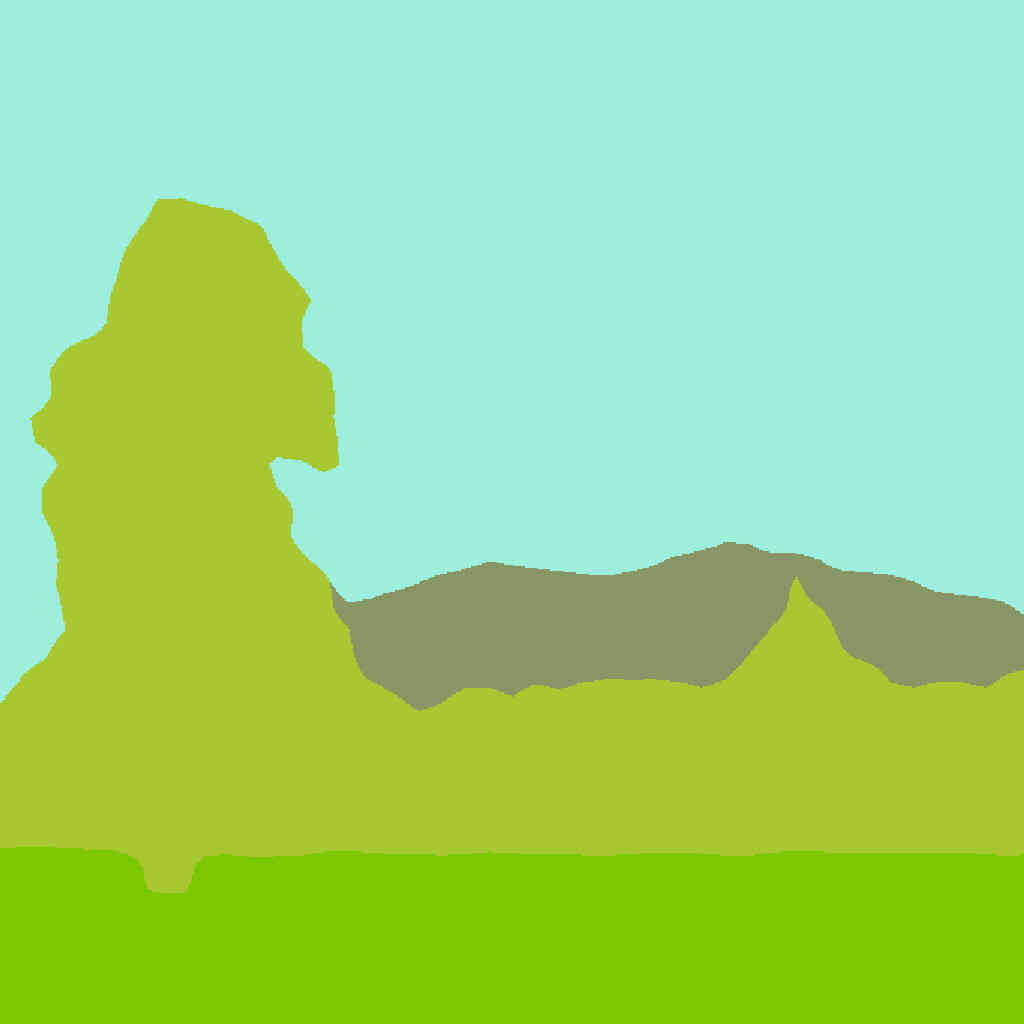}}}
	\end{overpic}
	\begin{overpic}[width=0.23\textwidth, height=0.23\textwidth]{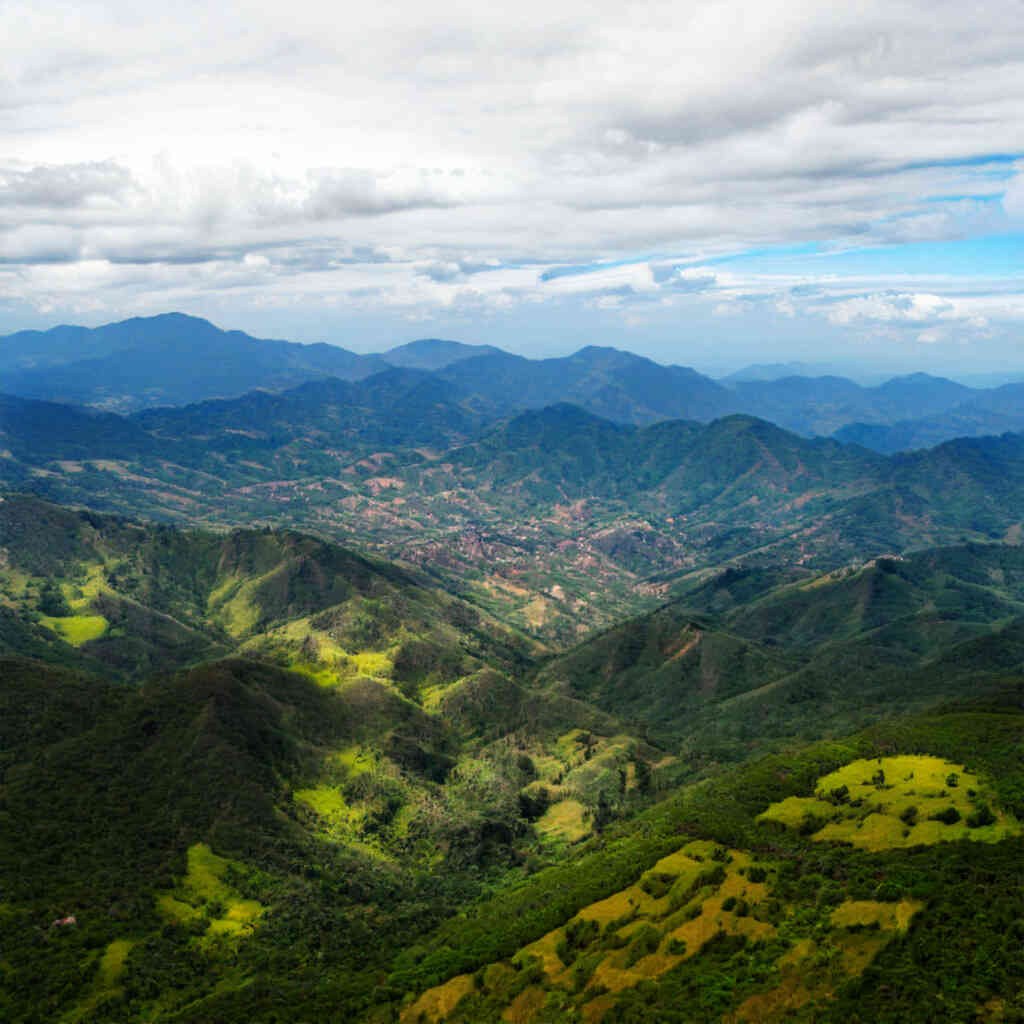}
		\put(62.8, 0.2){
		\frame{\includegraphics[width=0.08\textwidth, height=0.08\textwidth]{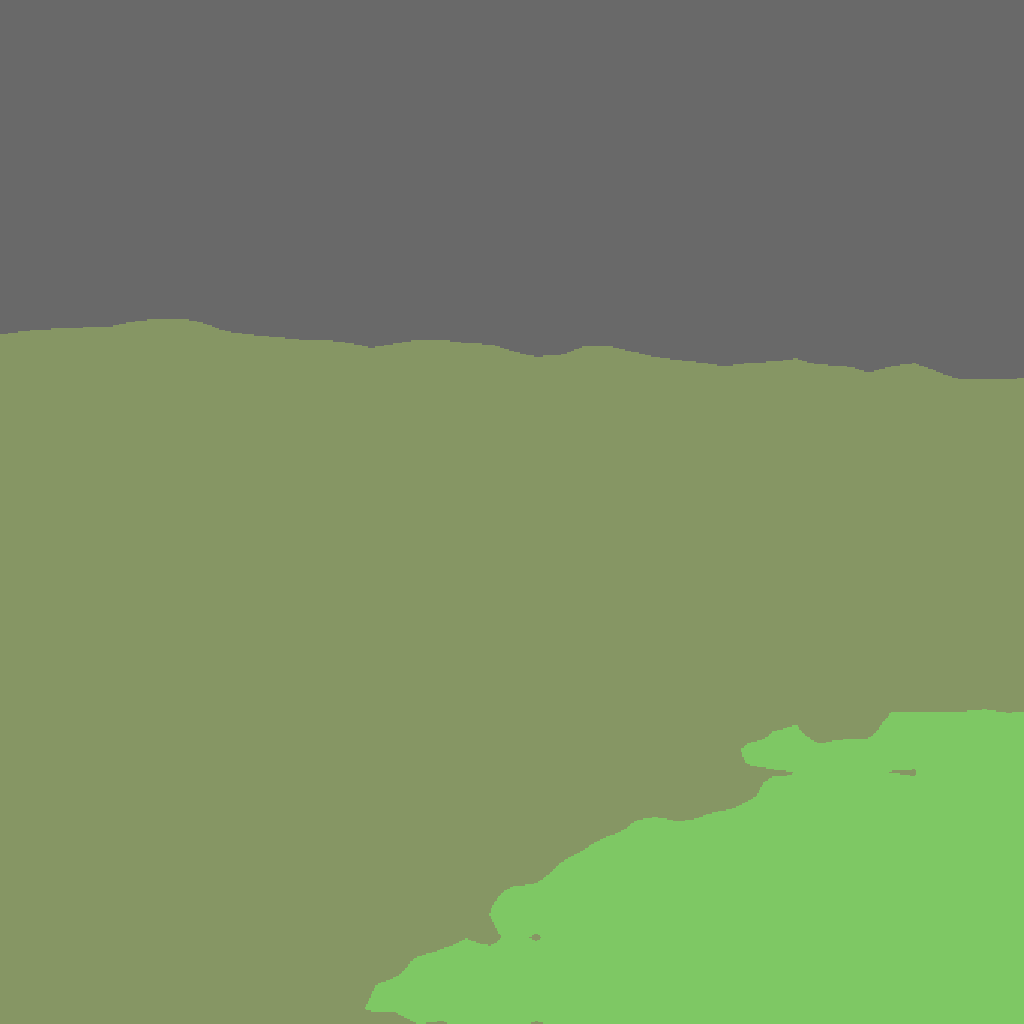}}}
	\end{overpic}
	\begin{minipage}[t]{0.23\textwidth}
		\centering
		Huge ocean waves clash \\ into rocks.
	\end{minipage}
	\begin{minipage}[t]{0.23\textwidth}
		\centering
		A beach with black sand \\ and palm trees.
	\end{minipage}
	\caption{Examples generated by PoE-GAN when conditioned on contradictory segmentation and text inputs. The text input is simply ignored.}
	\label{fig:limitation}
\end{figure}

\begin{figure}
	\centering
	\newcommand{\sizea}{0.185\linewidth}
	\setlength{\tabcolsep}{1.5pt}
	\renewcommand{\arraystretch}{0.8}
	\begin{tabular}{cc|ccccc}
		\footnotesize{Condition \#1} & \footnotesize{Condition \#2} &  \multicolumn{3}{c}{\footnotesize{Samples}}  \\
		\begin{minipage}[b]{\sizea}
			\scriptsize
			\centering This person is wearing hat, lipstick. She has big lips, and wavy hair.\vspace{1mm}
		\end{minipage}     &  \includegraphics[width=\sizea,height=\sizea]{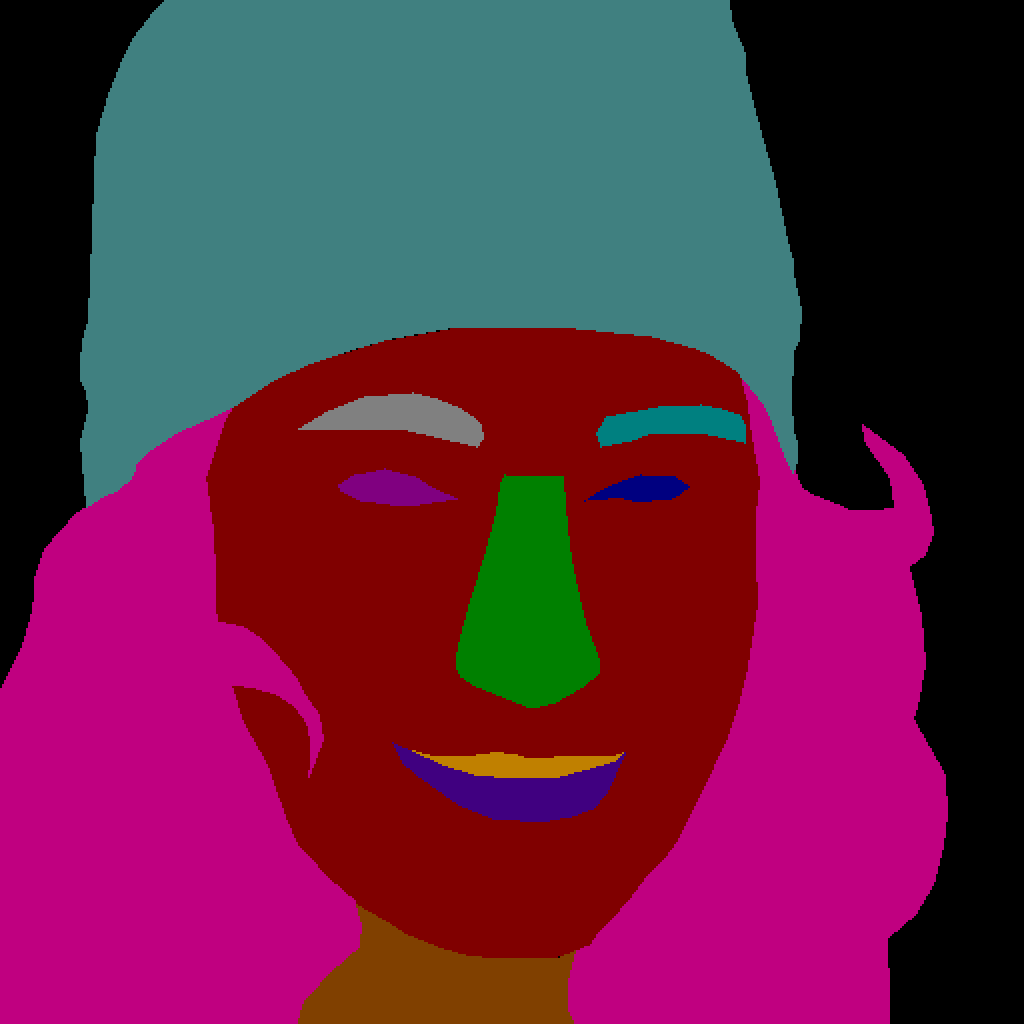}  & \includegraphics[width=\sizea,height=\sizea]{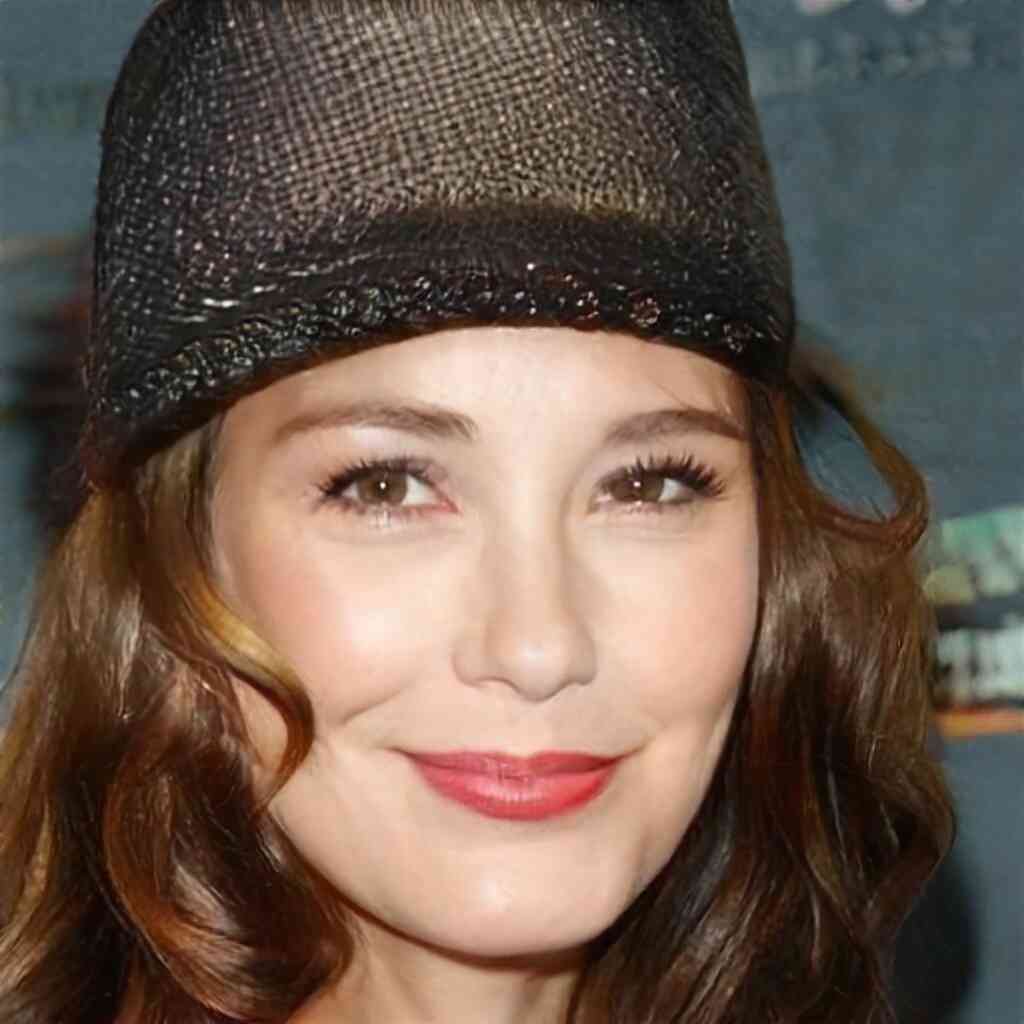}  & \includegraphics[width=\sizea,height=\sizea]{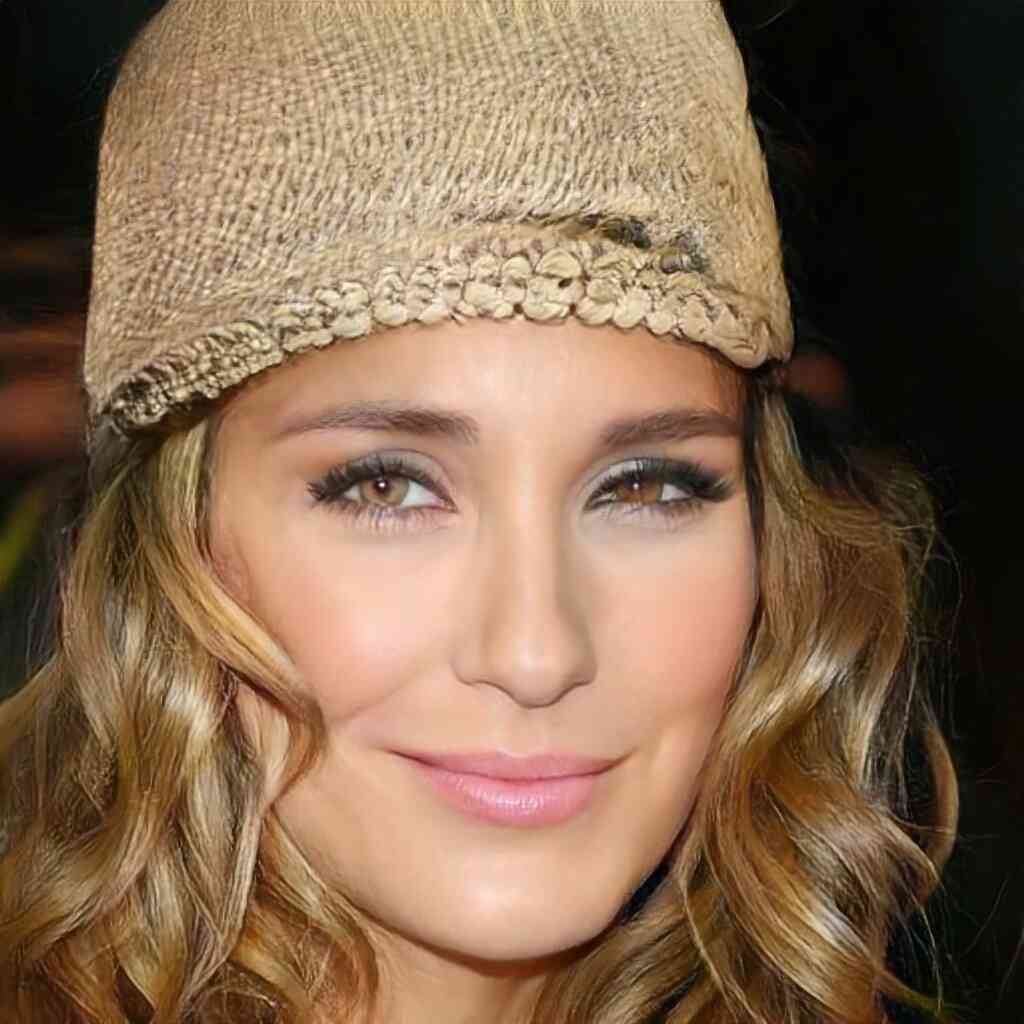} &  \includegraphics[width=\sizea,height=\sizea]{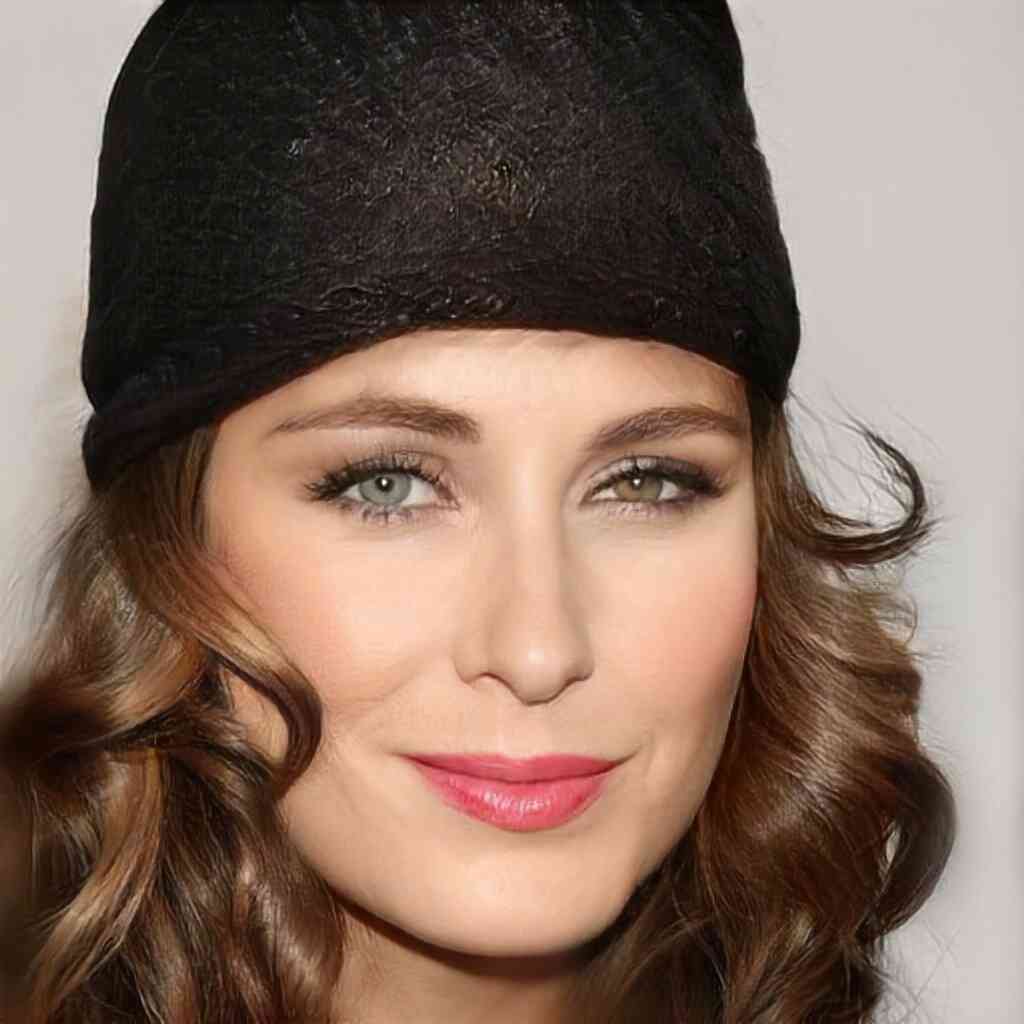} \\
		\begin{minipage}[b]{\sizea}
			\scriptsize
			\centering This smiling man has bushy eyebrows, and bags under eyes.\vspace{1mm}
		\end{minipage}     &  
	    \includegraphics[width=\sizea,height=\sizea]{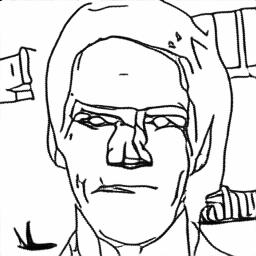} & \includegraphics[width=\sizea,height=\sizea]{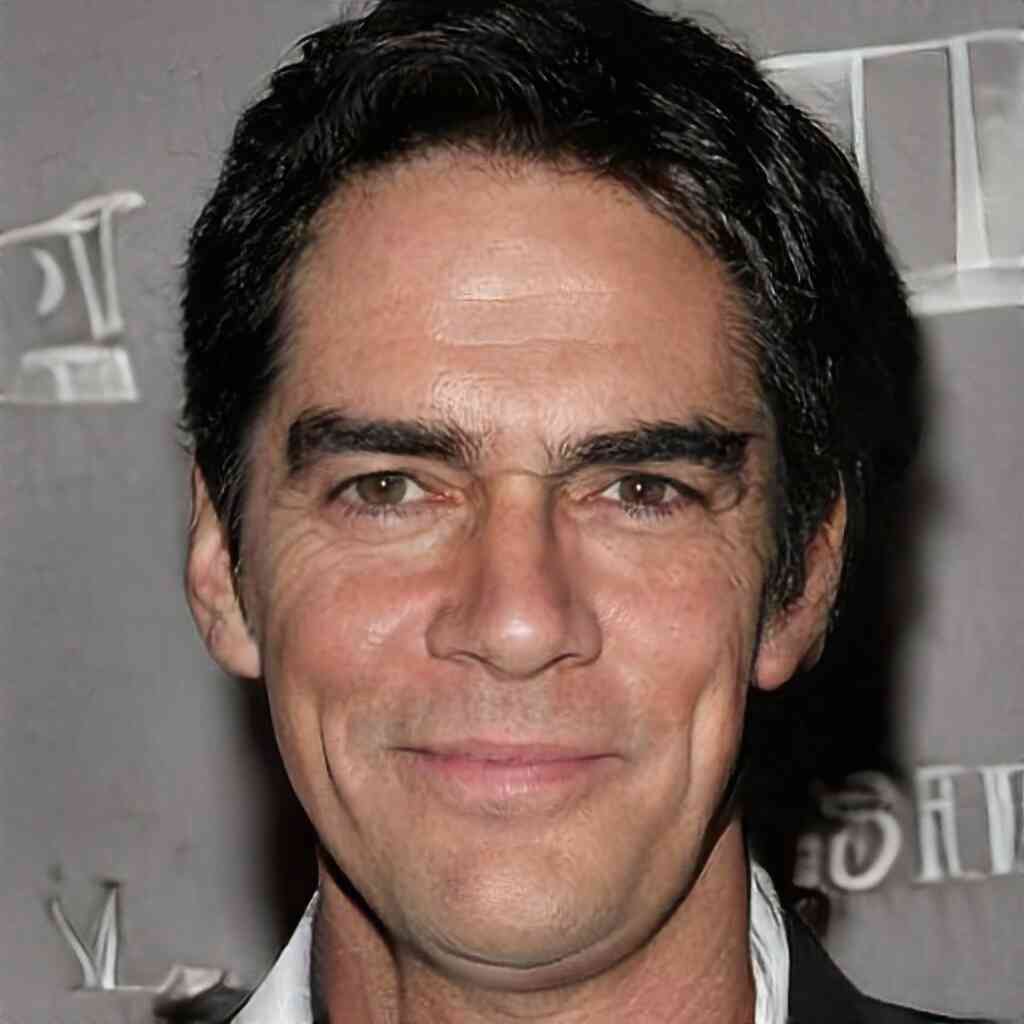}  & \includegraphics[width=\sizea,height=\sizea]{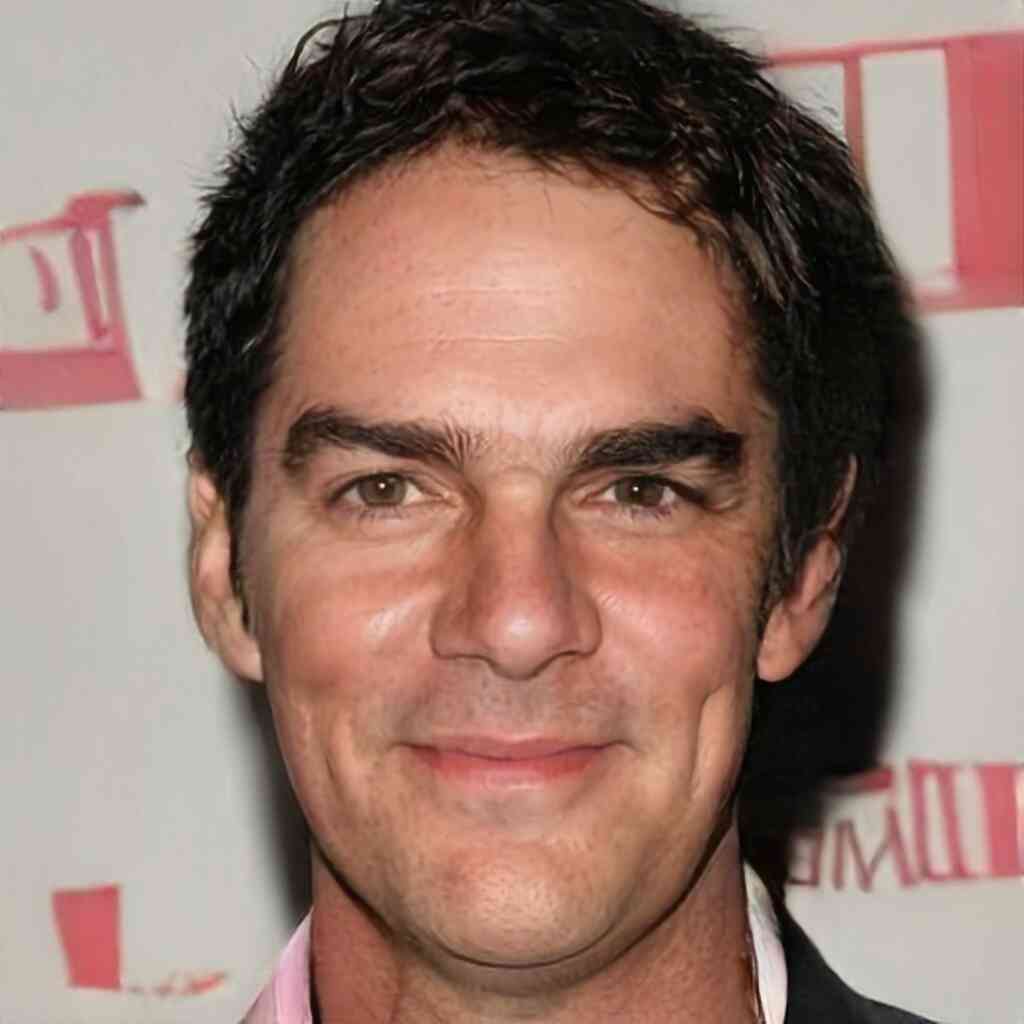}  & \includegraphics[width=\sizea,height=\sizea]{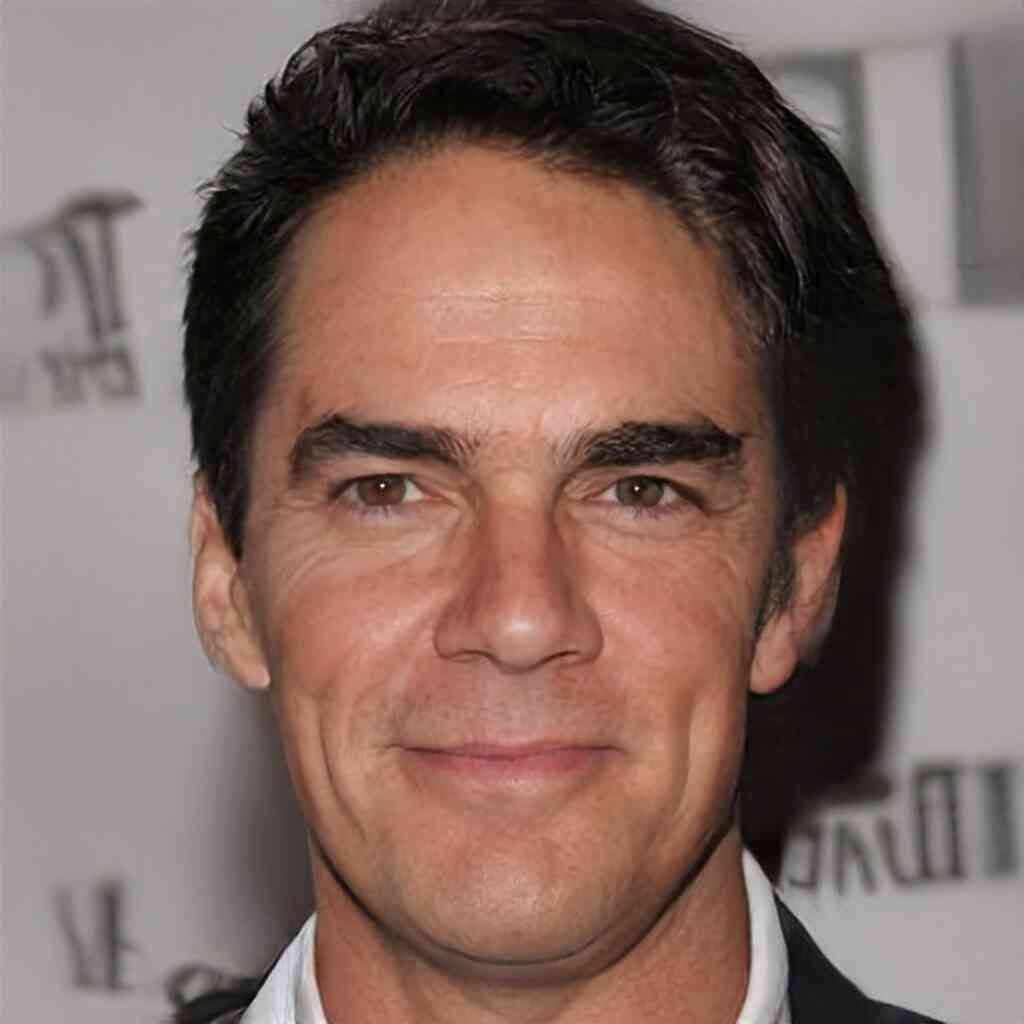} \\
		\includegraphics[width=\sizea,height=\sizea]{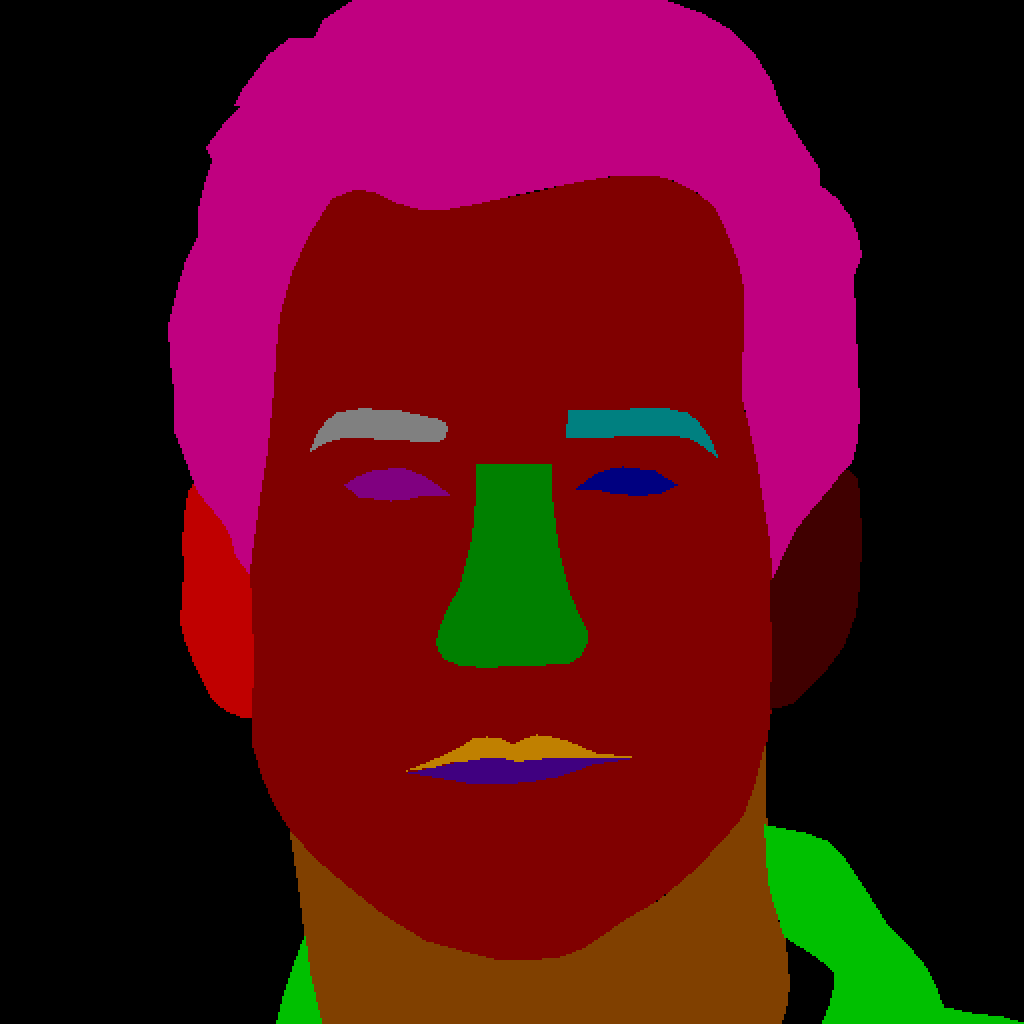} & \includegraphics[width=\sizea,height=\sizea]{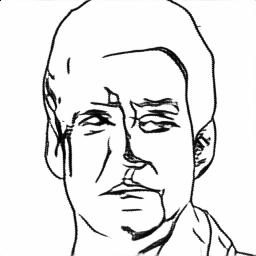} & \includegraphics[width=\sizea,height=\sizea]{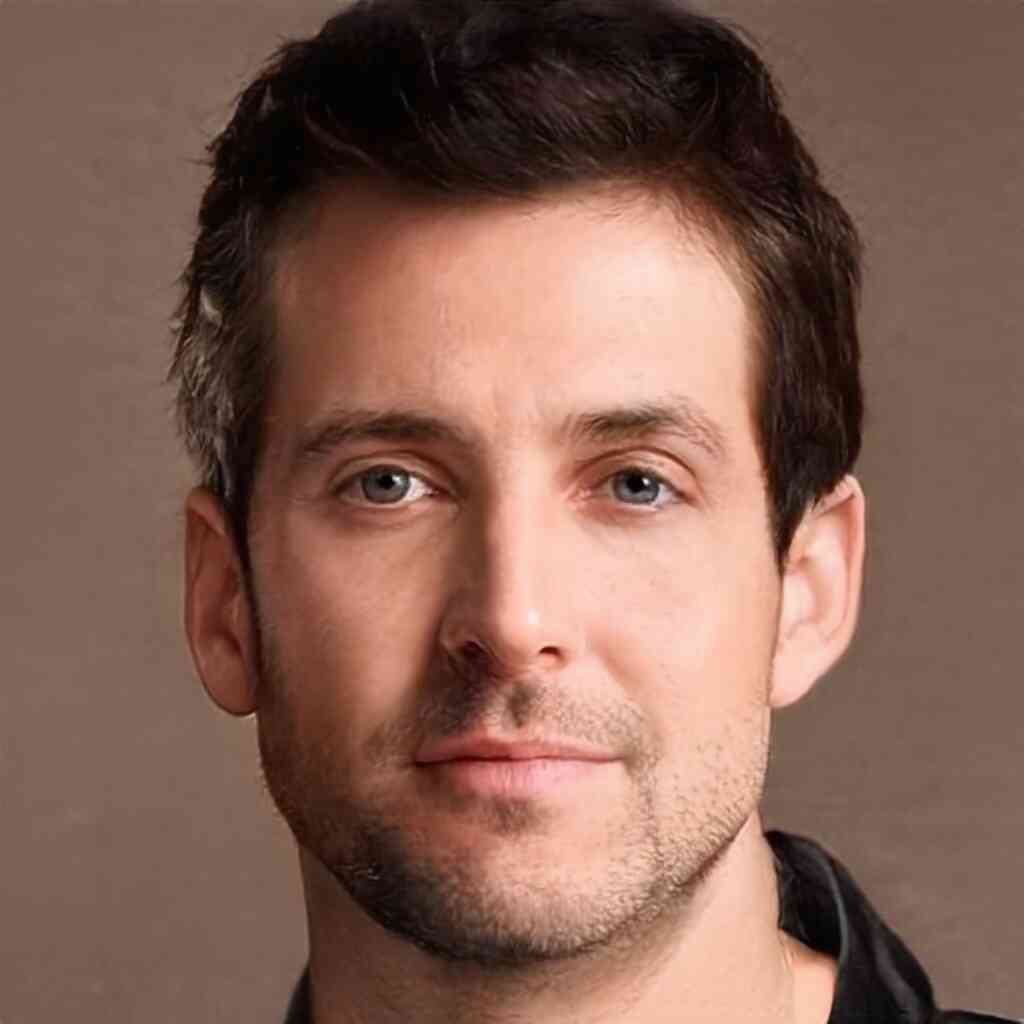}  & \includegraphics[width=\sizea,height=\sizea]{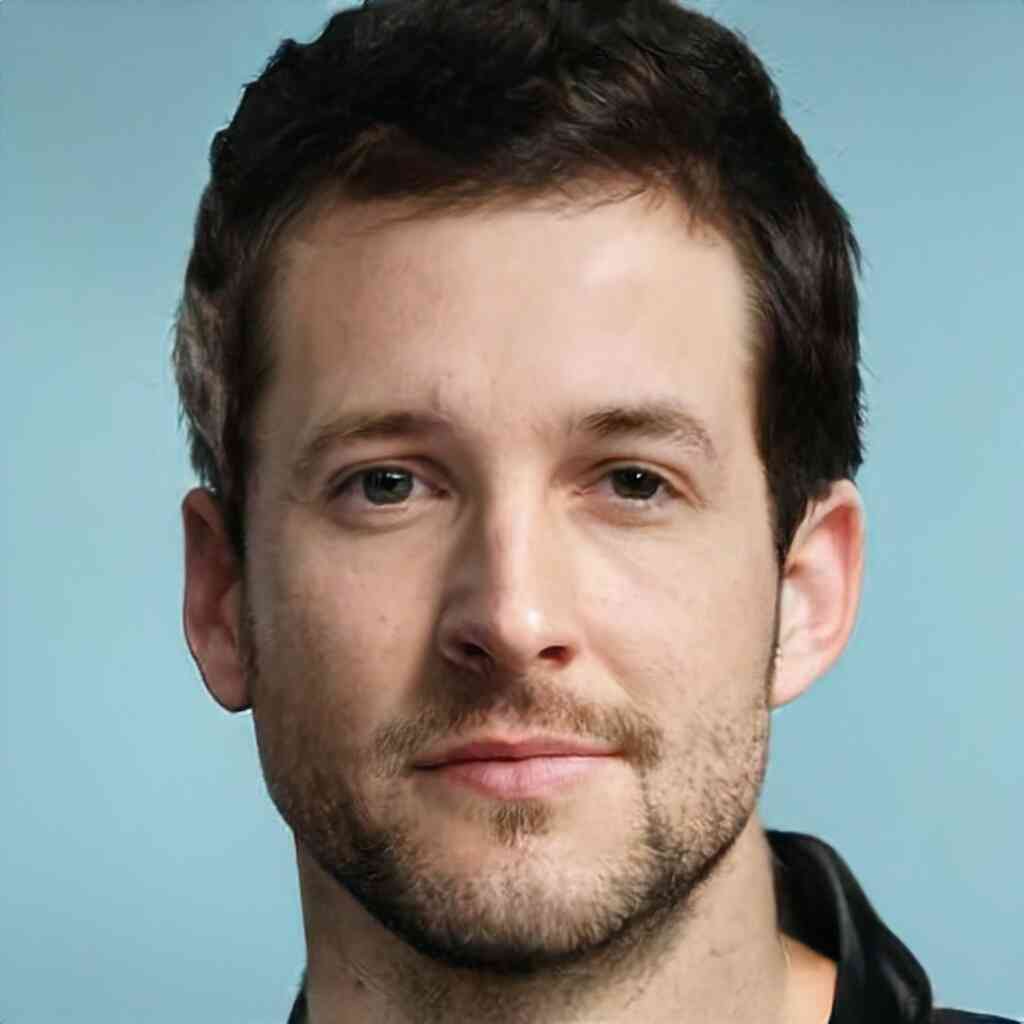}  & \includegraphics[width=\sizea,height=\sizea]{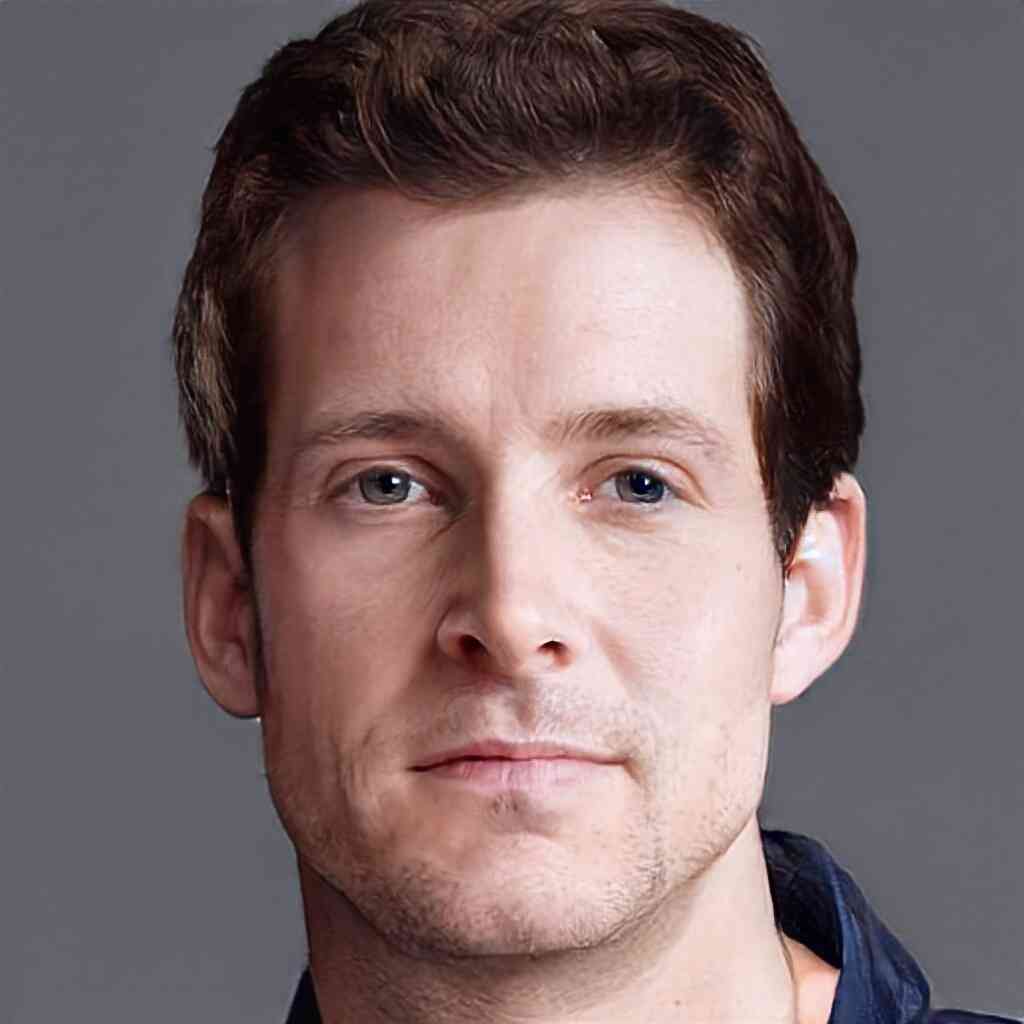} \\
		\includegraphics[width=\sizea,height=\sizea]{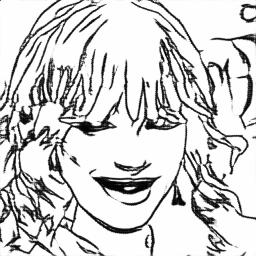} & \includegraphics[width=\sizea,height=\sizea]{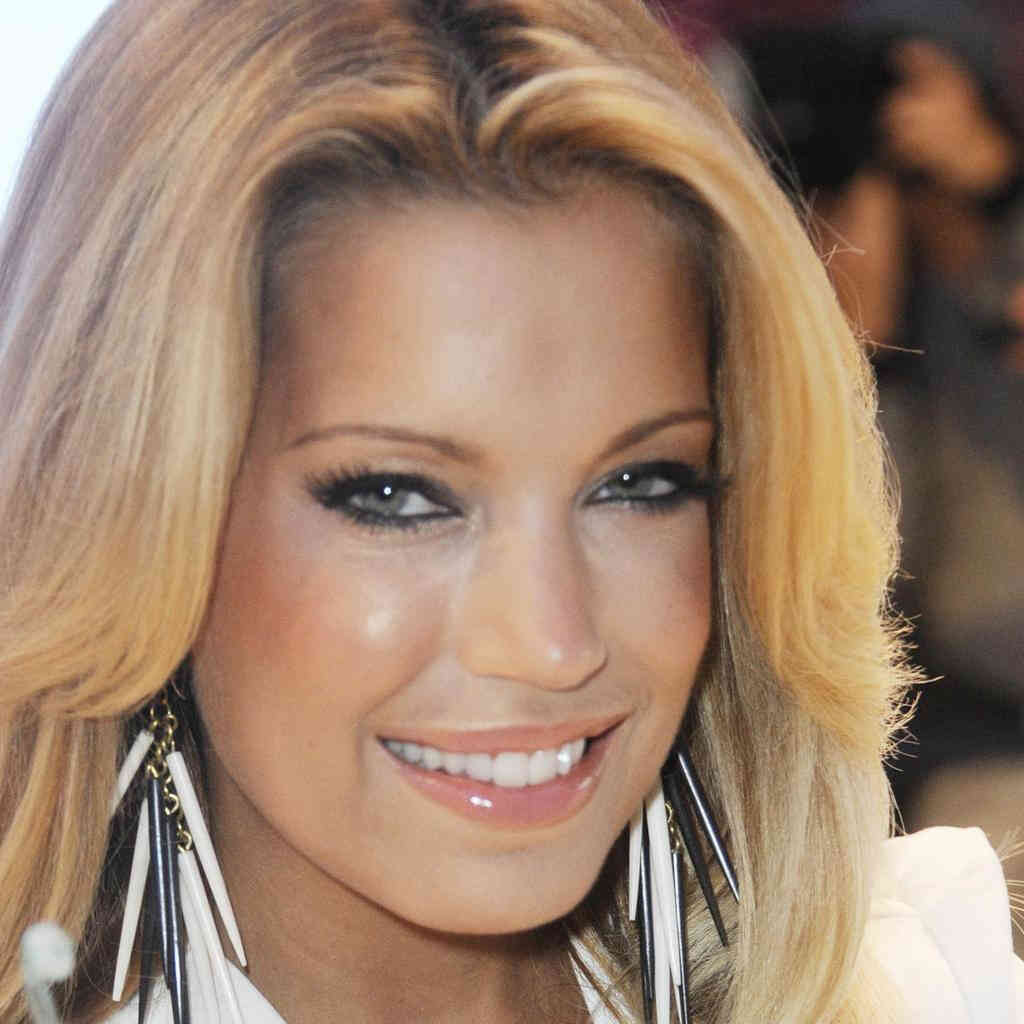} & \includegraphics[width=\sizea,height=\sizea]{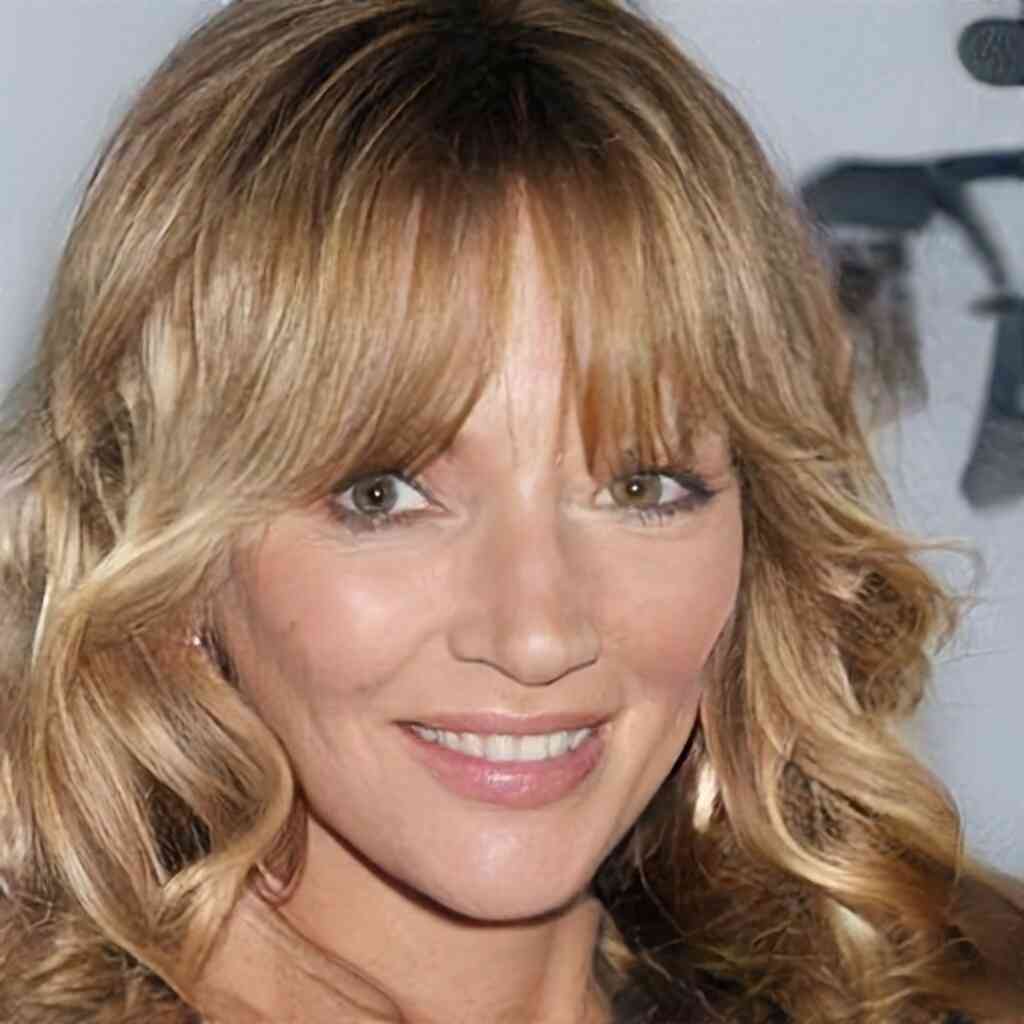}  & \includegraphics[width=\sizea,height=\sizea]{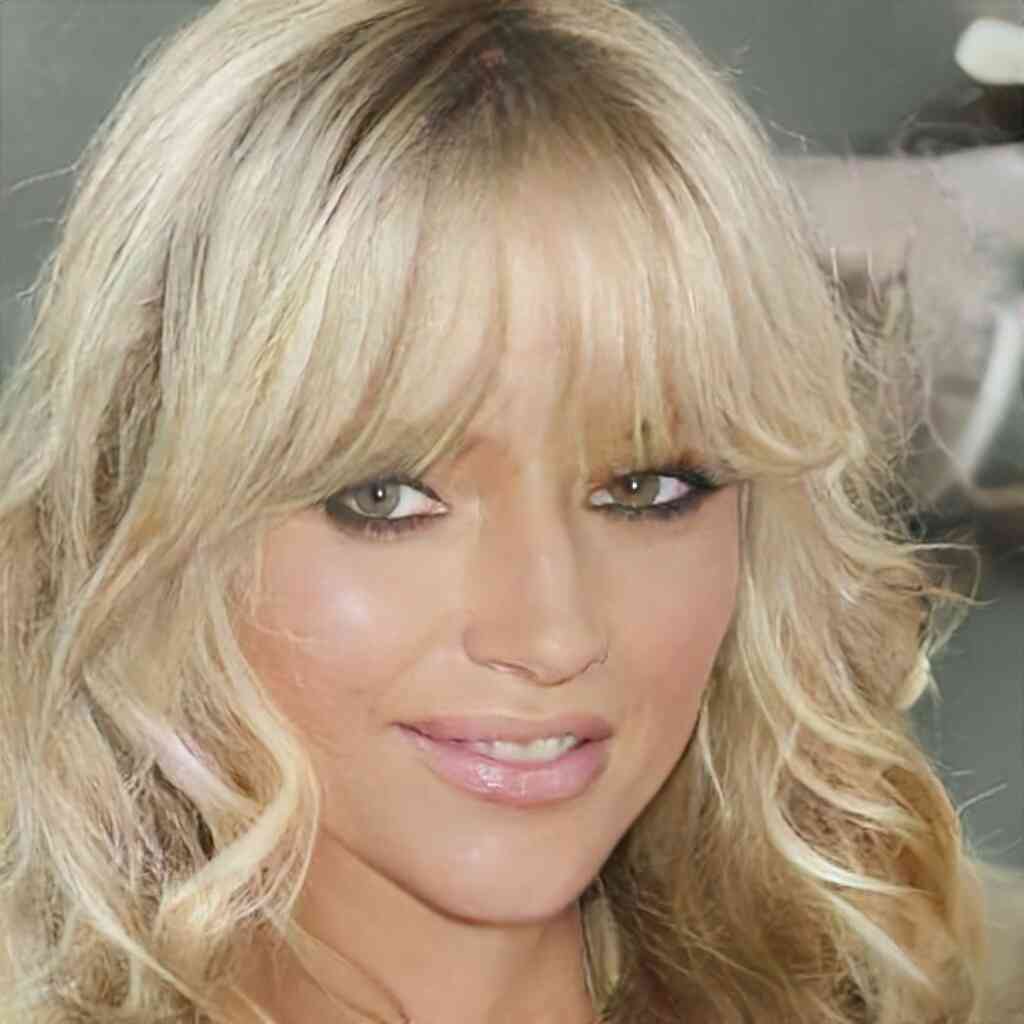}  & \includegraphics[width=\sizea,height=\sizea]{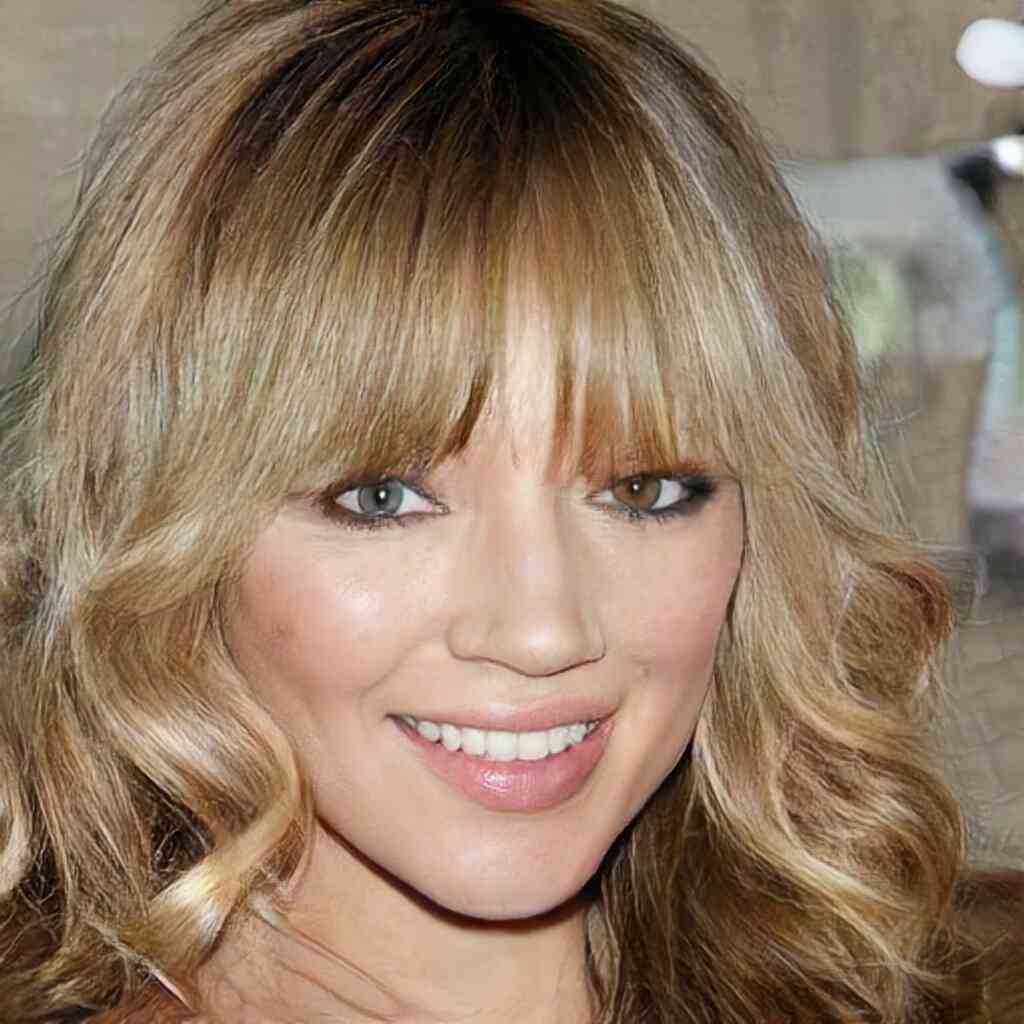} \\
		\includegraphics[width=\sizea,height=\sizea]{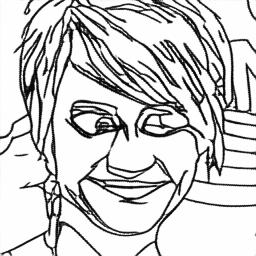} & \includegraphics[width=\sizea,height=\sizea]{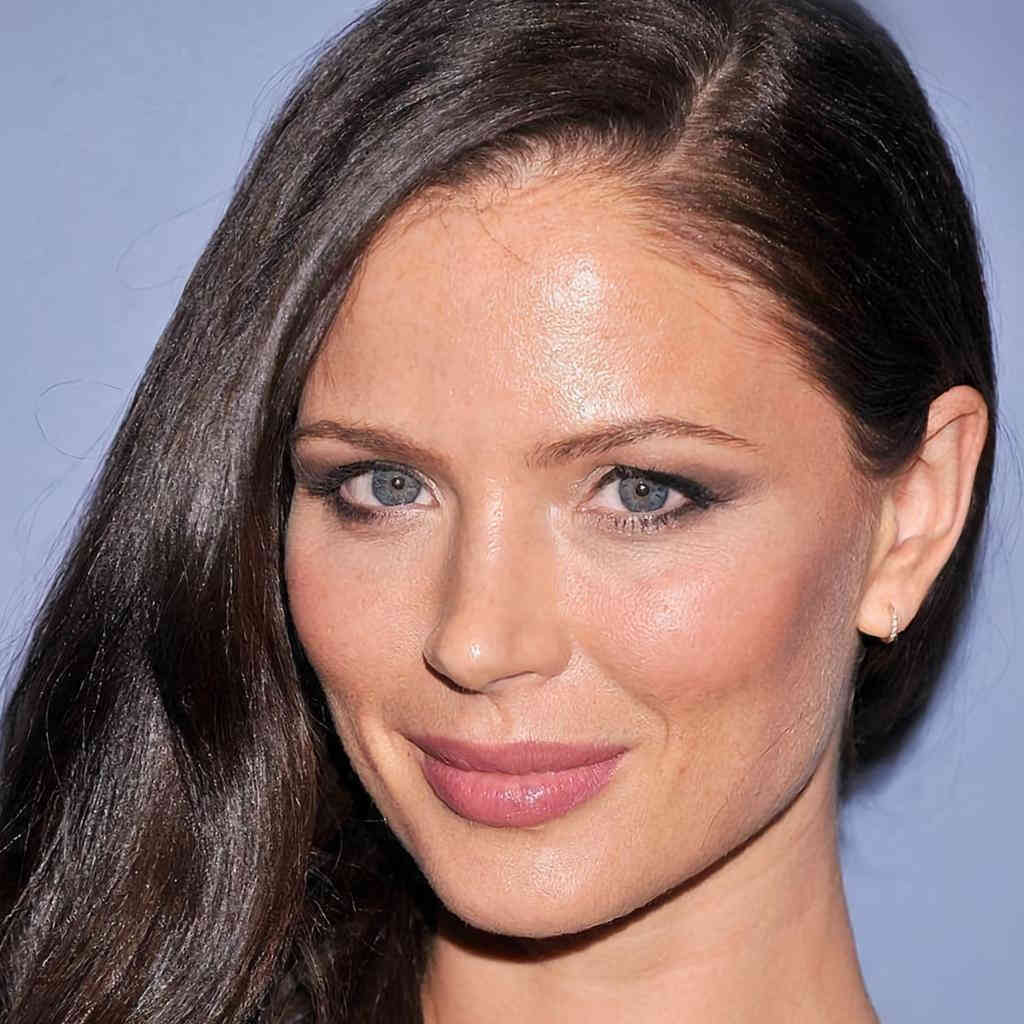} & \includegraphics[width=\sizea,height=\sizea]{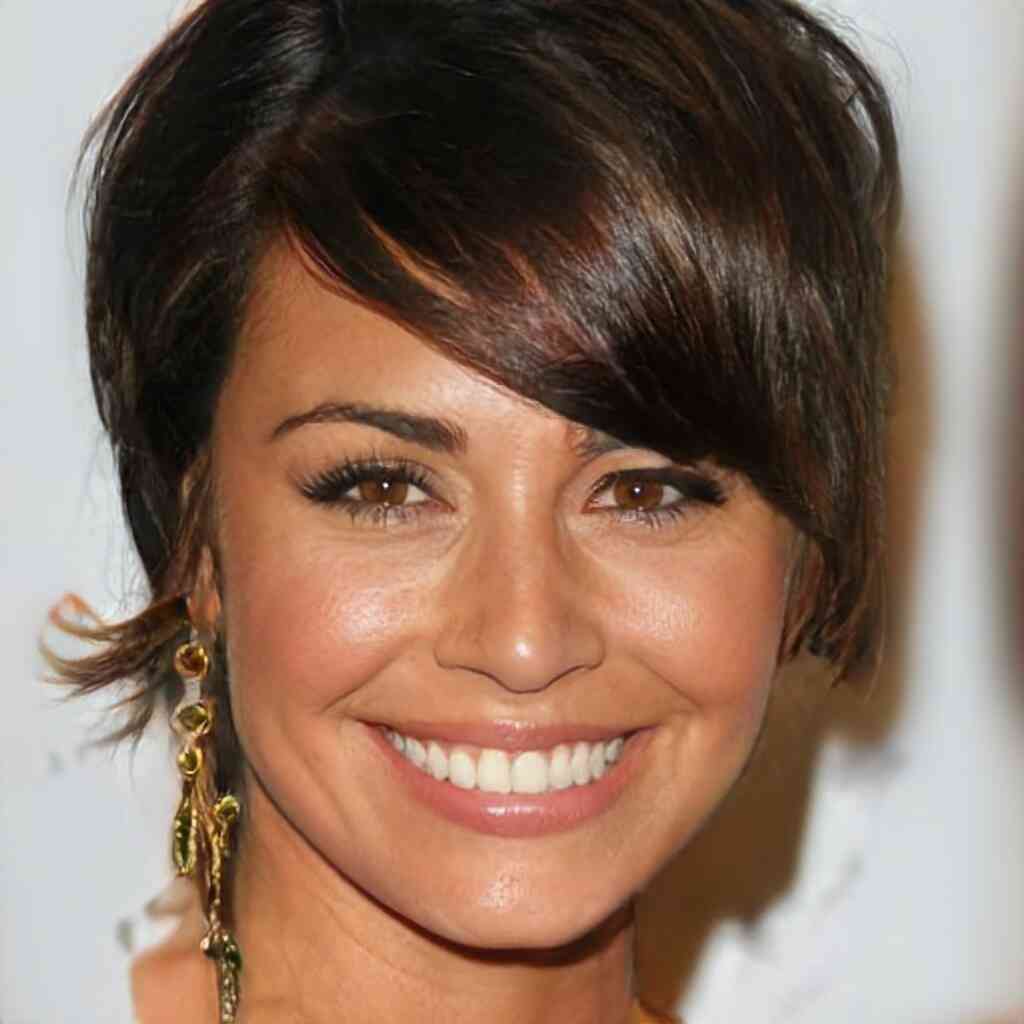}  & \includegraphics[width=\sizea,height=\sizea]{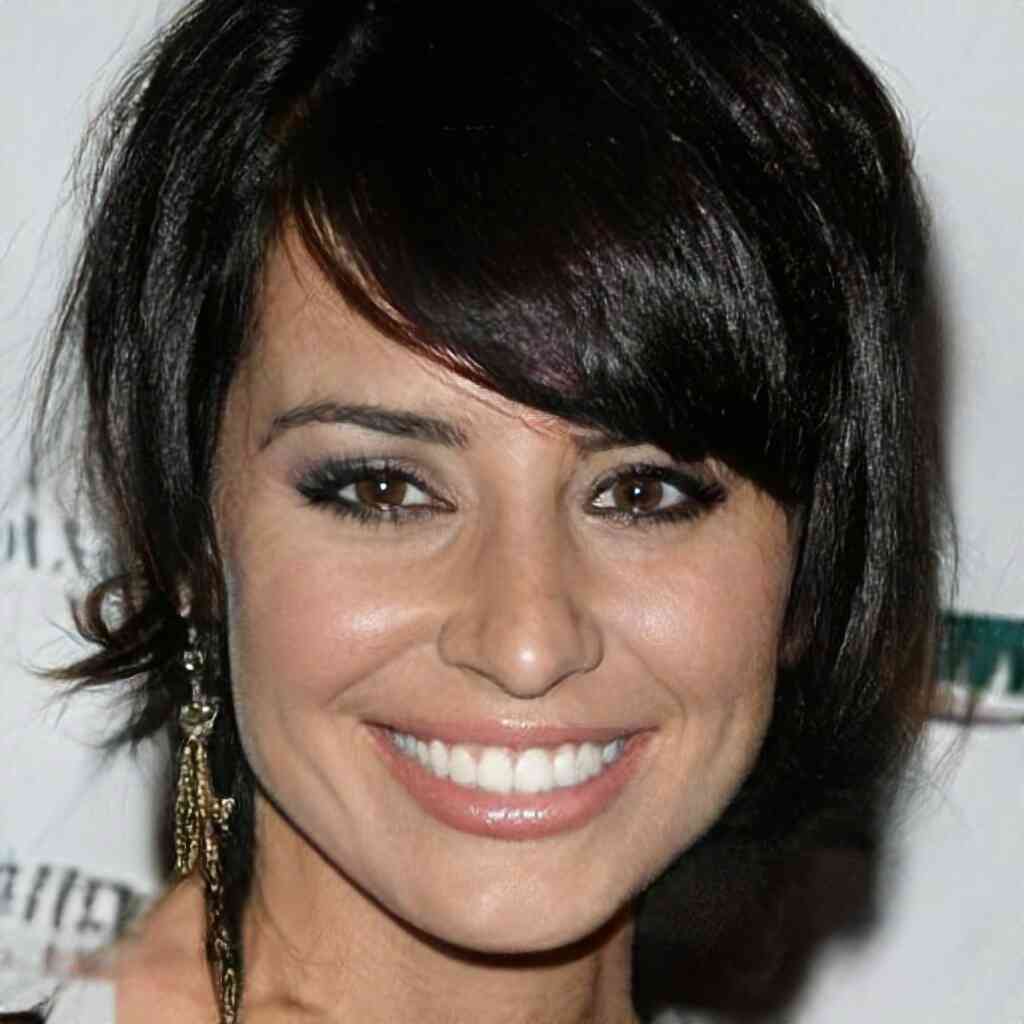}  & \includegraphics[width=\sizea,height=\sizea]{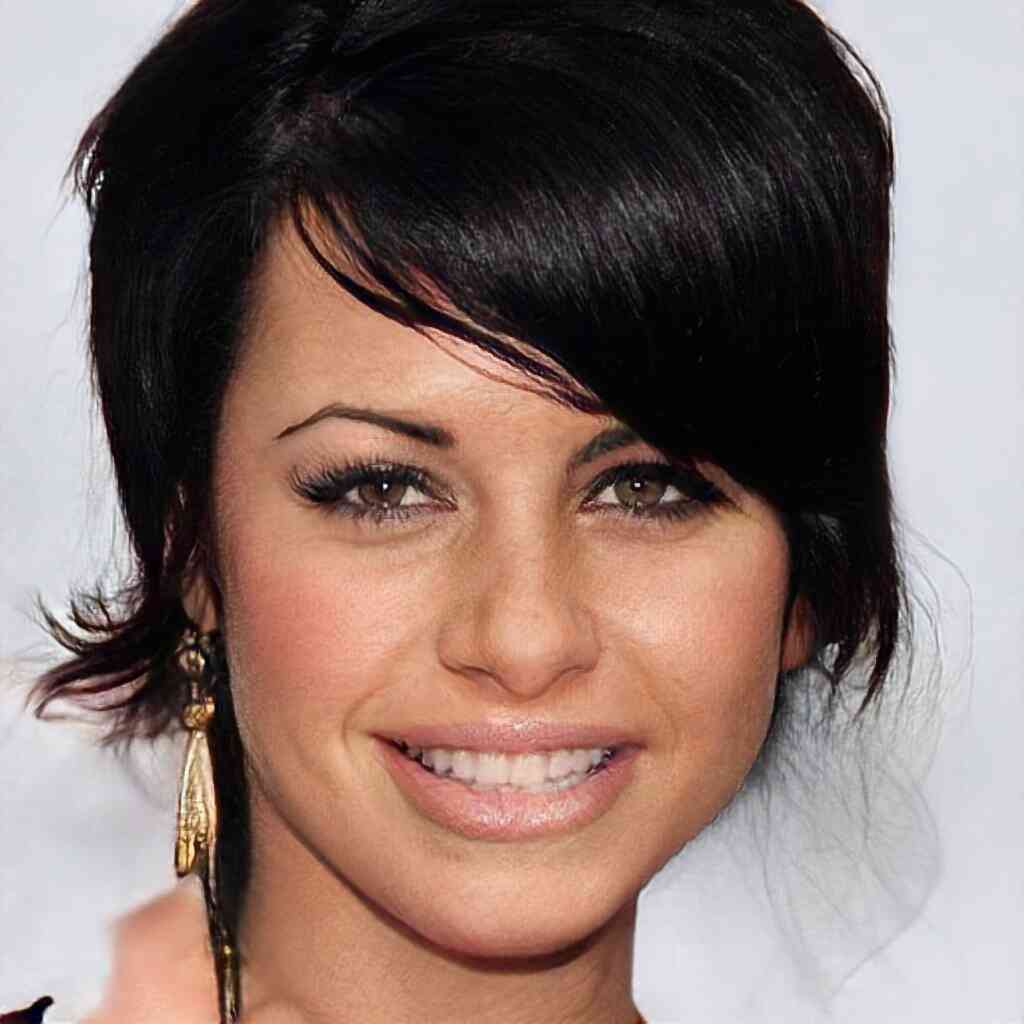} \\
		\begin{minipage}[b]{\sizea}
			\footnotesize
			\centering The person has goatee, mustache, and sideburns. \vspace{-2mm} 
		\end{minipage}     &
		\includegraphics[width=\sizea,height=\sizea]{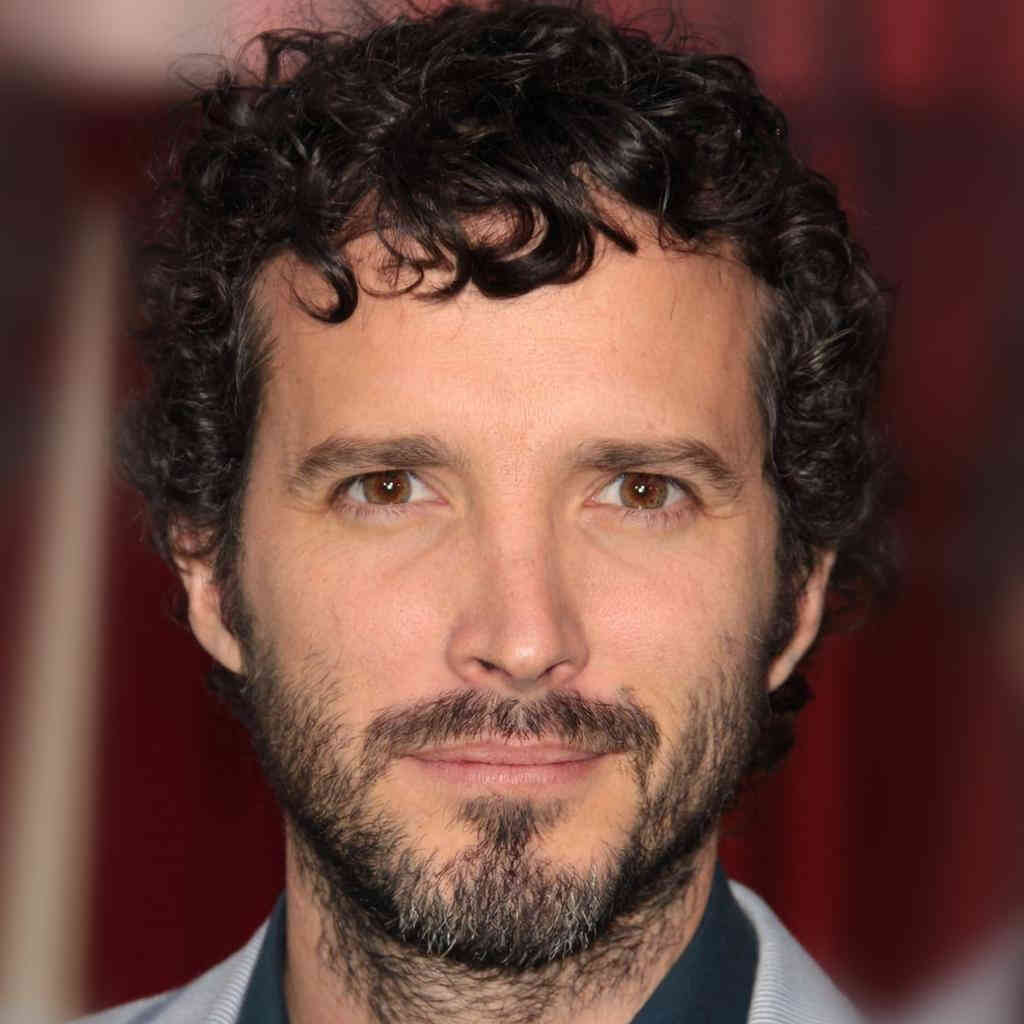} & \includegraphics[width=\sizea,height=\sizea]{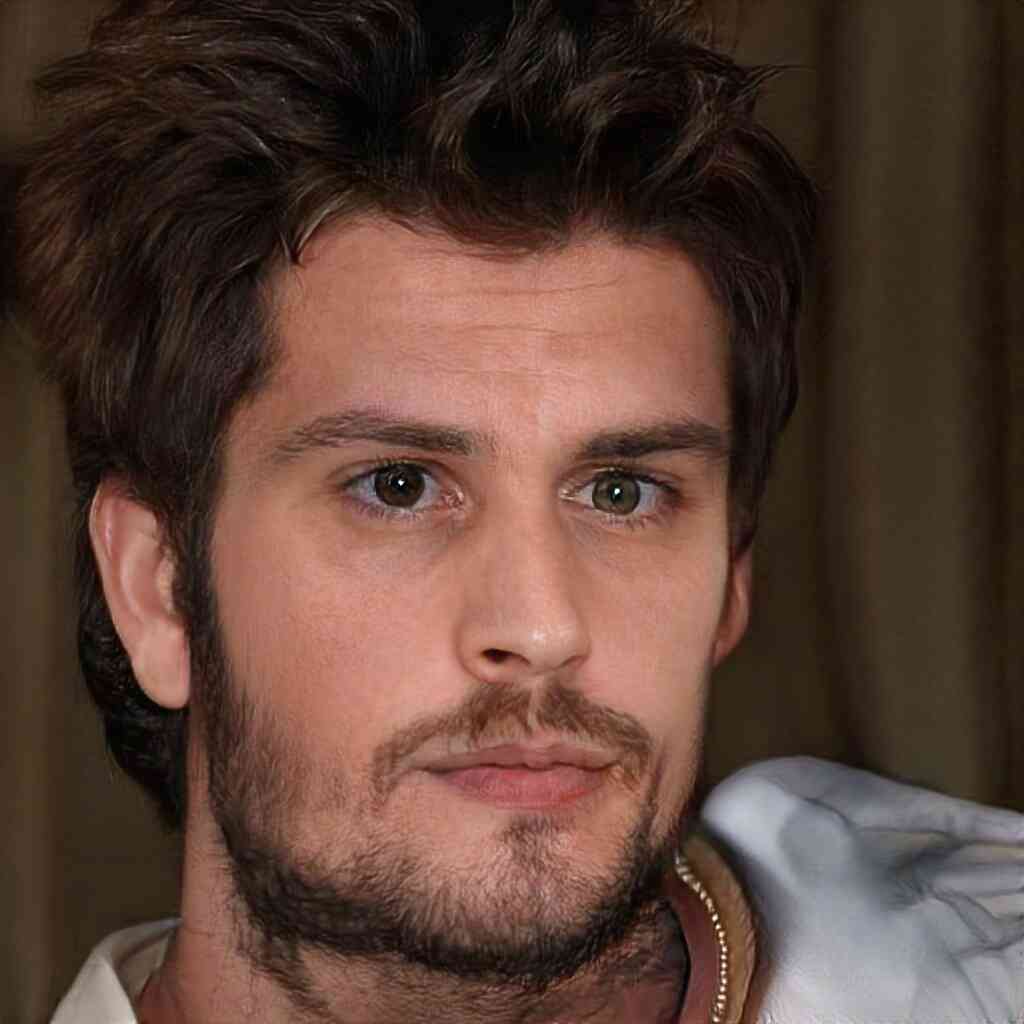}  & \includegraphics[width=\sizea,height=\sizea]{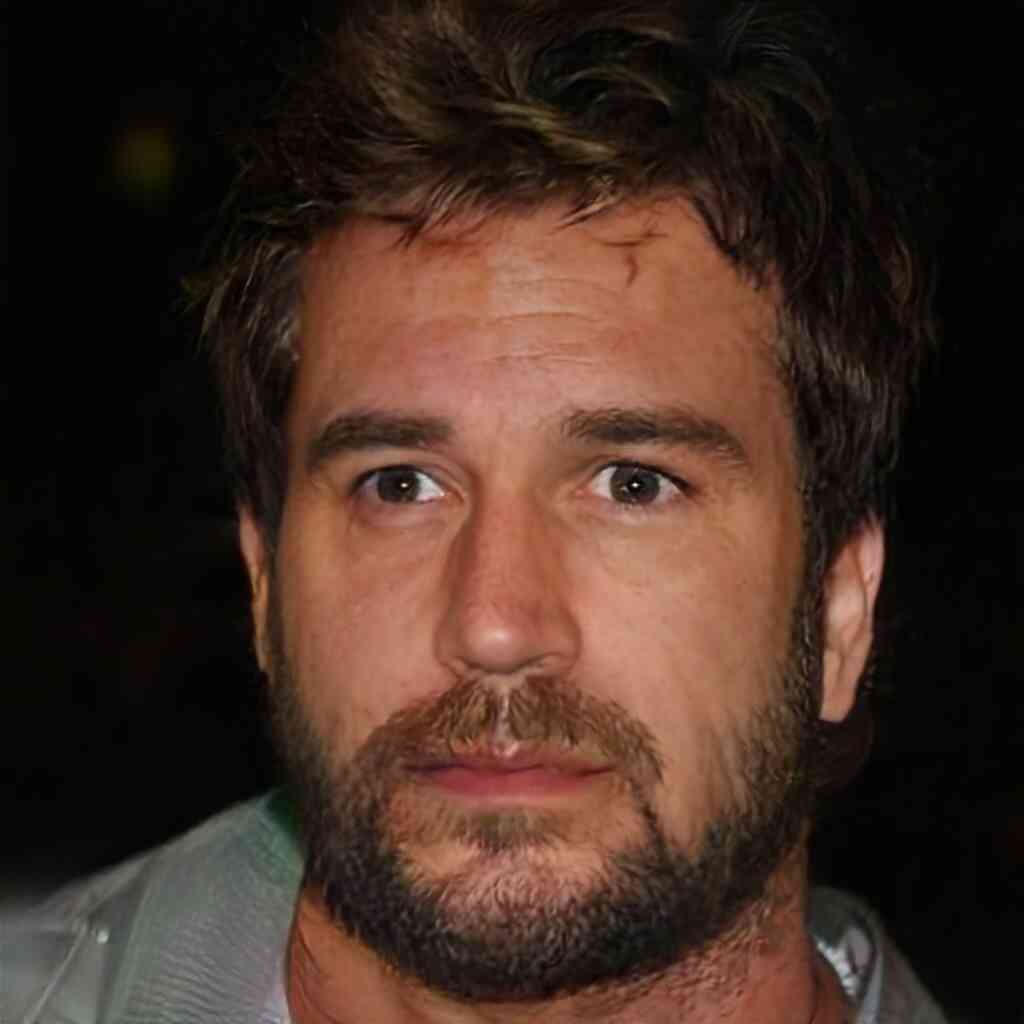}  & \includegraphics[width=\sizea,height=\sizea]{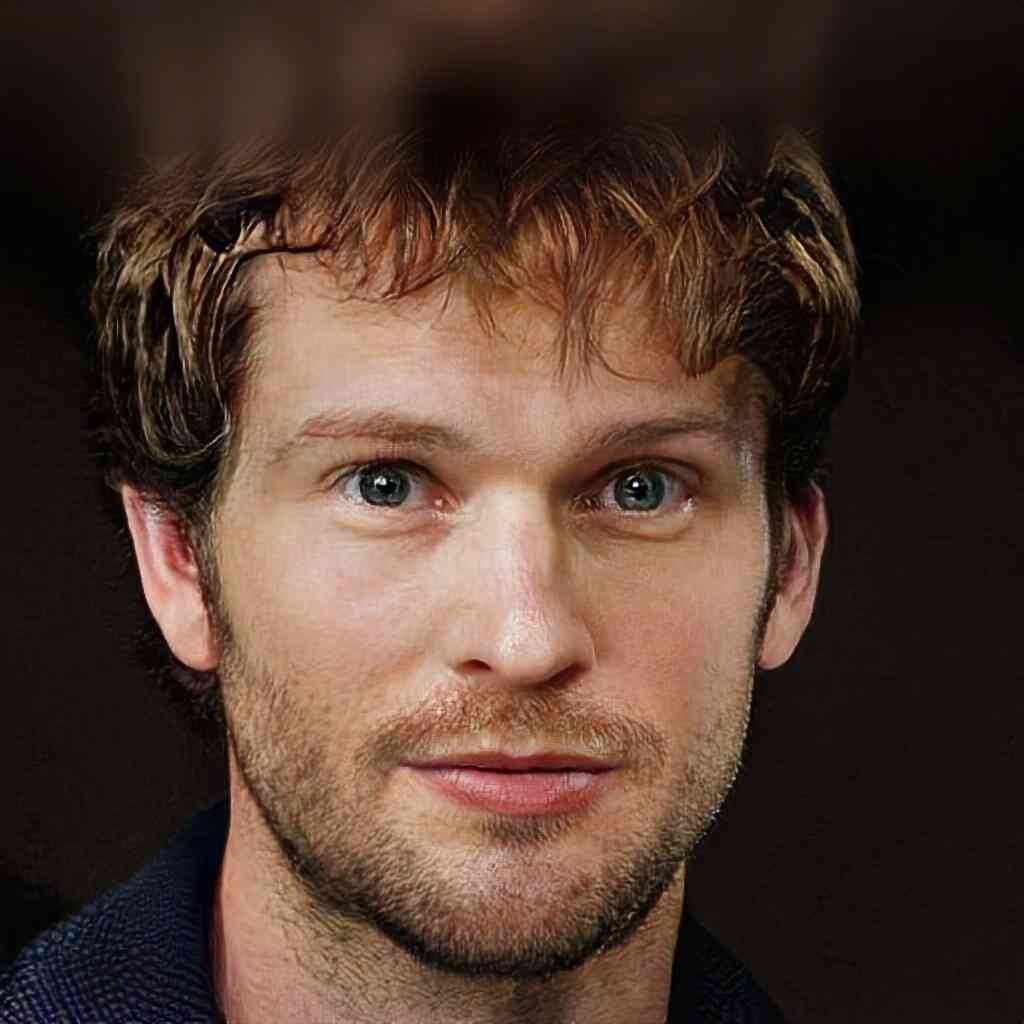} \\
	\end{tabular}

	\vspace{-2mm}
	\caption{Examples of multimodal conditional image synthesis on MM-CelebA-HQ. We show three random samples from PoE-GAN conditioned on two modalities.}
	\label{fig:celeba_mm_2}
	\vspace{-2mm}
\end{figure}

\begin{figure*}
	\centering
	\newcommand{\sizea}{0.16\linewidth}
	\setlength{\tabcolsep}{2pt}
	\begin{tabular}{cc|cccc}
		Condition \#1 & Condition \#2 &   \multicolumn{4}{c}{Samples}  \\
		\begin{minipage}[b]{\sizea}
			\small
			\centering Large mustard yellow commercial airplane parked in the airport.\vspace{6pt} 
		\end{minipage}     
		& \includegraphics[width=\sizea]{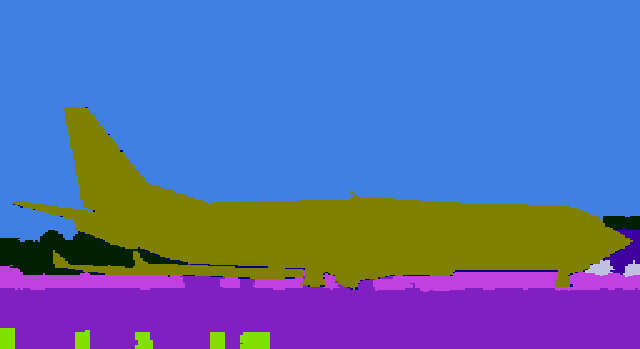}    & \includegraphics[width=\sizea]{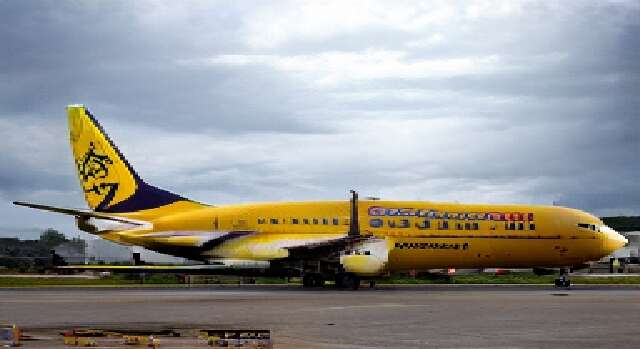}  & \includegraphics[width=\sizea]{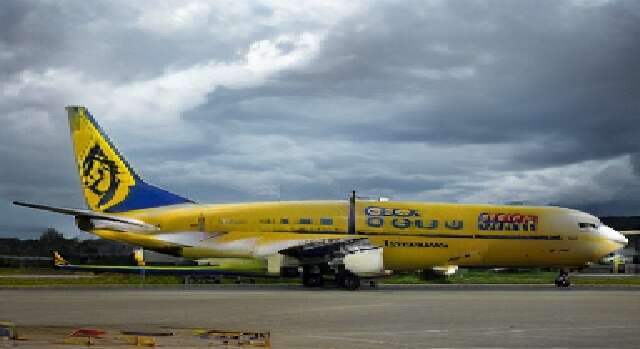}  & \includegraphics[width=\sizea]{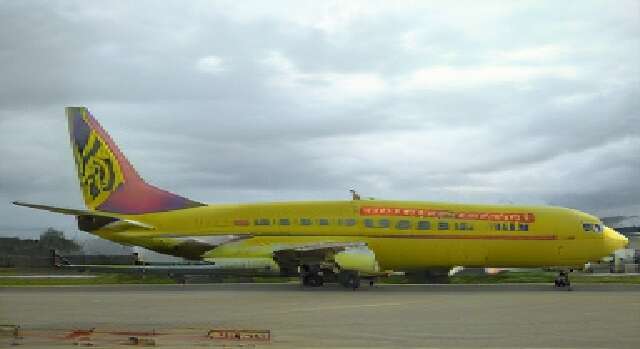}  & \includegraphics[width=\sizea]{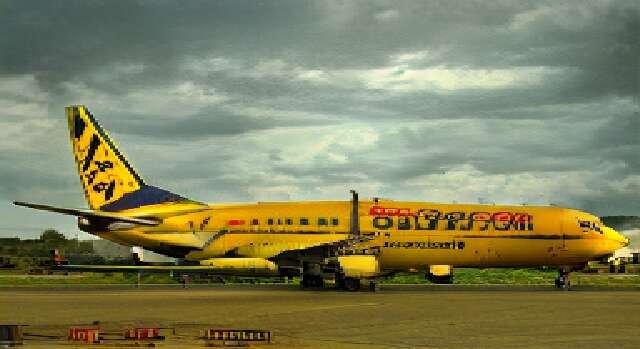} \\
		\begin{minipage}[b]{\sizea}
			\small
			\centering A rain covered terrain after a night of rain. \vspace{20pt} 
		\end{minipage}     
		& \includegraphics[width=\sizea]{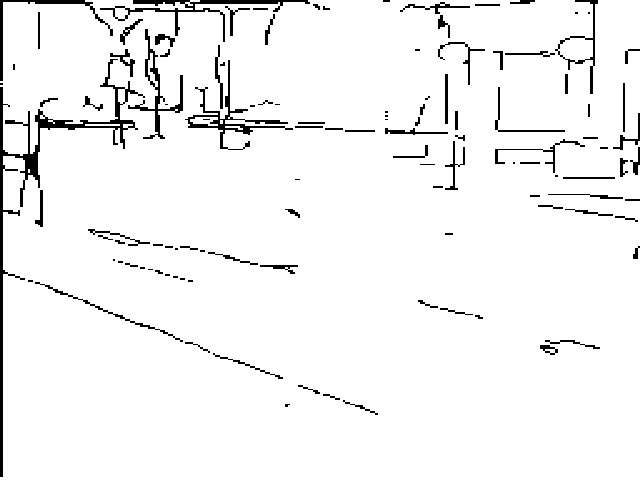}    & \includegraphics[width=\sizea]{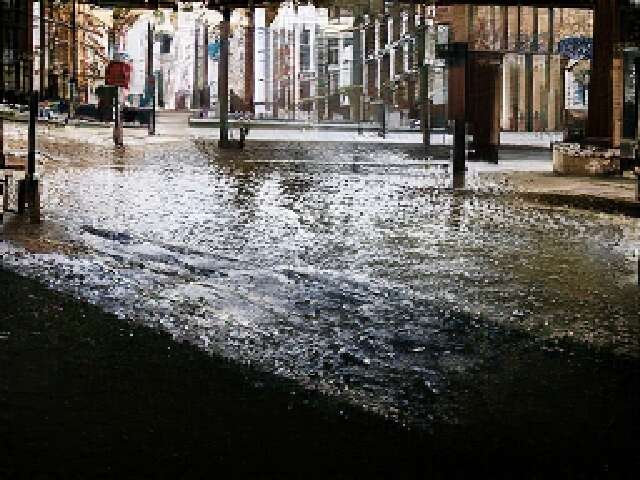}  & \includegraphics[width=\sizea]{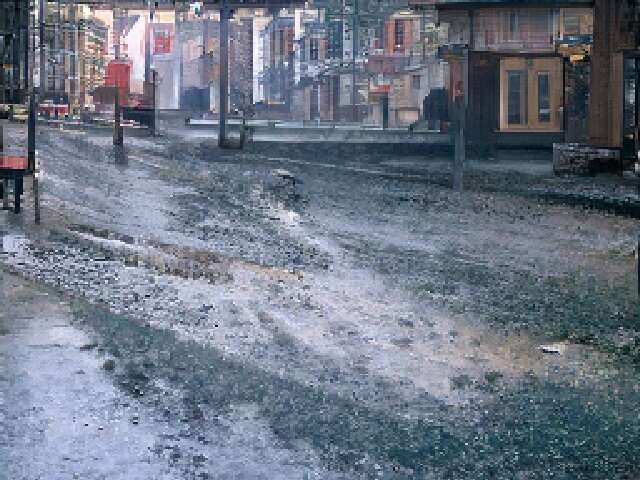}  & \includegraphics[width=\sizea]{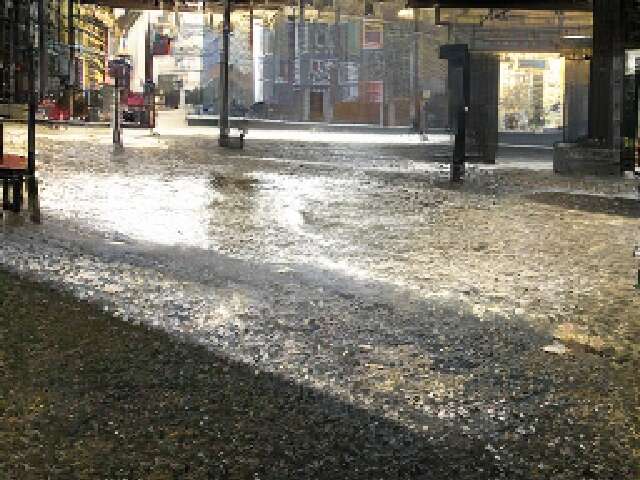}  & \includegraphics[width=\sizea]{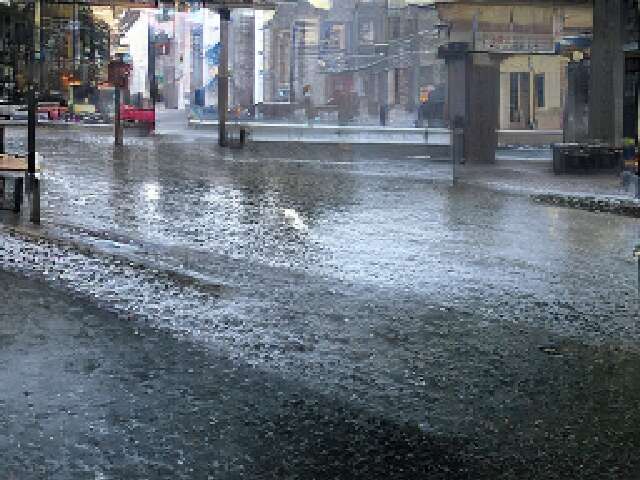} \\
		\includegraphics[width=\sizea]{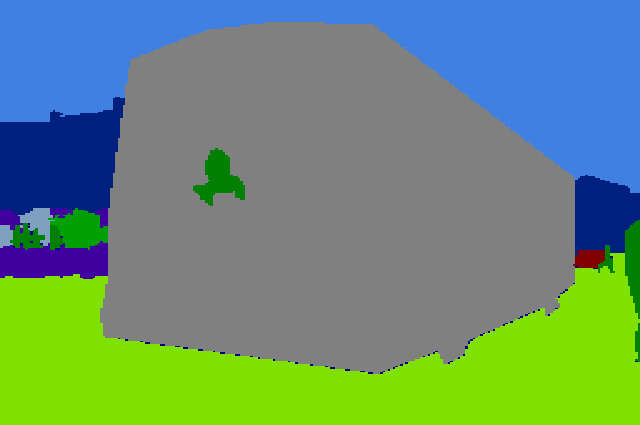} & \includegraphics[width=\sizea]{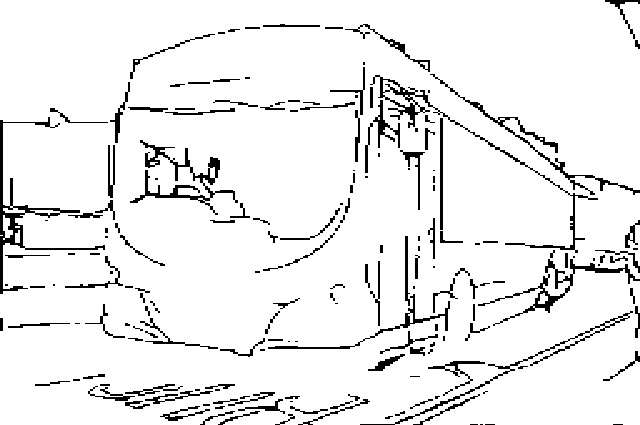} & \includegraphics[width=\sizea]{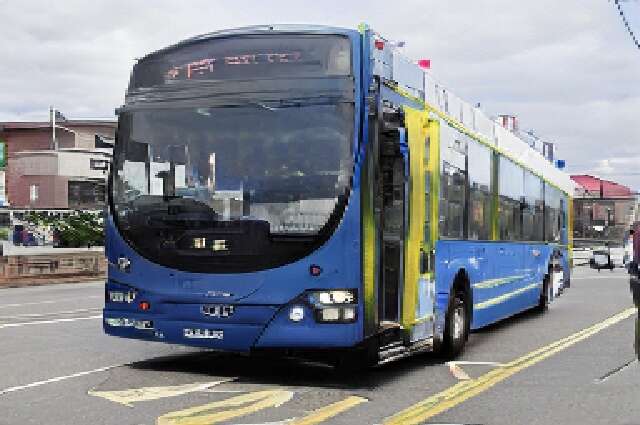}  & \includegraphics[width=\sizea]{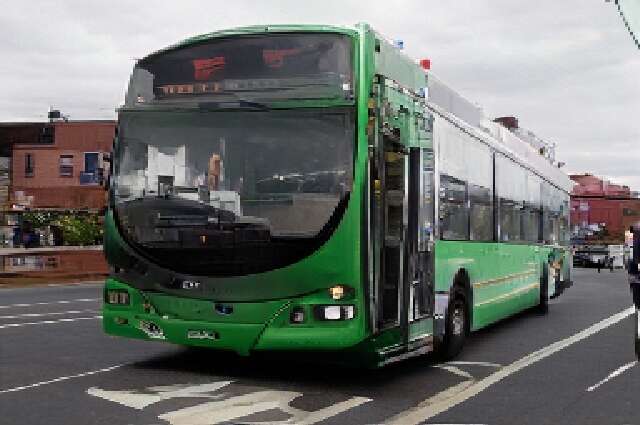}  & \includegraphics[width=\sizea]{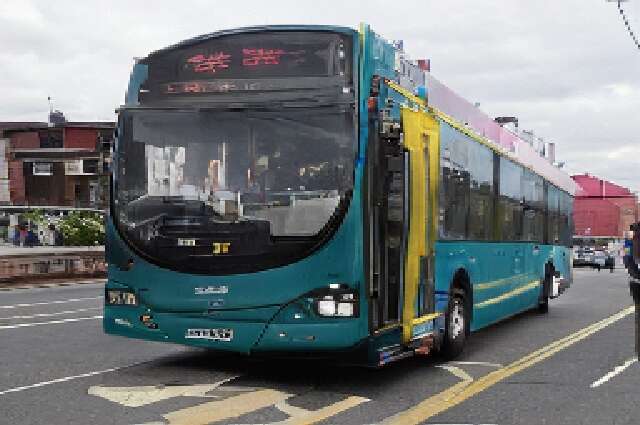}  & \includegraphics[width=\sizea]{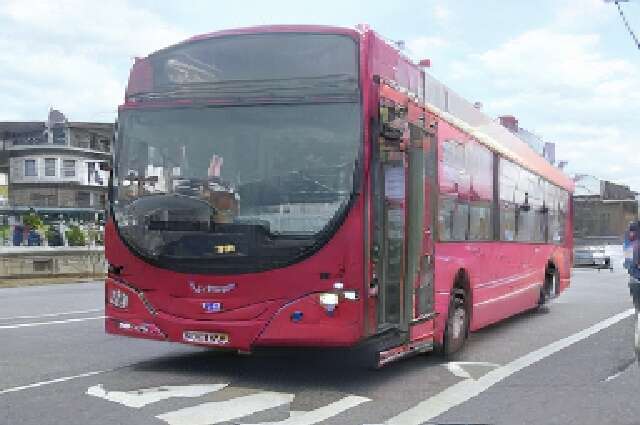}
	\end{tabular}
	\vspace{-2mm}
	\caption{Examples of multimodal conditional image synthesis on MS-COCO. We show three random samples from PoE-GAN conditioned on two modalities~(from top to bottom: text + segmentation, text + sketch, and segmentation + sketch).}
	\vspace{-2mm}
	\label{fig:coco_mm_2}
\end{figure*}

\begin{figure*}
\centering
\newcommand{\sizea}{0.16\linewidth}
\setlength{\tabcolsep}{2pt}
\begin{tabular}{cc|cccc}
	Condition \#1 & Condition \#2~(Style)  &   \multicolumn{4}{c}{Samples}  \\
		\includegraphics[width=\sizea]{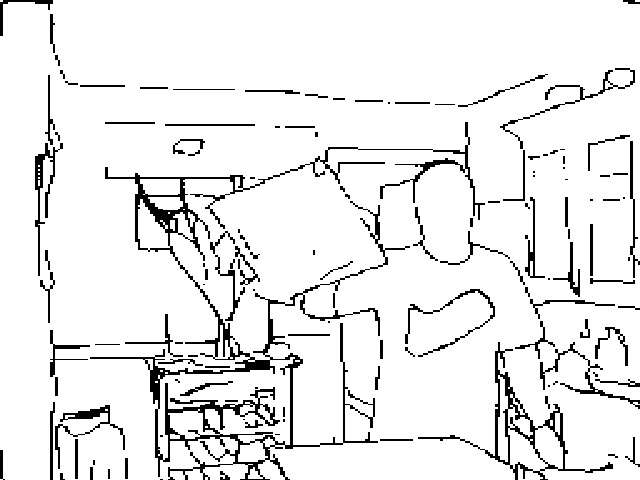} & \includegraphics[width=\sizea]{figures/coco_mm/000000357888} & \includegraphics[width=\sizea]{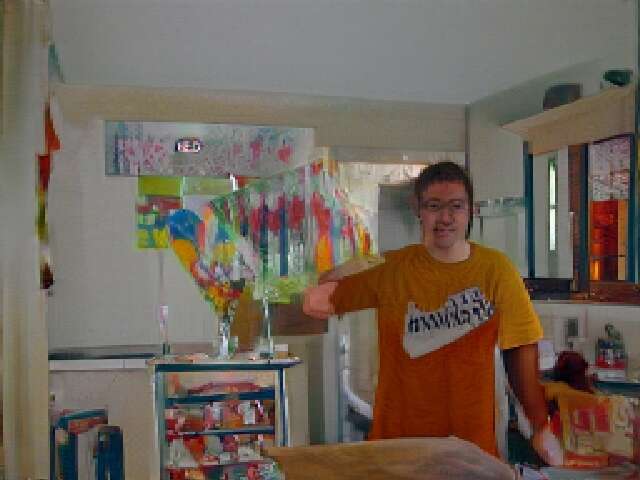}  & \includegraphics[width=\sizea]{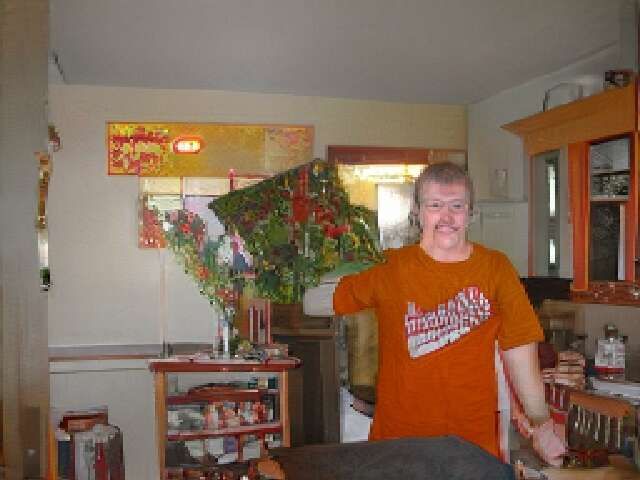}  & \includegraphics[width=\sizea]{figures/coco_mm/000000357888_17}  & \includegraphics[width=\sizea]{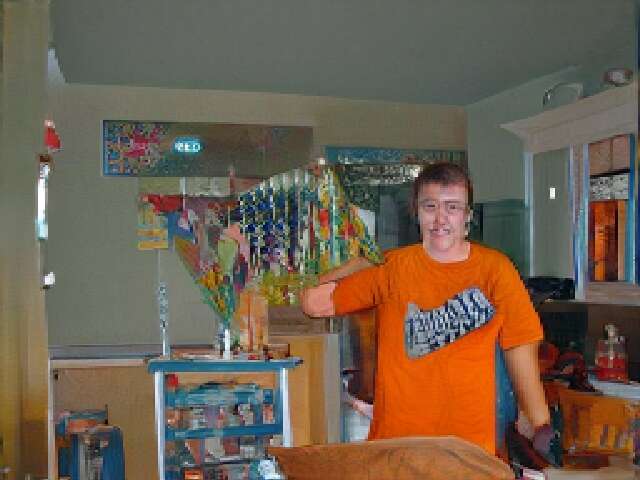} \\
		\includegraphics[width=\sizea]{figures/coco_mm/edge_000000357888} & \includegraphics[width=\sizea]{figures/coco_mm/000000005193} & \includegraphics[width=\sizea]{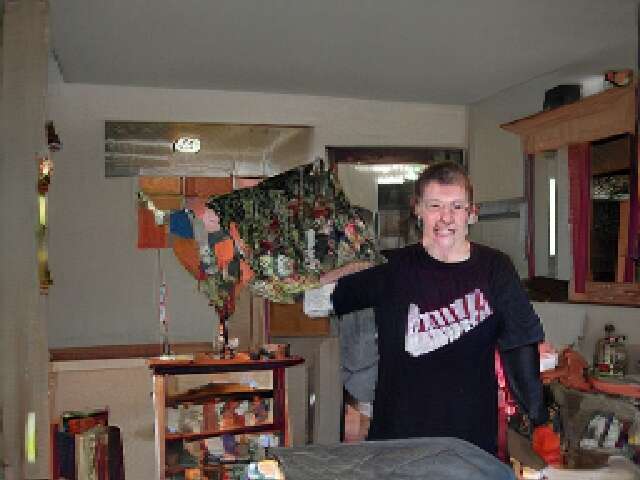}  & \includegraphics[width=\sizea]{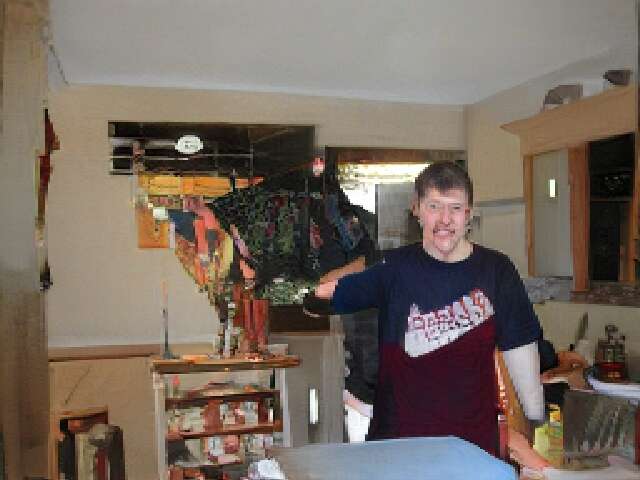}  & \includegraphics[width=\sizea]{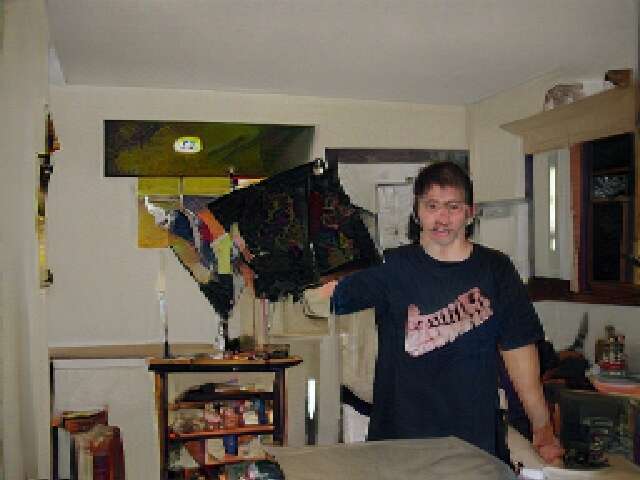}  & \includegraphics[width=\sizea]{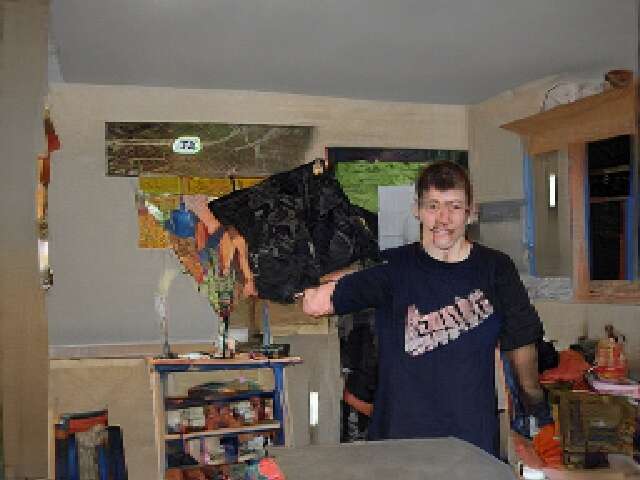} \\
	\includegraphics[width=\sizea]{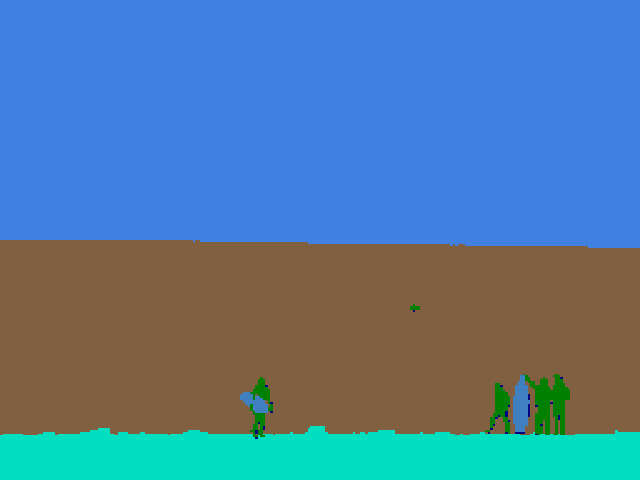} & \includegraphics[width=\sizea]{figures/coco_mm/000000515445} & \includegraphics[width=\sizea]{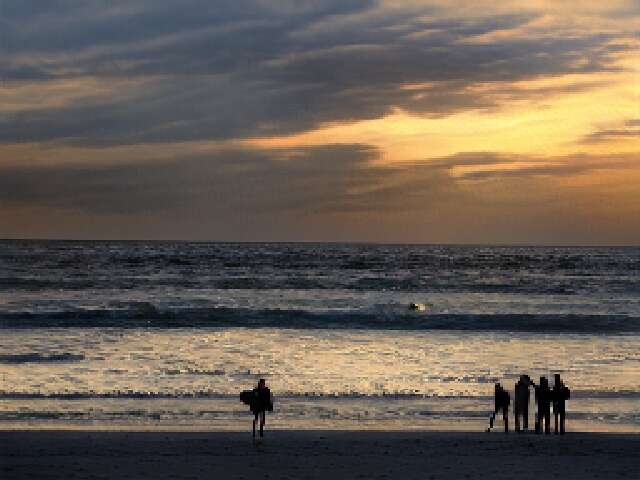}  & \includegraphics[width=\sizea]{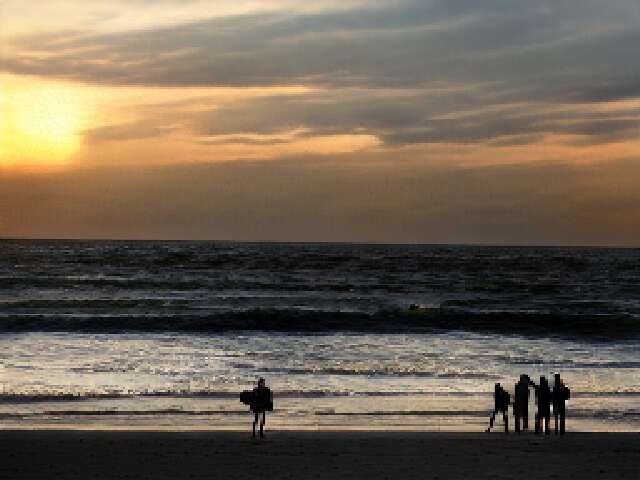}  & \includegraphics[width=\sizea]{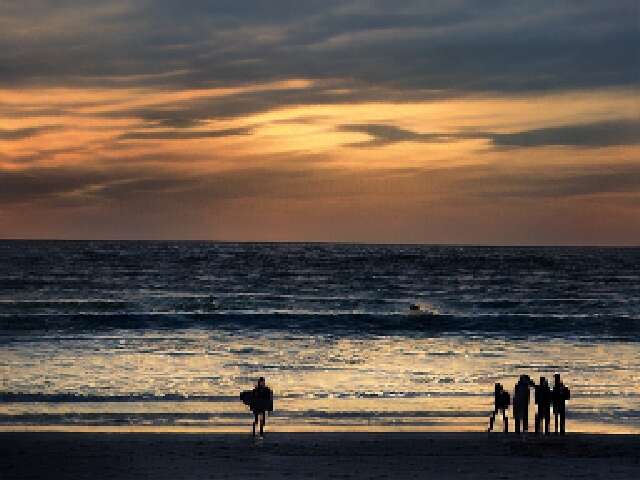}  & \includegraphics[width=\sizea]{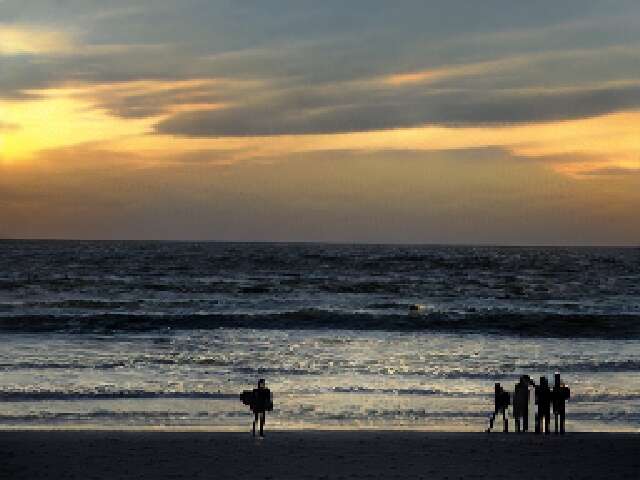} \\
	\includegraphics[width=\sizea]{figures/coco_mm/seg_000000515445} & \includegraphics[width=\sizea]{figures/coco_mm/000000007511}  & \includegraphics[width=\sizea]{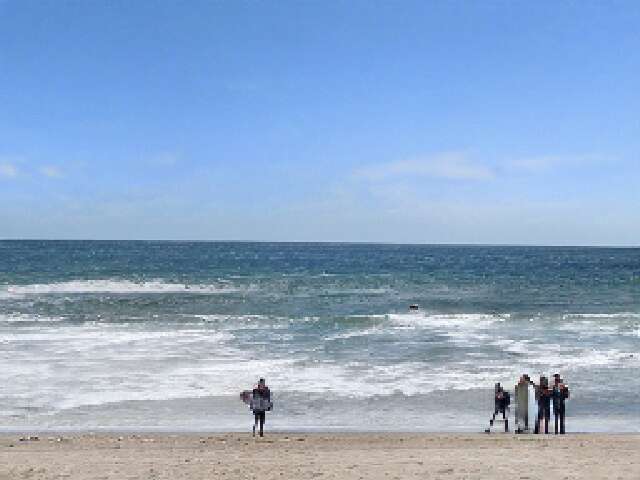}  & \includegraphics[width=\sizea]{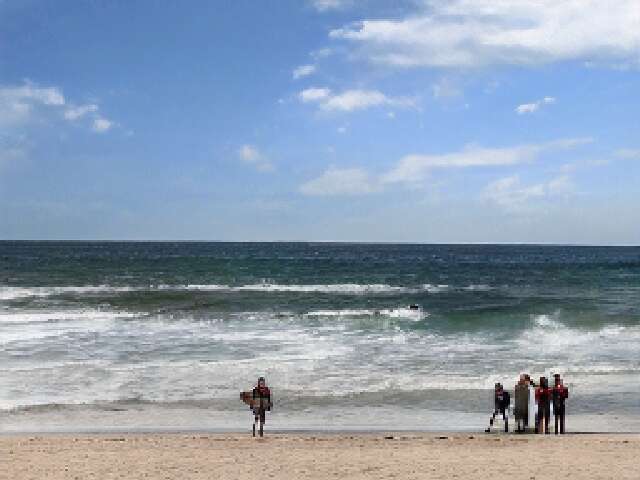}  & \includegraphics[width=\sizea]{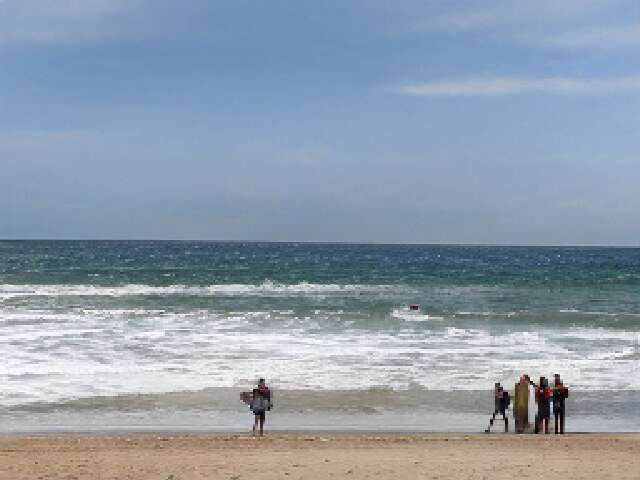}  & \includegraphics[width=\sizea]{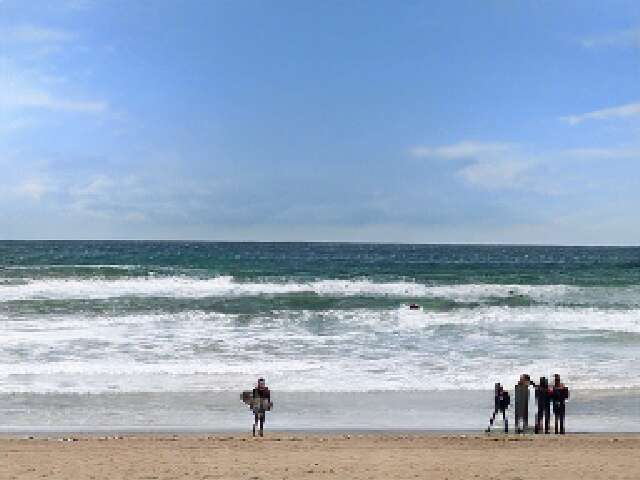} \\
	\begin{minipage}[b]{\sizea}
		\small
		\centering A person on a surf board riding a wave. \vspace{18pt} 
	\end{minipage}     
	& \includegraphics[width=\sizea]{figures/coco_mm/000000441286}    & \includegraphics[width=\sizea]{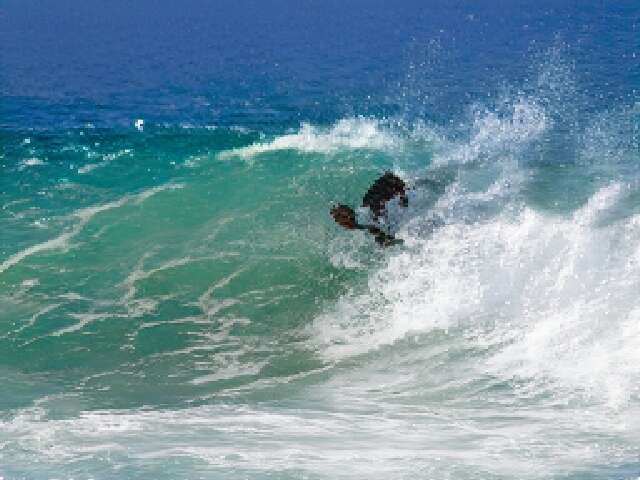}  & \includegraphics[width=\sizea]{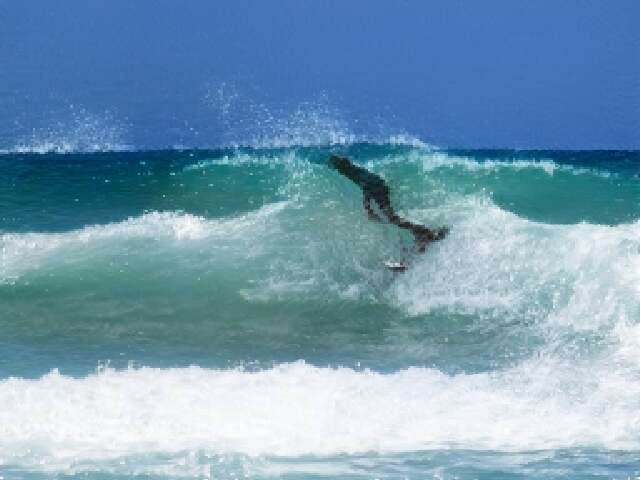}  & \includegraphics[width=\sizea]{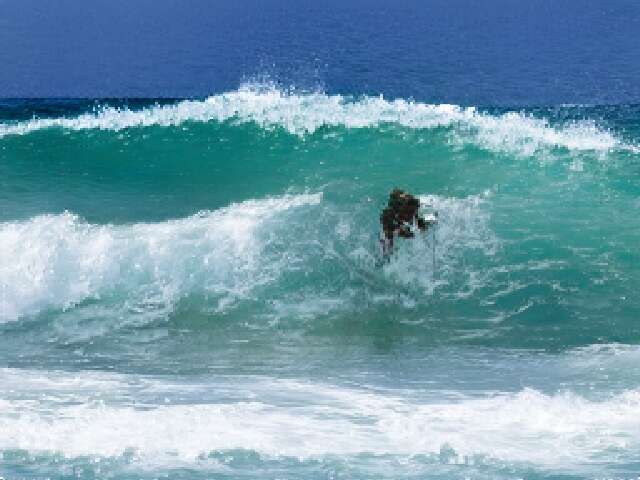}  & \includegraphics[width=\sizea]{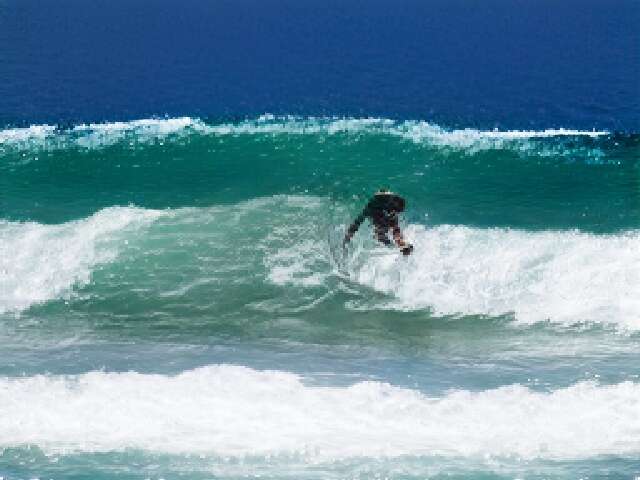} \\
	\begin{minipage}[b]{\sizea}
		\small
		\centering A person on a surf board riding a wave. \vspace{18pt} 
	\end{minipage}     
	& \includegraphics[width=\sizea]{figures/coco_mm/000000006460}    & \includegraphics[width=\sizea]{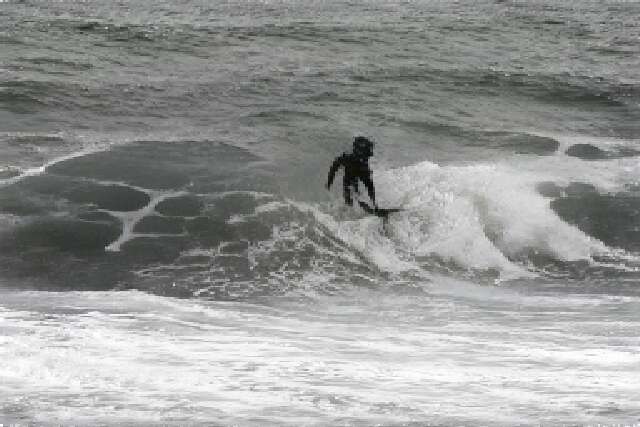}  & \includegraphics[width=\sizea]{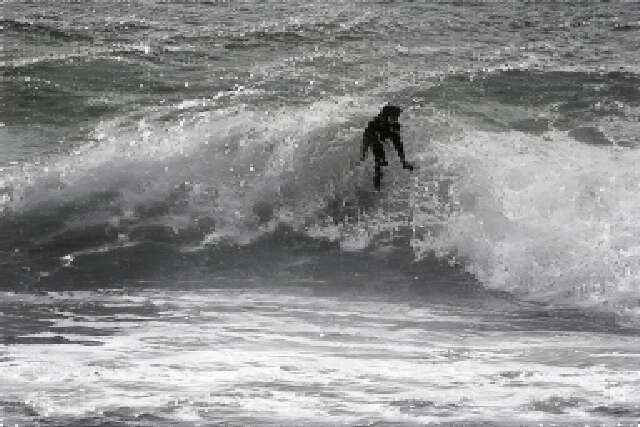}  & \includegraphics[width=\sizea]{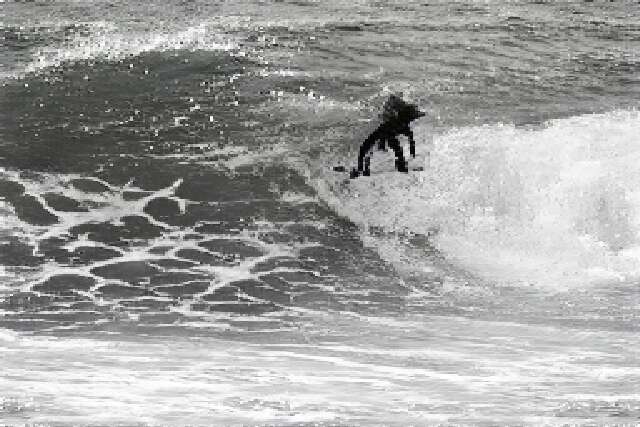}  & \includegraphics[width=\sizea]{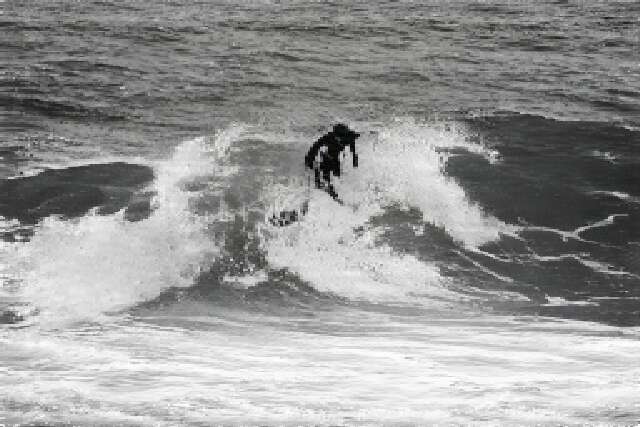} 
\end{tabular}

\vspace{-2mm}
\caption{Examples of multimodal conditional image synthesis on MS-COCO. We show three random samples from PoE-GAN conditioned on two modalities, one being text/segmentation/sketch and another being style reference.}
\label{fig:coco_mm_2_style}
\vspace{-2mm}
\end{figure*}

\begin{figure*}
	\centering
	\newcommand{\sizea}{0.19\linewidth}
	\setlength{\tabcolsep}{2pt}
	\begin{tabular}{cc|ccc}
		Condition \#1 & Condition \#2 &   \multicolumn{3}{c}{Samples}  \\
		\begin{minipage}[b]{\sizea}
			\centering Trees near a lake \\in an autumn rainy day.\vspace{35pt} 
		\end{minipage}     
		& \includegraphics[width=\sizea,height=\sizea]{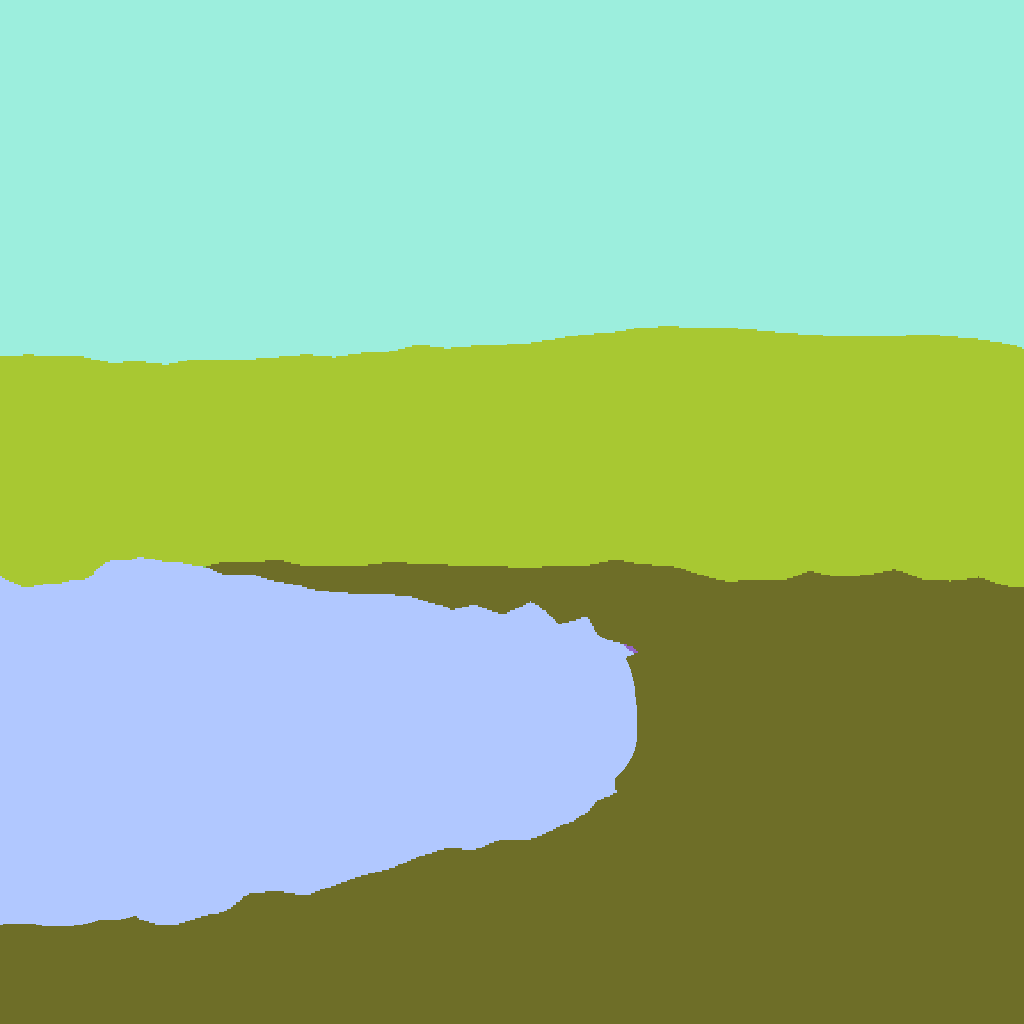}    & \includegraphics[width=\sizea,height=\sizea]{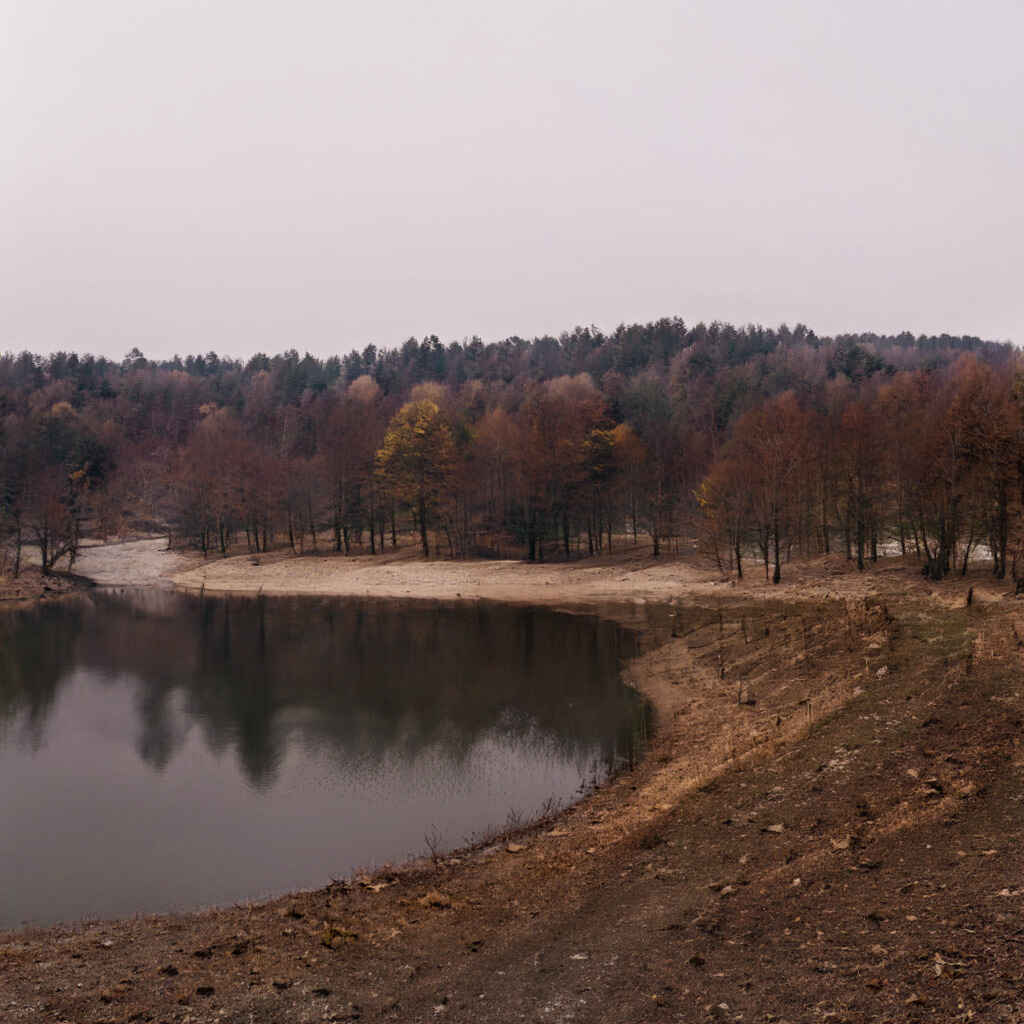}  & \includegraphics[width=\sizea,height=\sizea]{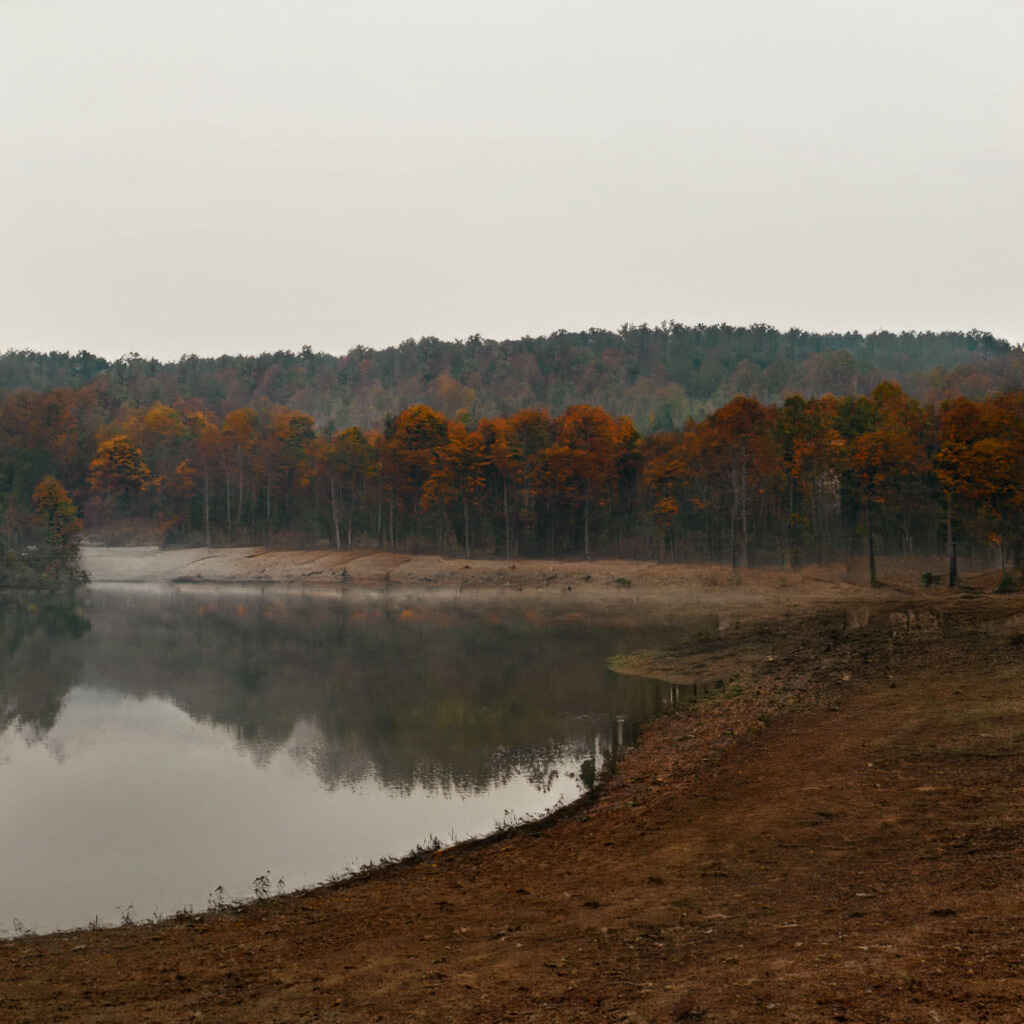}  & \includegraphics[width=\sizea,height=\sizea]{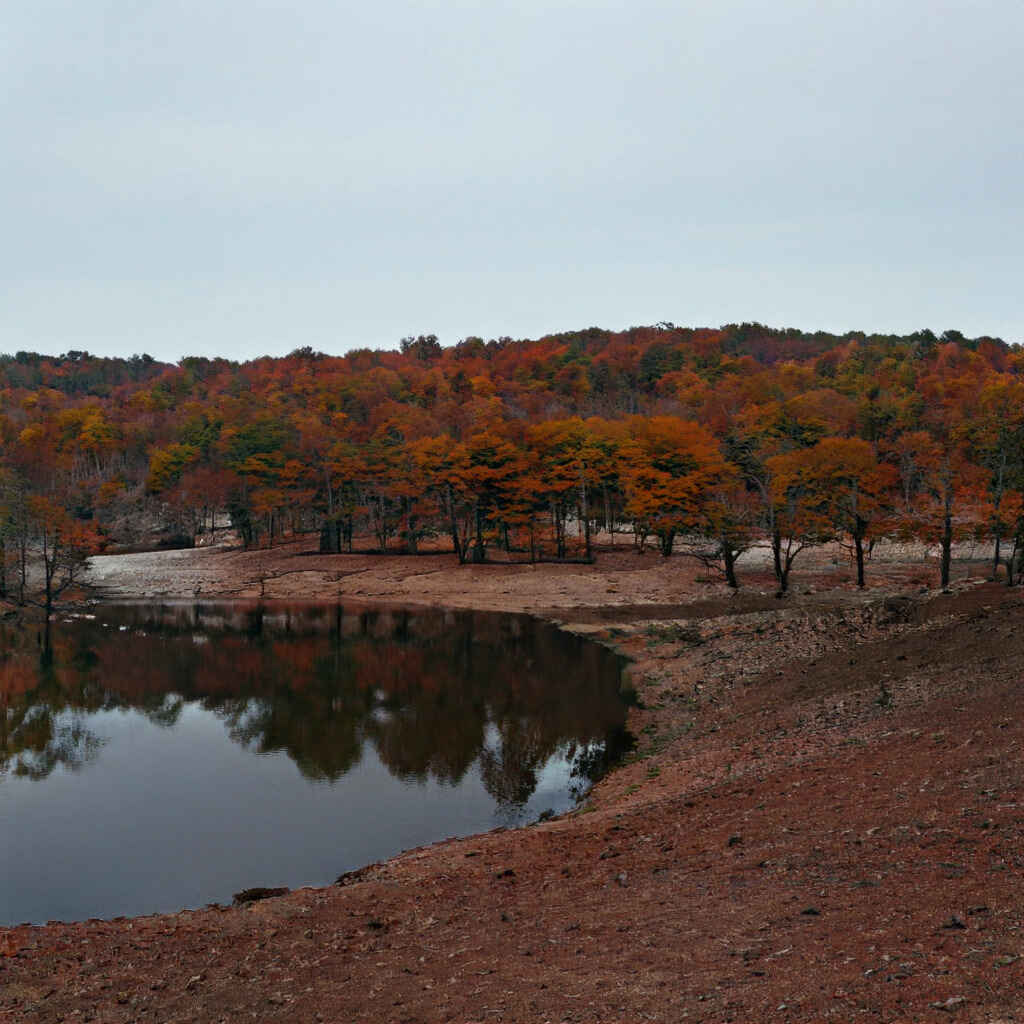} \\
		\begin{minipage}[b]{\sizea}
			\centering Huge ocean waves crash into rocks in a day with colorful clouds.\vspace{34pt} 
		\end{minipage}     
		& \includegraphics[width=\sizea,height=\sizea]{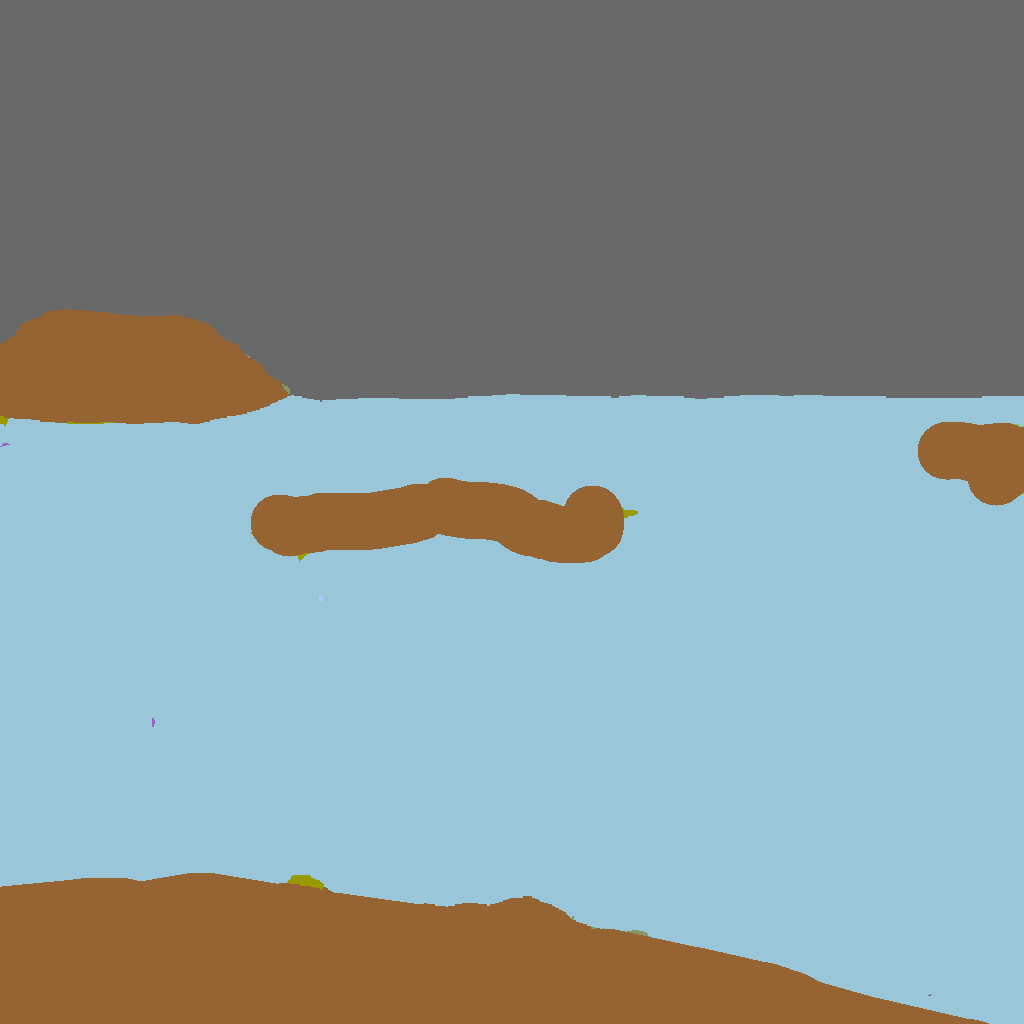}    & \includegraphics[width=\sizea,height=\sizea]{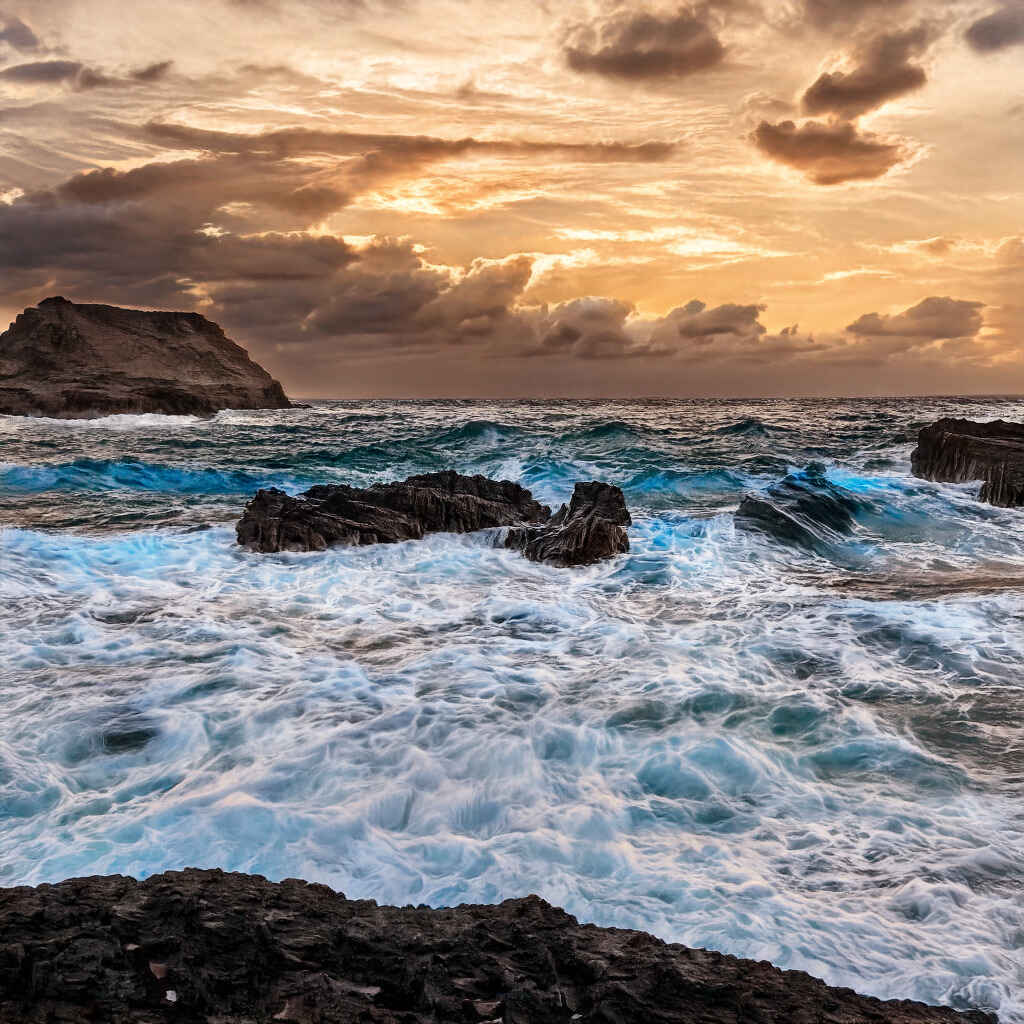}  & \includegraphics[width=\sizea,height=\sizea]{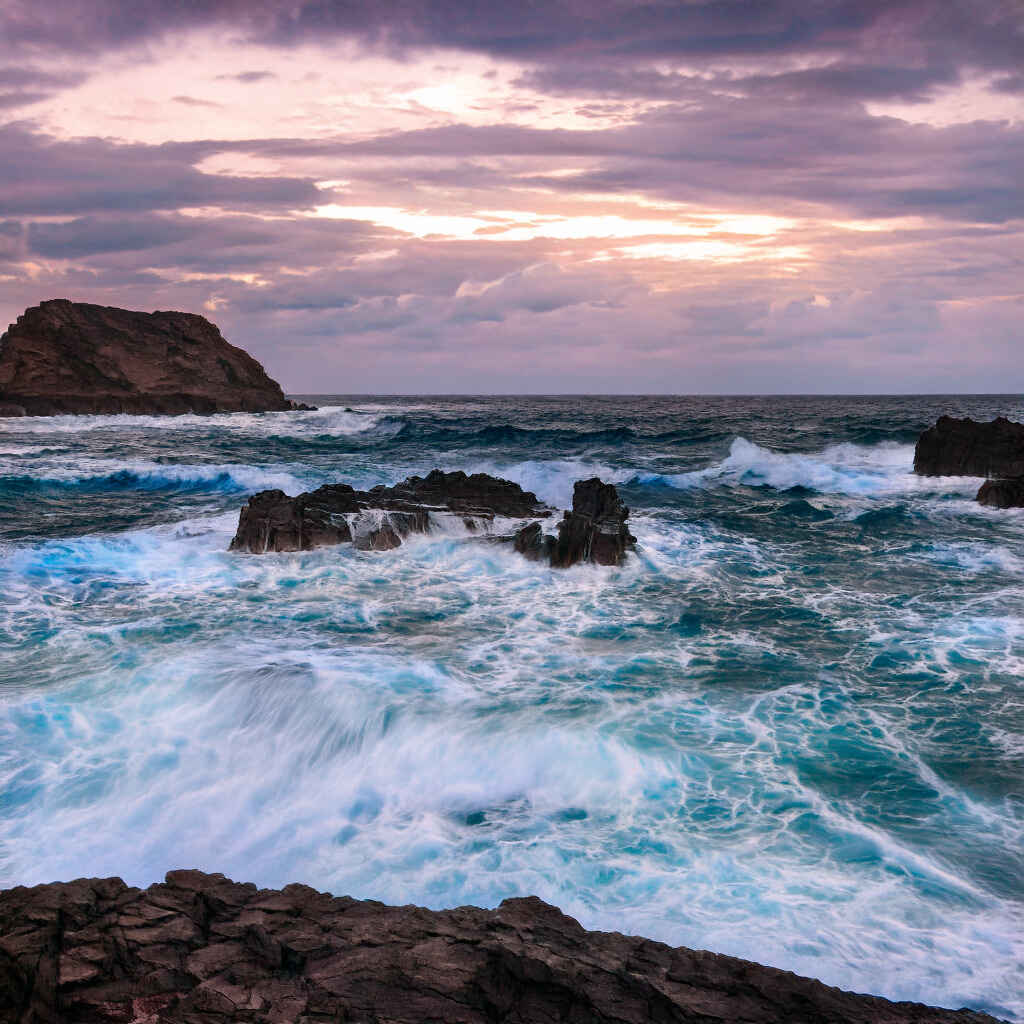}  & \includegraphics[width=\sizea,height=\sizea]{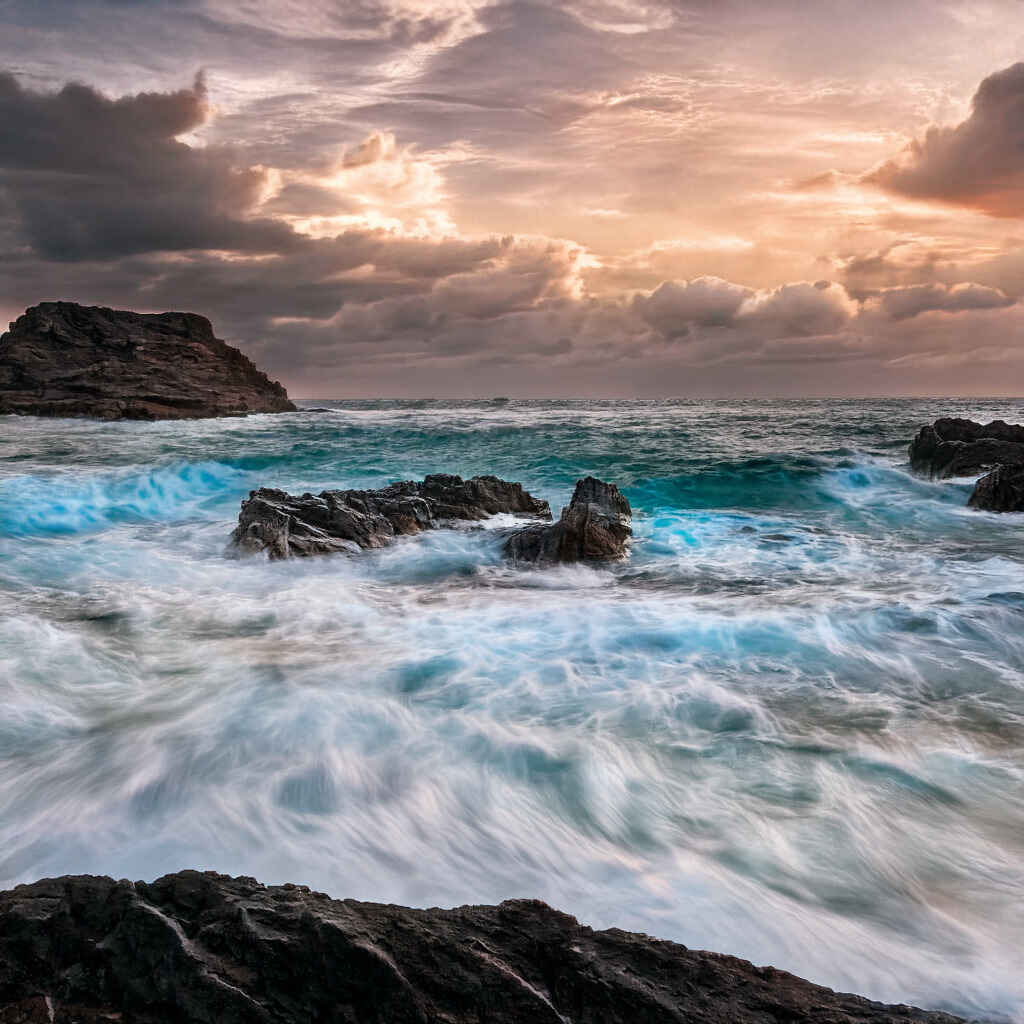} \\
		\begin{minipage}[b]{\sizea}
			\centering Waterfall and river between mountains. \vspace{35pt} 
		\end{minipage}     
		& \includegraphics[width=\sizea,height=\sizea]{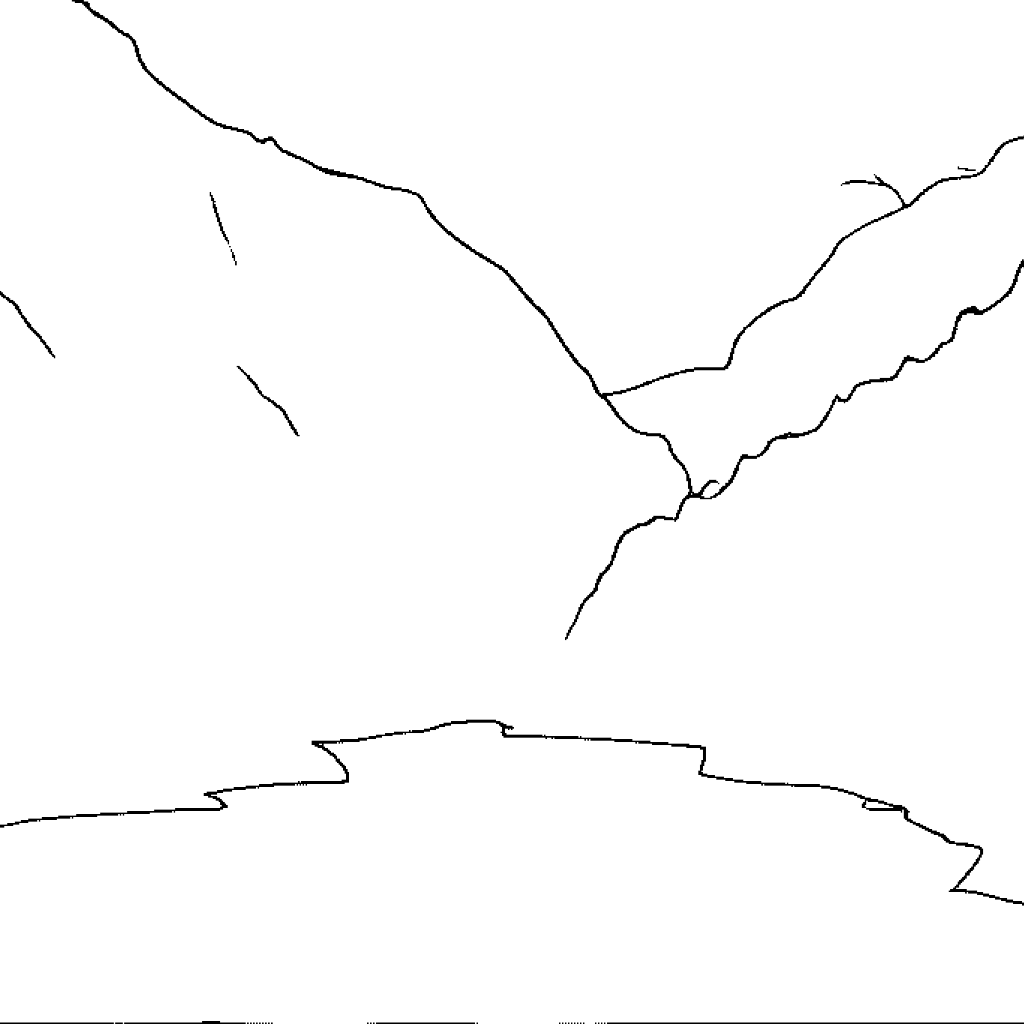}    & \includegraphics[width=\sizea,height=\sizea]{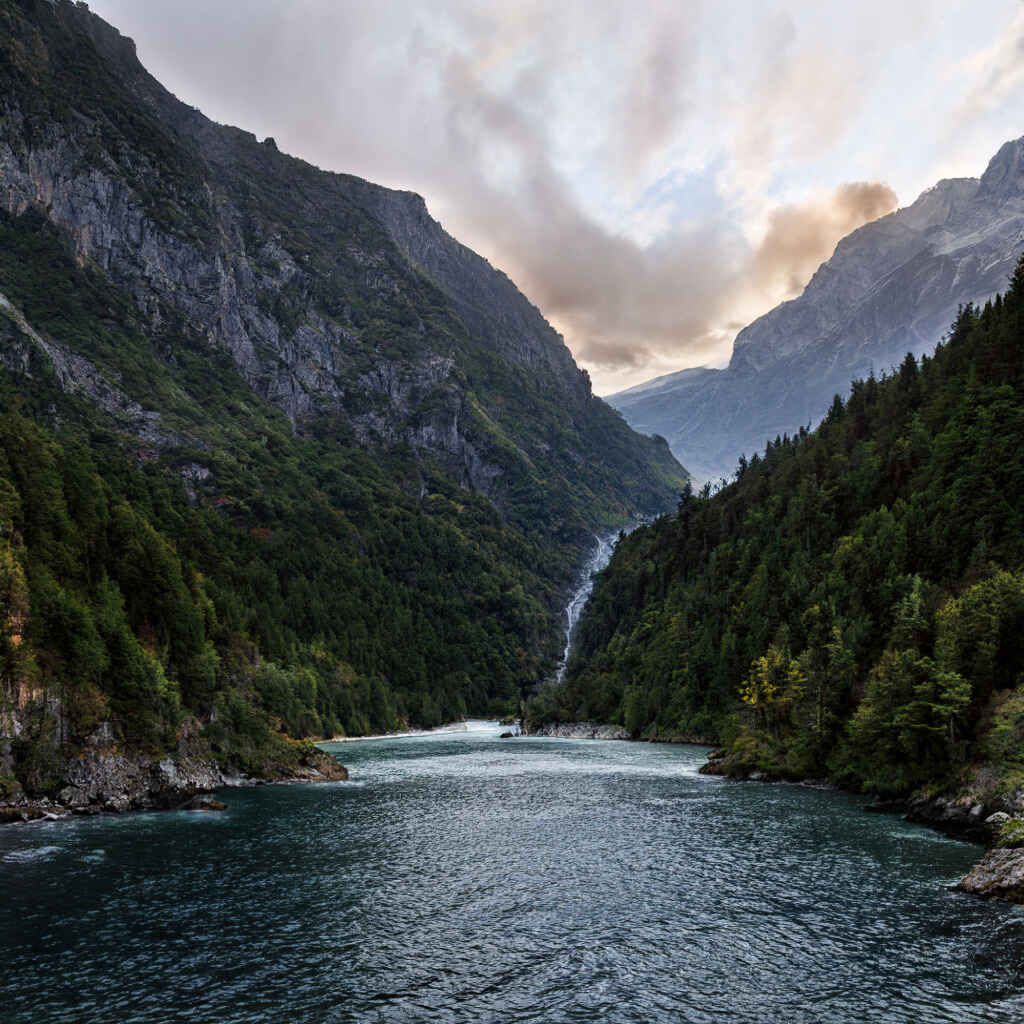}  & \includegraphics[width=\sizea,height=\sizea]{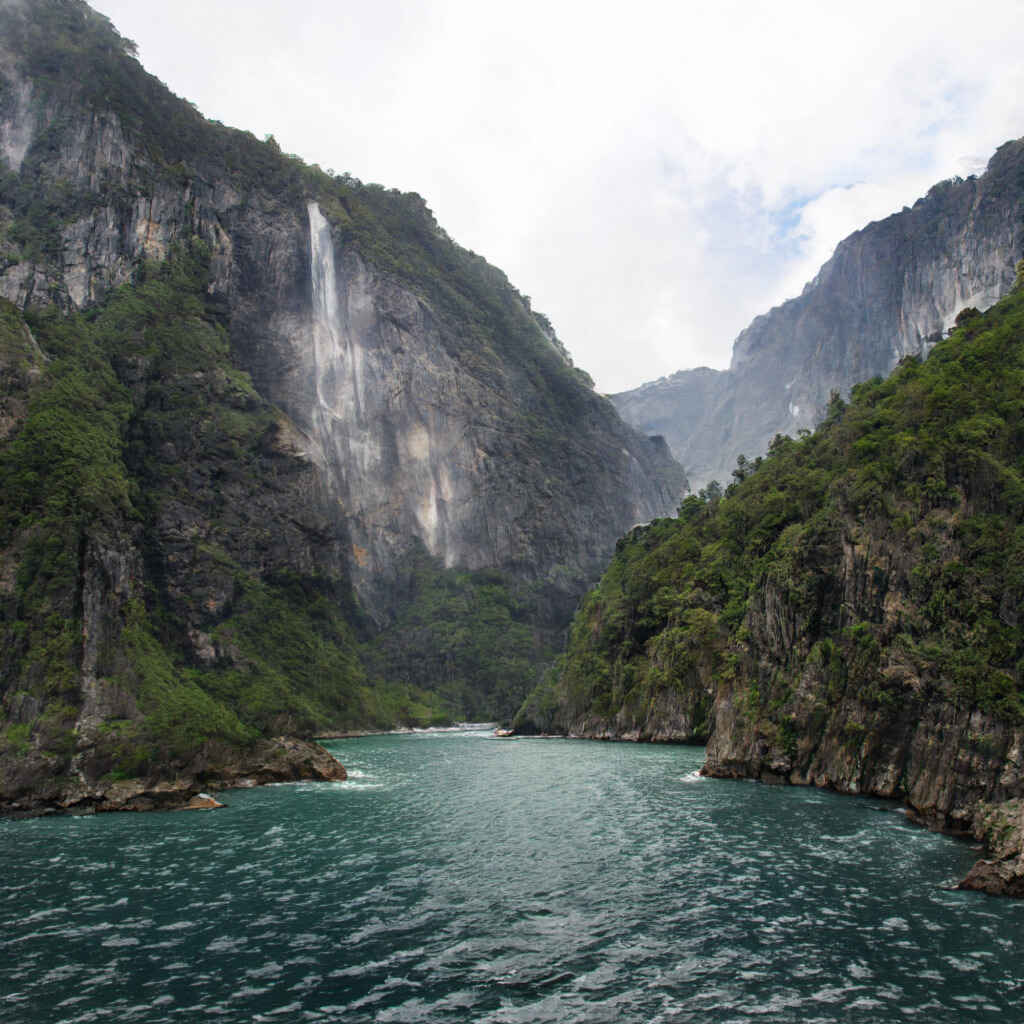}  & \includegraphics[width=\sizea,height=\sizea]{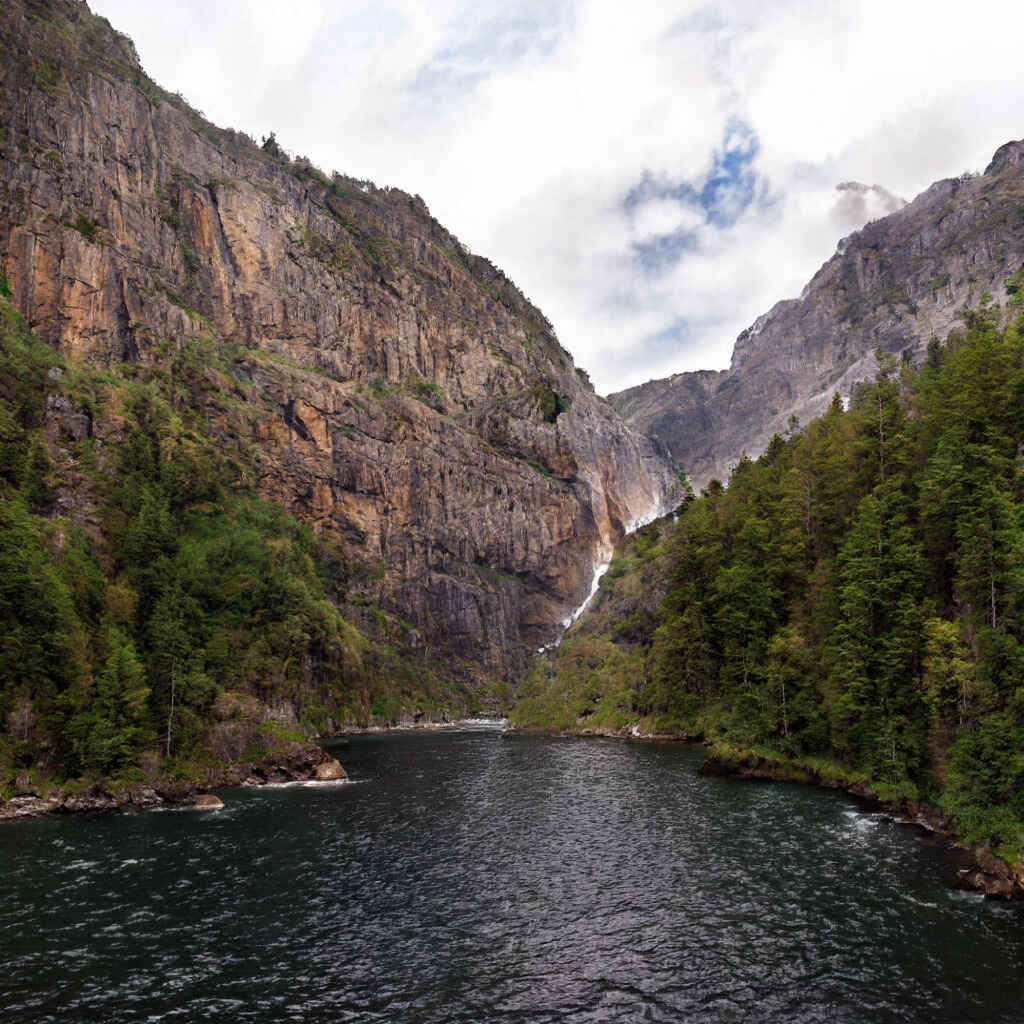} \\
		\begin{minipage}[b]{\sizea}
			\centering A white sand beach near cyan ocean.\vspace{35pt} 
		\end{minipage}     
		& \includegraphics[width=\sizea,height=\sizea]{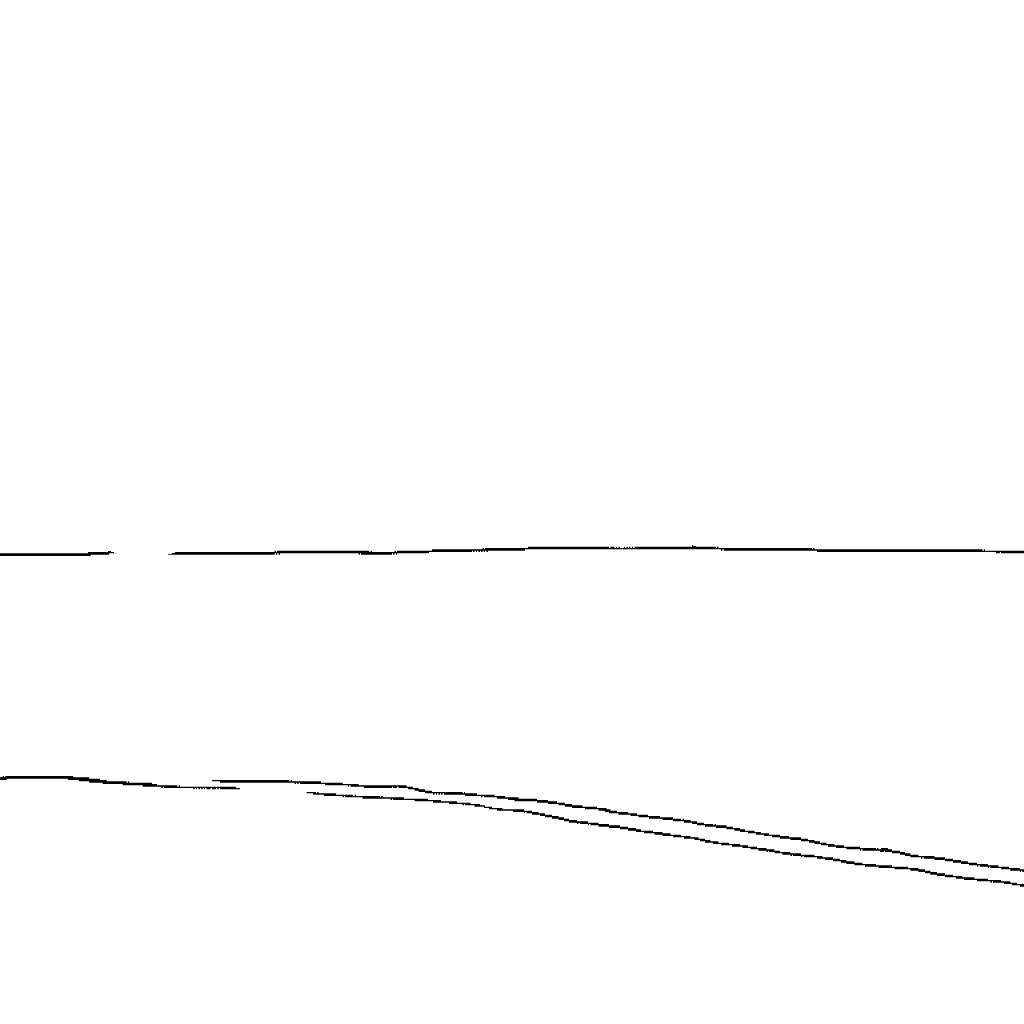}    & \includegraphics[width=\sizea,height=\sizea]{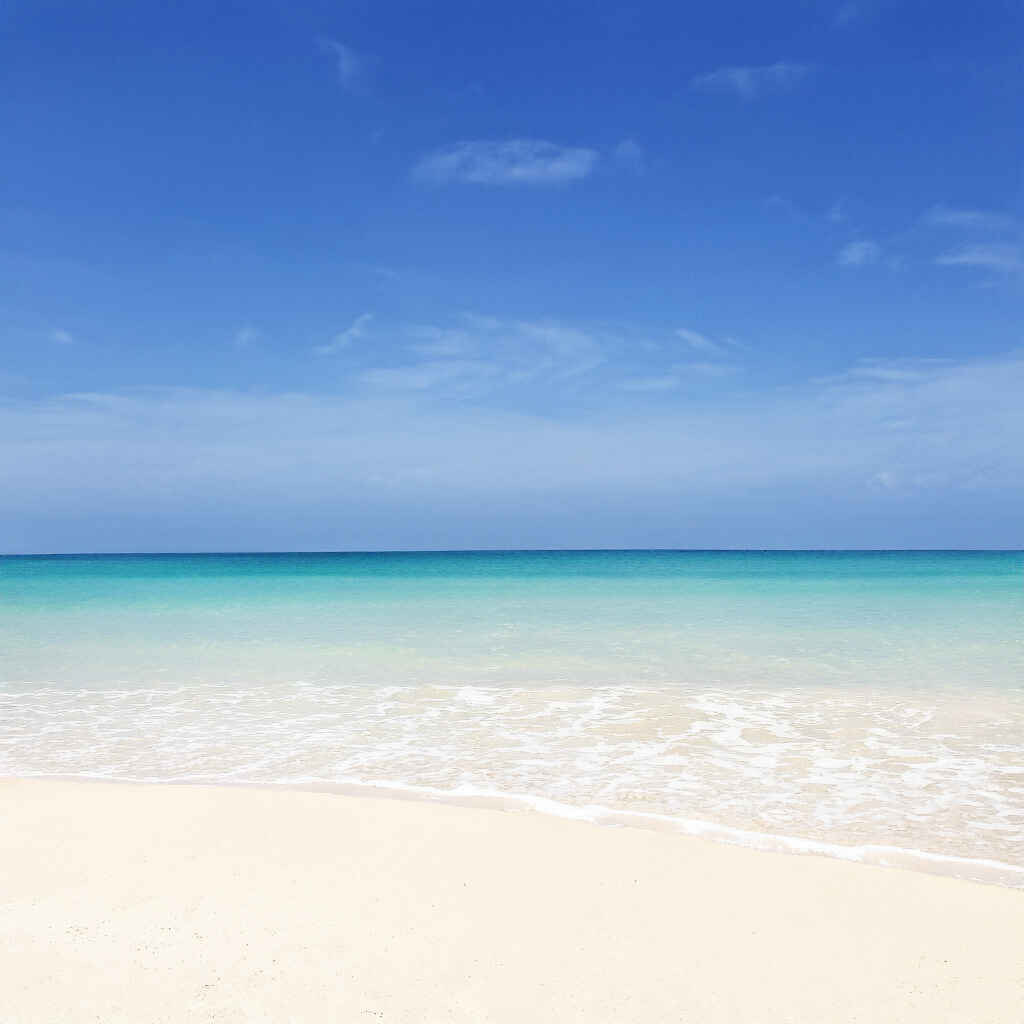}  & \includegraphics[width=\sizea,height=\sizea]{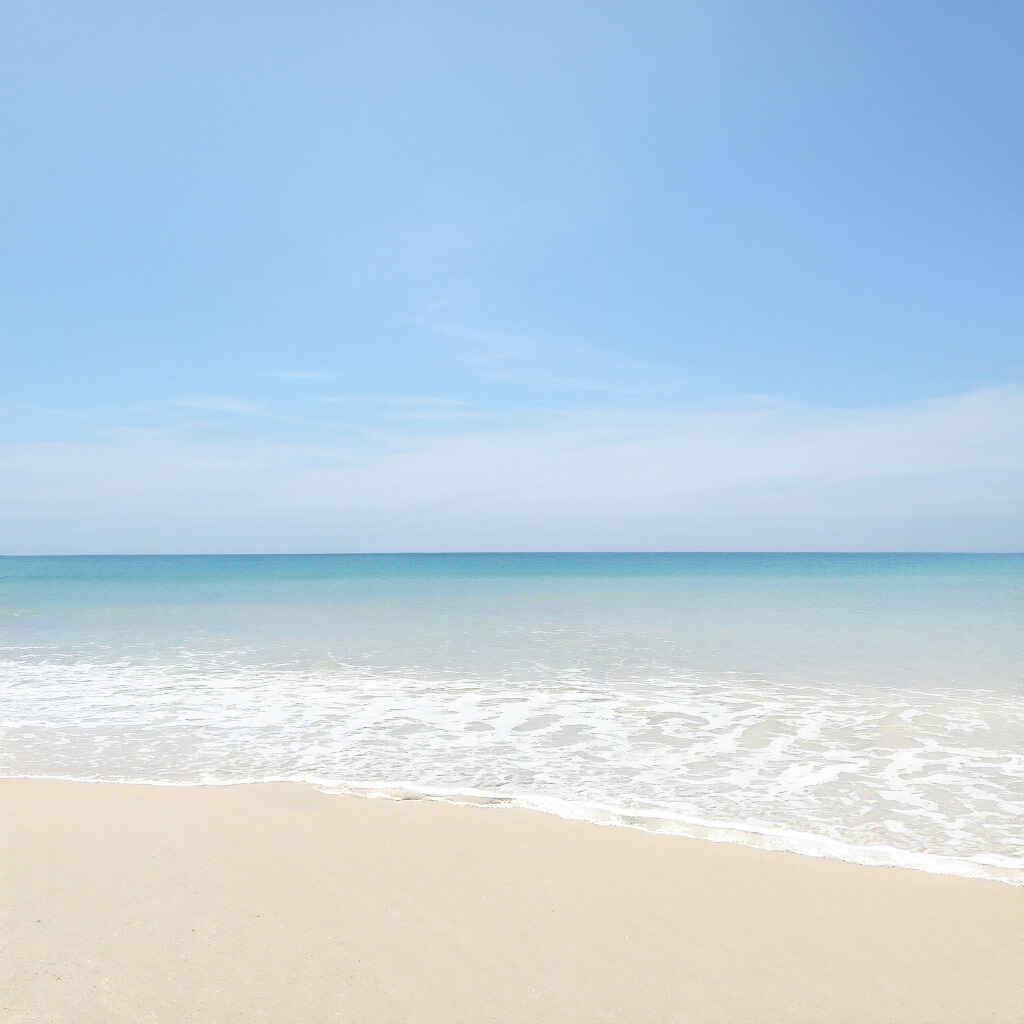}  & \includegraphics[width=\sizea,height=\sizea]{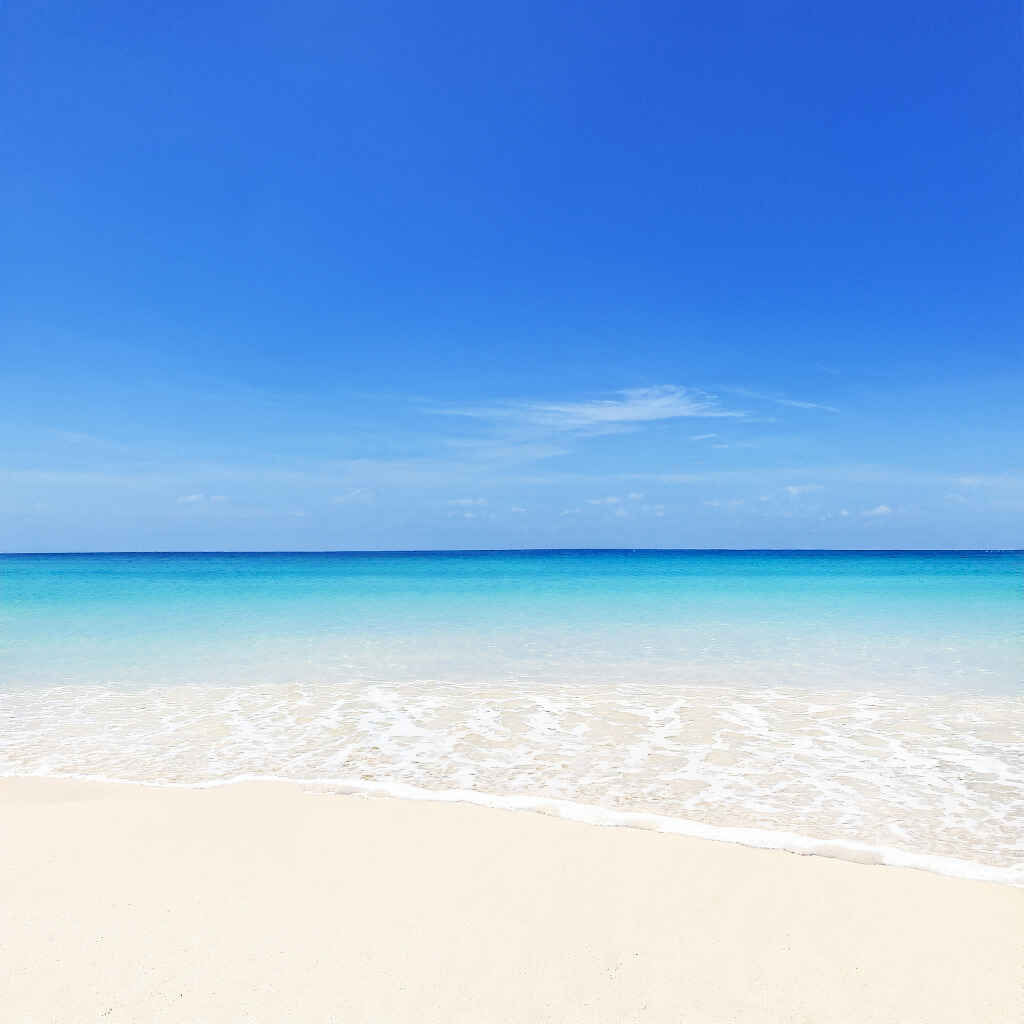} \\
		\includegraphics[width=\sizea,height=\sizea]{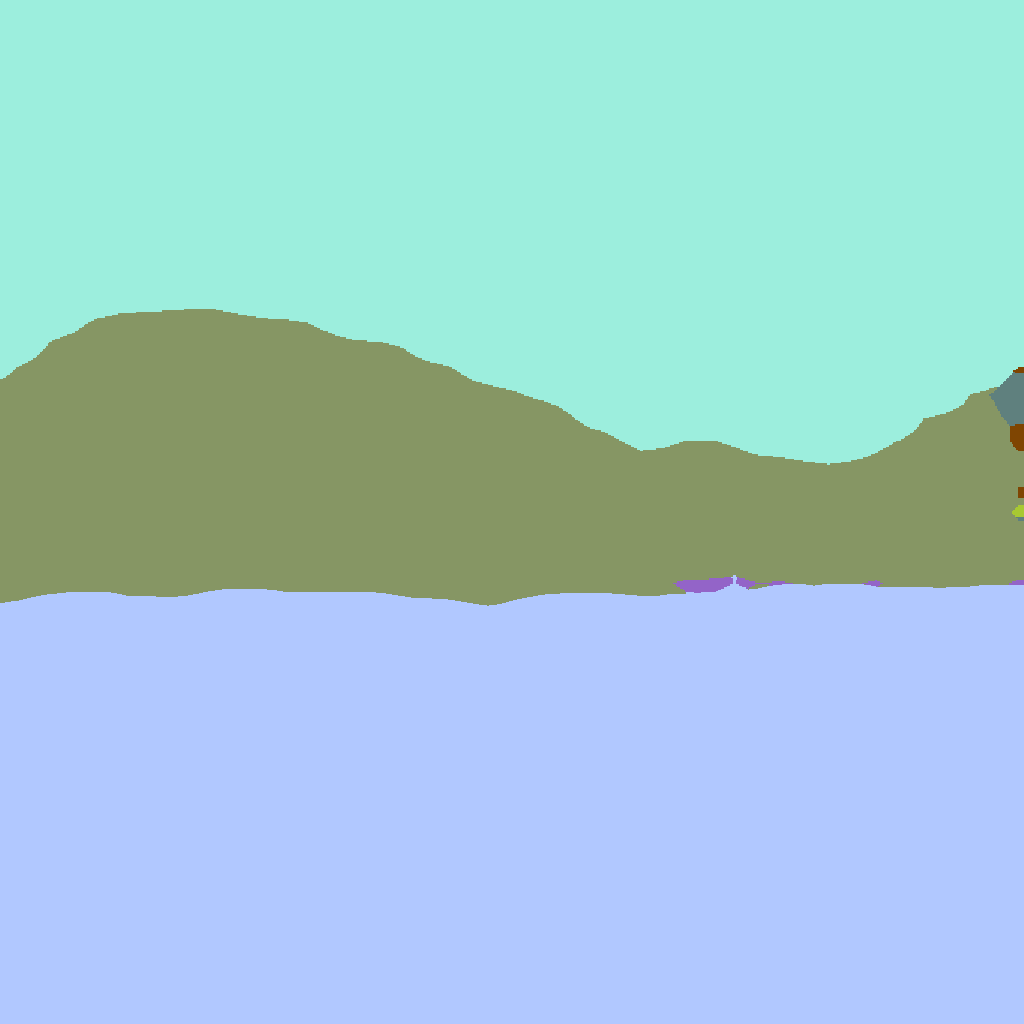}    &
		\includegraphics[width=\sizea,height=\sizea]{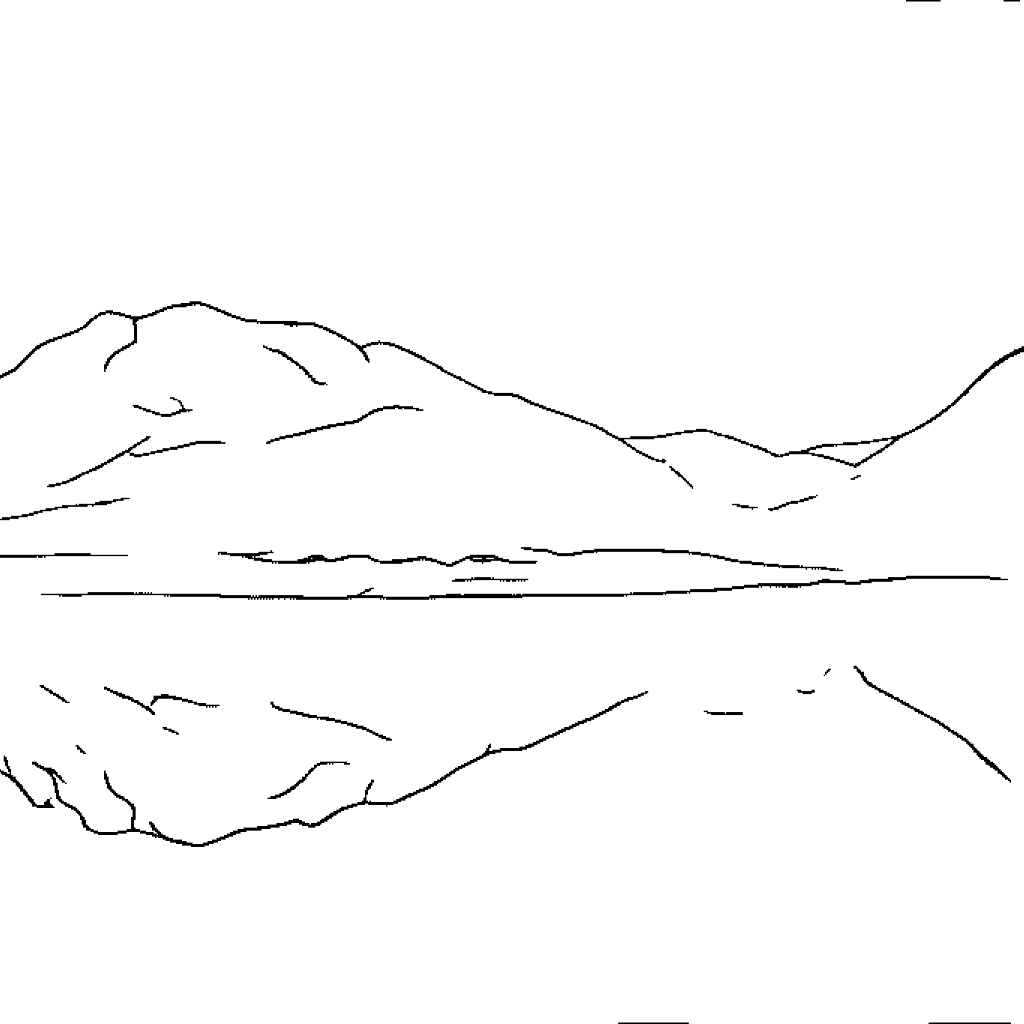}    & \includegraphics[width=\sizea,height=\sizea]{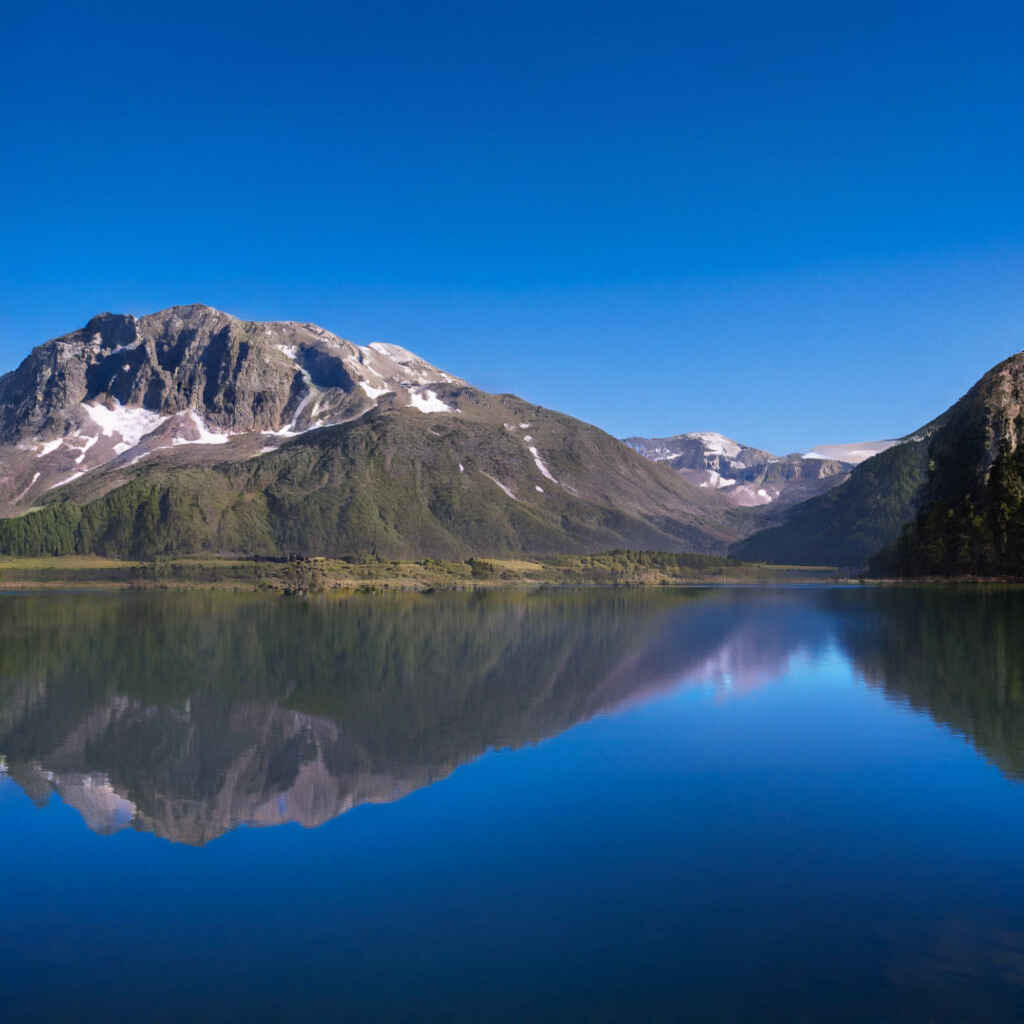}  & \includegraphics[width=\sizea,height=\sizea]{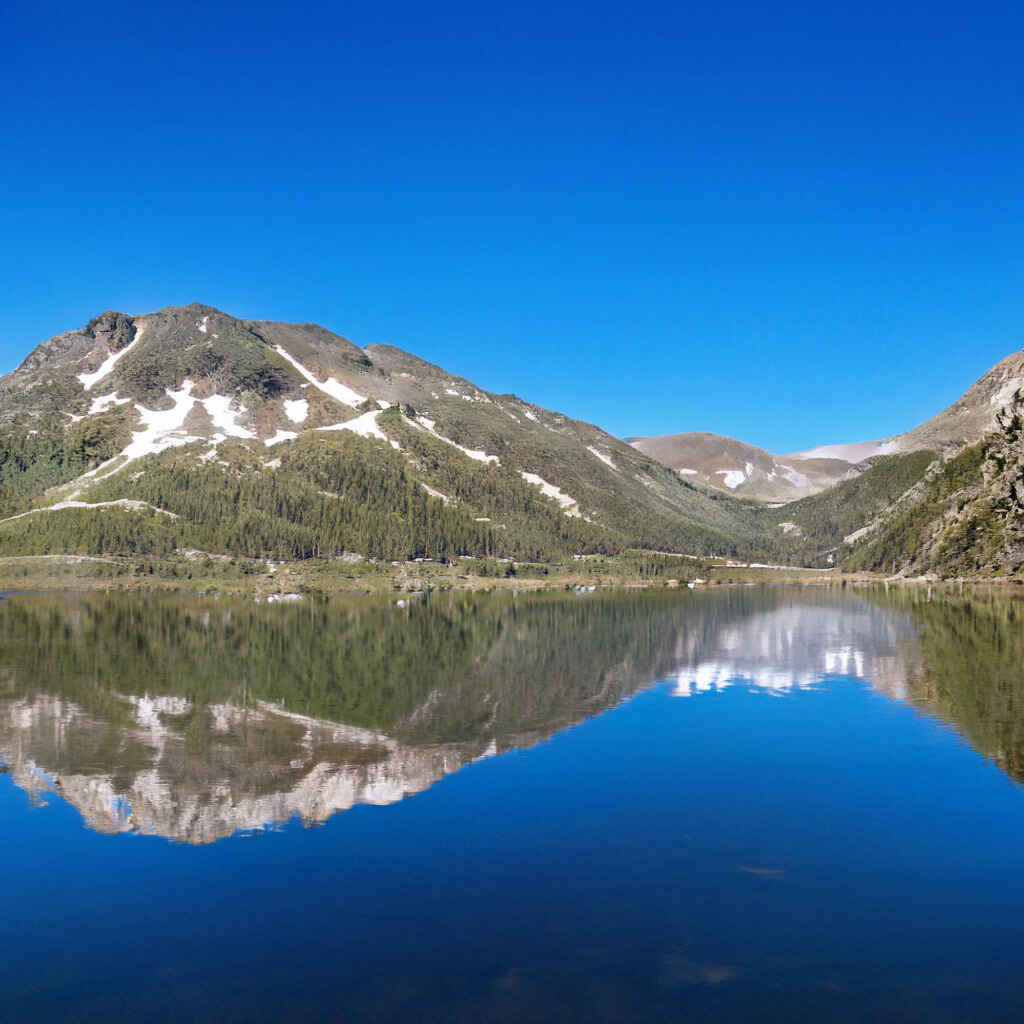}  & \includegraphics[width=\sizea,height=\sizea]{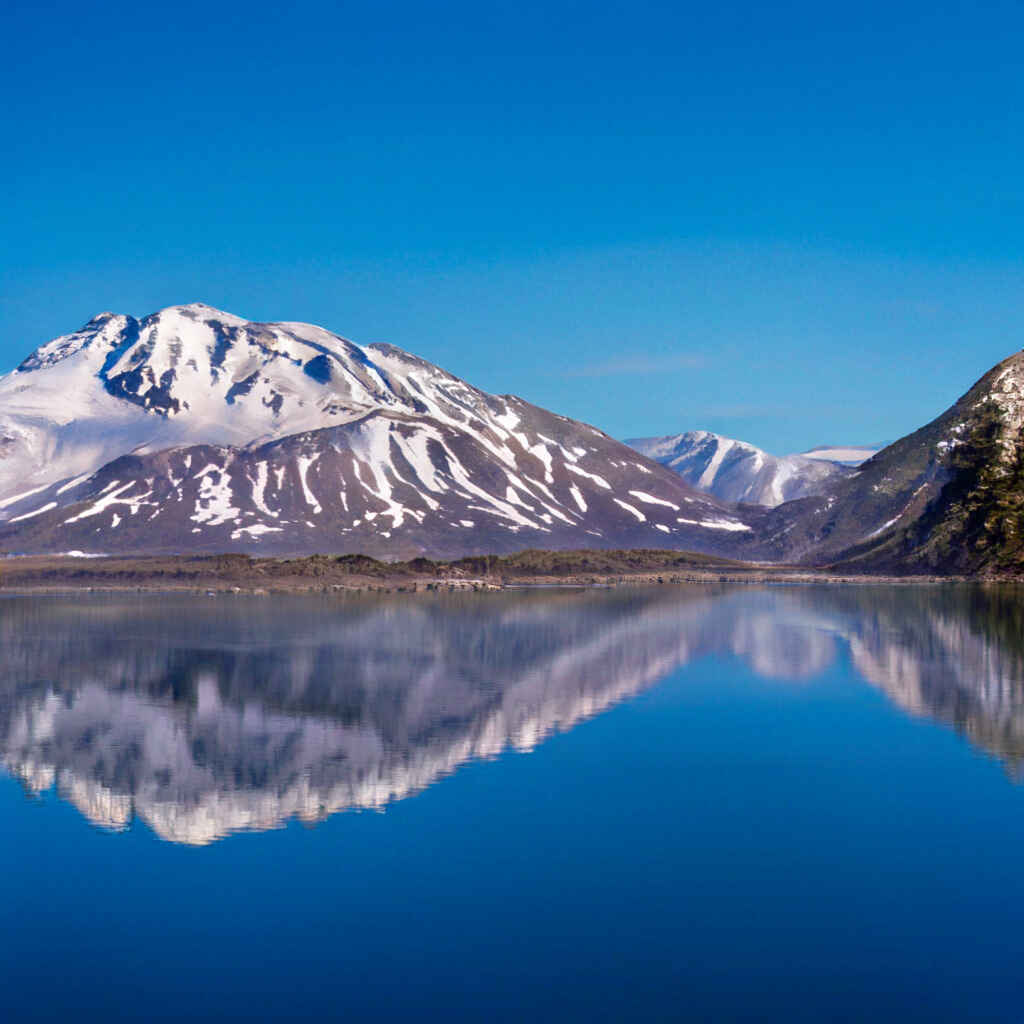} \\
		\includegraphics[width=\sizea,height=\sizea]{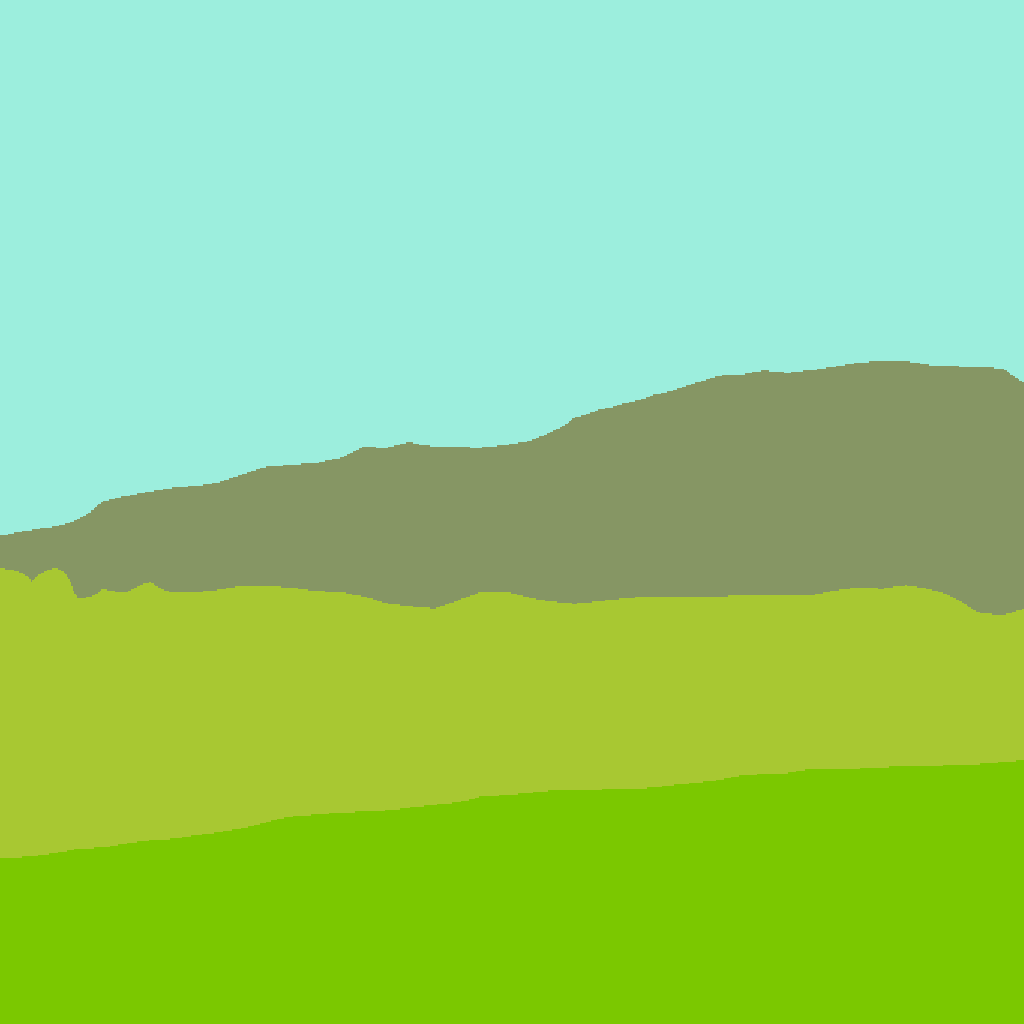}    &
		\includegraphics[width=\sizea,height=\sizea]{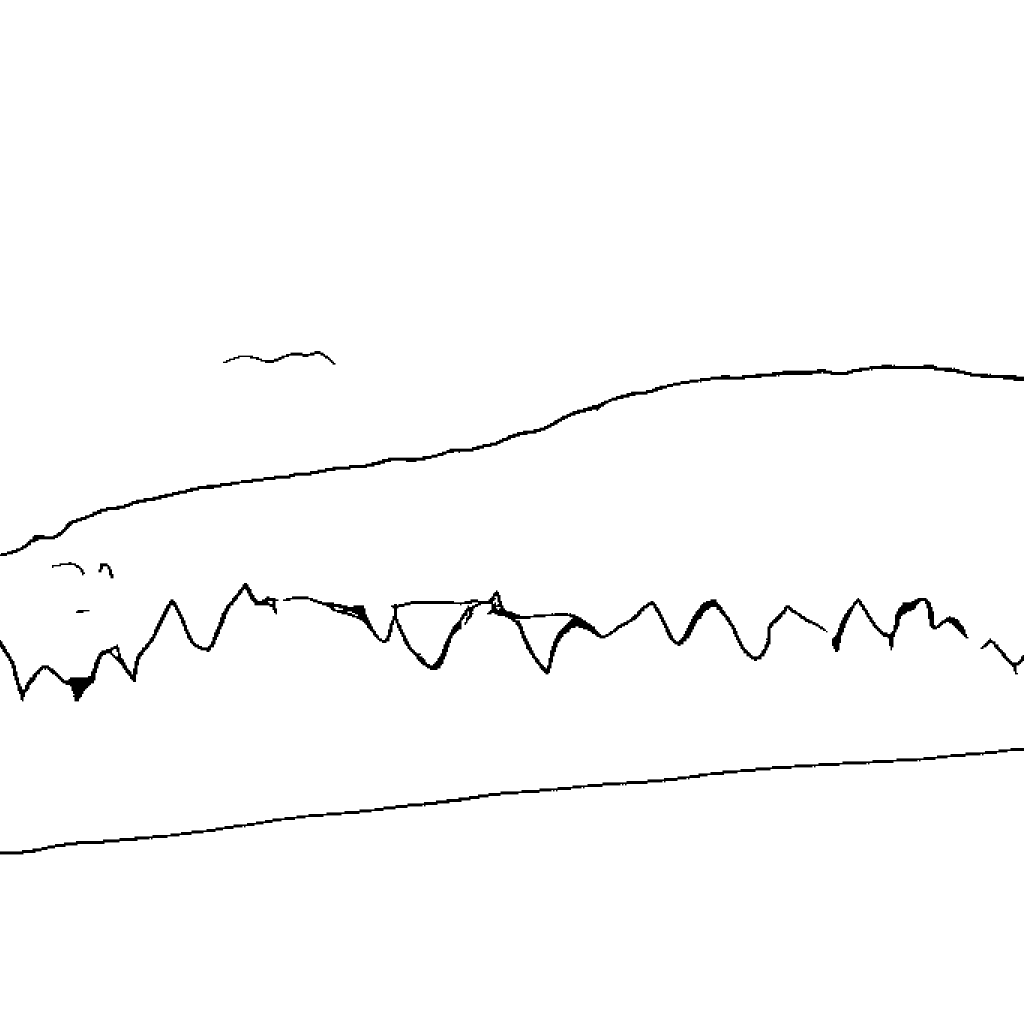}    & \includegraphics[width=\sizea,height=\sizea]{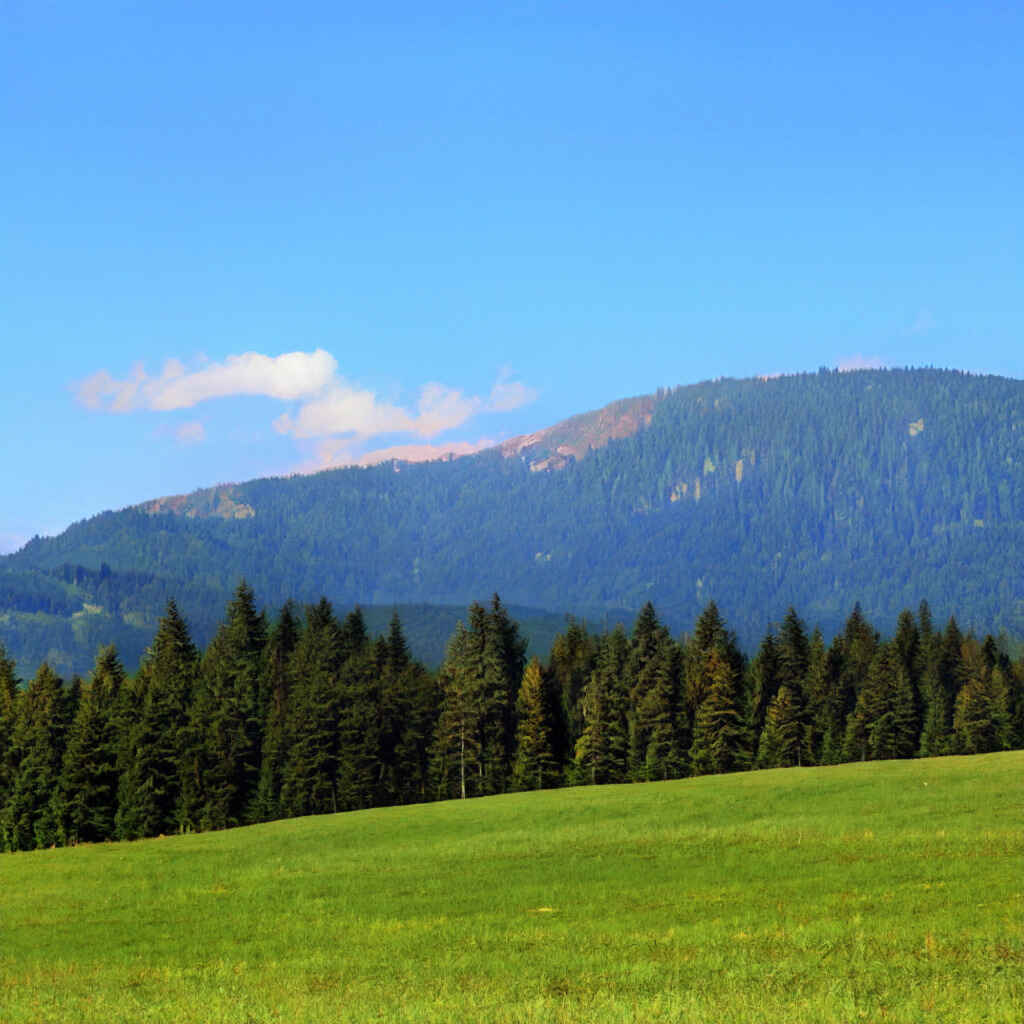}  & \includegraphics[width=\sizea,height=\sizea]{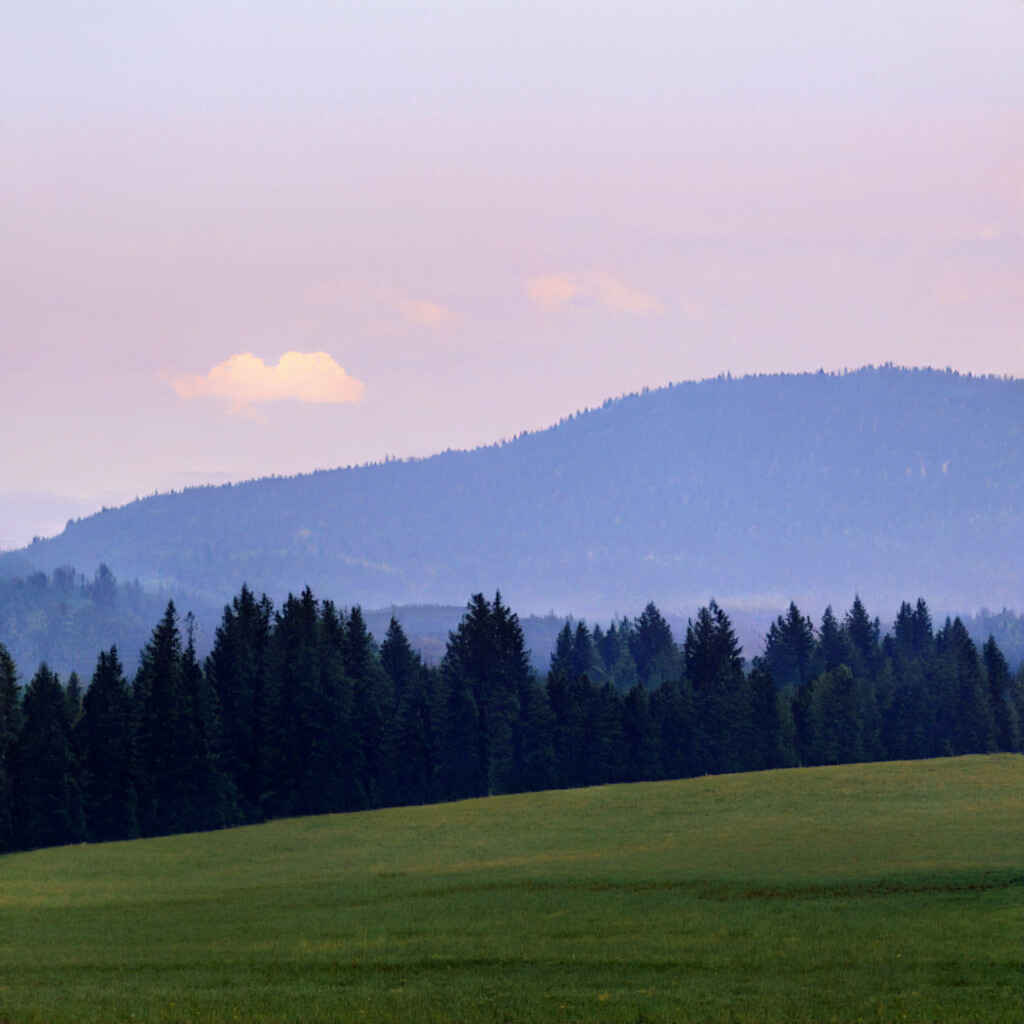}  & \includegraphics[width=\sizea,height=\sizea]{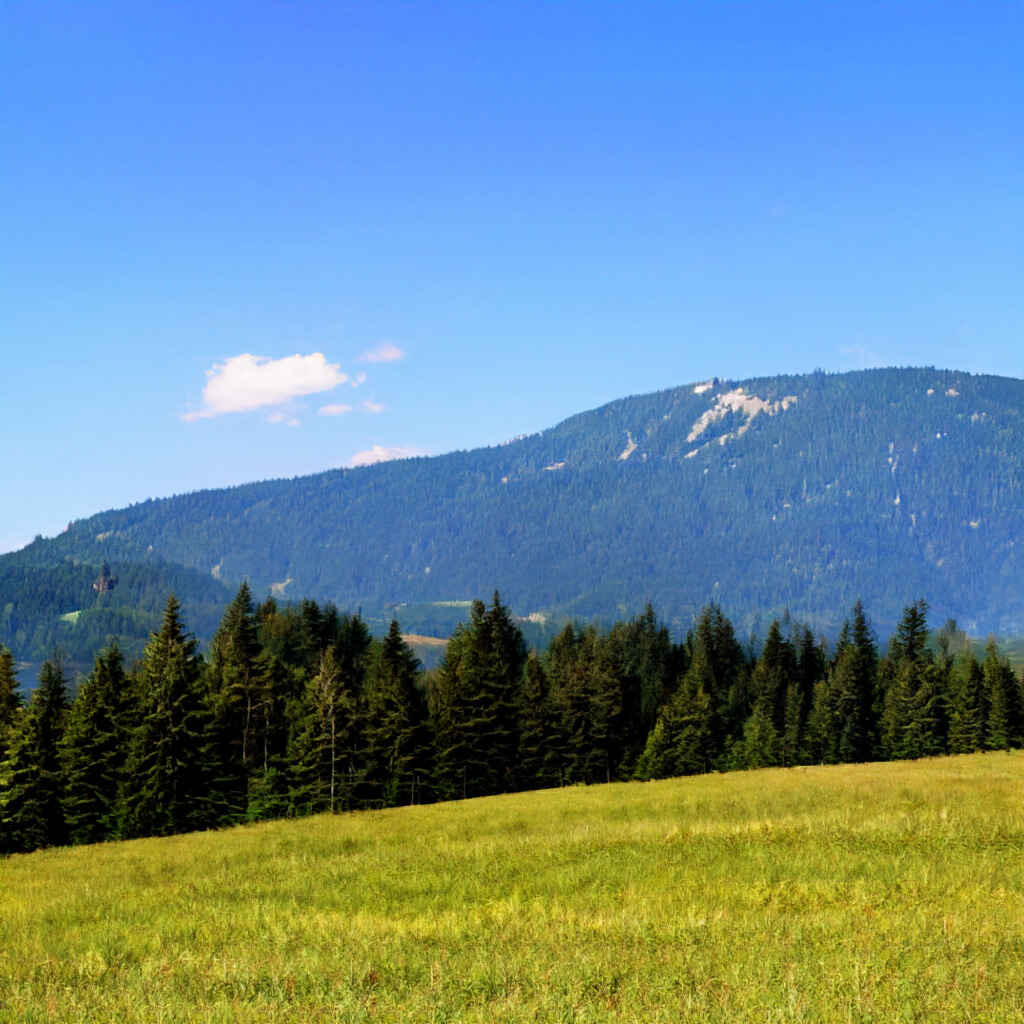} \\
	\end{tabular}
	\vspace{-2mm}
	\caption{Examples of multimodal conditional image synthesis on Landscape. We show three random samples from PoE-GAN conditioned on two modalities~(from top to bottom: text + segmentation, text + sketch, and segmentation + sketch).}
	\vspace{-2mm}
	\label{fig:getty_mm}
\end{figure*}
\begin{figure*}
\centering
\newcommand{\sizea}{0.19\linewidth}
\setlength{\tabcolsep}{2pt}
\begin{tabular}{cc|ccc}
	Condition \#1 & Condition \#2~(Style) &  & Samples &  \\ \includegraphics[width=\sizea,height=\sizea]{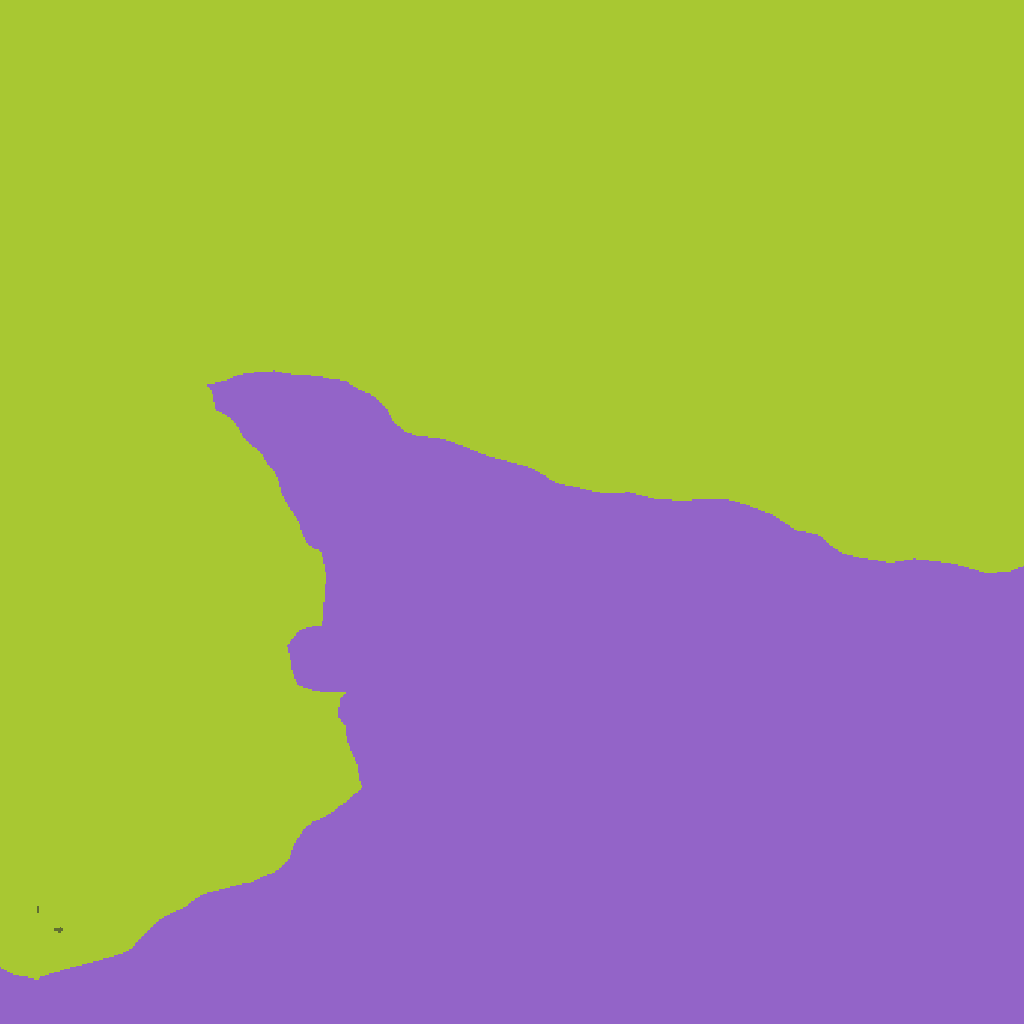} &
	\includegraphics[width=\sizea,height=\sizea]{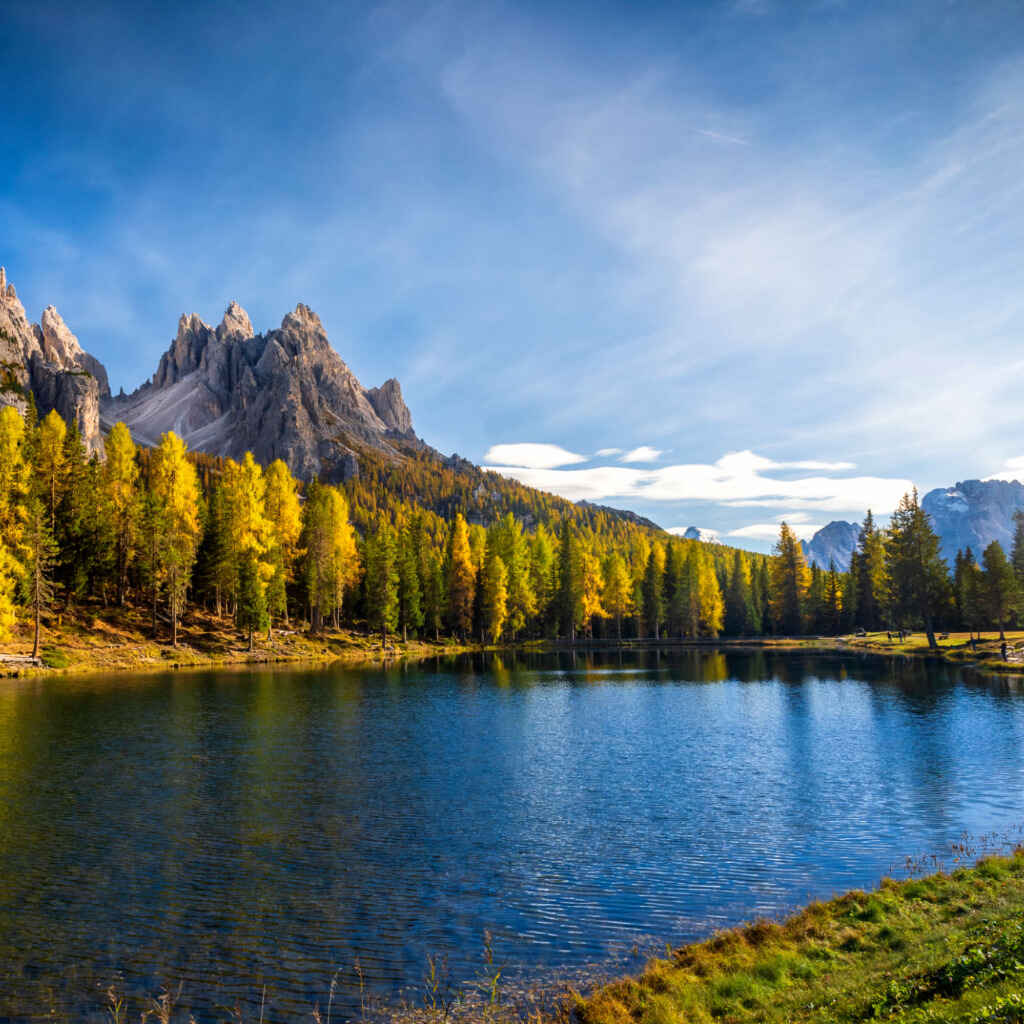} & \includegraphics[width=\sizea,height=\sizea]{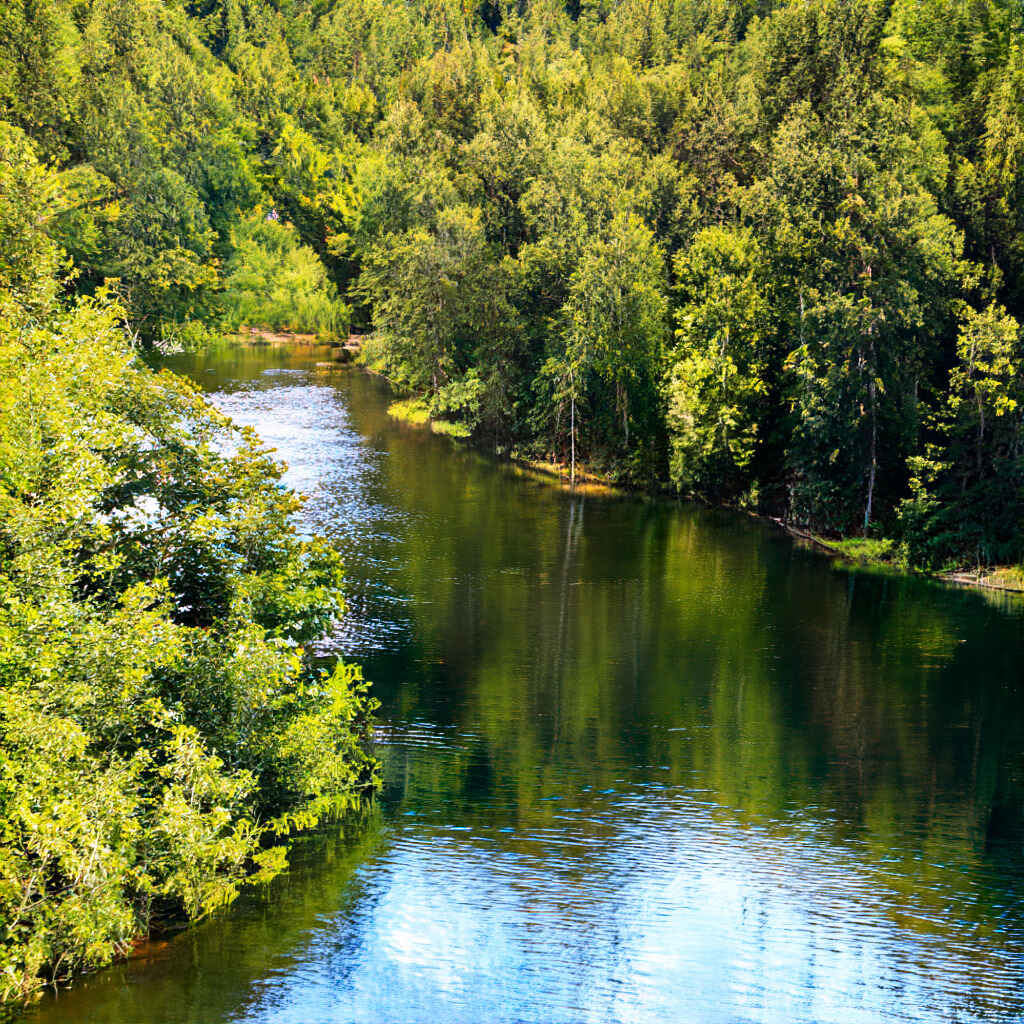}  & \includegraphics[width=\sizea,height=\sizea]{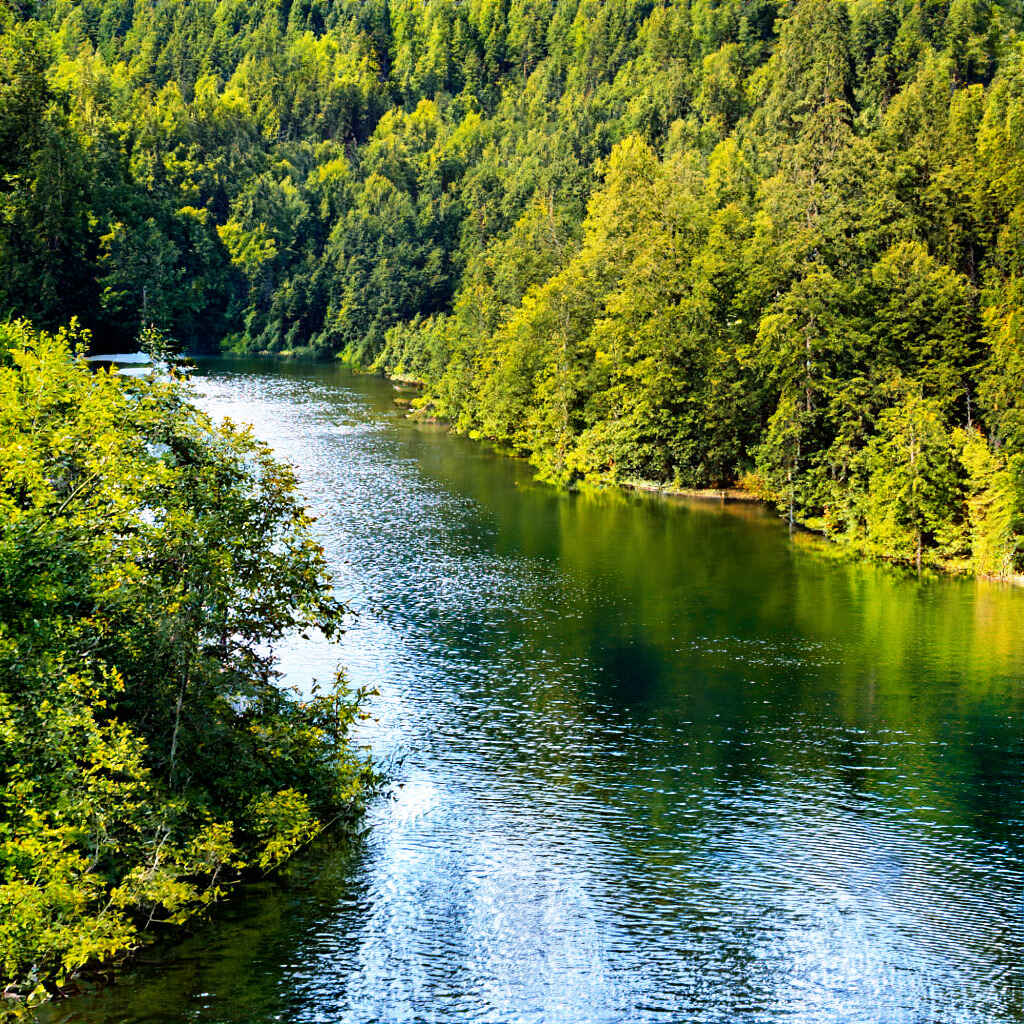}  & \includegraphics[width=\sizea,height=\sizea]{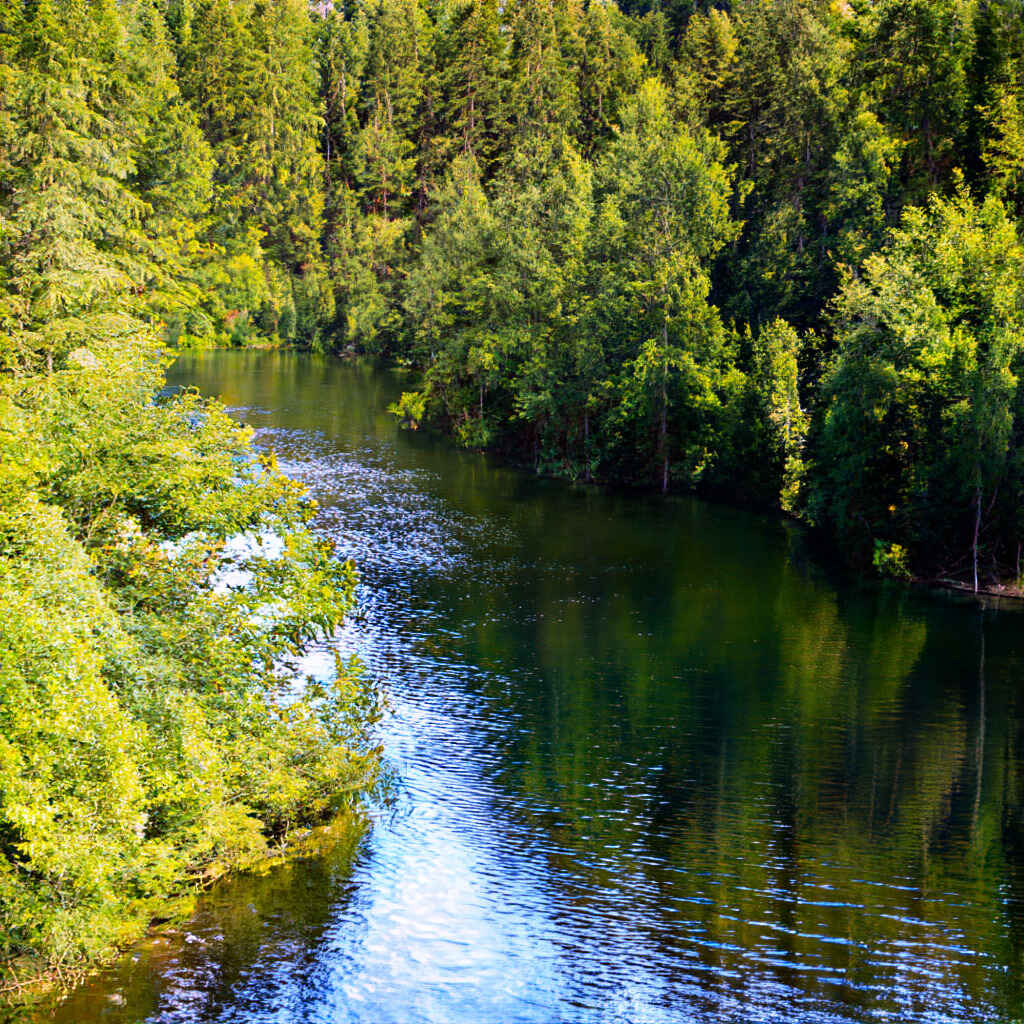} \\
	\includegraphics[width=\sizea,height=\sizea]{figures/getty_mm/seg_6} & \includegraphics[width=\sizea,height=\sizea]{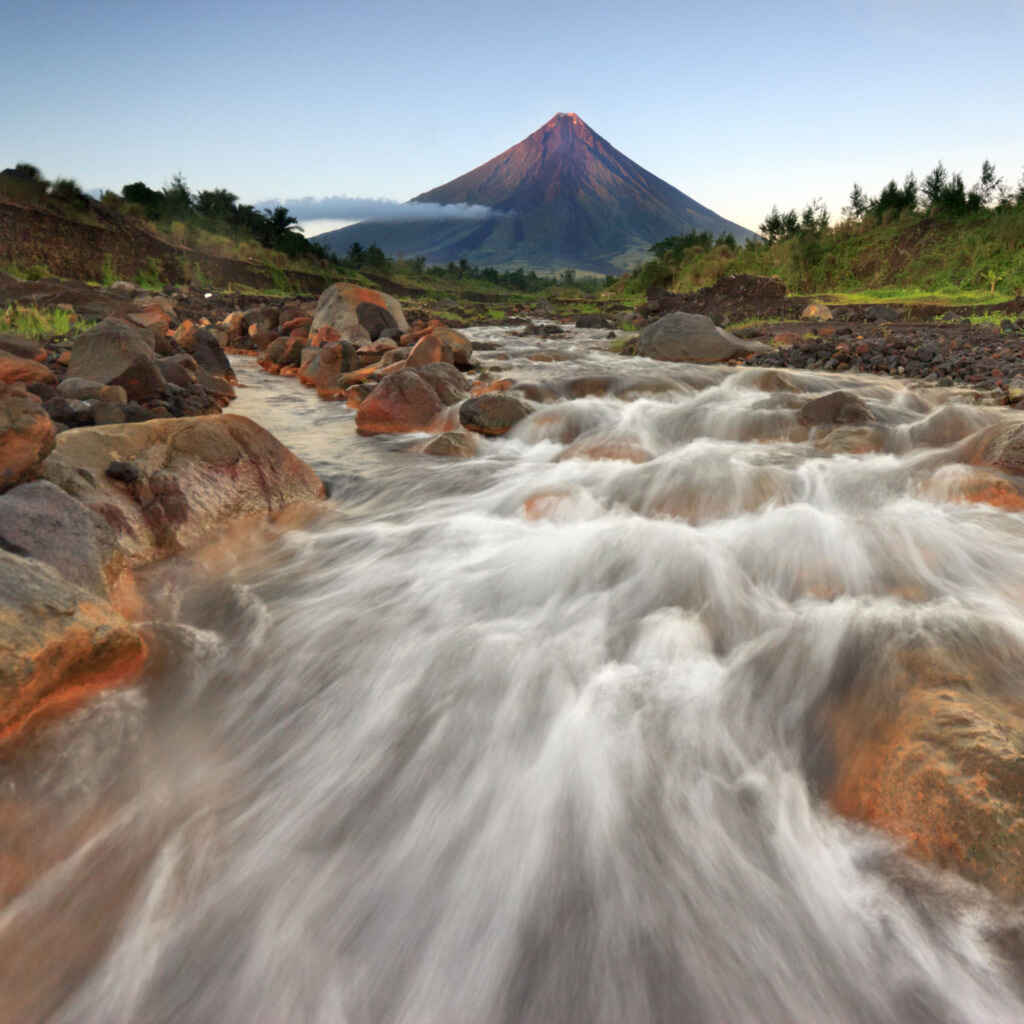} & \includegraphics[width=\sizea,height=\sizea]{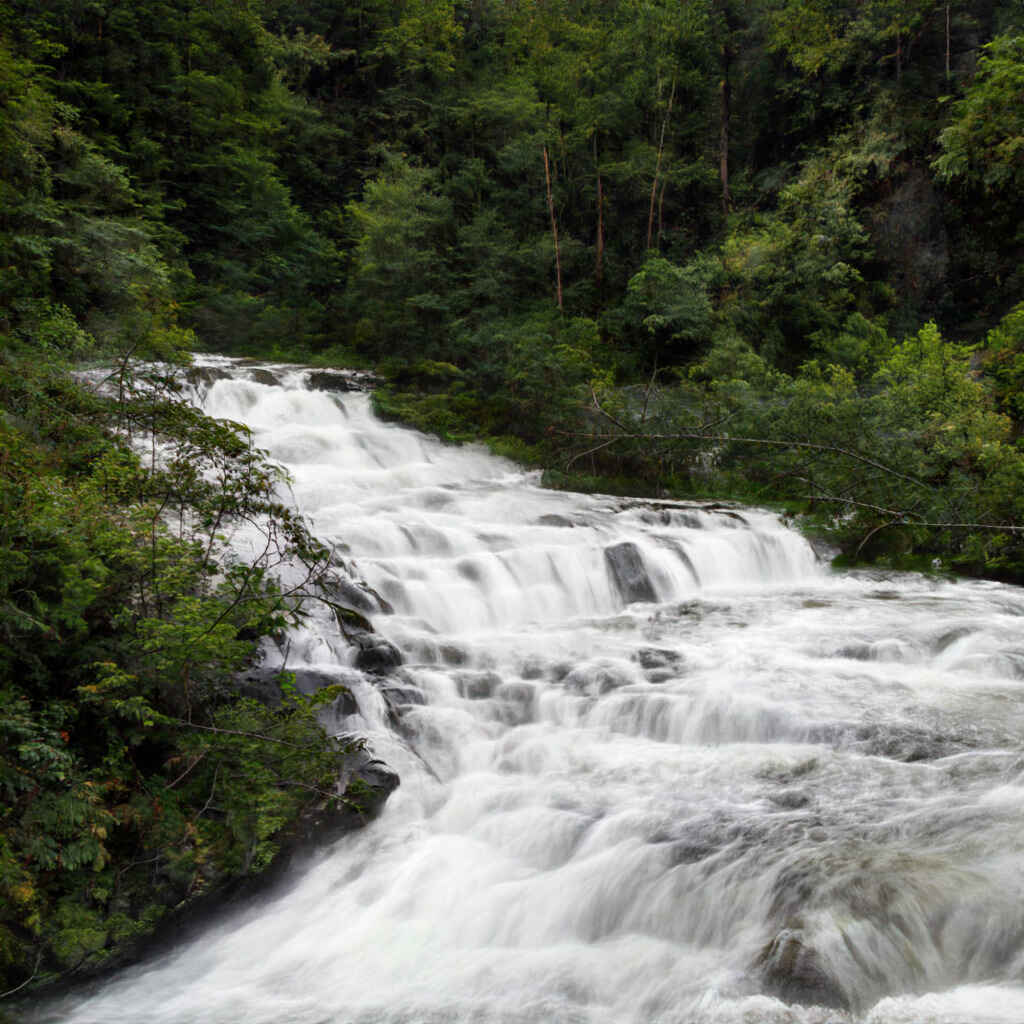}  & \includegraphics[width=\sizea,height=\sizea]{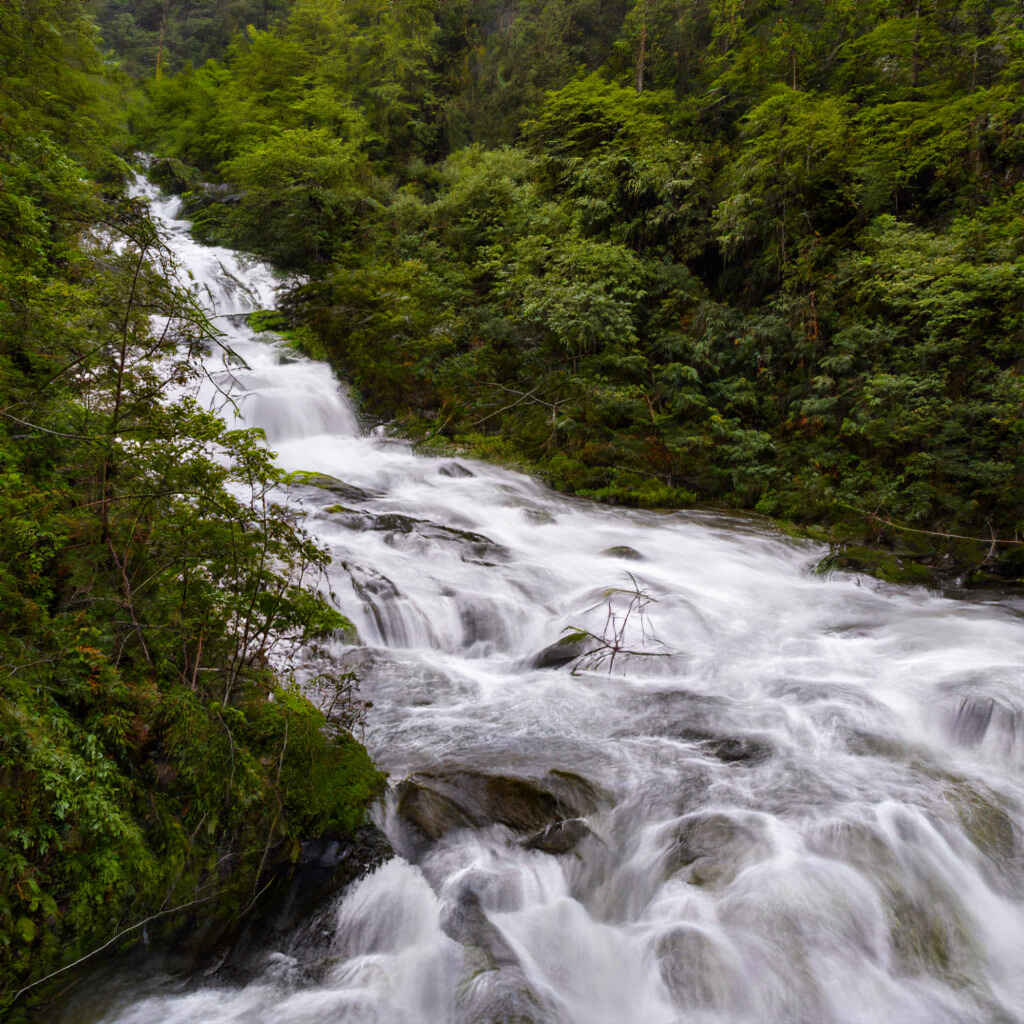}  & \includegraphics[width=\sizea,height=\sizea]{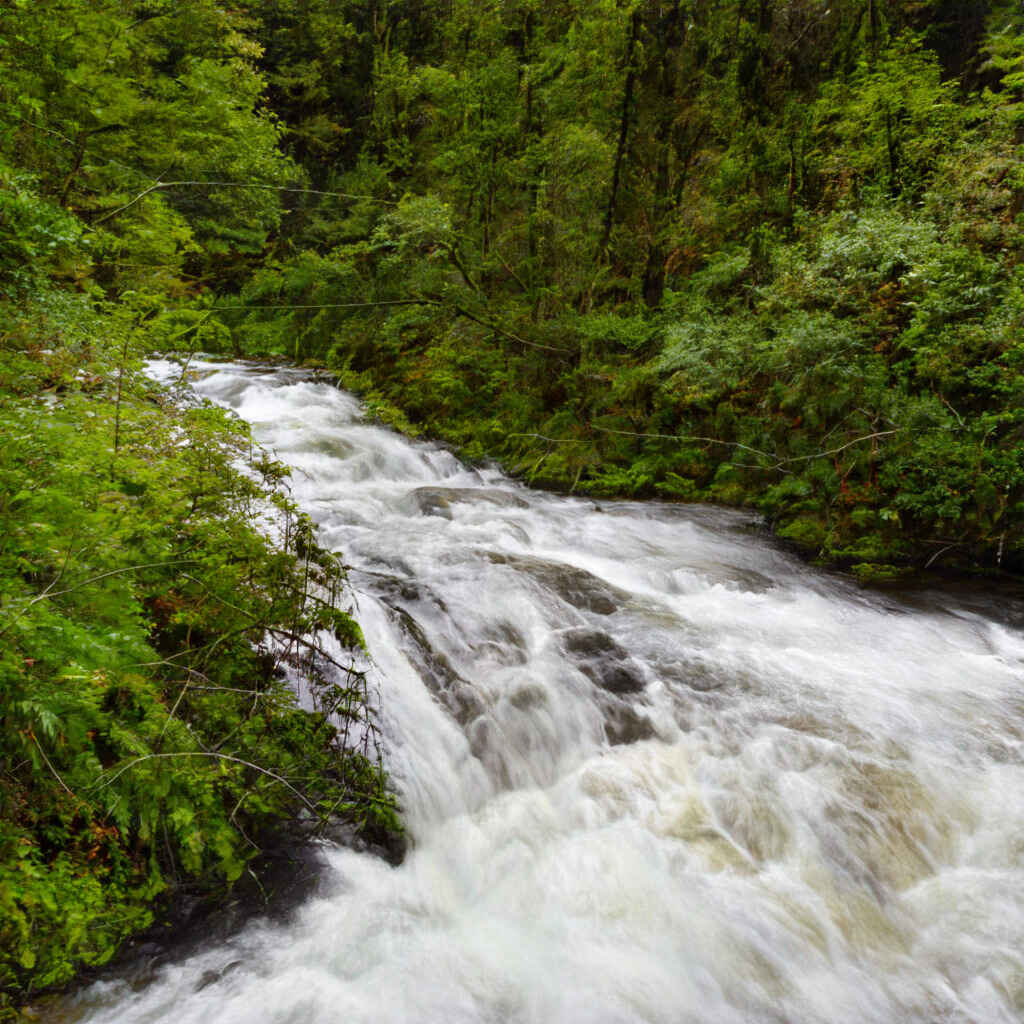} \\
	\includegraphics[width=\sizea,height=\sizea]{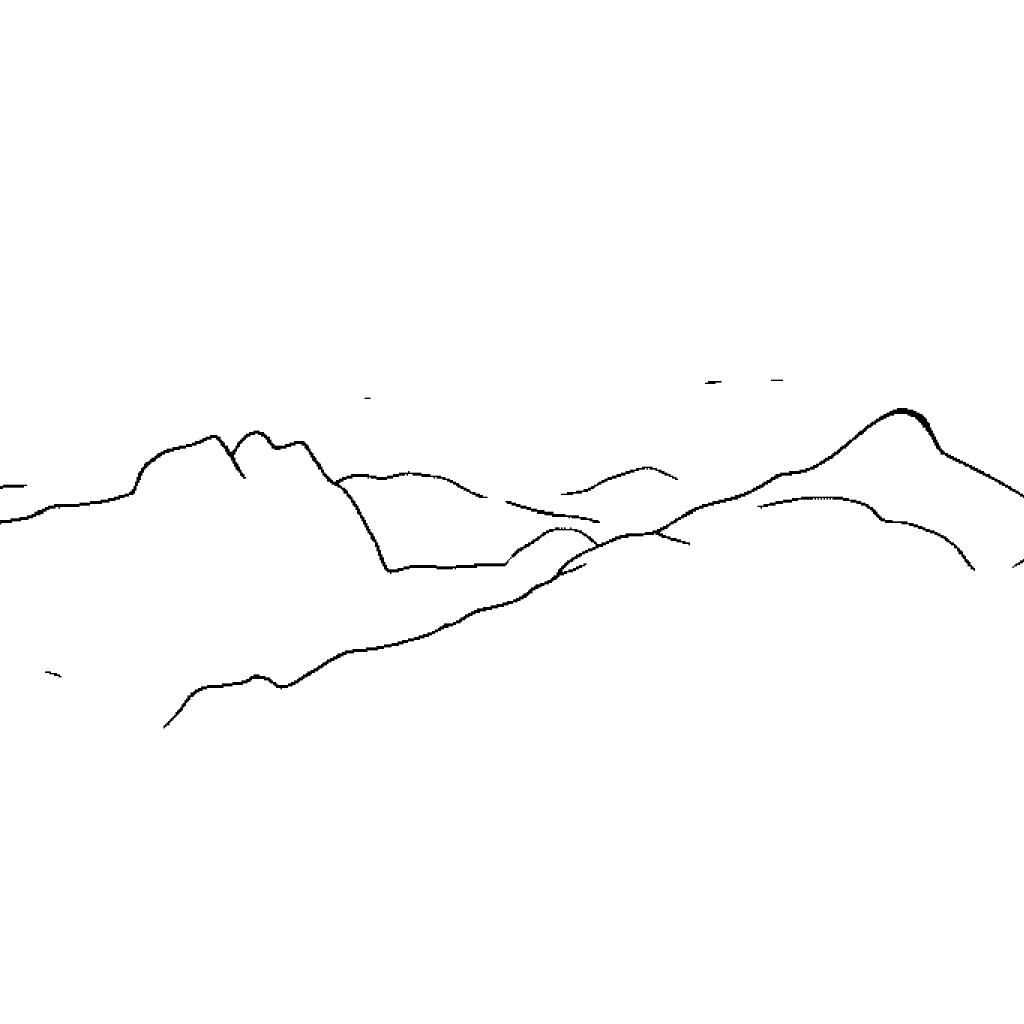} & \includegraphics[width=\sizea,height=\sizea]{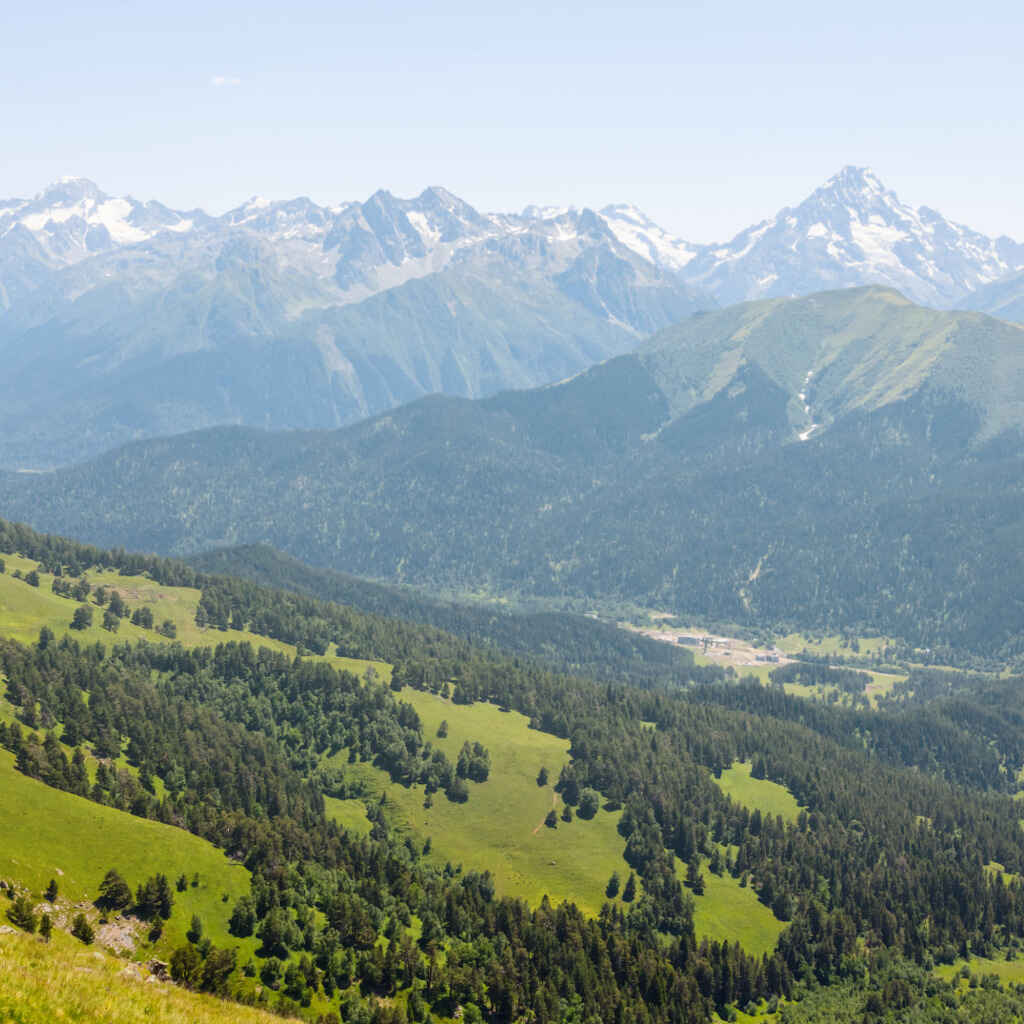} & \includegraphics[width=\sizea,height=\sizea]{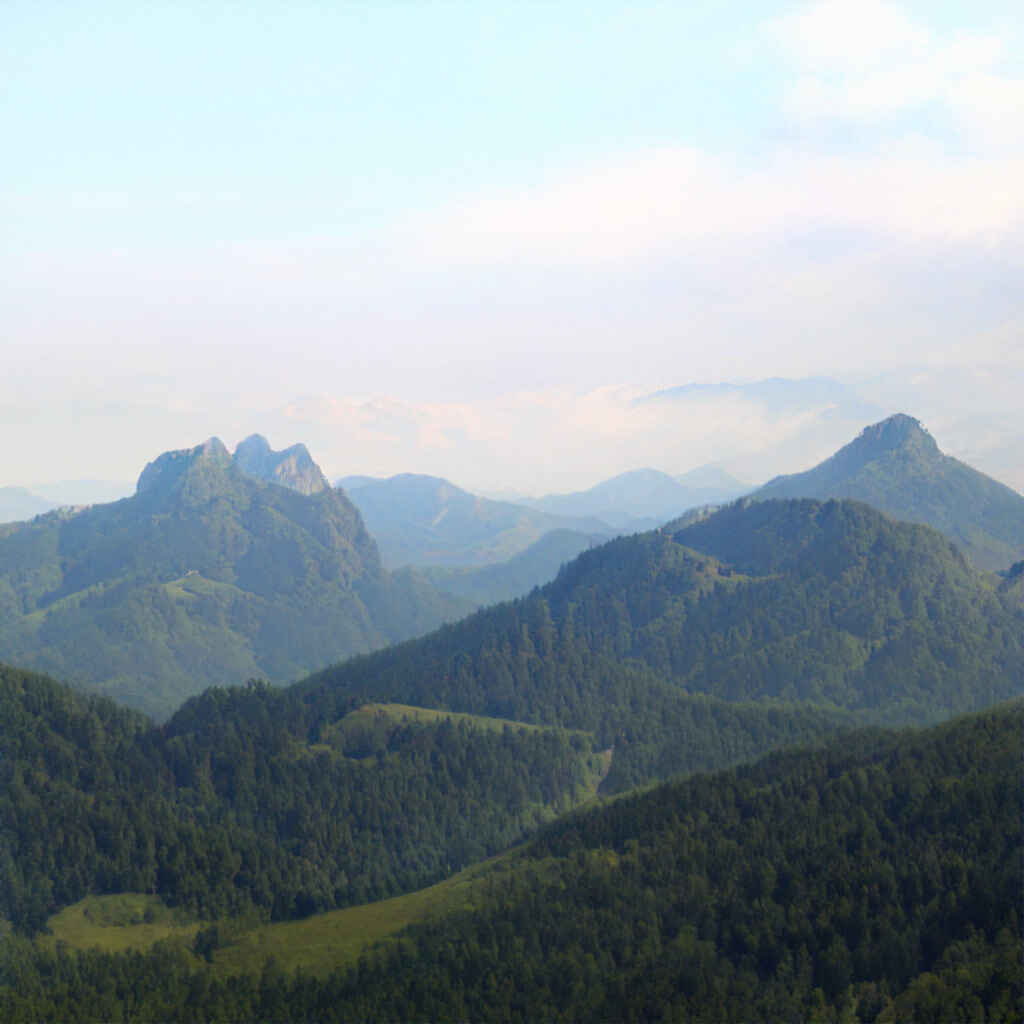}  & \includegraphics[width=\sizea,height=\sizea]{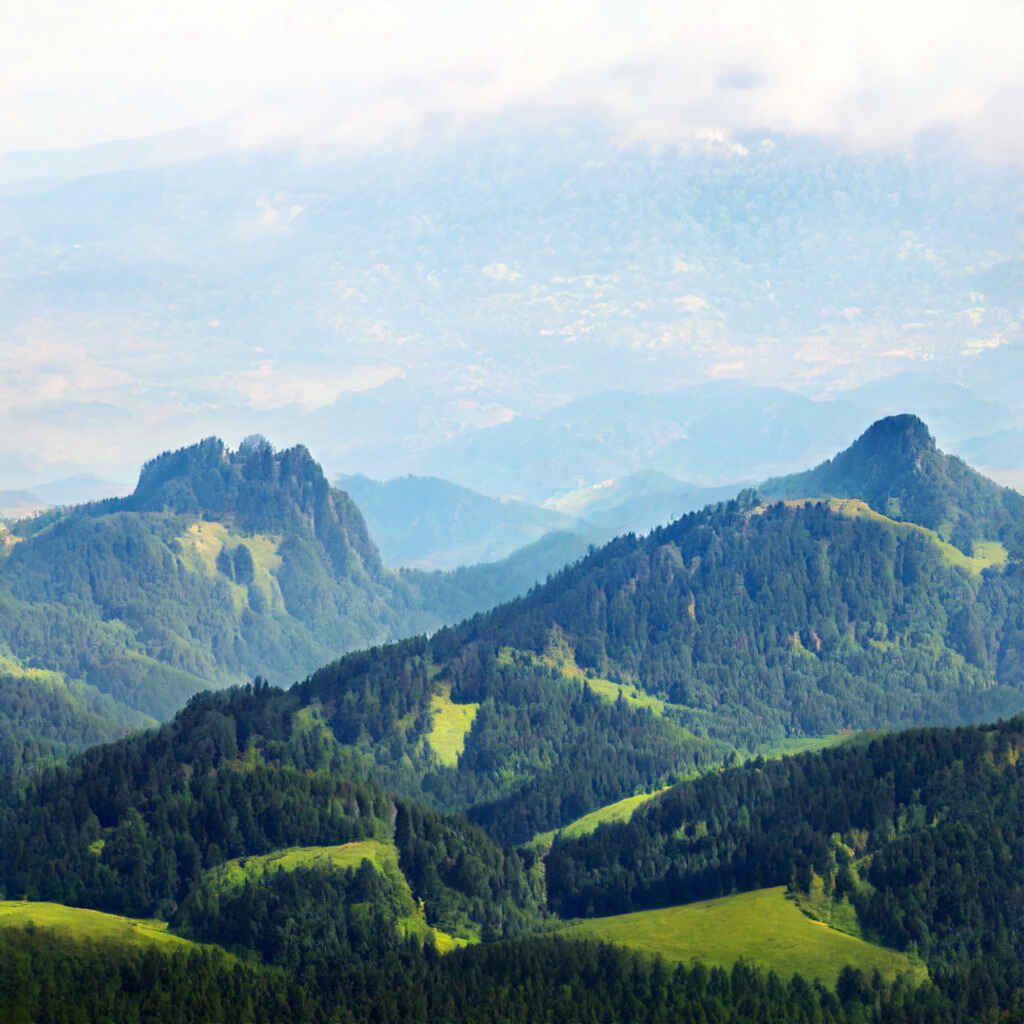}  & \includegraphics[width=\sizea,height=\sizea]{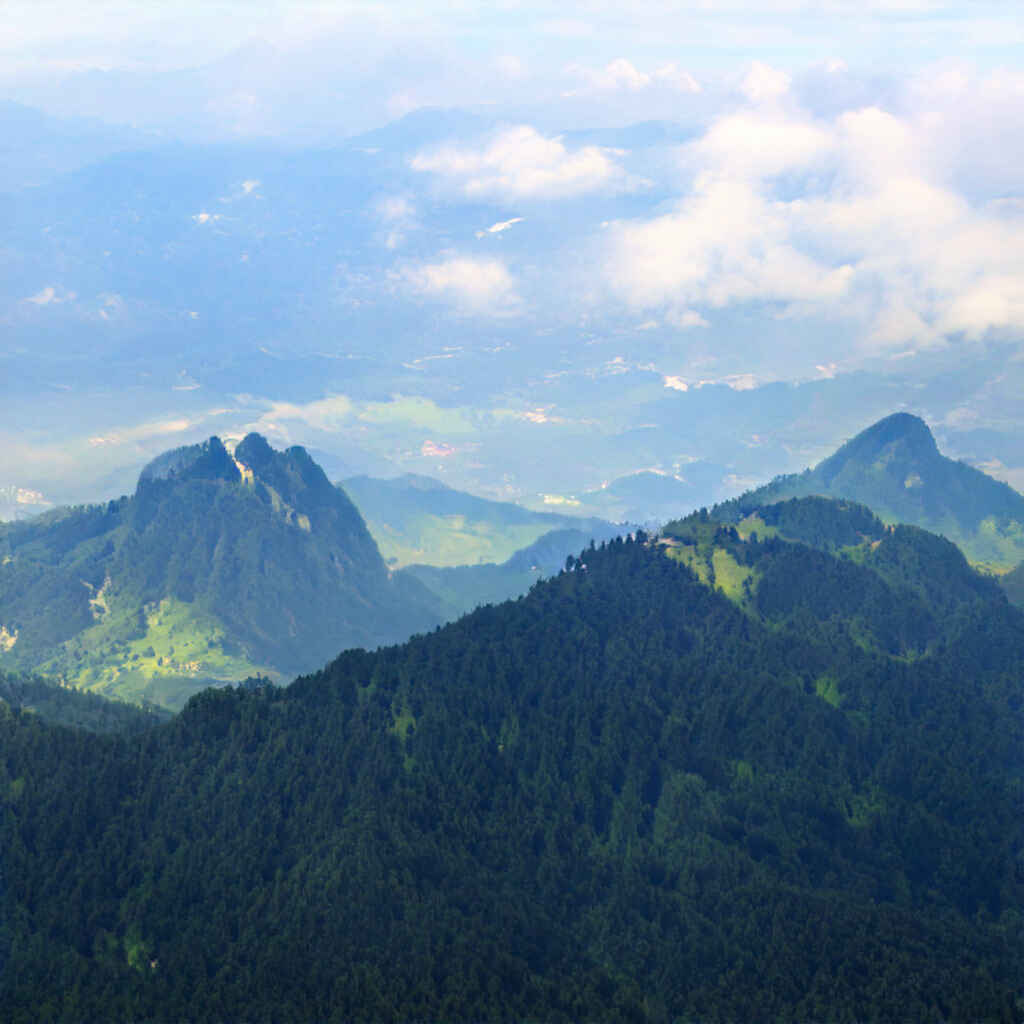} \\
	\includegraphics[width=\sizea,height=\sizea]{figures/getty_mm/sketch_6} & \includegraphics[width=\sizea,height=\sizea]{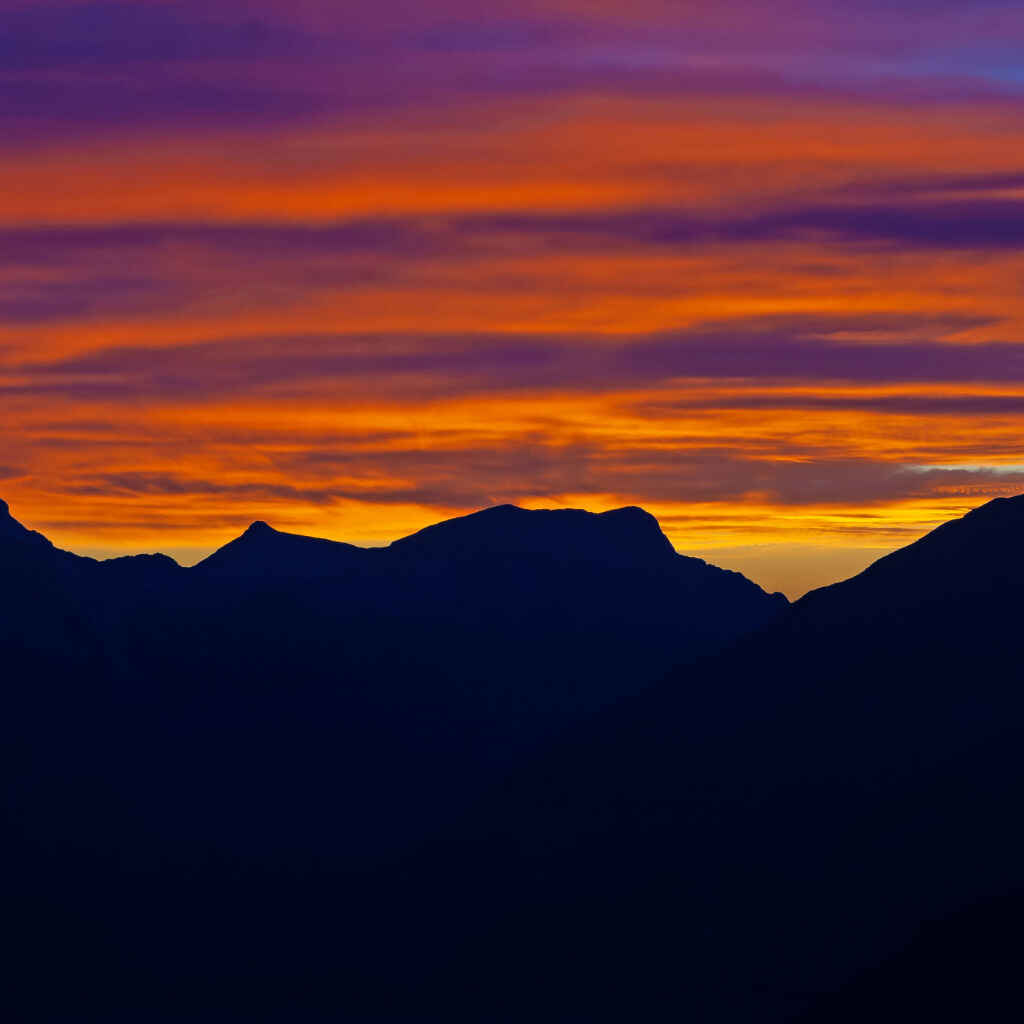}  & \includegraphics[width=\sizea,height=\sizea]{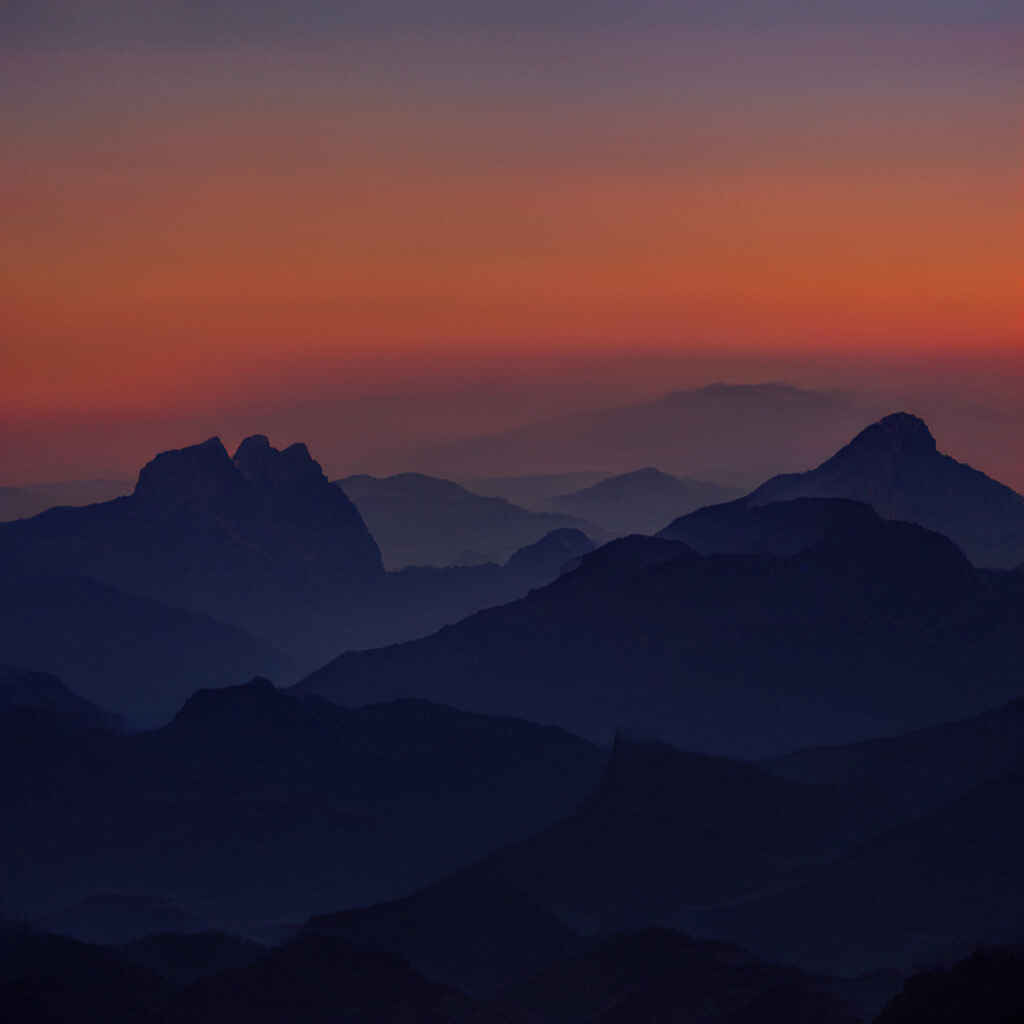}  & \includegraphics[width=\sizea,height=\sizea]{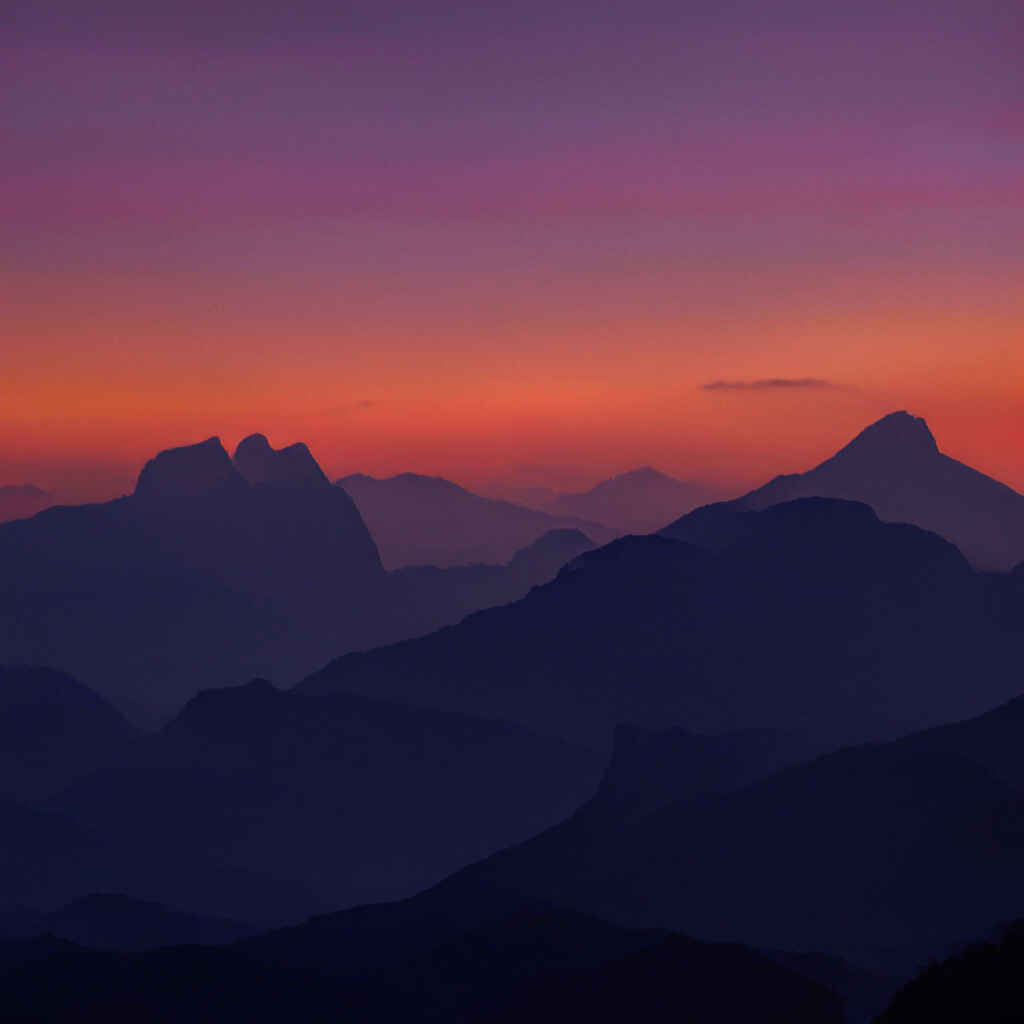}  & \includegraphics[width=\sizea,height=\sizea]{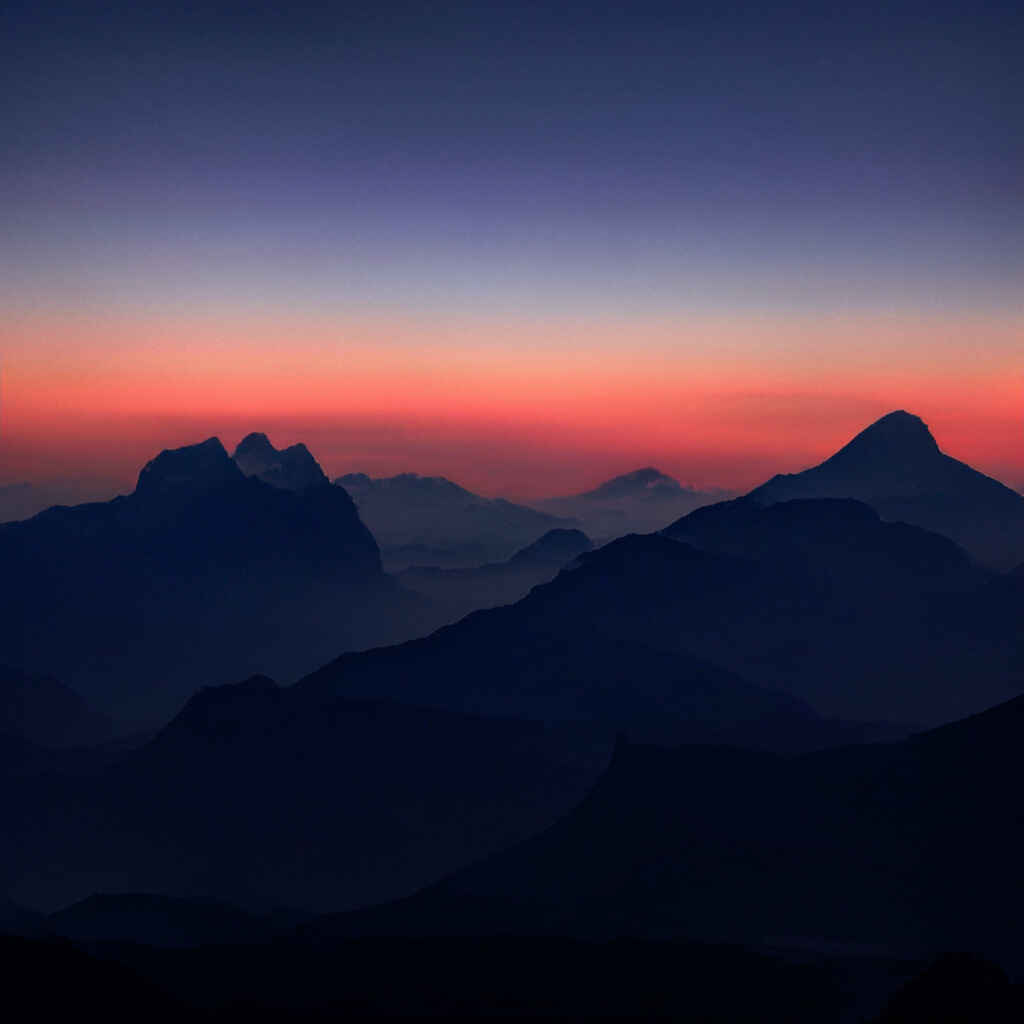} \\
	\begin{minipage}[b]{\sizea}
		\small
		\centering A lake surrounded by mountains. \vspace{35pt} 
	\end{minipage}     
	& \includegraphics[width=\sizea,height=\sizea]{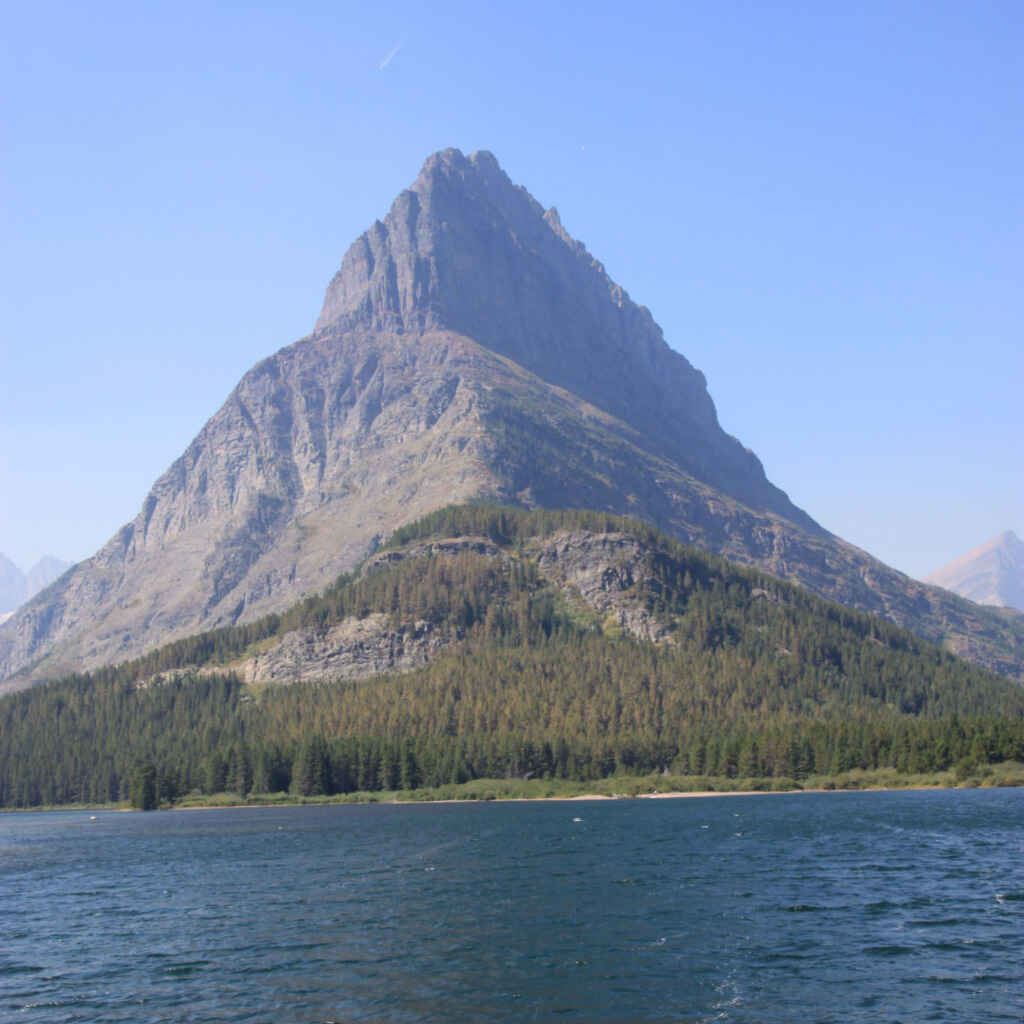}    & \includegraphics[width=\sizea,height=\sizea]{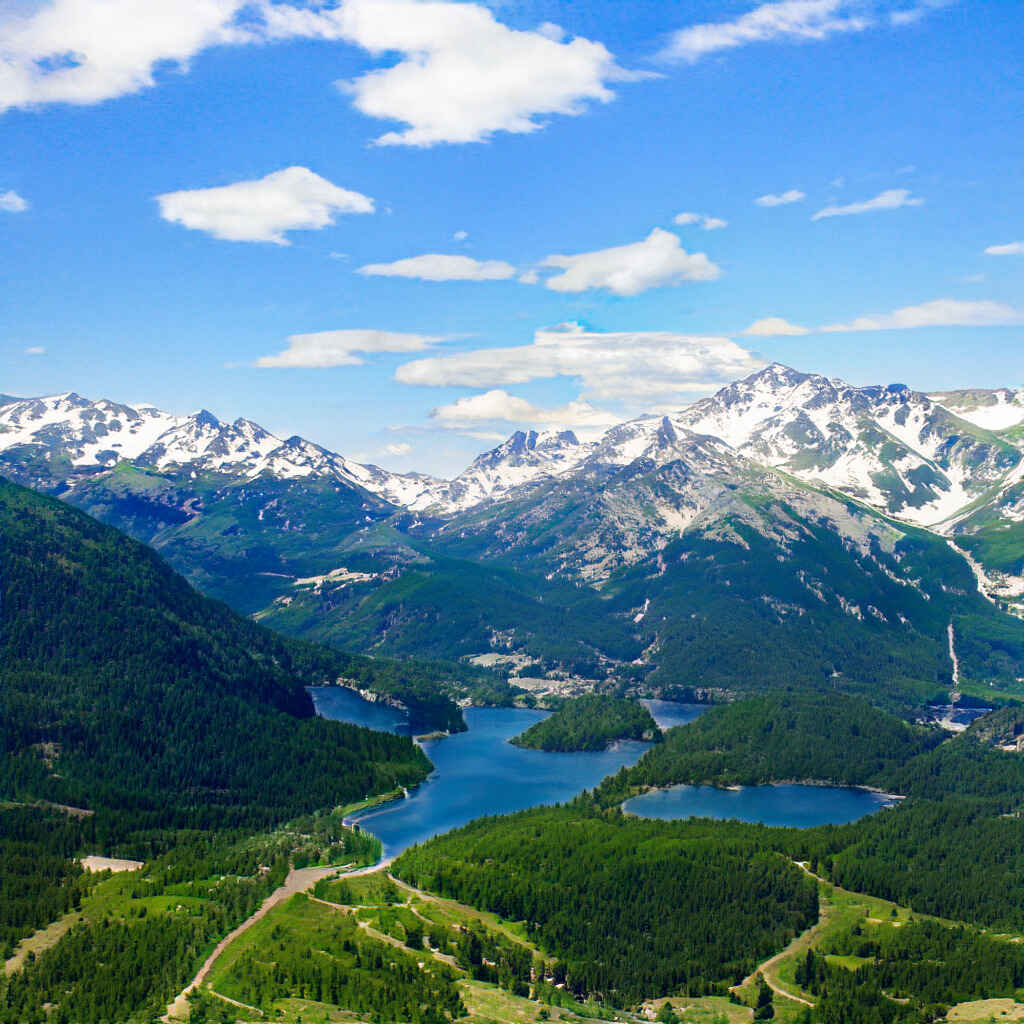}  & \includegraphics[width=\sizea,height=\sizea]{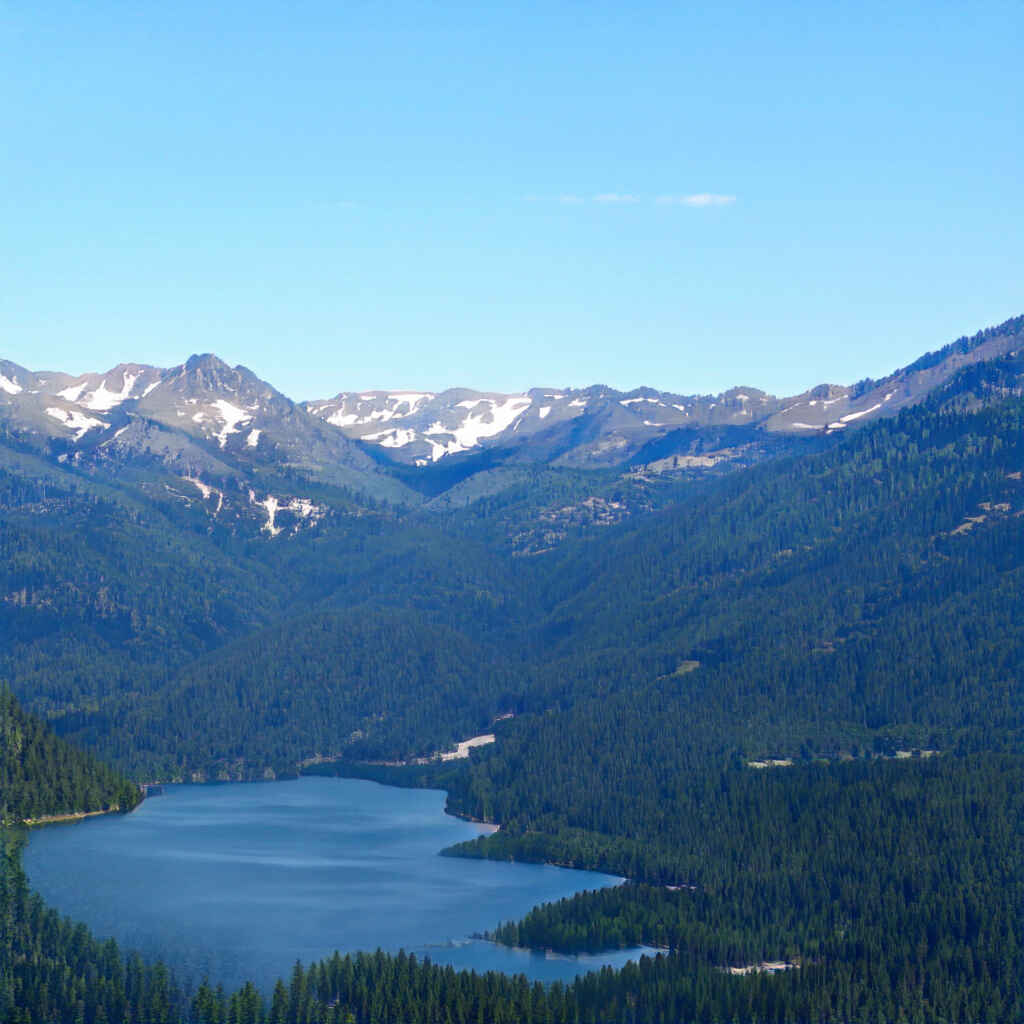}  & \includegraphics[width=\sizea,height=\sizea]{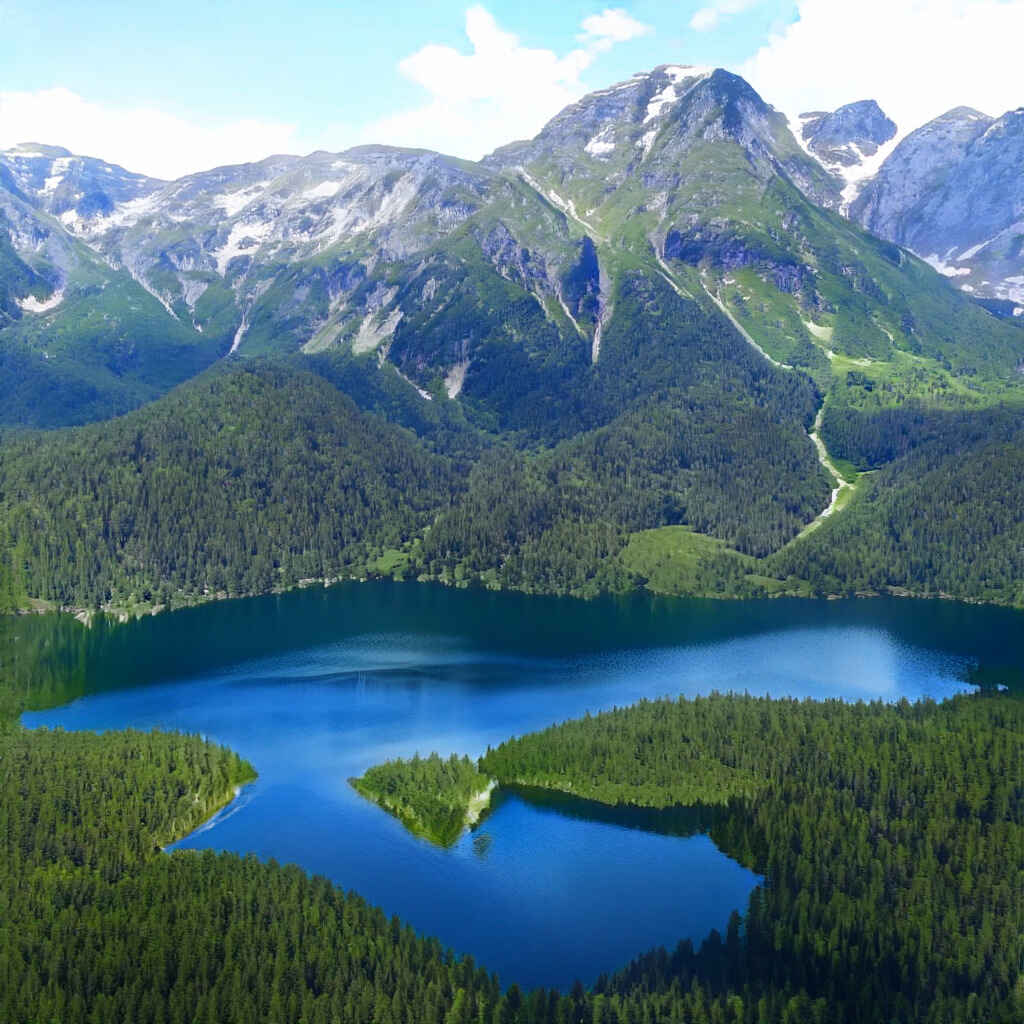} \\
	\begin{minipage}[b]{\sizea}
		\small
		\centering A lake surrounded by mountains. \vspace{35pt}  
	\end{minipage}     
	& \includegraphics[width=\sizea,height=\sizea]{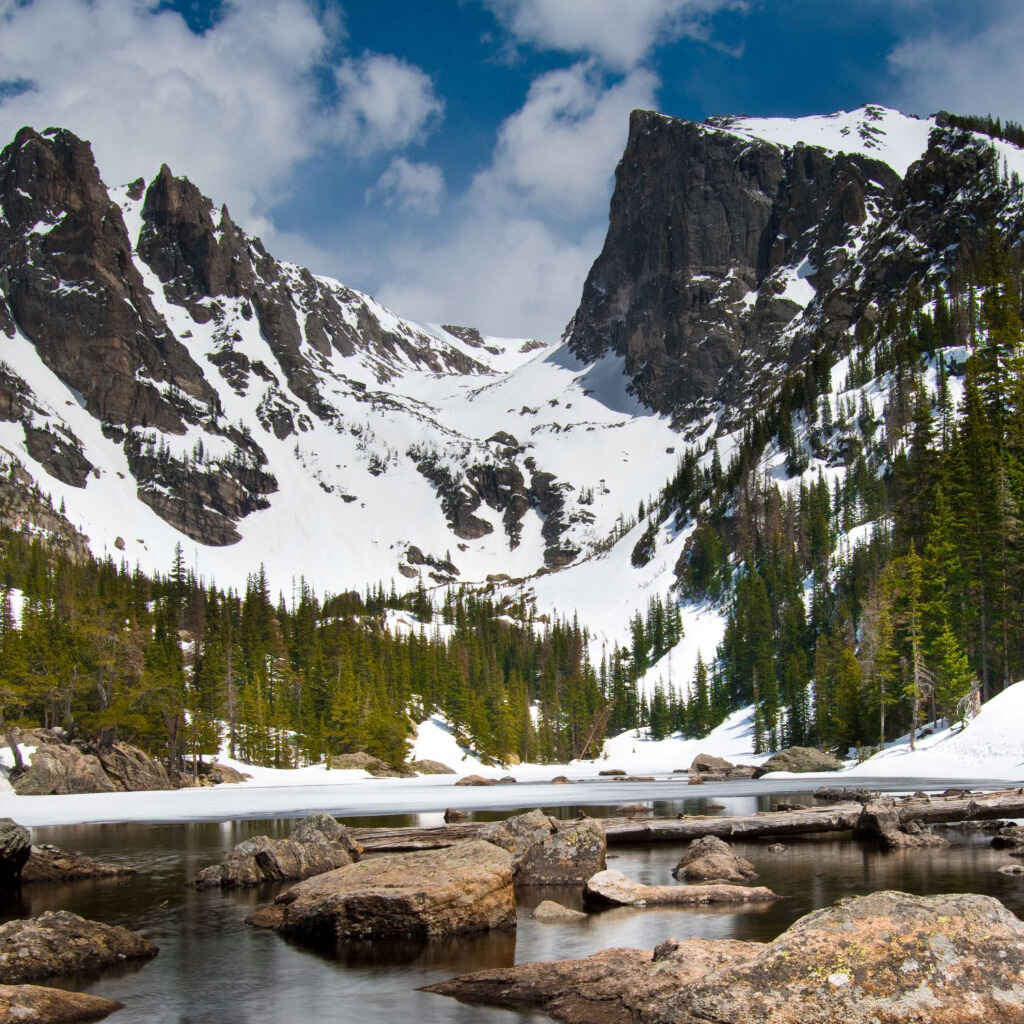}    & \includegraphics[width=\sizea,height=\sizea]{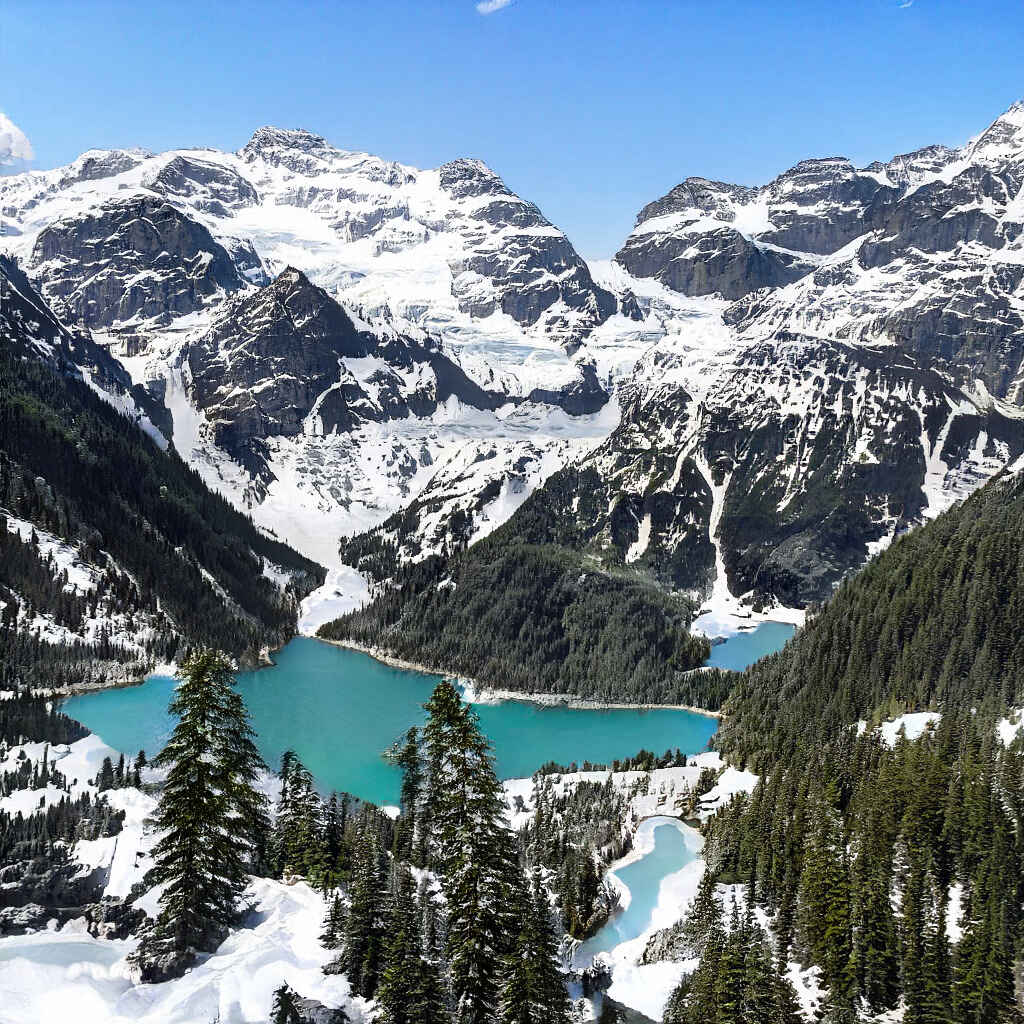}  & \includegraphics[width=\sizea,height=\sizea]{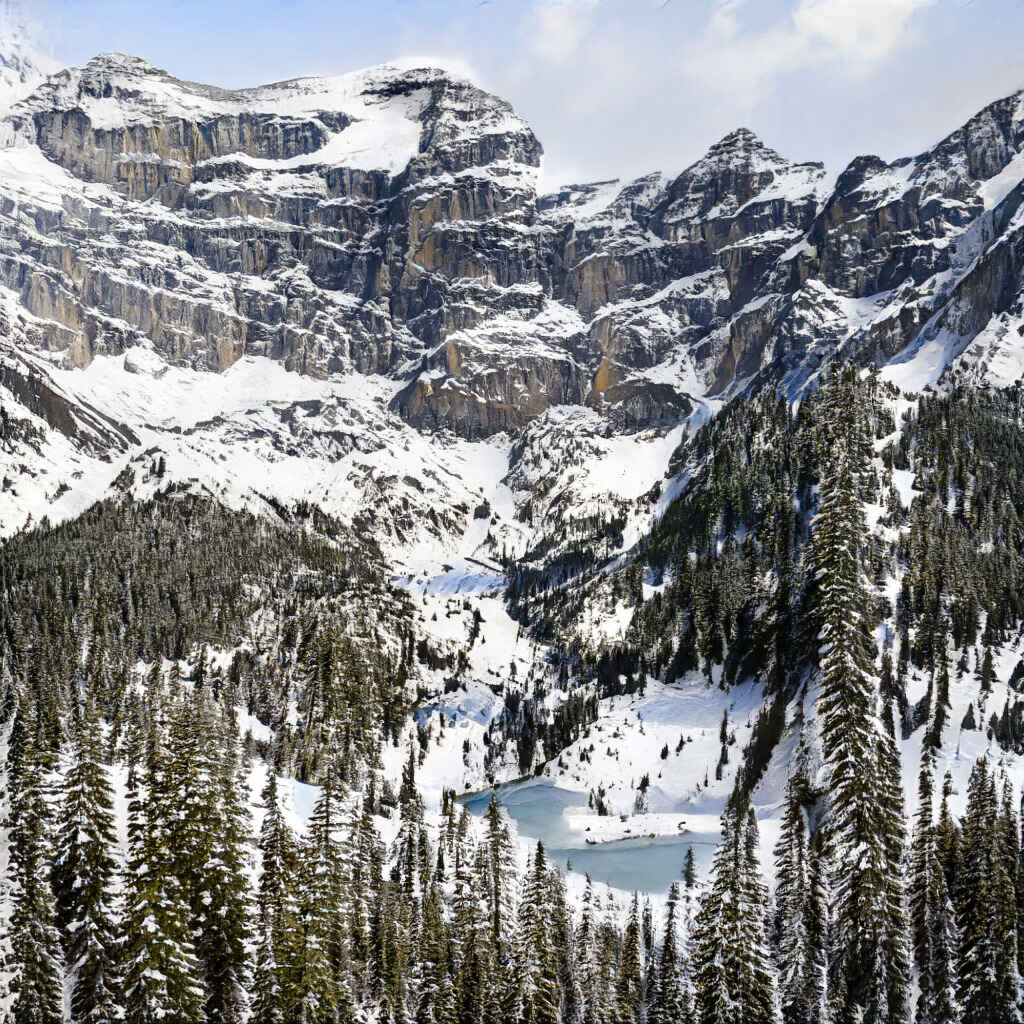}  & \includegraphics[width=\sizea,height=\sizea]{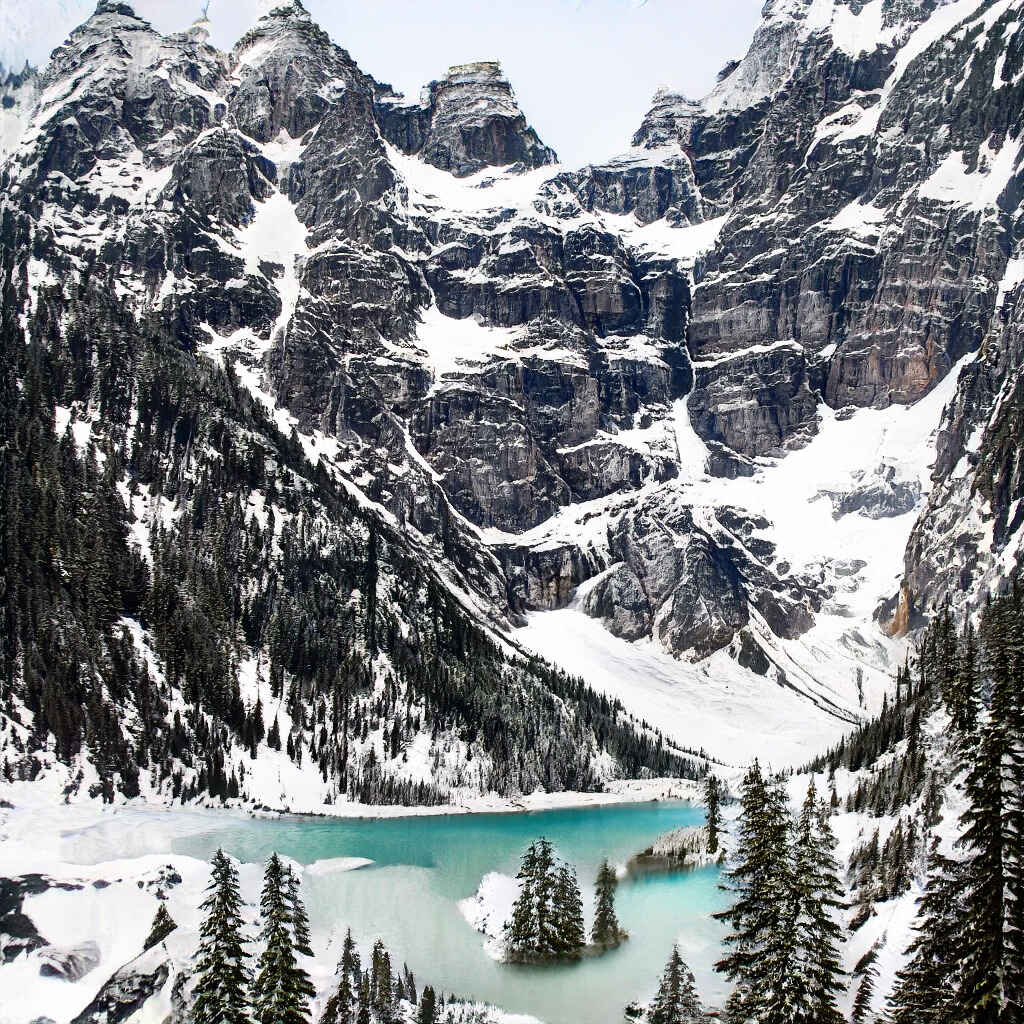} 
\end{tabular}

\vspace{-2mm}
\caption{Examples of multimodal conditional image synthesis on Landscape. We show three random samples from PoE-GAN conditioned on two modalities, one being text/segmentation/sketch and another being style reference.}
\label{fig:getty_mm_style}
\vspace{-2mm}
\end{figure*}

\begin{figure*}
	\centering
	\newcommand{\sizea}{0.19\linewidth}
	\setlength{\tabcolsep}{2pt}
	\begin{tabular}{ccccc}
		Text & Ground truth & DF-GAN & DM-GAN + CL & PoE-GAN~(Ours) \\
		\begin{minipage}[b]{\sizea}
			\centering A kitchen with a counter and a table with chairs.\vspace{18pt} 
		\end{minipage}     
		& \includegraphics[width=\sizea]{figures/coco_text/gt_000000034873.jpg}    & \includegraphics[width=\sizea]{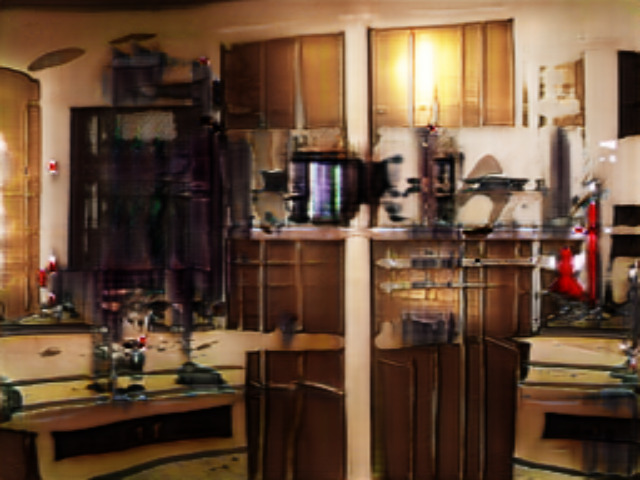}  & \includegraphics[width=\sizea]{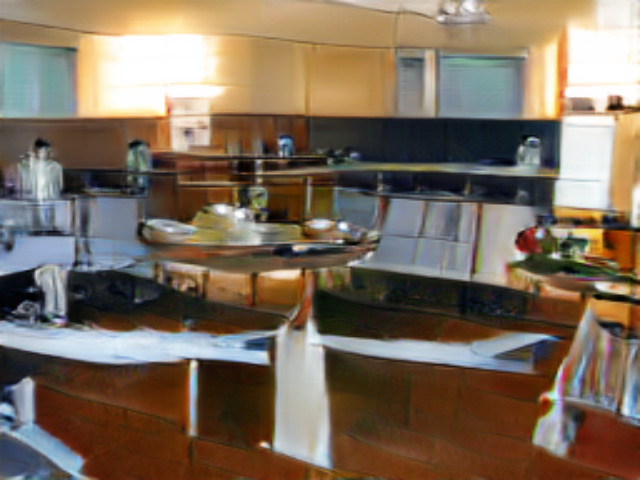}  & \includegraphics[width=\sizea]{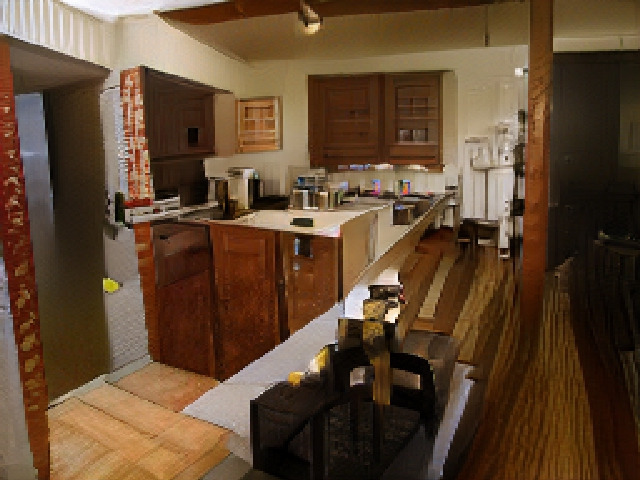} \\
		\begin{minipage}[b]{\sizea}
			\centering A view of mountains from the window of a jet airplane.\vspace{18pt} 
		\end{minipage}     
		& \includegraphics[width=\sizea]{figures/coco_text/gt_000000208208.jpg}    & \includegraphics[width=\sizea]{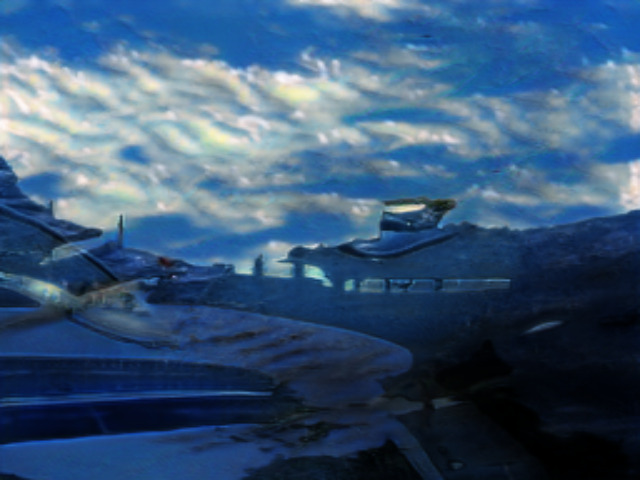}  & \includegraphics[width=\sizea]{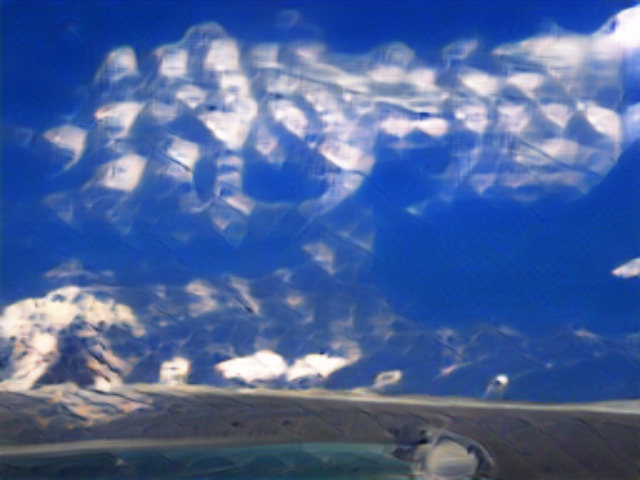}  & \includegraphics[width=\sizea]{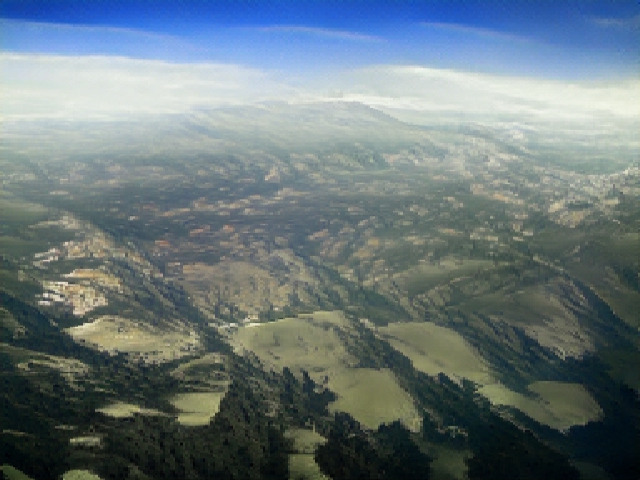} \\
		\begin{minipage}[b]{\sizea}
			\centering A blueberry cake is on a plate and is topped with butter.\vspace{18pt} 
		\end{minipage}     
		& \includegraphics[width=\sizea]{figures/coco_text/gt_000000184384.jpg}    & \includegraphics[width=\sizea]{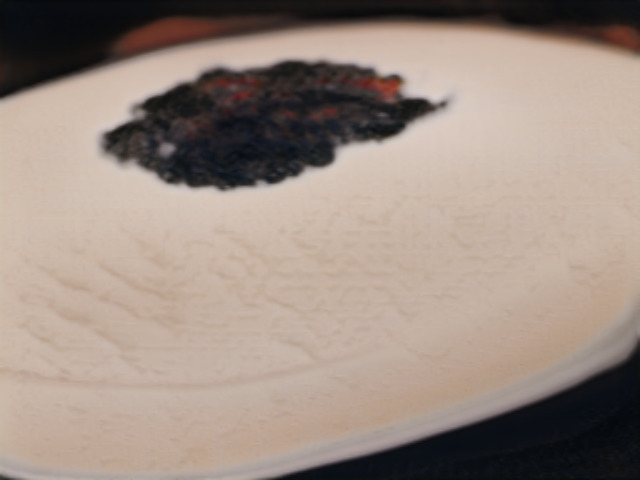}  & \includegraphics[width=\sizea]{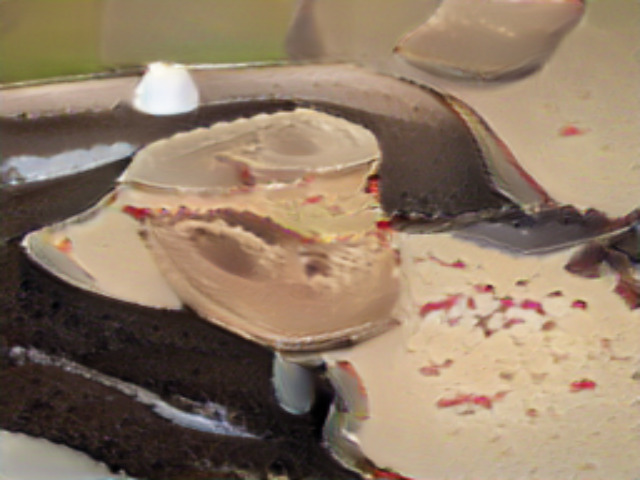}  & \includegraphics[width=\sizea]{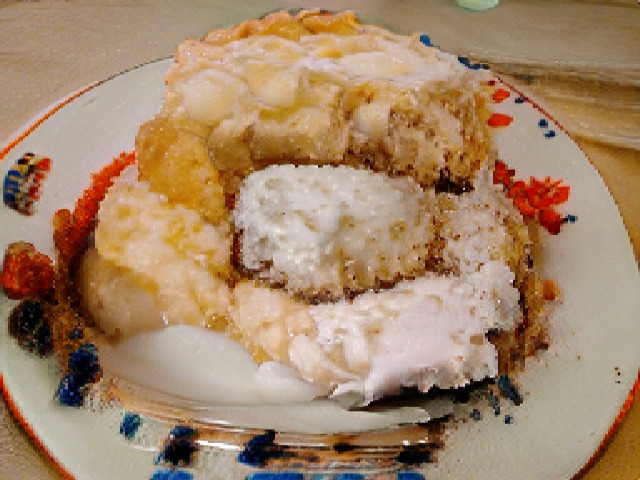} \\
		\begin{minipage}[b]{\sizea}
			\centering Three men each holding something and posing for a picture.\vspace{18pt} 
		\end{minipage}     
		& \includegraphics[width=\sizea]{figures/coco_text/gt_000000571598.jpg}    & \includegraphics[width=\sizea]{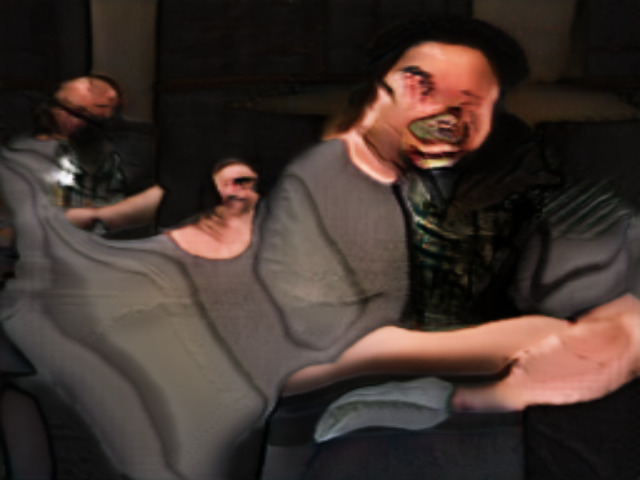}  & \includegraphics[width=\sizea]{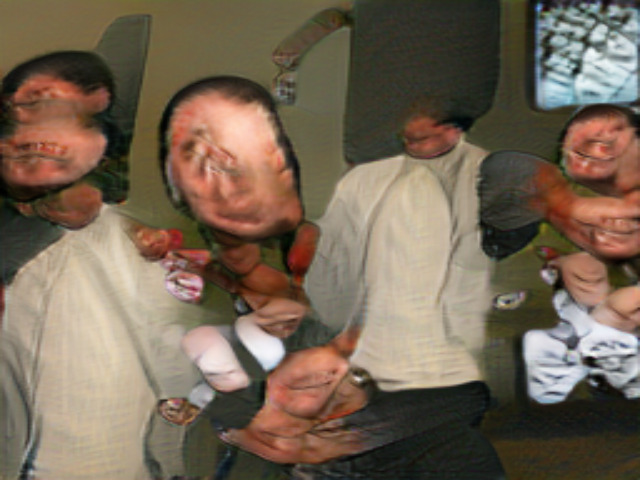}  & \includegraphics[width=\sizea]{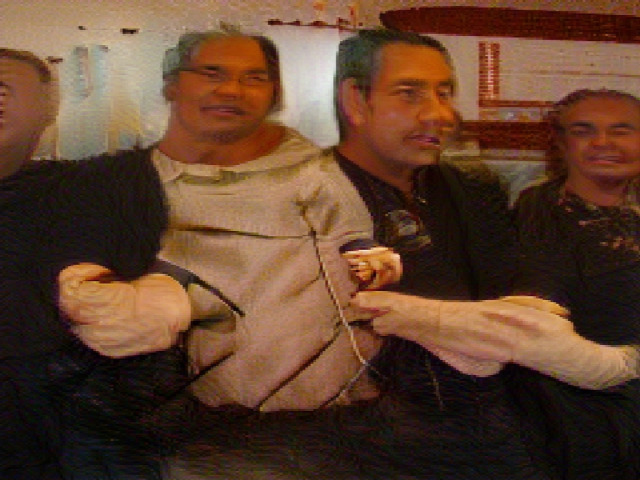} \\
		\begin{minipage}[b]{\sizea}
			\centering A plate holds a good size portion of a cooked, mixed dish that includes broccoli and pasta.\vspace{7pt} 
		\end{minipage}     
		& \includegraphics[width=\sizea]{figures/coco_text/gt_000000407524.jpg}    & \includegraphics[width=\sizea]{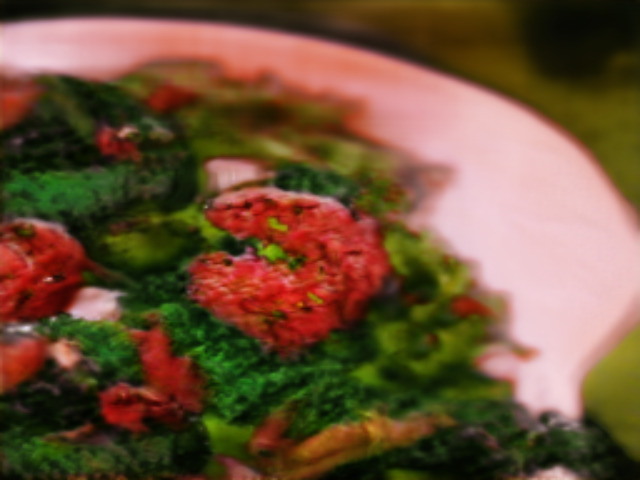}  & \includegraphics[width=\sizea]{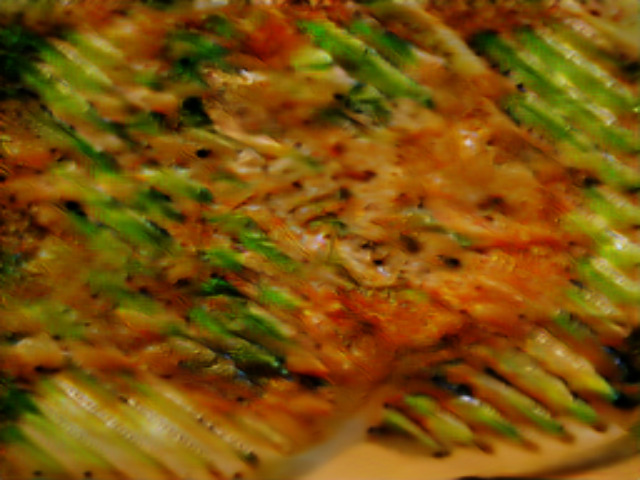}  & \includegraphics[width=\sizea]{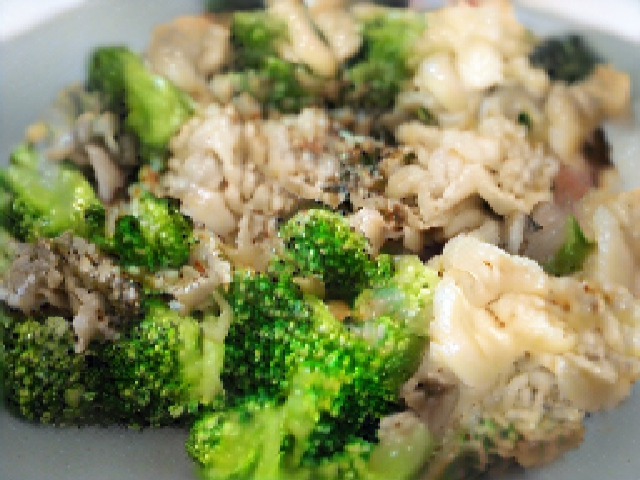} \\
		\begin{minipage}[b]{\sizea}
			\centering A red blue and yellow train and some people on a platform.\vspace{18pt} 
		\end{minipage}     
		& \includegraphics[width=\sizea]{figures/coco_text/gt_000000439593.jpg}    & \includegraphics[width=\sizea]{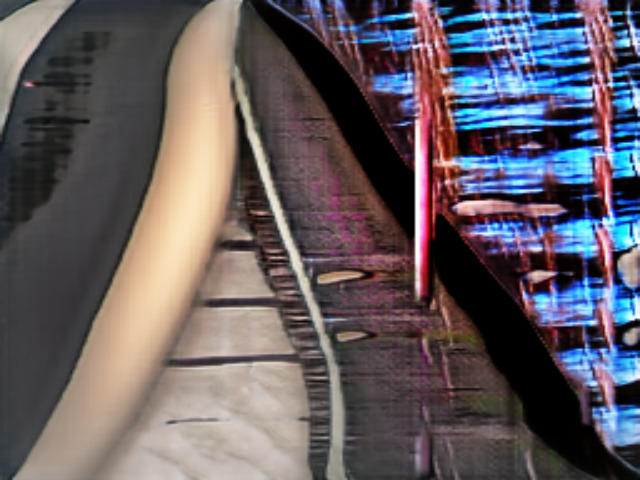}  & \includegraphics[width=\sizea]{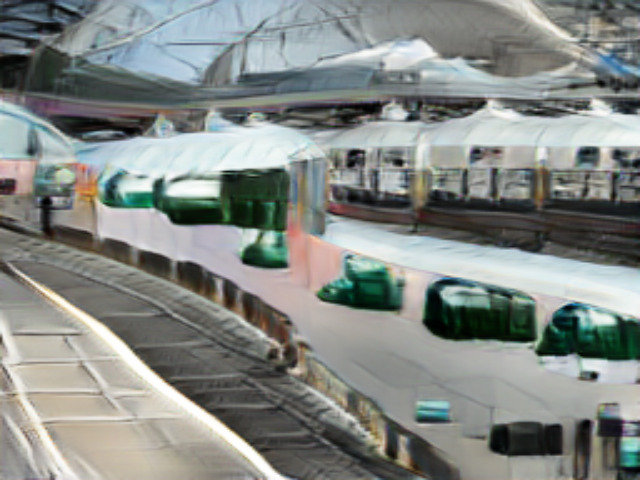}  & \includegraphics[width=\sizea]{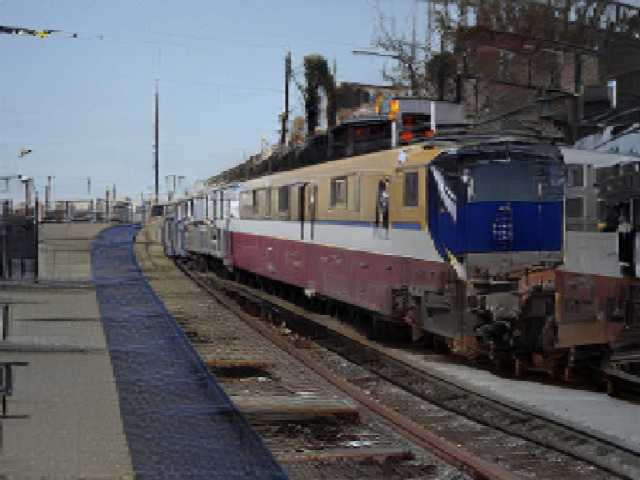} \\
		\begin{minipage}[b]{\sizea}
			\centering A man blowing out candles on a birthday cake.\vspace{18pt} 
		\end{minipage}     
		& \includegraphics[width=\sizea]{figures/coco_text/gt_000000206579.jpg}    & \includegraphics[width=\sizea]{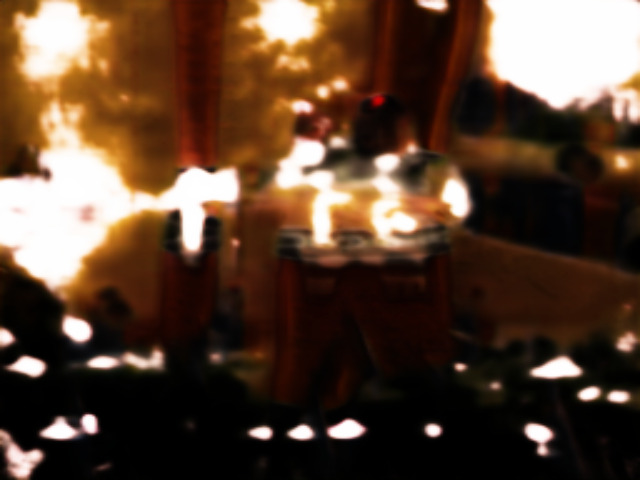}  & \includegraphics[width=\sizea]{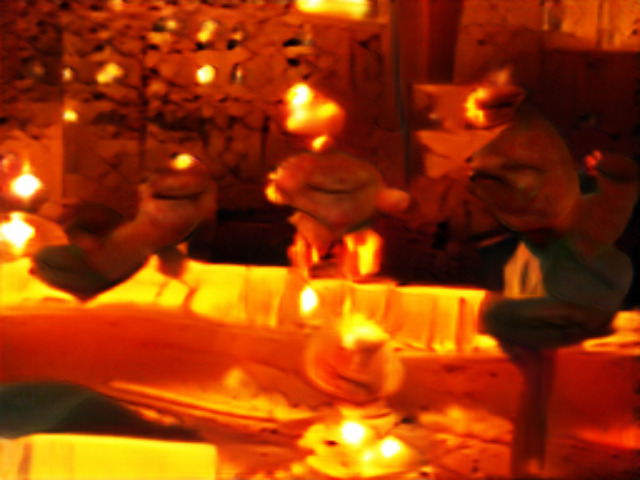}  & \includegraphics[width=\sizea]{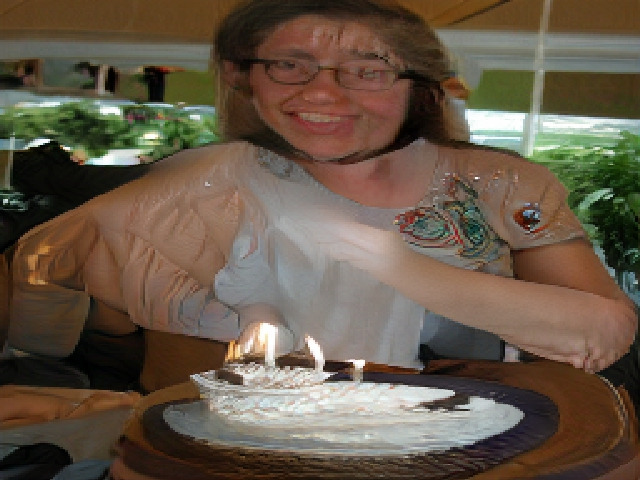} \\
		\begin{minipage}[b]{\sizea}
			\centering A desk set up as a workstation with a laptop.\vspace{18pt} 
		\end{minipage}     
		& \includegraphics[width=\sizea]{figures/coco_text/gt_000000513688.jpg}    & \includegraphics[width=\sizea]{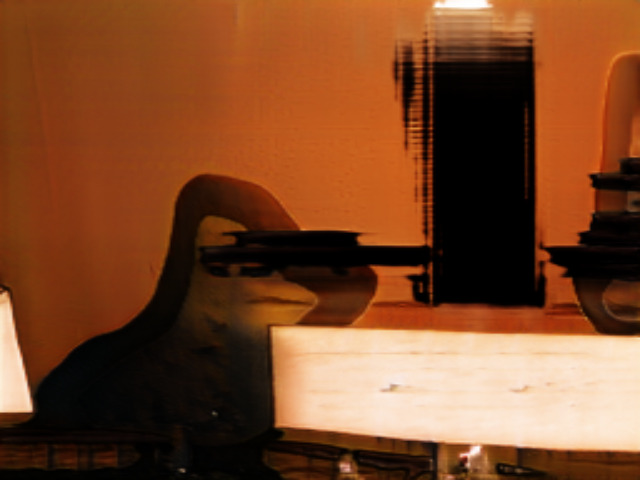}  & \includegraphics[width=\sizea]{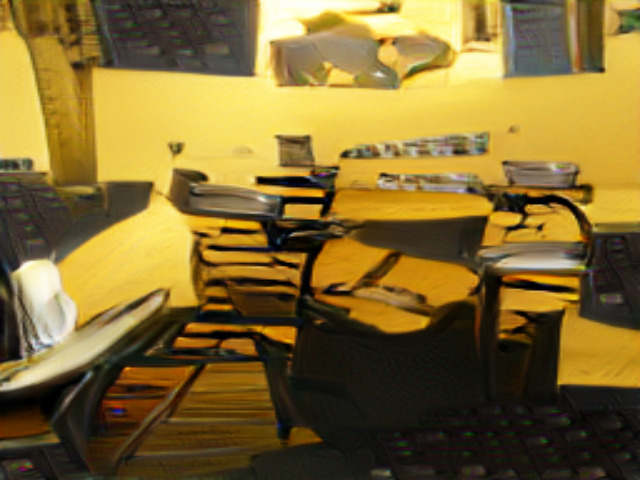}  & \includegraphics[width=\sizea]{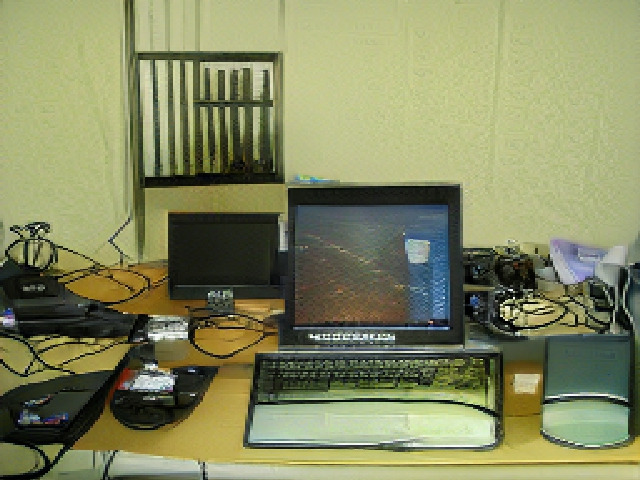}
	\end{tabular}
	\caption{Additional qualitative comparison of text-to-image synthesis on MS-COCO 2017.}
	\label{fig:coco_text_more}
\end{figure*}

\begin{figure*}
	\centering
	\newcommand{\sizea}{0.19\linewidth}
	\setlength{\tabcolsep}{2pt}
	\begin{tabular}{ccccc}
		Segmentation & Ground truth & SPADE & OASIS & PoE-GAN~(Ours) \\
		\includegraphics[width=\sizea]{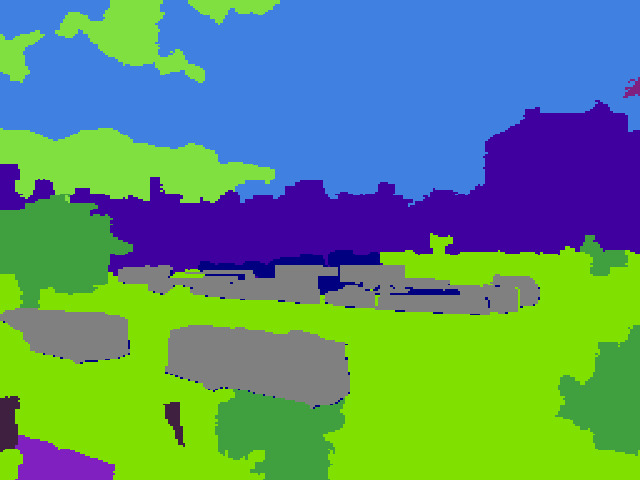}    & \includegraphics[width=\sizea]{figures/coco_seg/gt_000000015517.jpg}     & \includegraphics[width=\sizea]{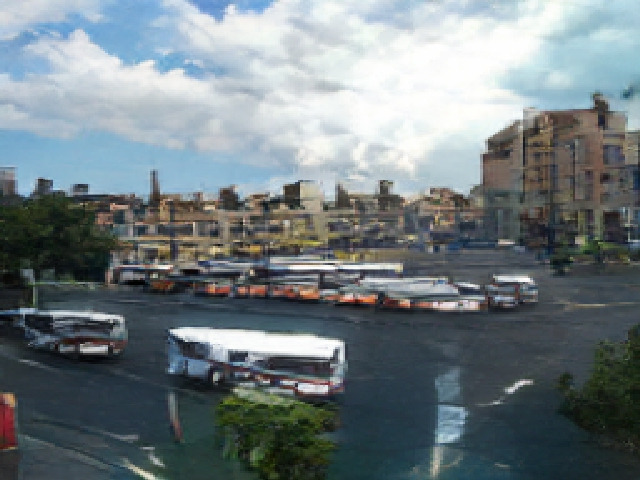}  & \includegraphics[width=\sizea]{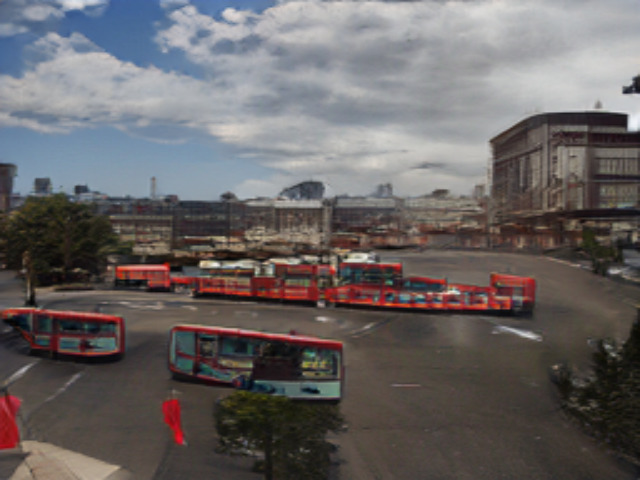}  & \includegraphics[width=\sizea]{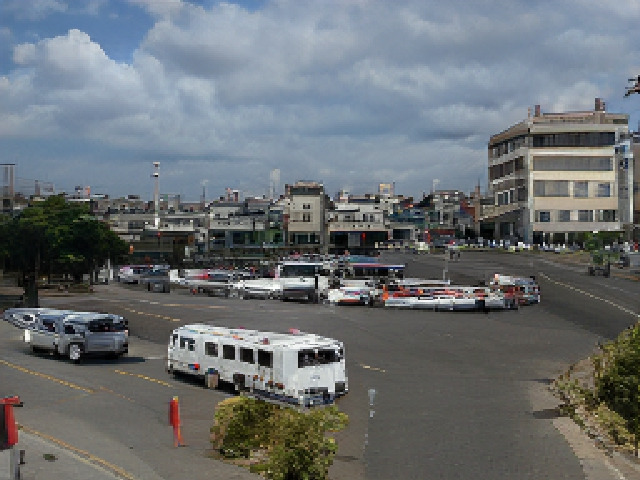} \\
		\includegraphics[width=\sizea]{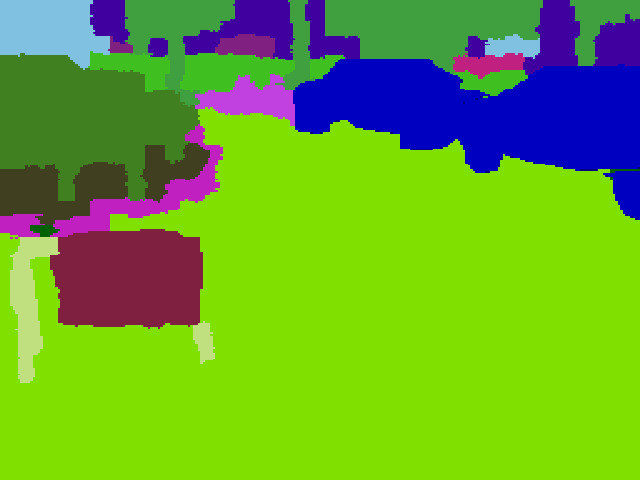}    & \includegraphics[width=\sizea]{figures/coco_seg/gt_000000463647.jpg}     & \includegraphics[width=\sizea]{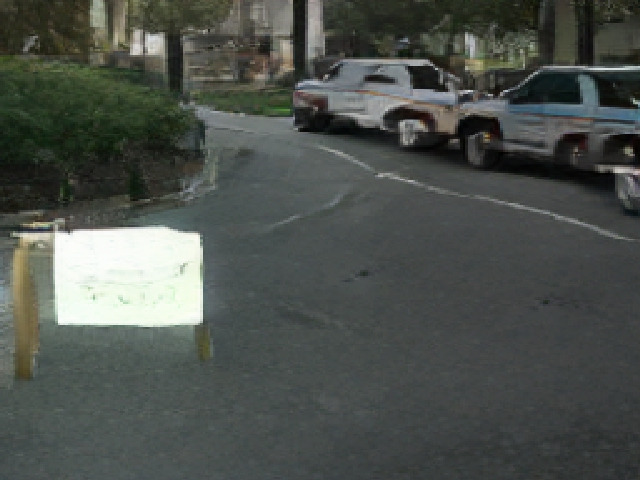}  & \includegraphics[width=\sizea]{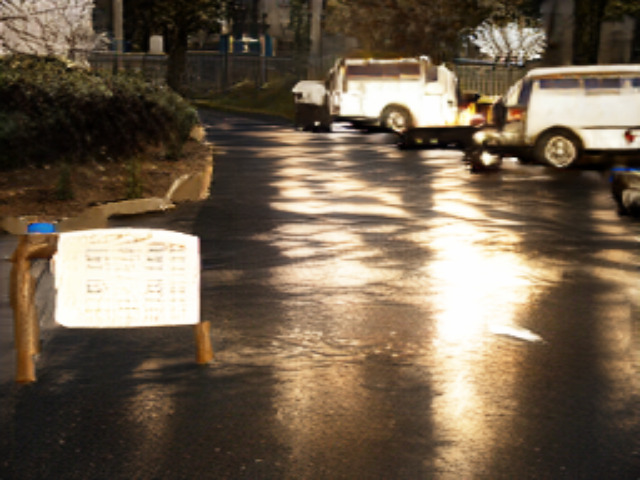}  & \includegraphics[width=\sizea]{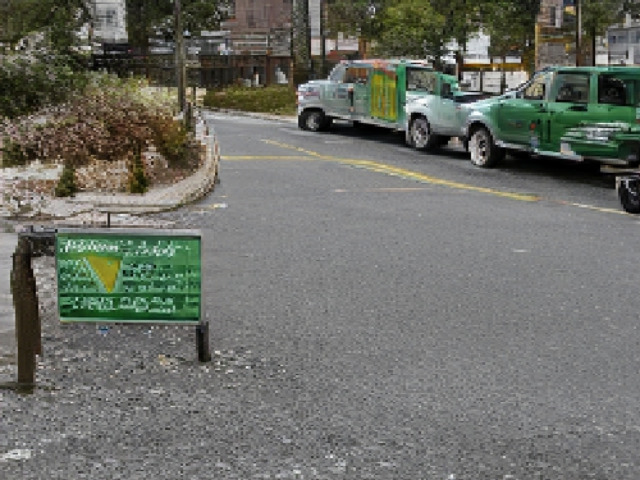} \\
		\includegraphics[width=\sizea]{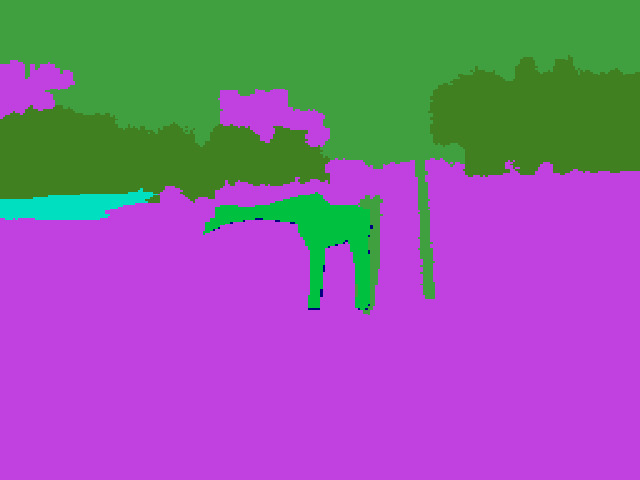}    & \includegraphics[width=\sizea]{figures/coco_seg/gt_000000459757.jpg}     & \includegraphics[width=\sizea]{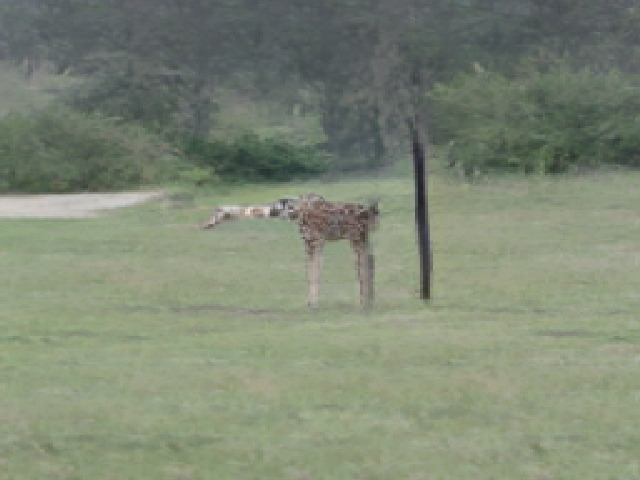}  & \includegraphics[width=\sizea]{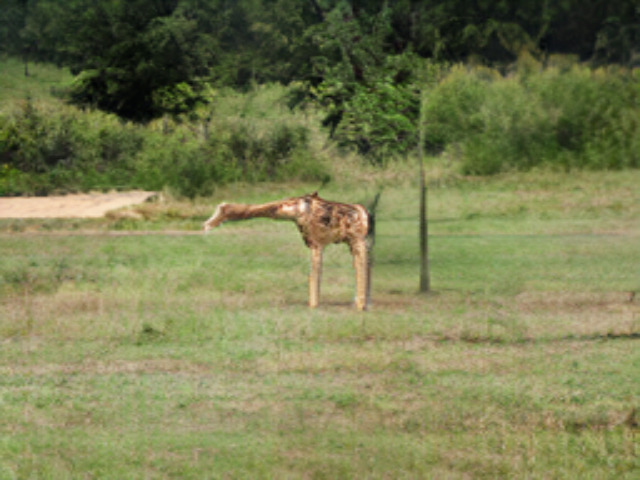}  & \includegraphics[width=\sizea]{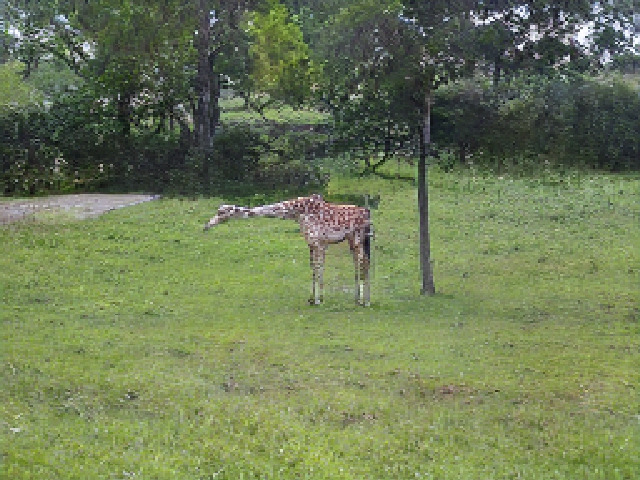} \\
		\includegraphics[width=\sizea]{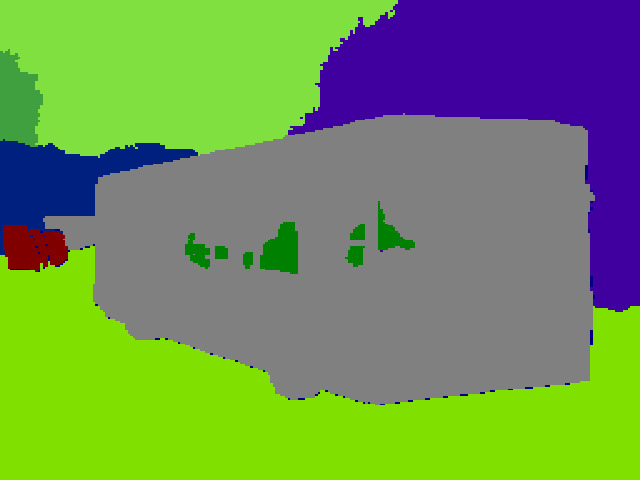}    & \includegraphics[width=\sizea]{figures/coco_seg/gt_000000344029.jpg}     & \includegraphics[width=\sizea]{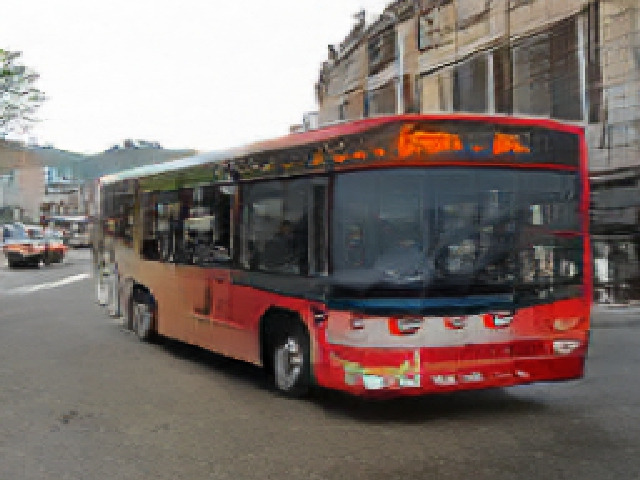}  & \includegraphics[width=\sizea]{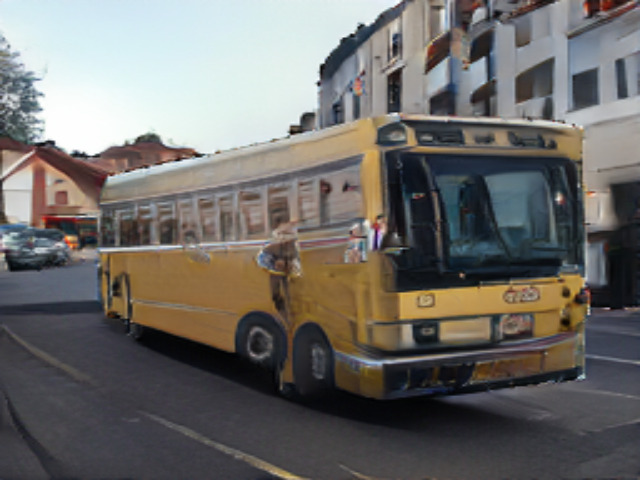}  & \includegraphics[width=\sizea]{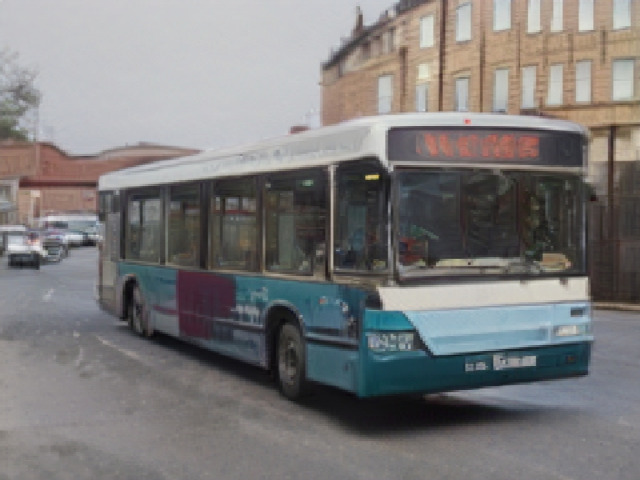} \\
		\includegraphics[width=\sizea]{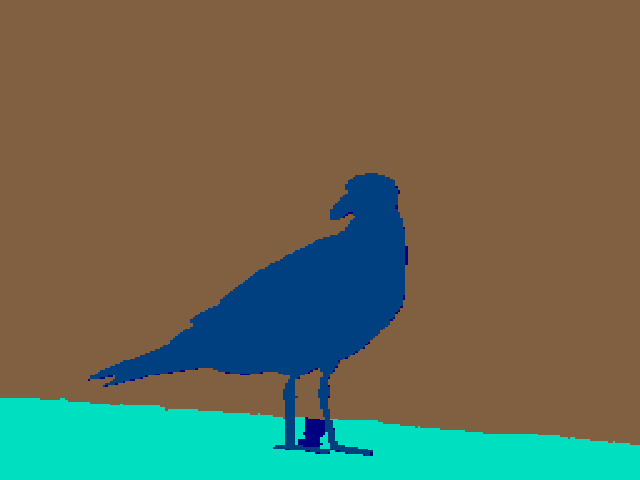}    & \includegraphics[width=\sizea]{figures/coco_seg/gt_000000404568.jpg}     & \includegraphics[width=\sizea]{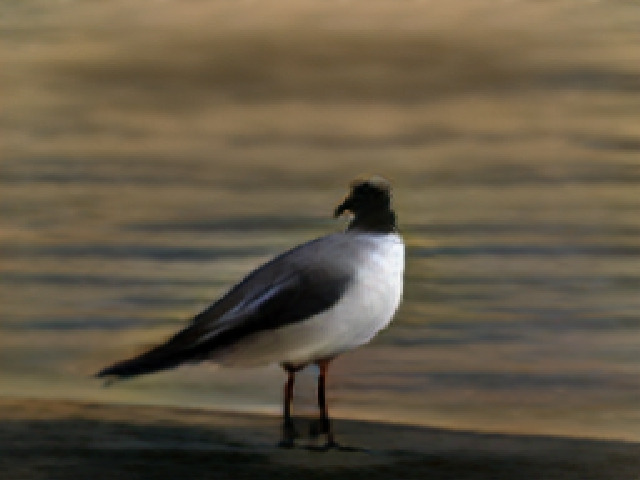}  & \includegraphics[width=\sizea]{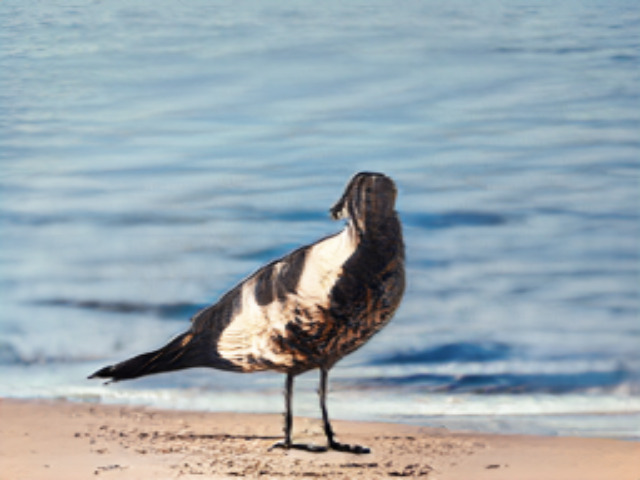}  & \includegraphics[width=\sizea]{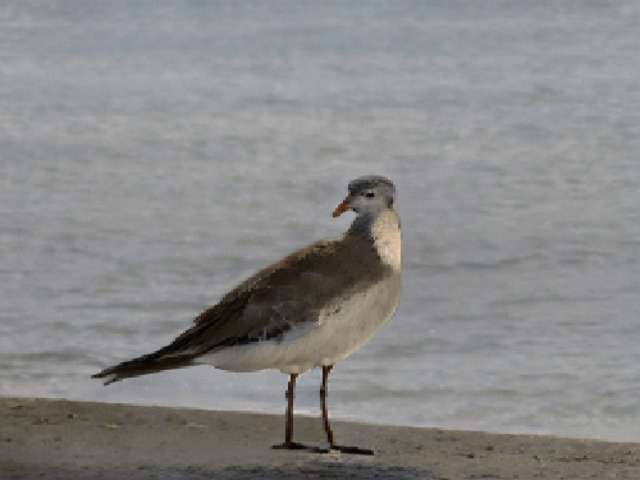} \\
		\includegraphics[width=\sizea]{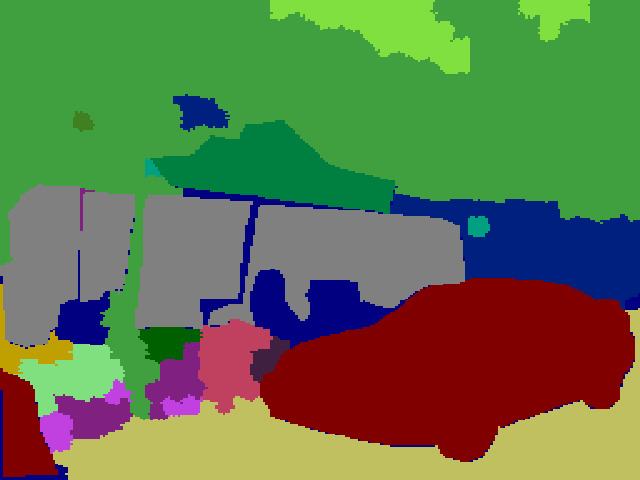}    & \includegraphics[width=\sizea]{figures/coco_seg/gt_000000543043.jpg}     & \includegraphics[width=\sizea]{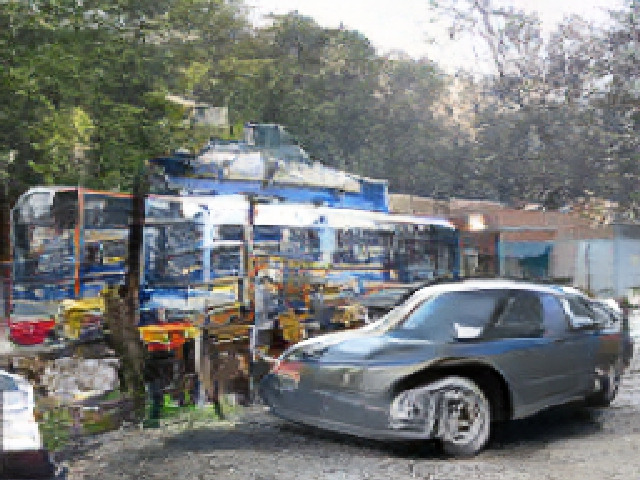}  & \includegraphics[width=\sizea]{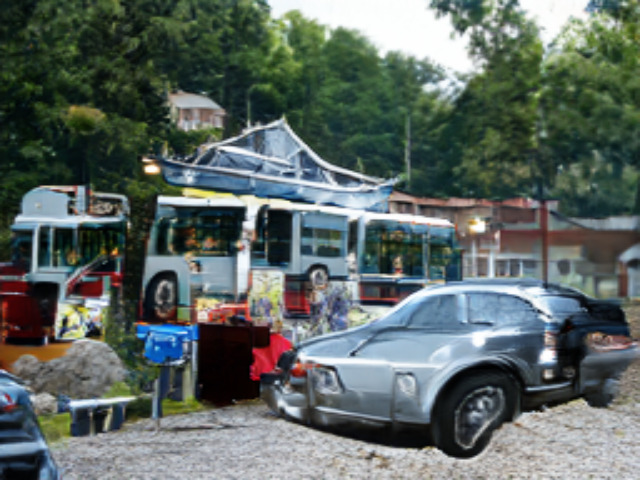}  & \includegraphics[width=\sizea]{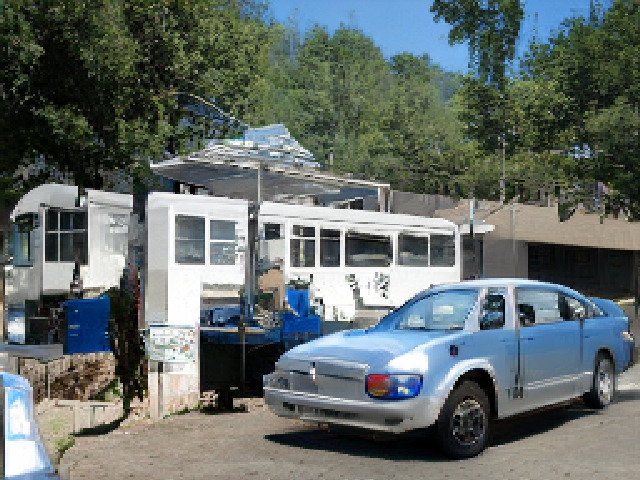} \\
		\includegraphics[width=\sizea]{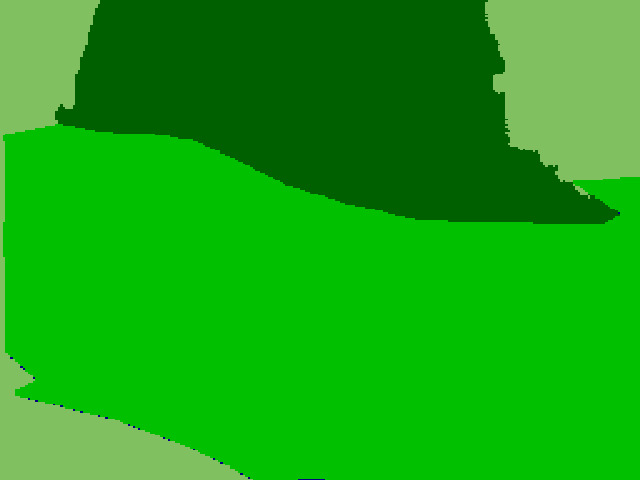}     & \includegraphics[width=\sizea]{figures/coco_seg/gt_000000475732.jpg}    & \includegraphics[width=\sizea]{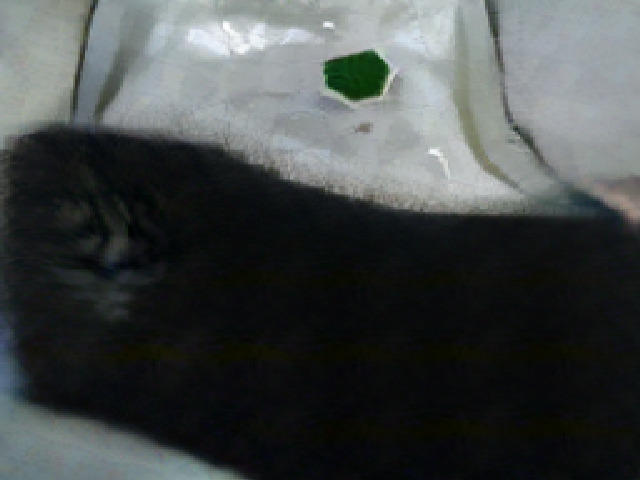}  & \includegraphics[width=\sizea]{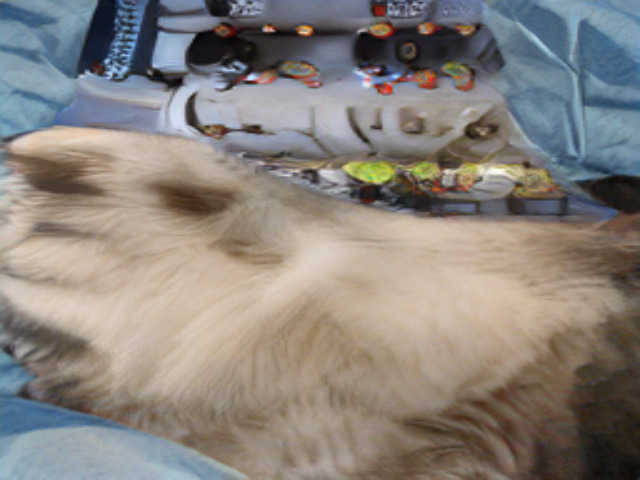}  & \includegraphics[width=\sizea]{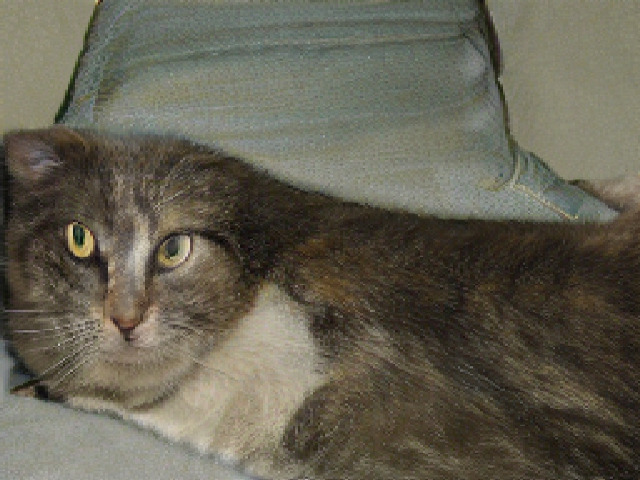} \\
		\includegraphics[width=\sizea]{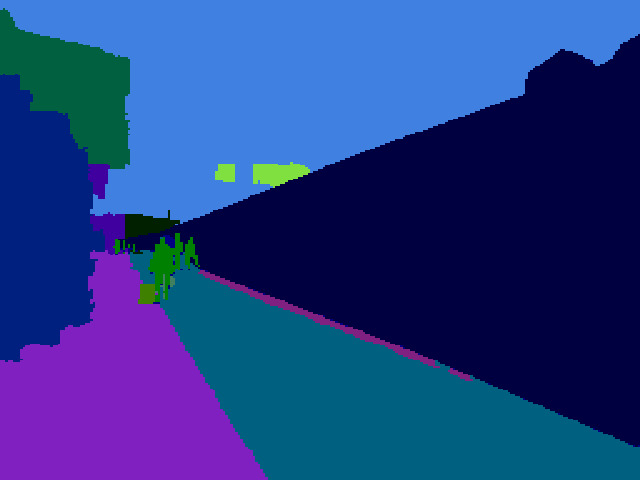}     & \includegraphics[width=\sizea]{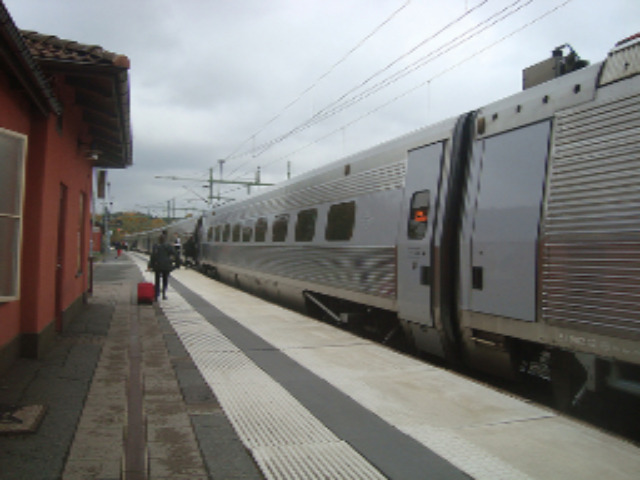}    & \includegraphics[width=\sizea]{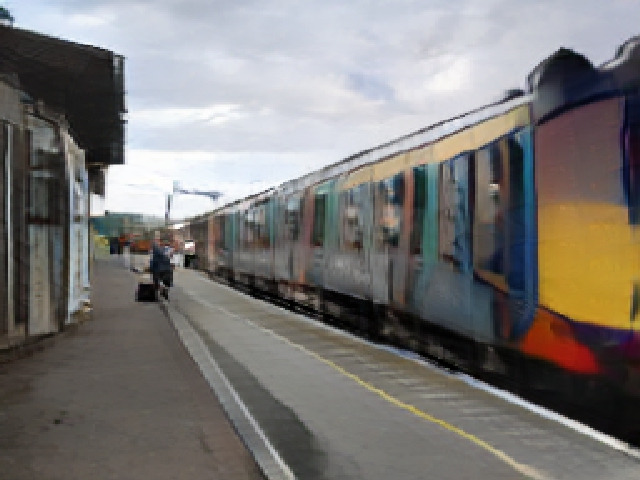}  & \includegraphics[width=\sizea]{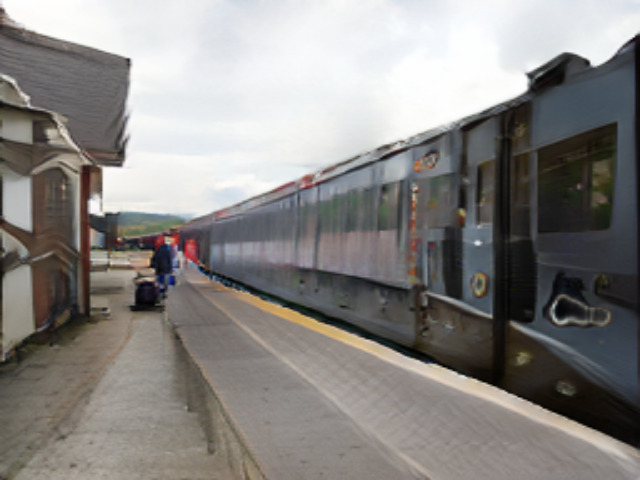}  & \includegraphics[width=\sizea]{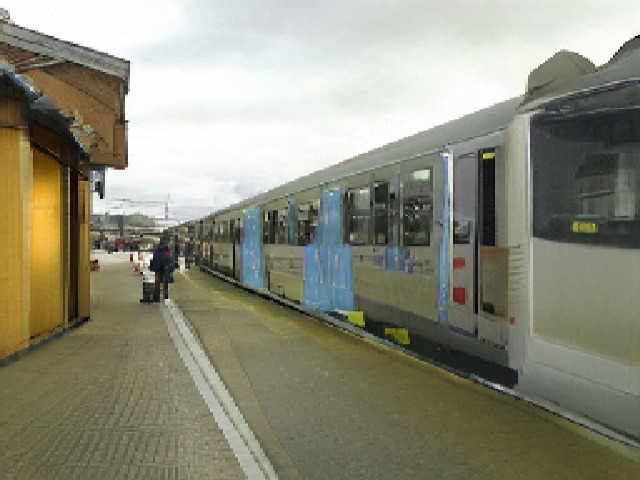}
	\end{tabular}
	\caption{Additional qualitative comparison of segmentation-to-image synthesis on MS-COCO 2017.}
	\label{fig:coco_seg_more}
\end{figure*}

\begin{figure*}[!t]
	\centering
	\newcommand{\sizea}{0.125\linewidth}
	\setlength{\lineskip}{0pt}
	\includegraphics[width=\sizea]{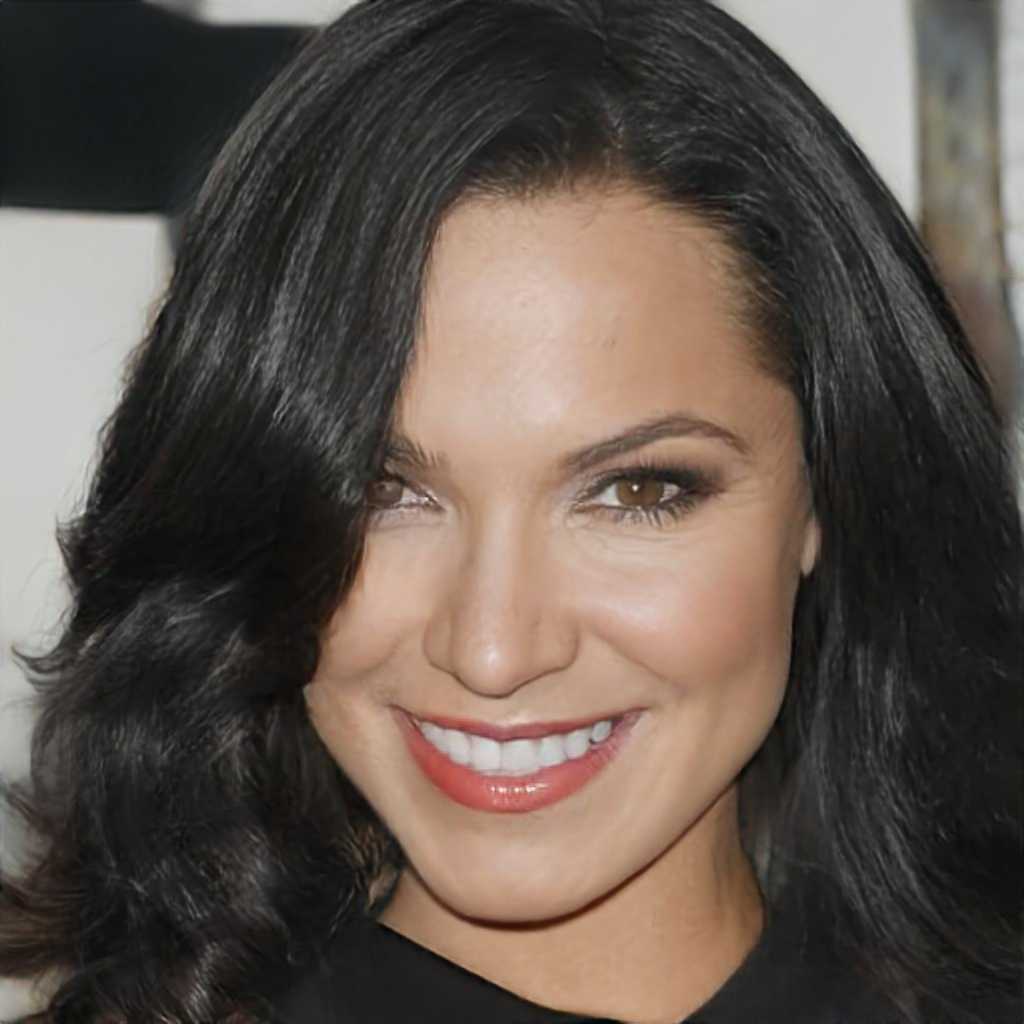}\linebreak[0]%
	\includegraphics[width=\sizea]{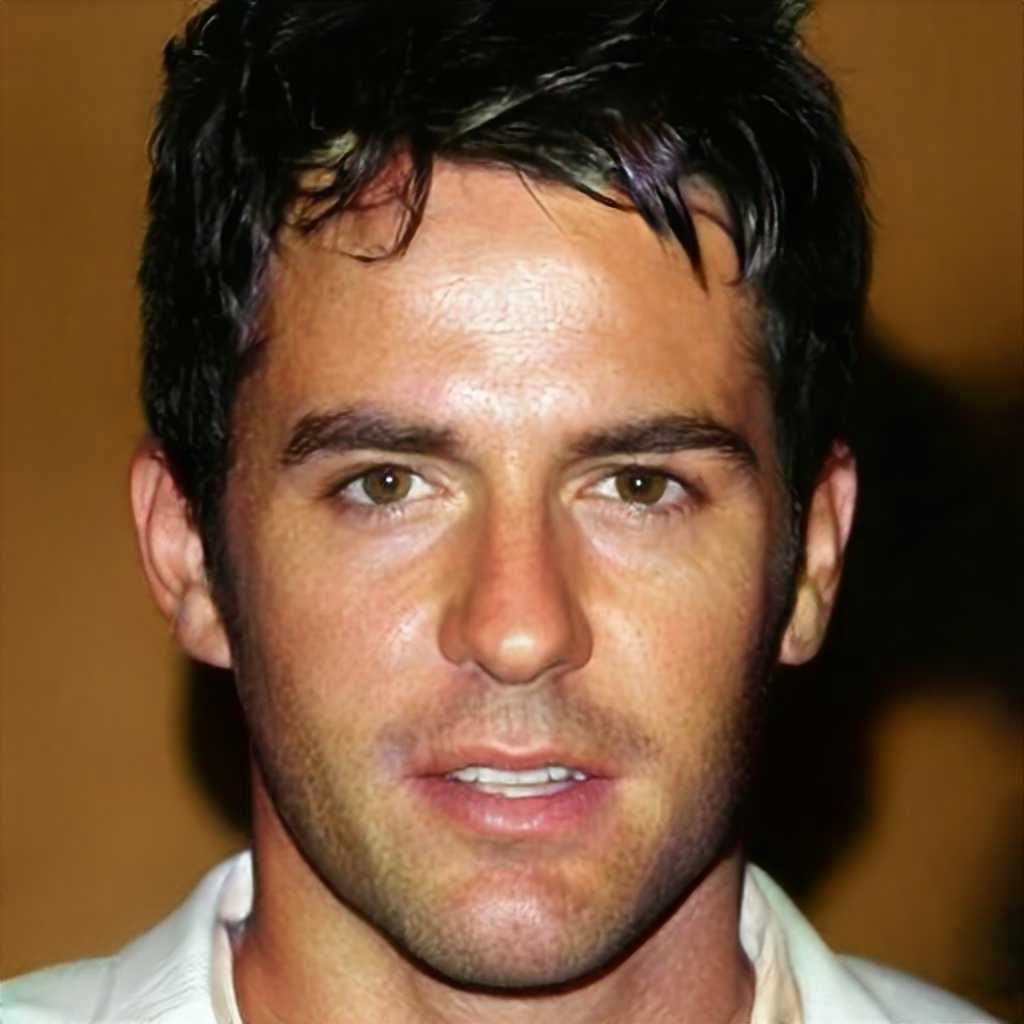}\linebreak[0]%
	\includegraphics[width=\sizea]{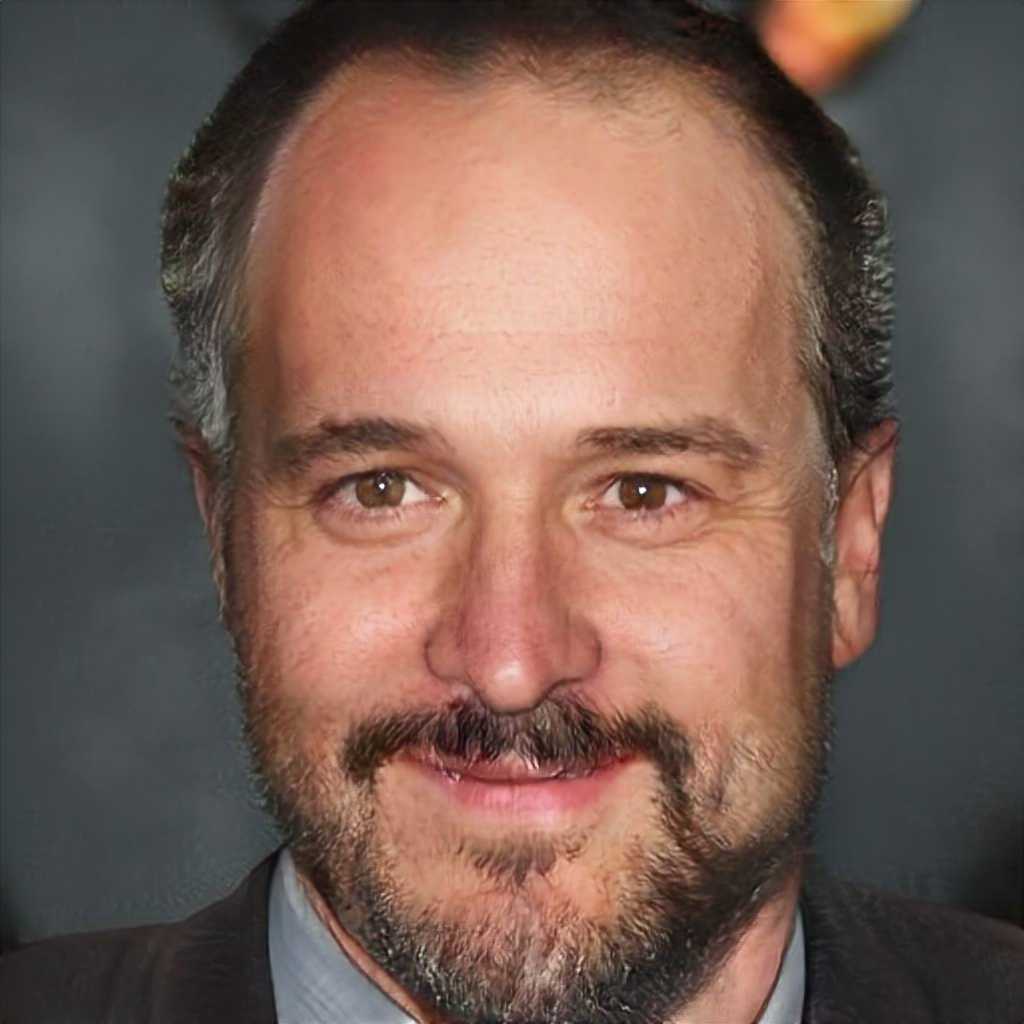}\linebreak[0]%
	\includegraphics[width=\sizea]{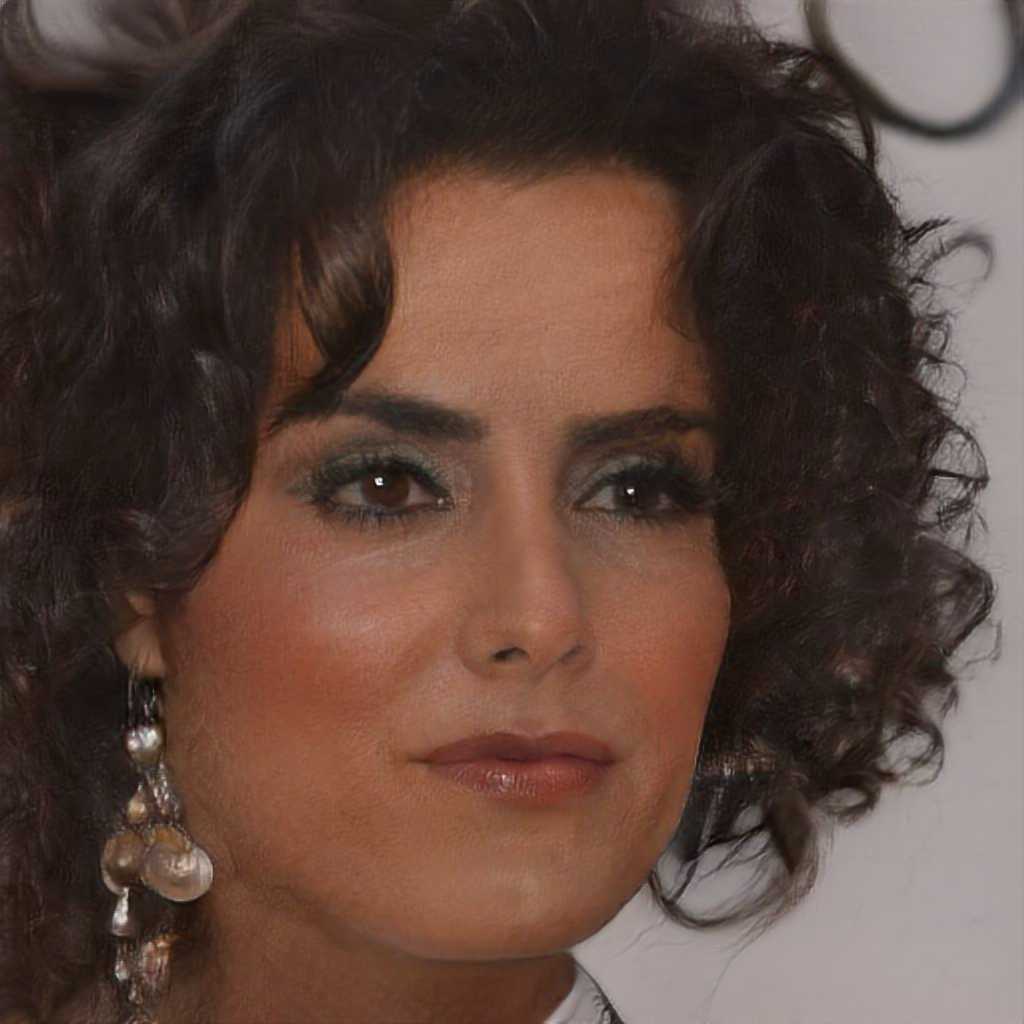}\linebreak[0]%
	\includegraphics[width=\sizea]{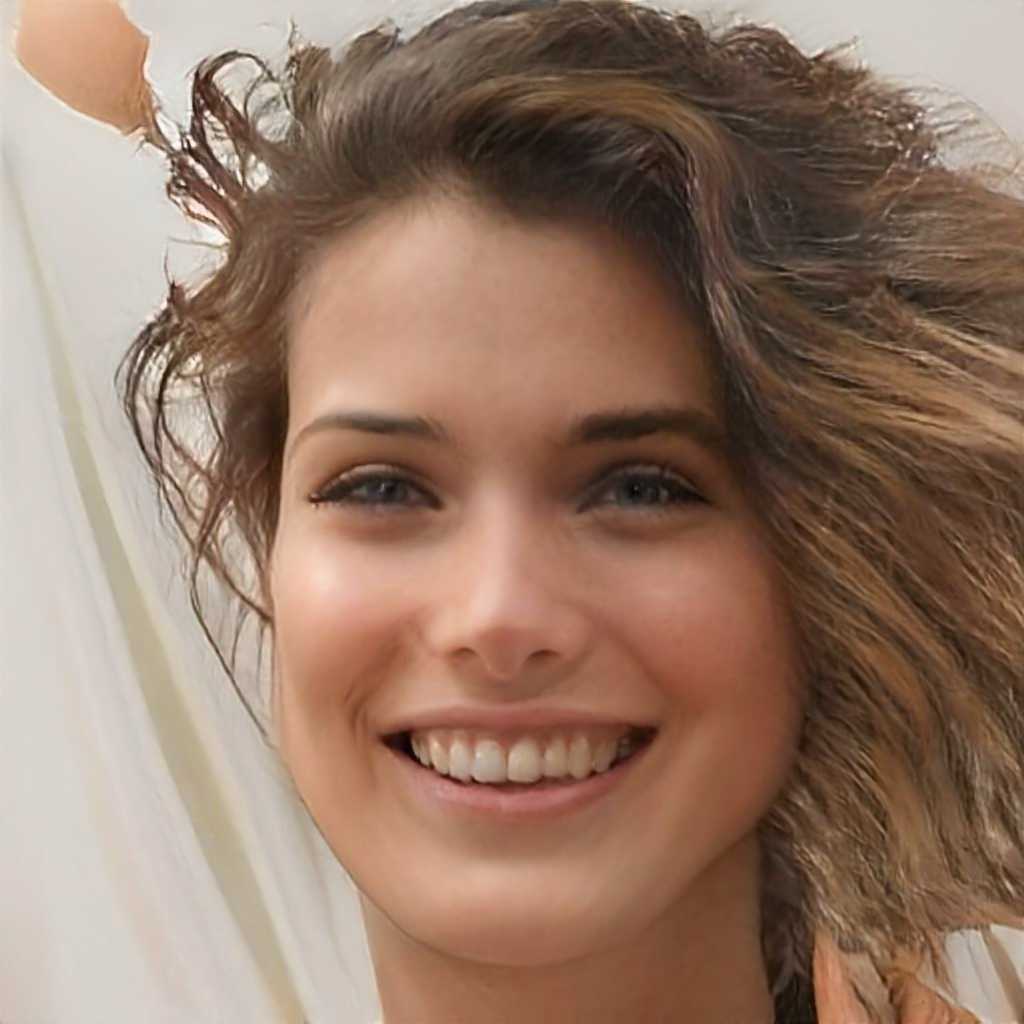}\linebreak[0]%
	\includegraphics[width=\sizea]{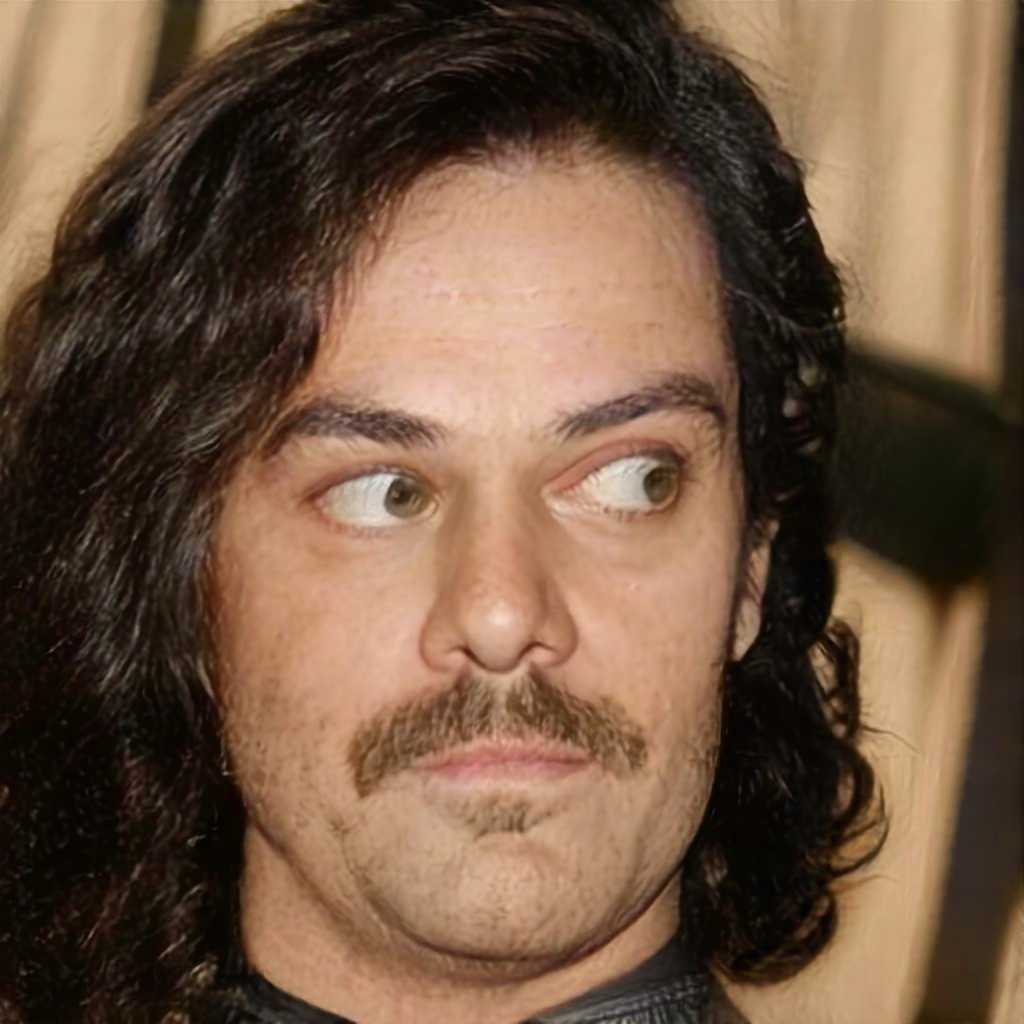}\linebreak[0]%
	\includegraphics[width=\sizea]{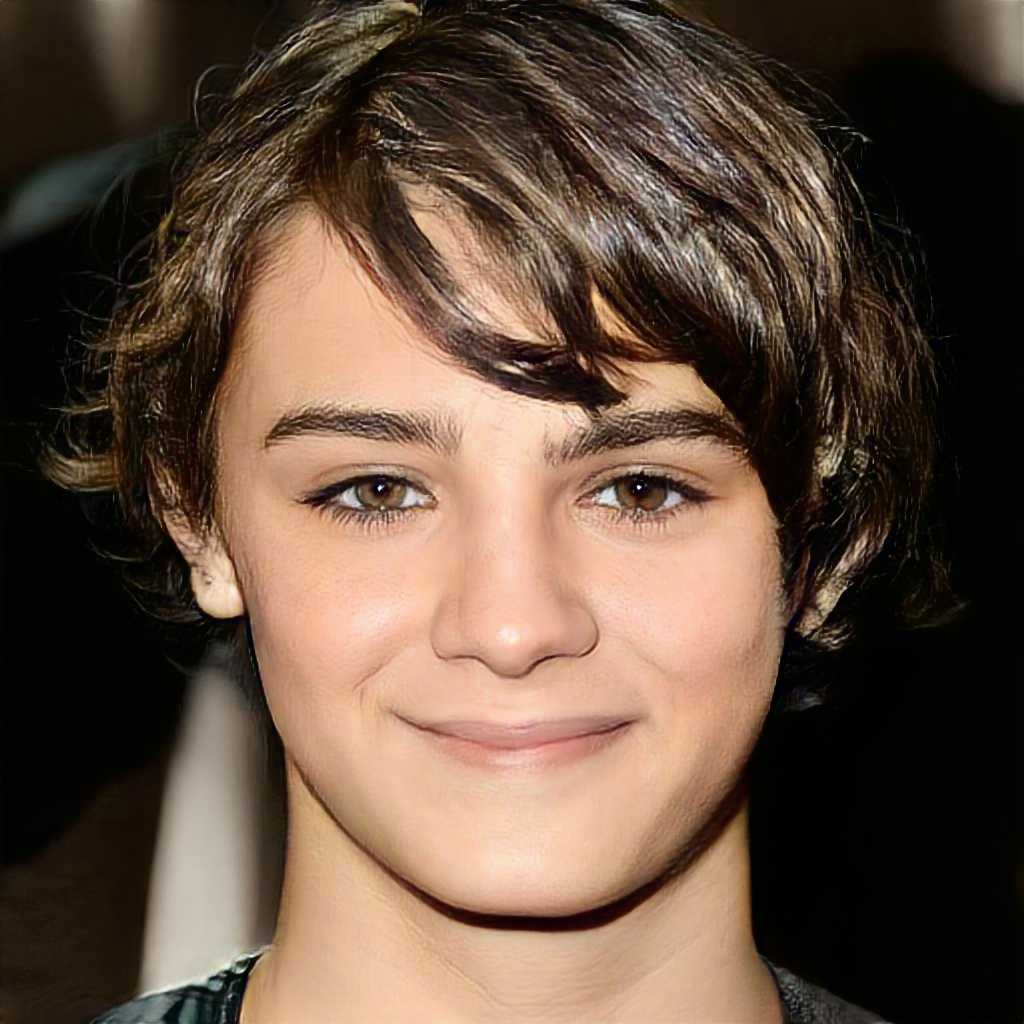}\linebreak[0]%
	\includegraphics[width=\sizea]{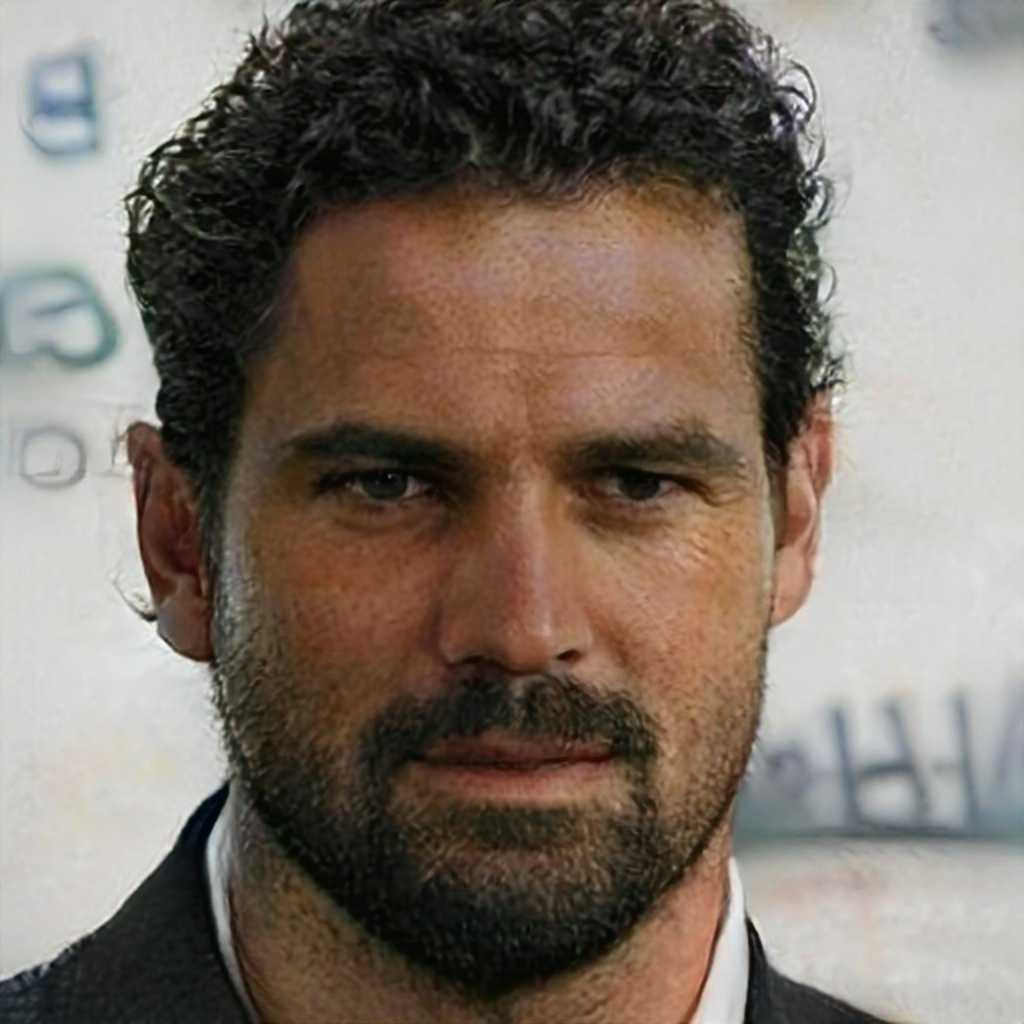}\linebreak[0]%
	\includegraphics[width=\sizea]{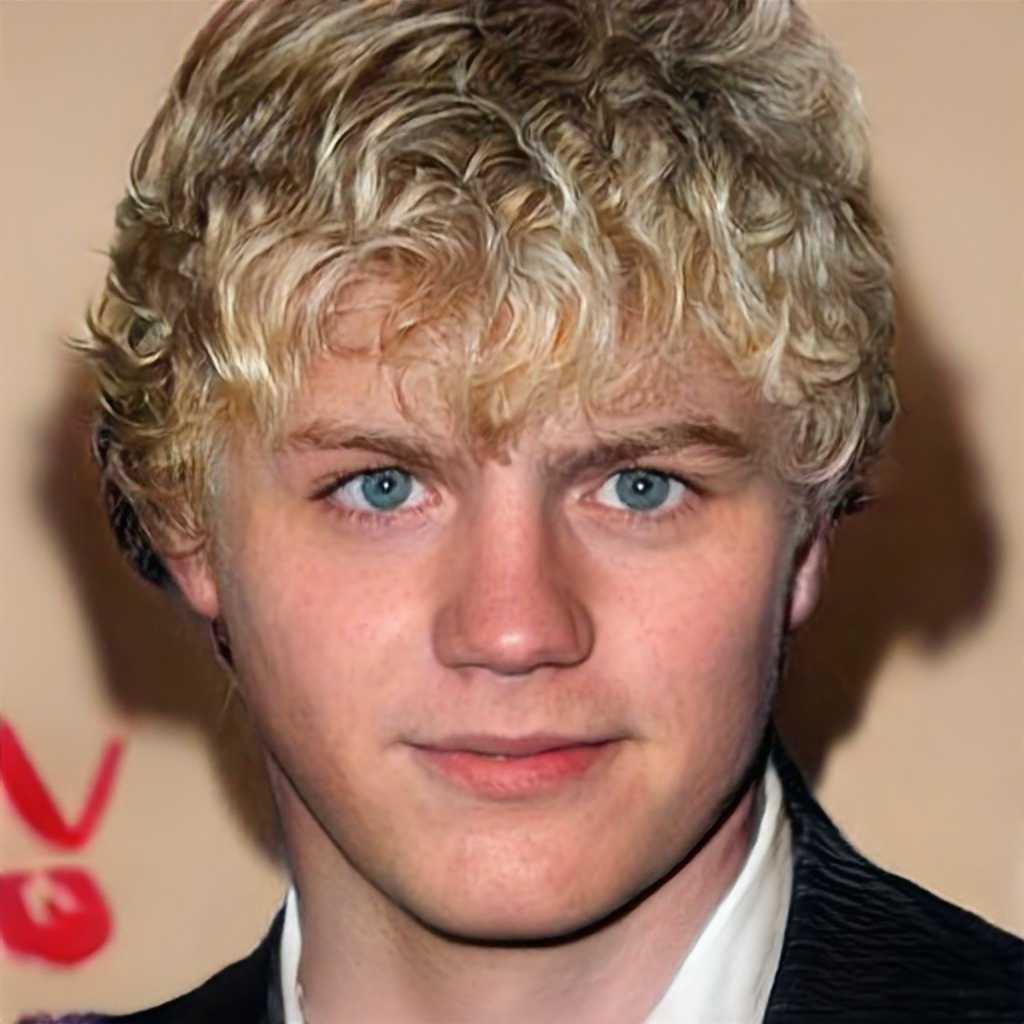}\linebreak[0]%
	\includegraphics[width=\sizea]{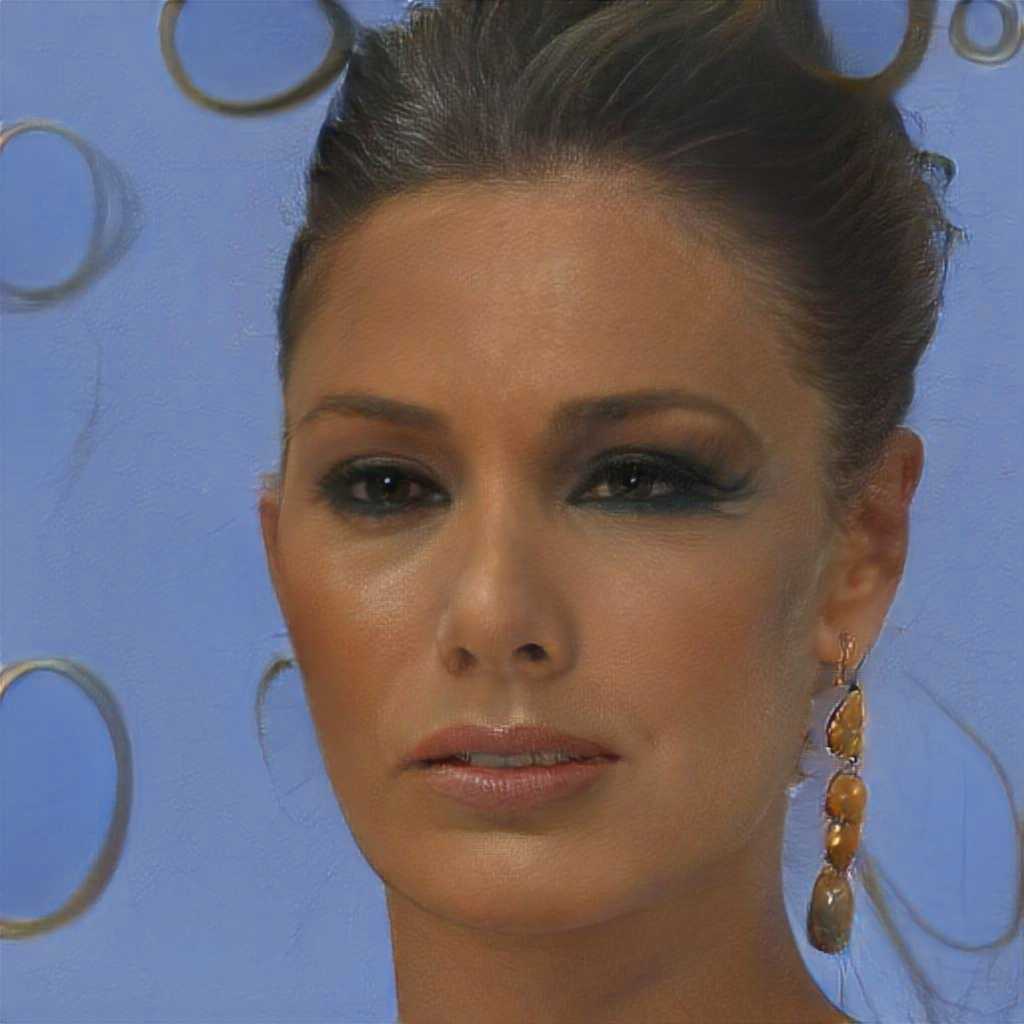}\linebreak[0]%
	\includegraphics[width=\sizea]{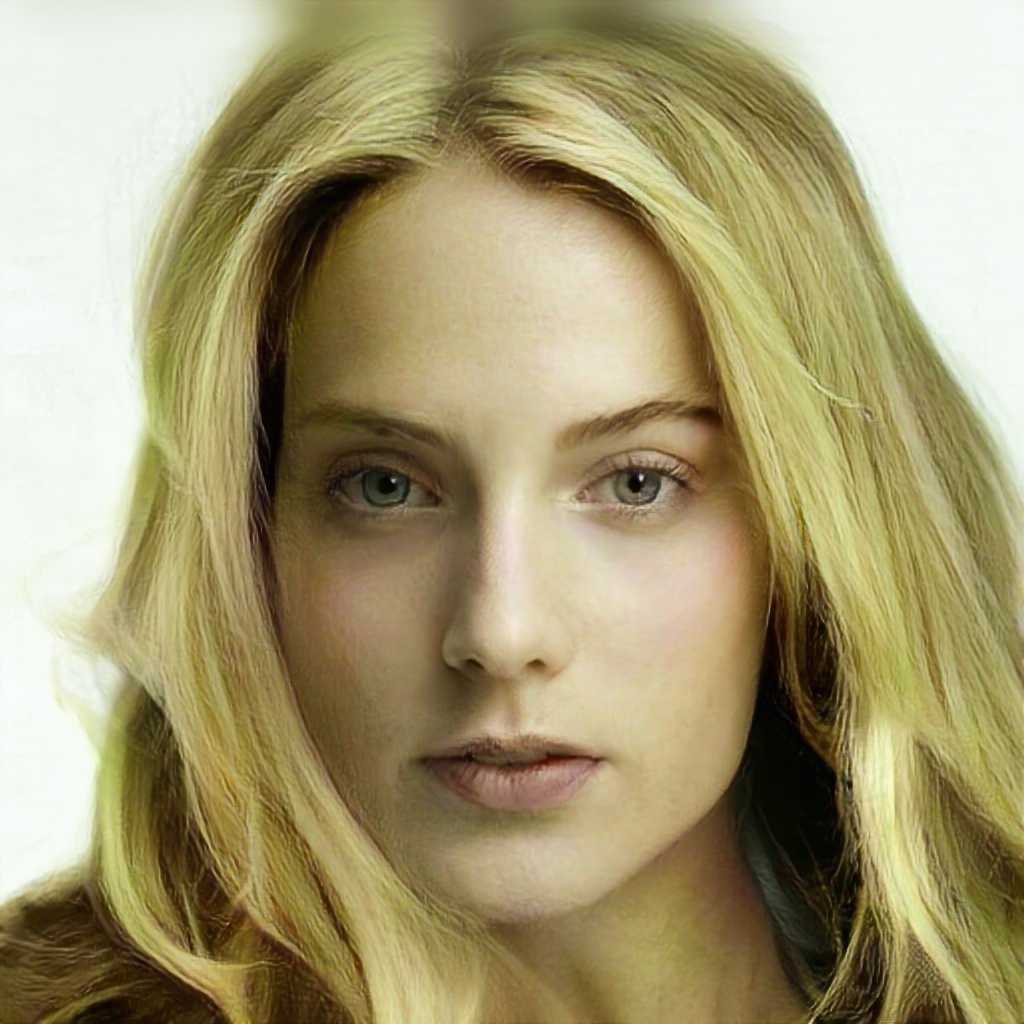}\linebreak[0]%
	\includegraphics[width=\sizea]{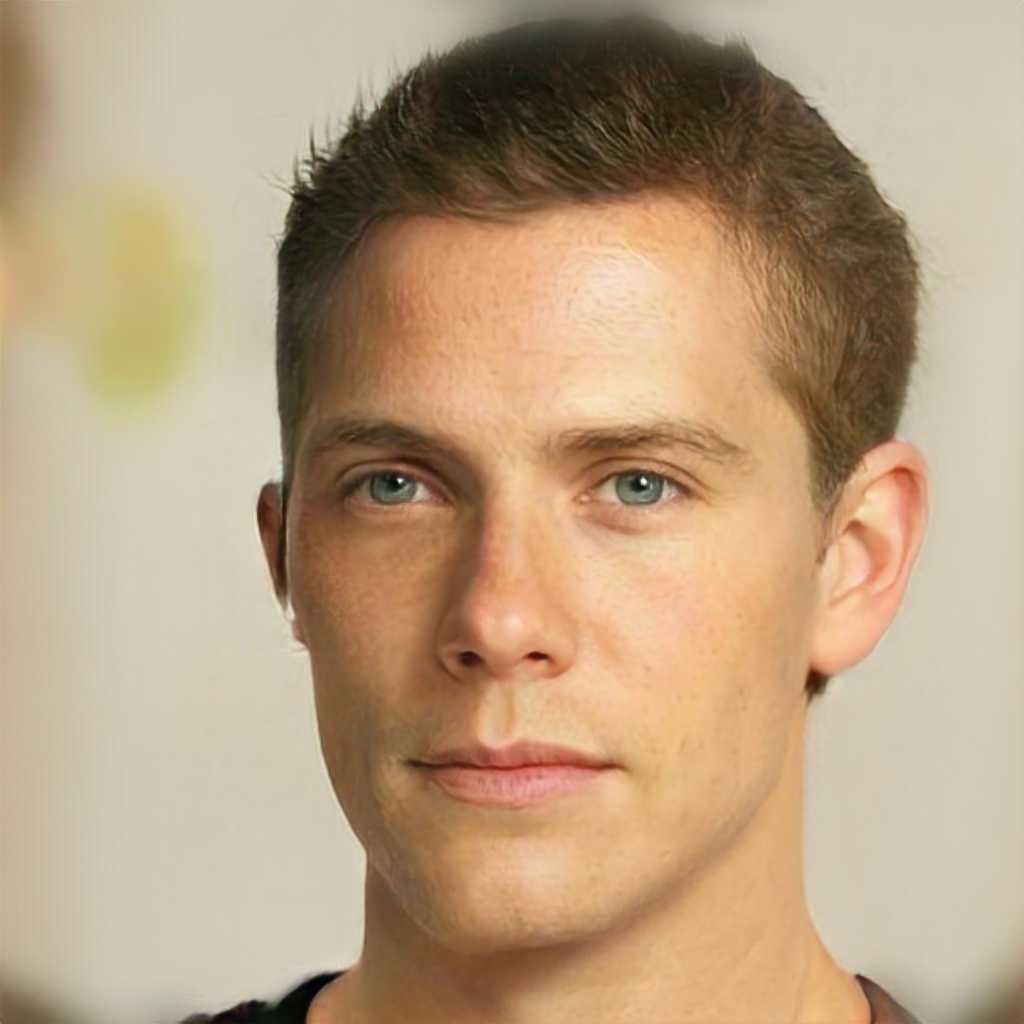}\linebreak[0]%
	\includegraphics[width=\sizea]{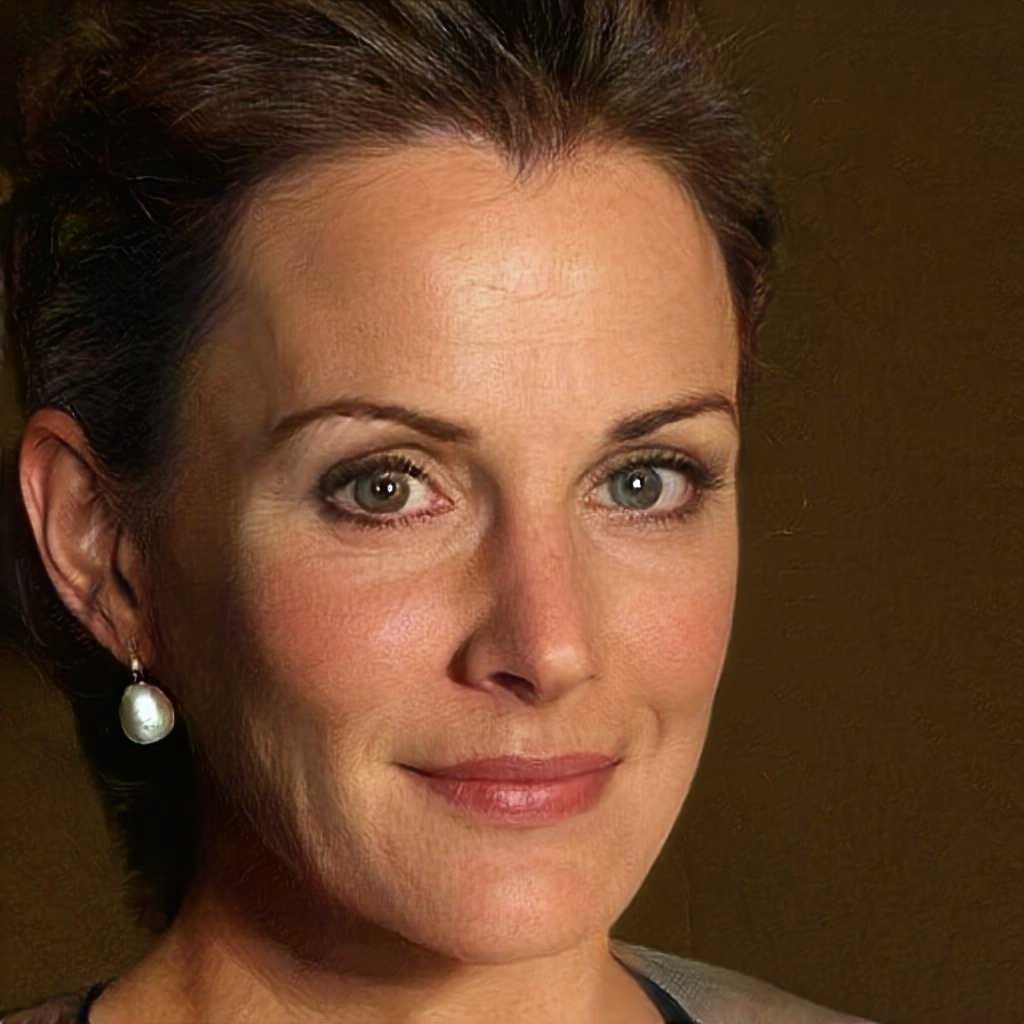}\linebreak[0]%
	\includegraphics[width=\sizea]{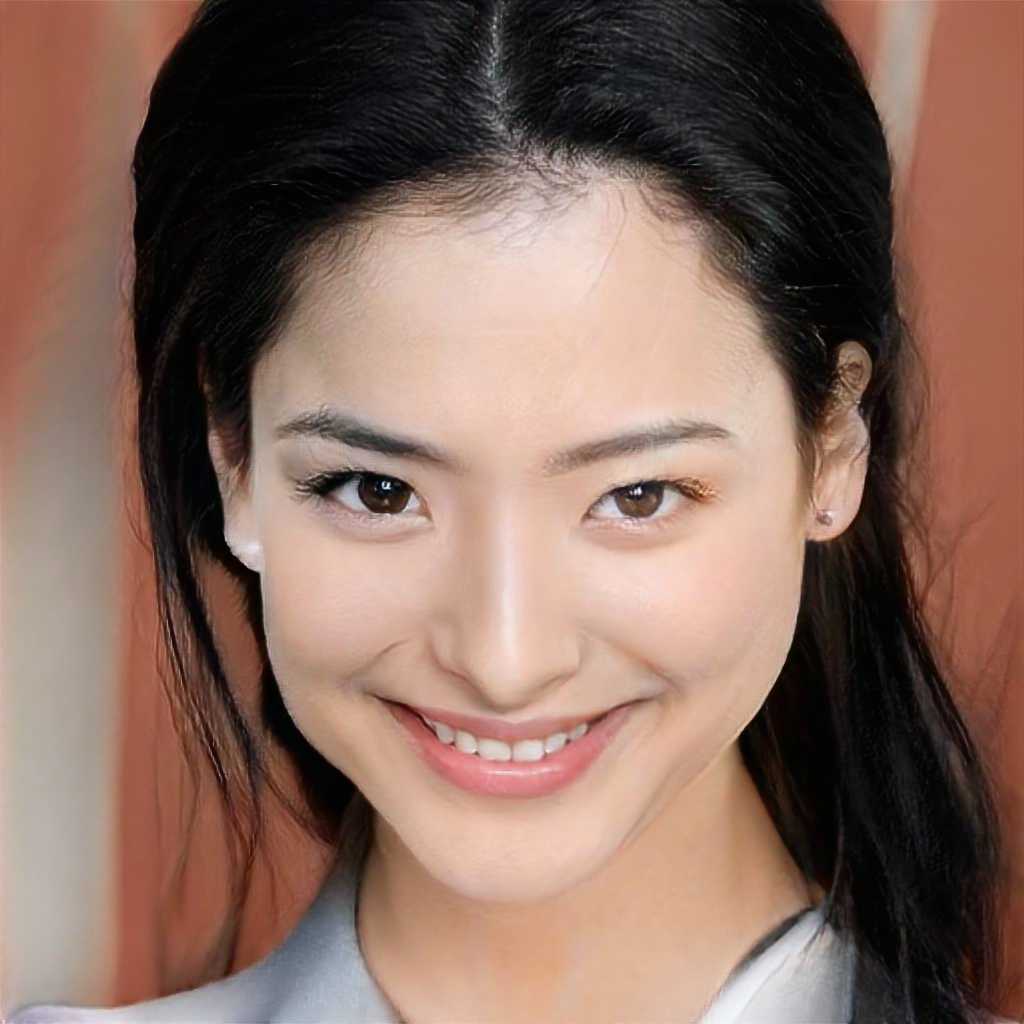}\linebreak[0]%
	\includegraphics[width=\sizea]{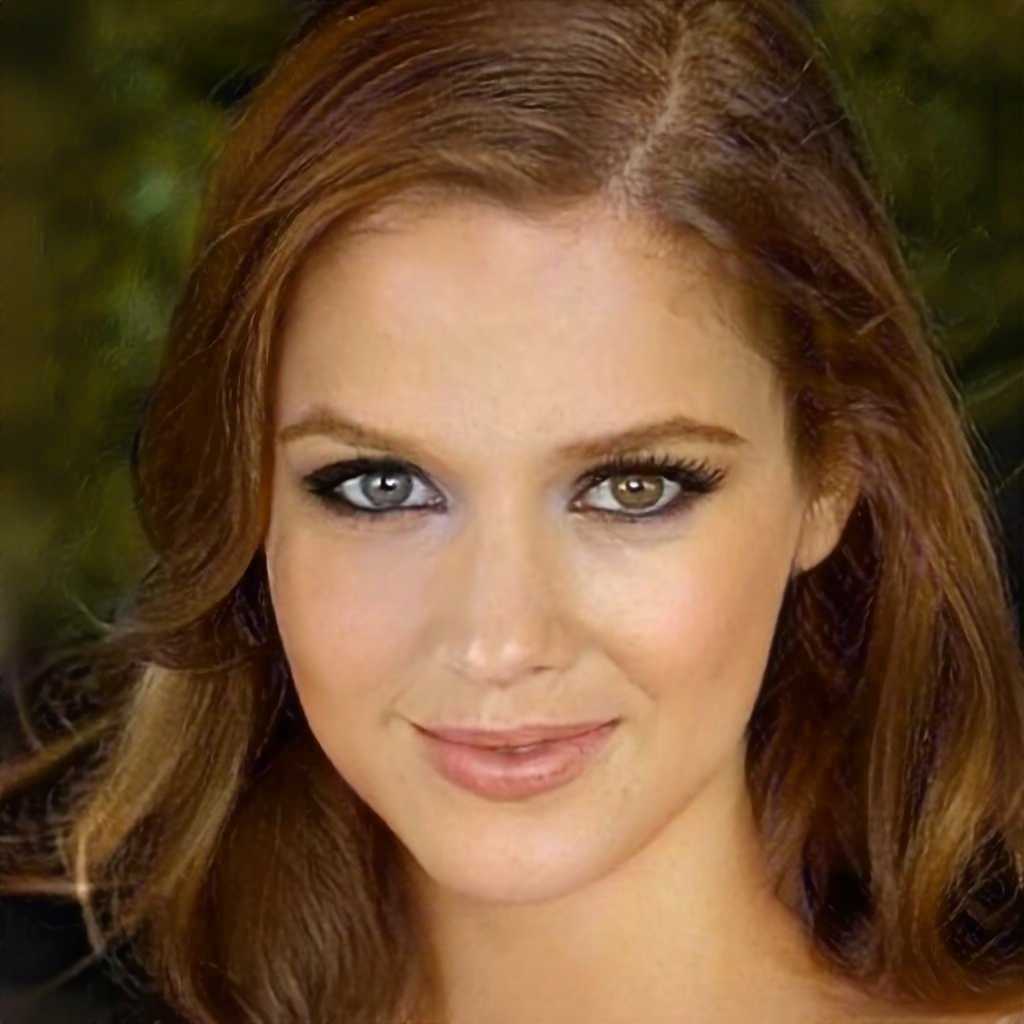}\linebreak[0]%
	\includegraphics[width=\sizea]{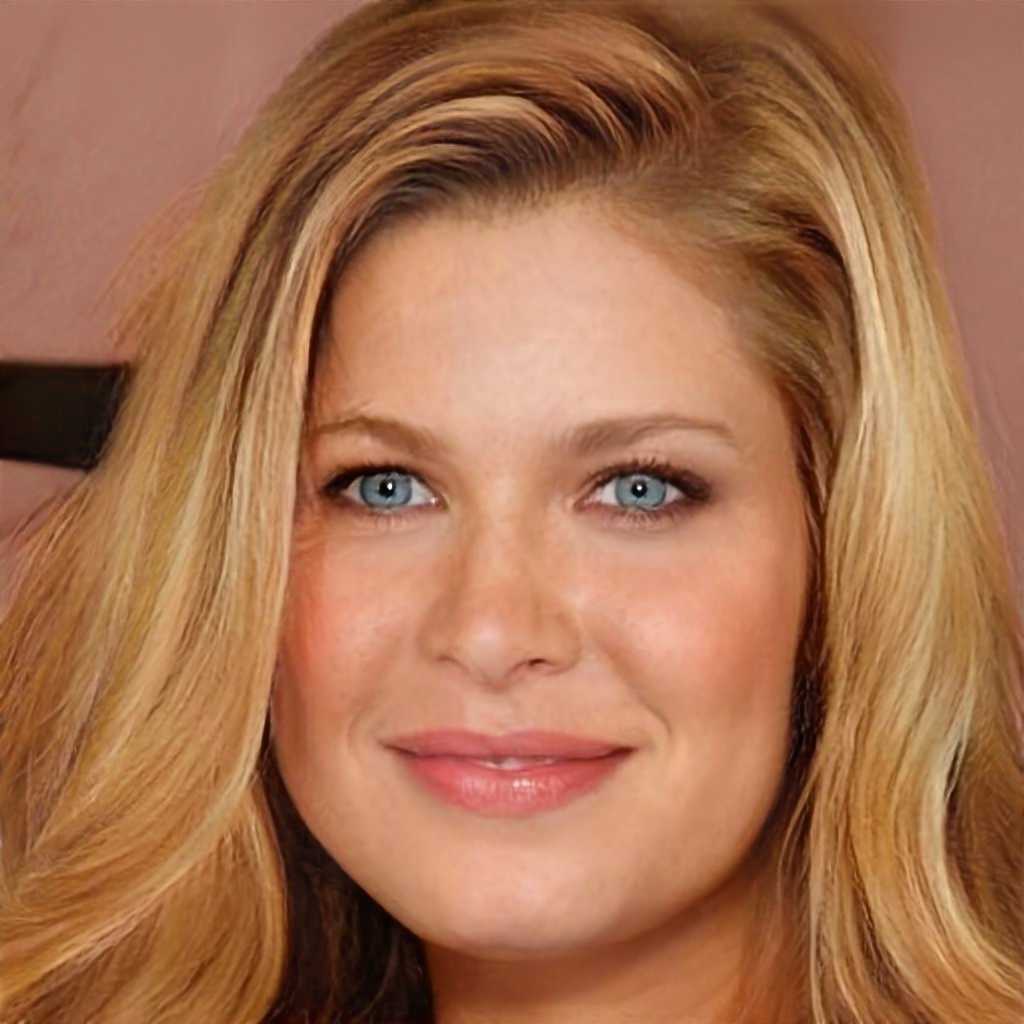}\linebreak[0]%
	\includegraphics[width=\sizea]{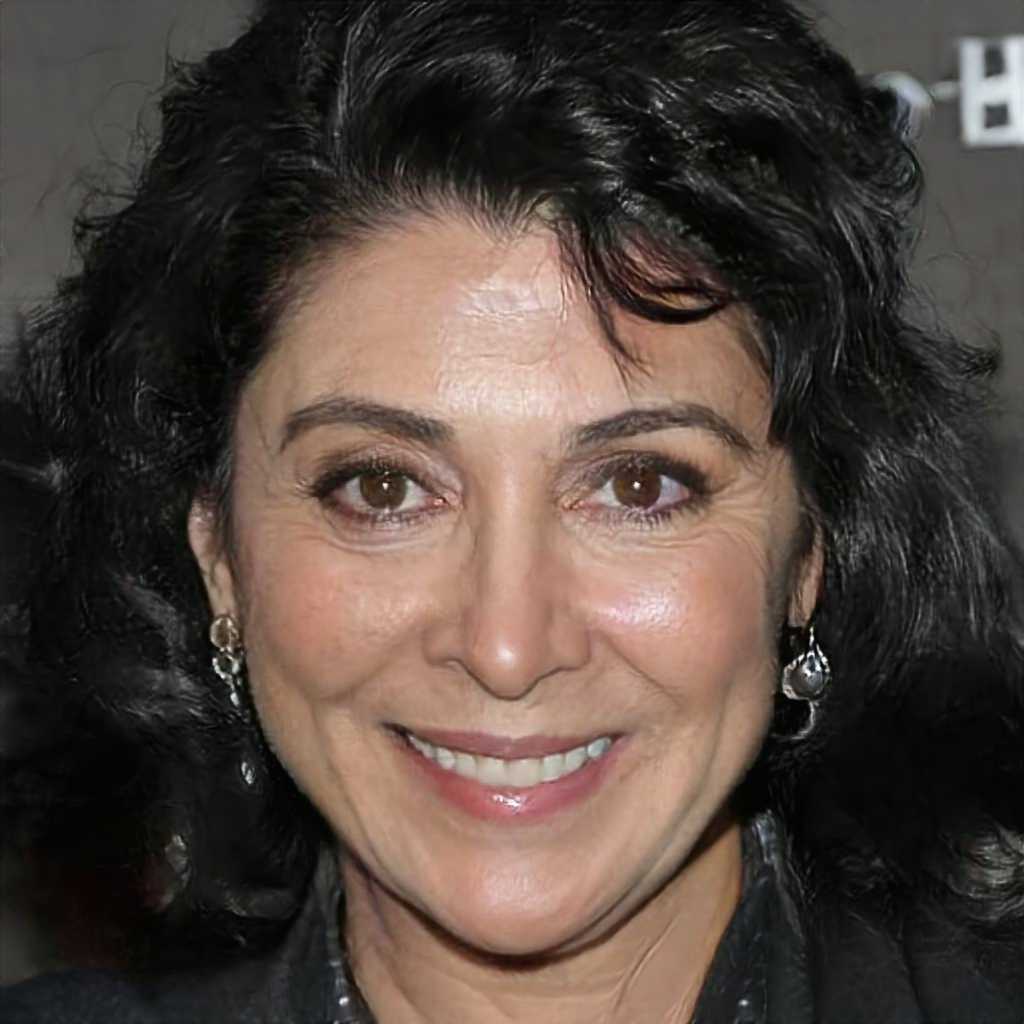}\linebreak[0]%
	\includegraphics[width=\sizea]{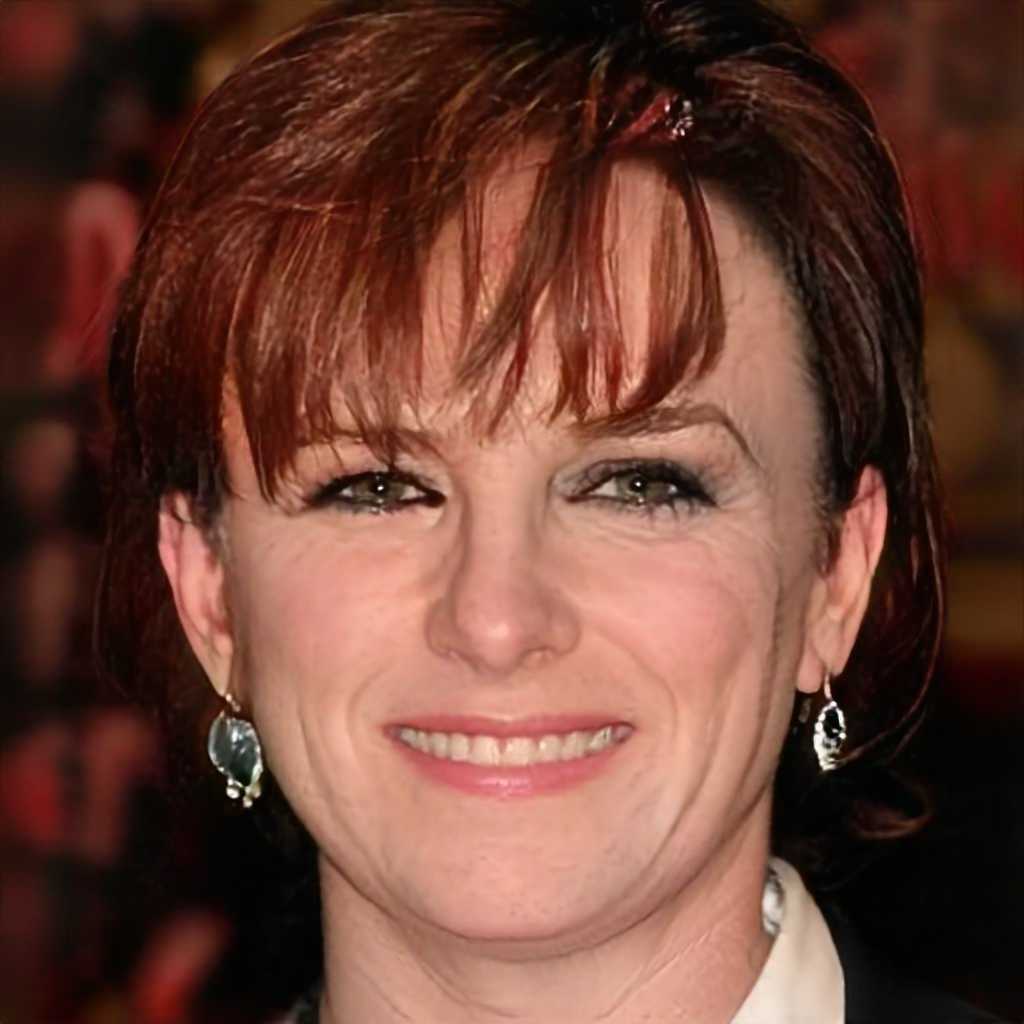}\linebreak[0]%
	\includegraphics[width=\sizea]{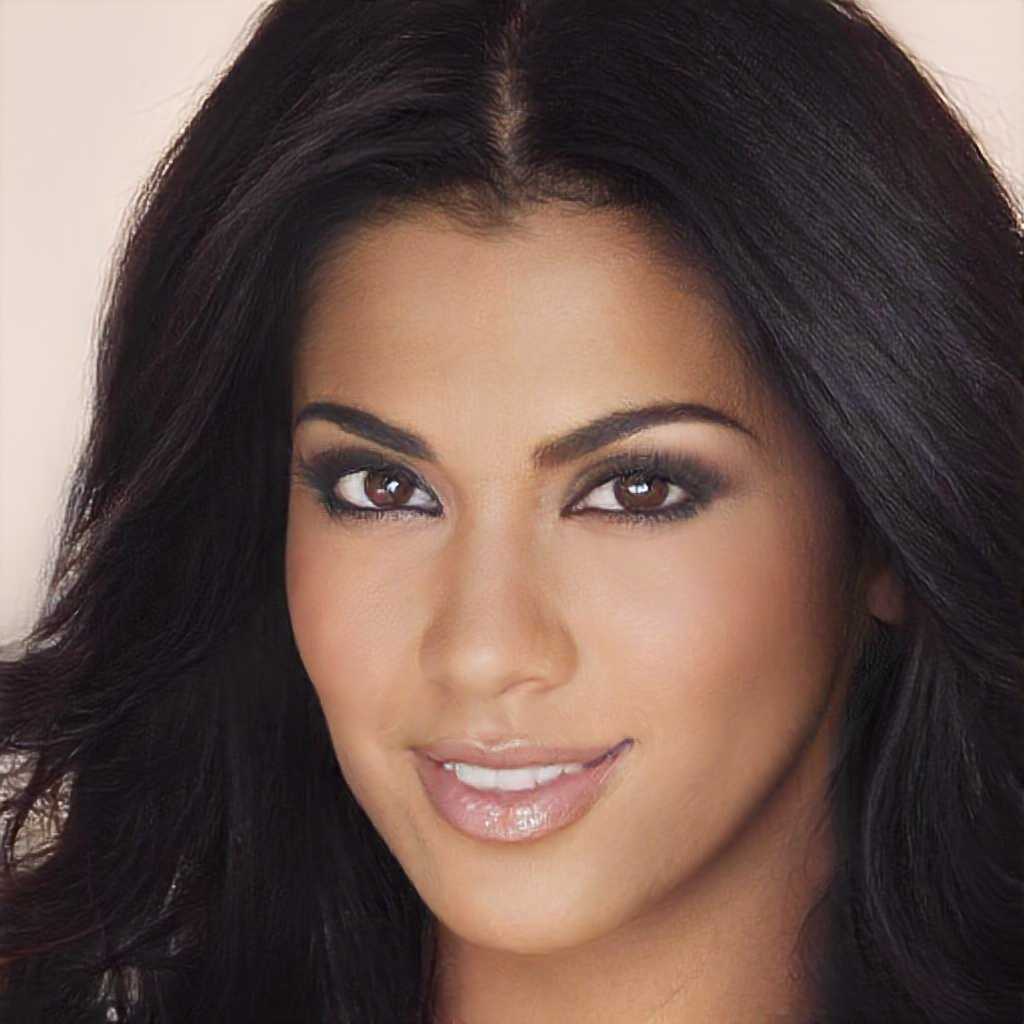}\linebreak[0]%
	\includegraphics[width=\sizea]{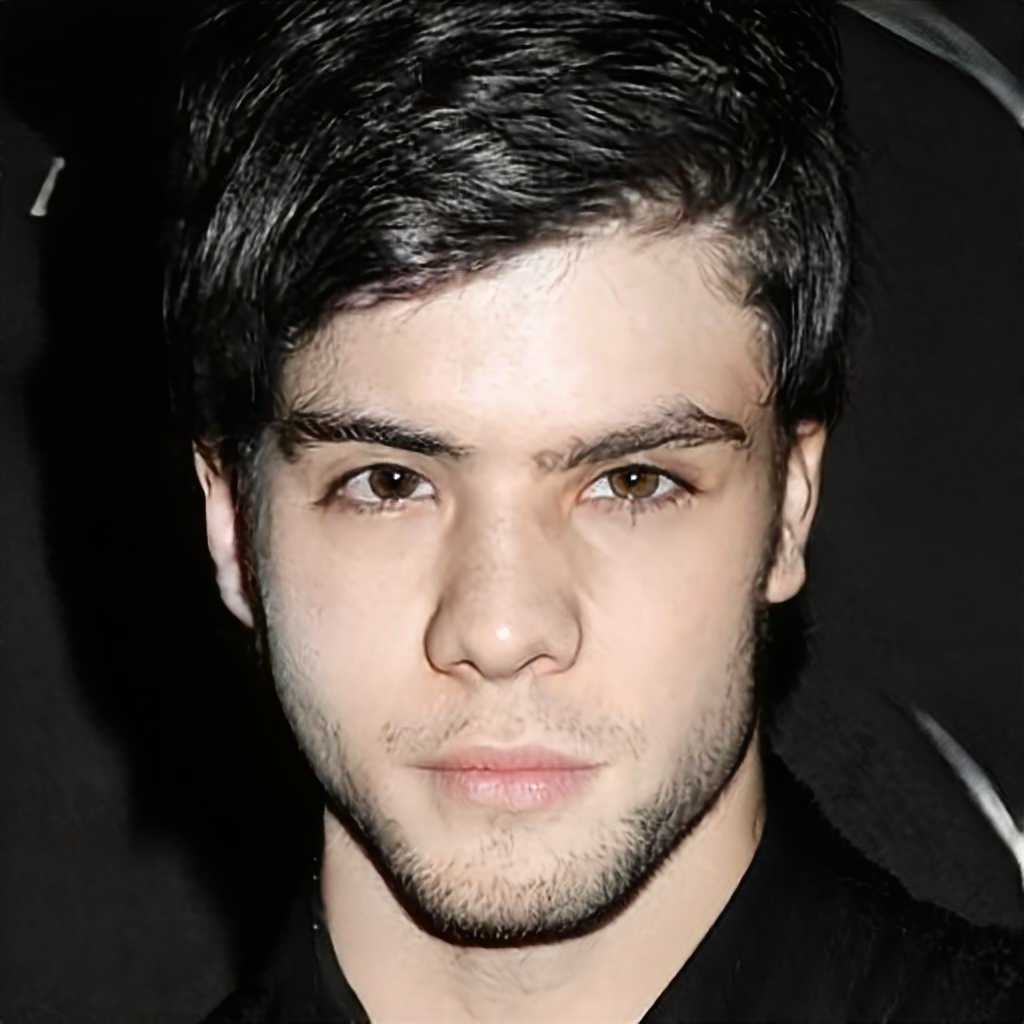}\linebreak[0]%
	\includegraphics[width=\sizea]{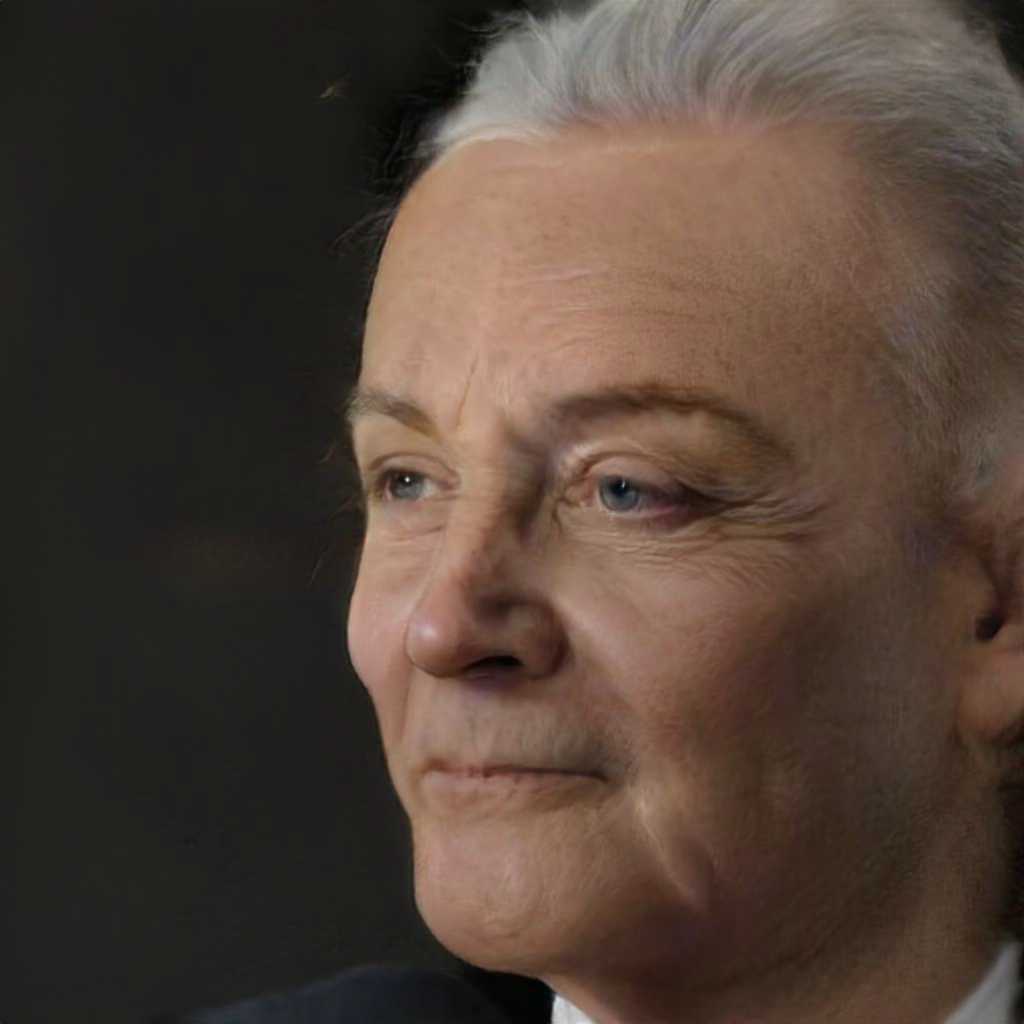}\linebreak[0]%
	\includegraphics[width=\sizea]{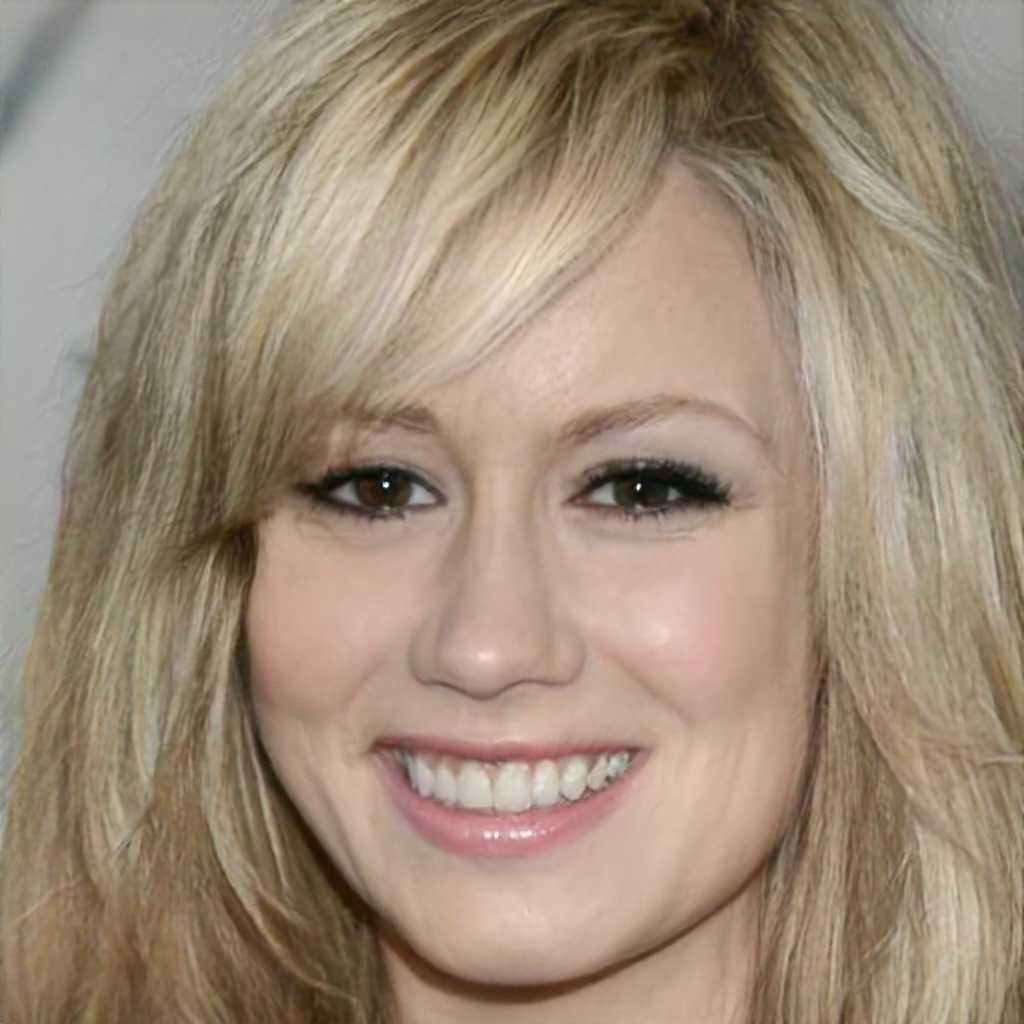}\linebreak[0]%
	\includegraphics[width=\sizea]{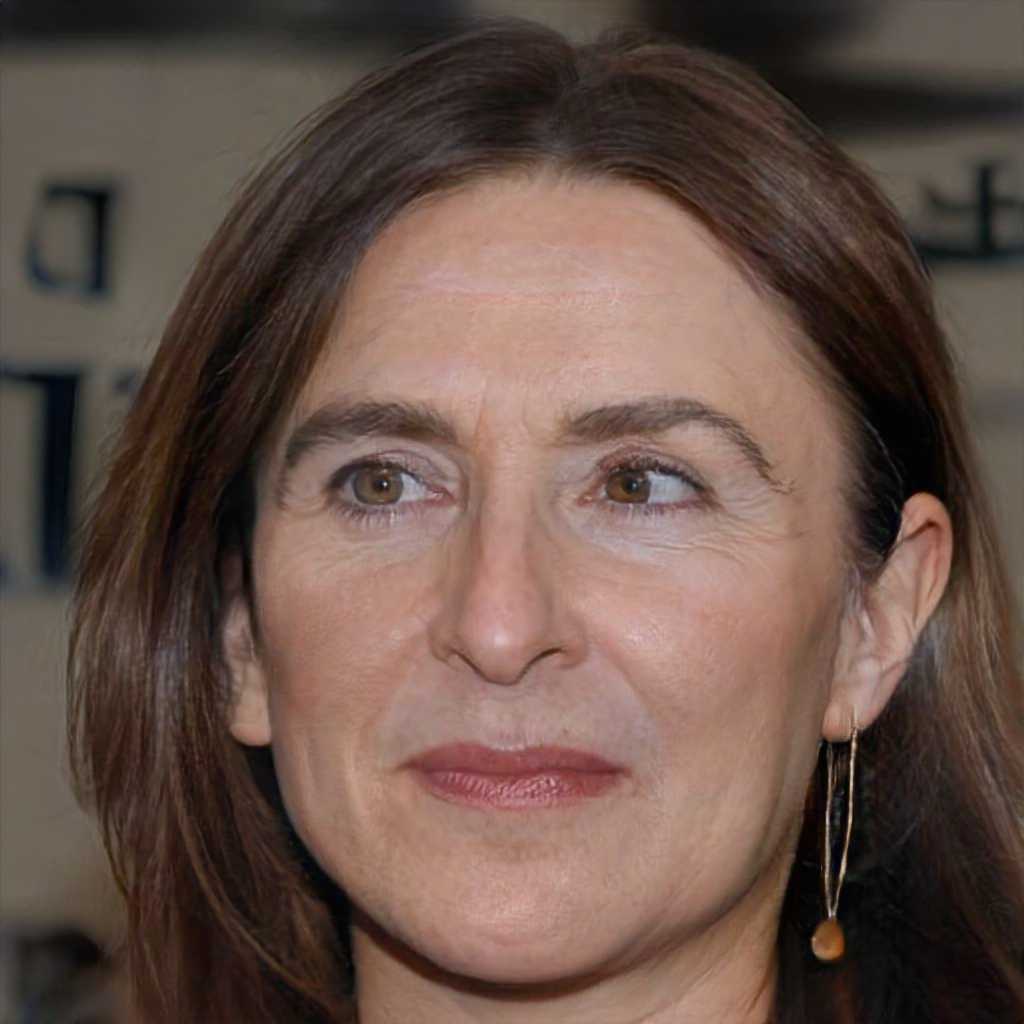}\linebreak[0]%
	\includegraphics[width=\sizea]{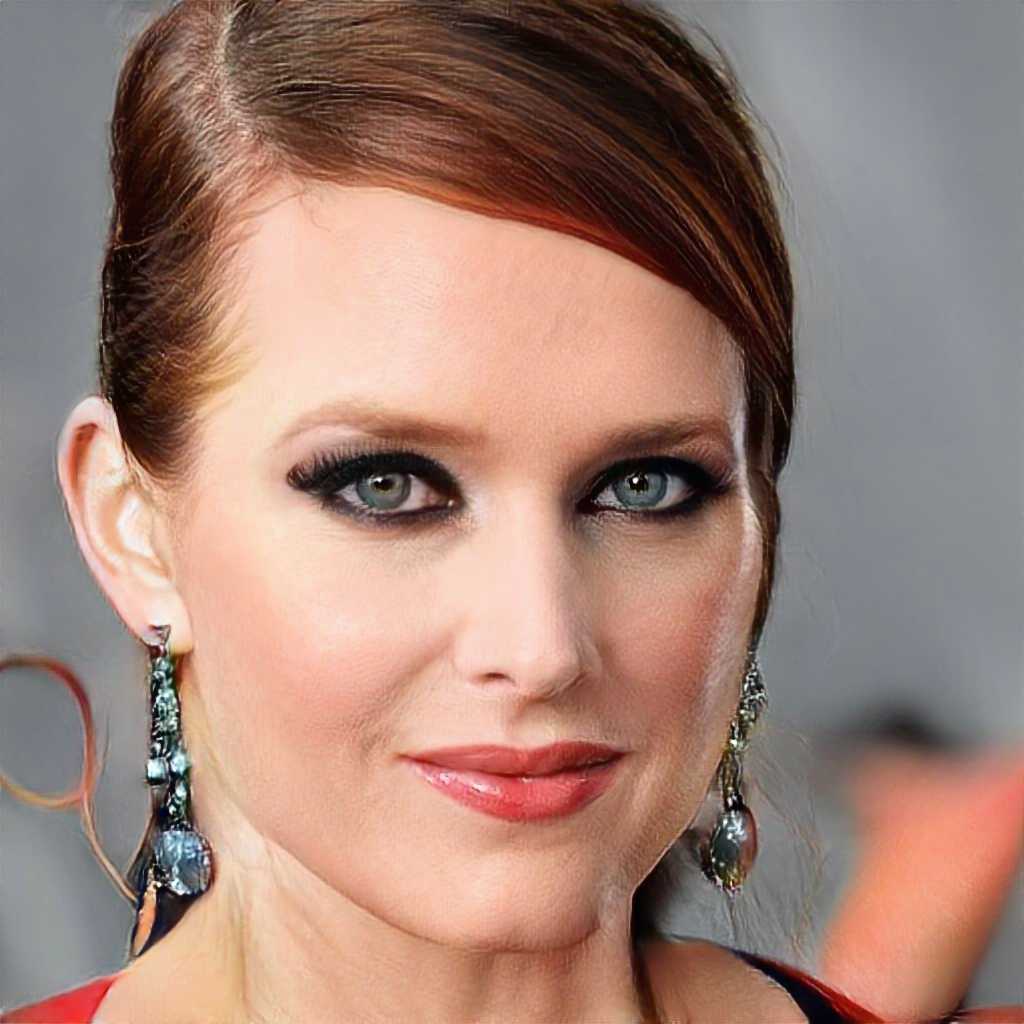}\linebreak[0]%
	\includegraphics[width=\sizea]{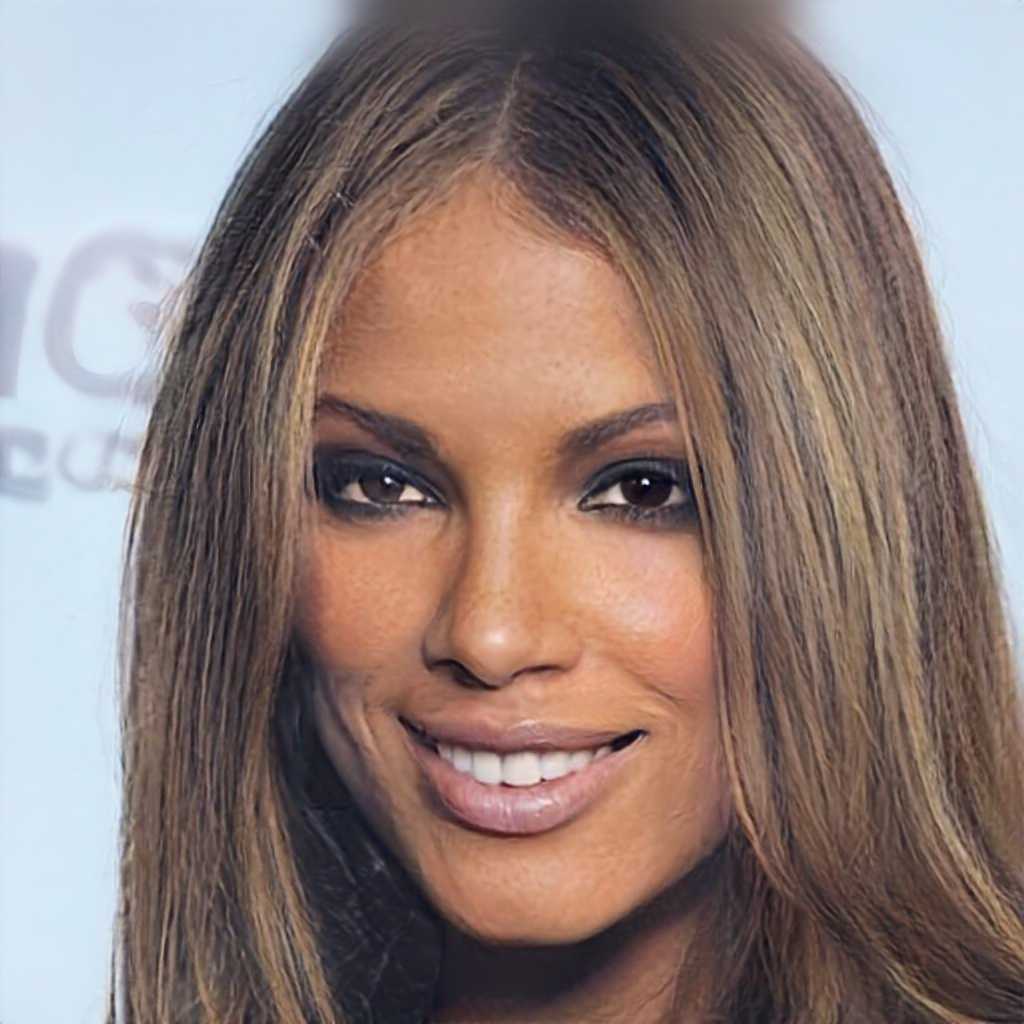}\linebreak[0]%
	\includegraphics[width=\sizea]{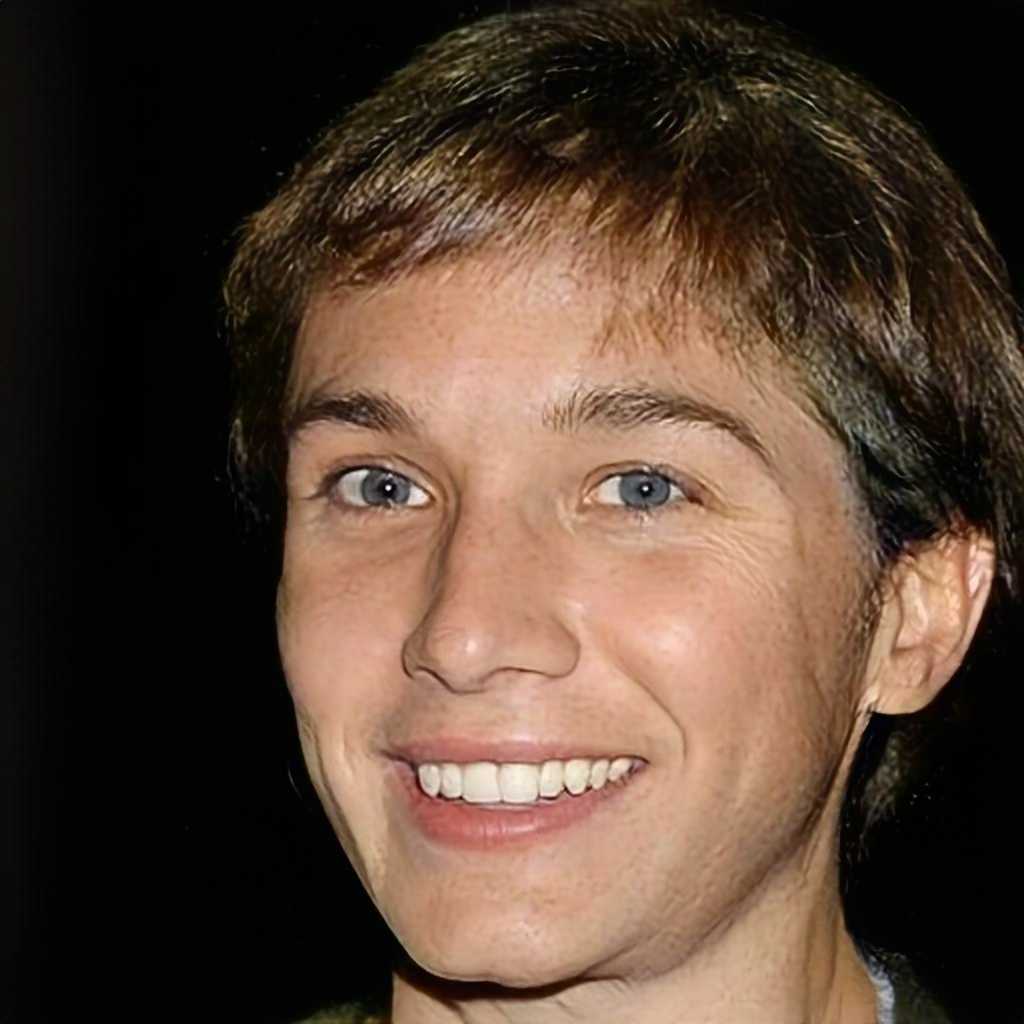}\linebreak[0]%
	\includegraphics[width=\sizea]{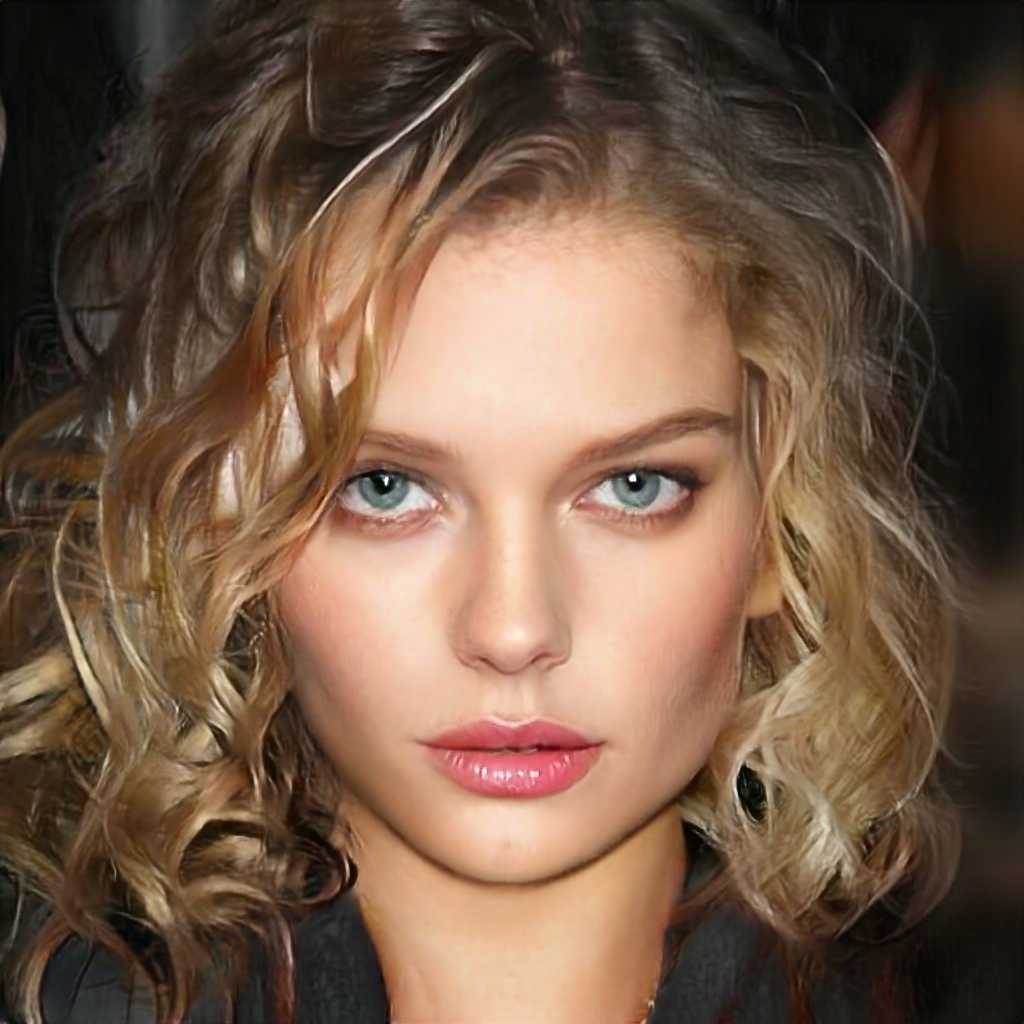}\linebreak[0]%
	\includegraphics[width=\sizea]{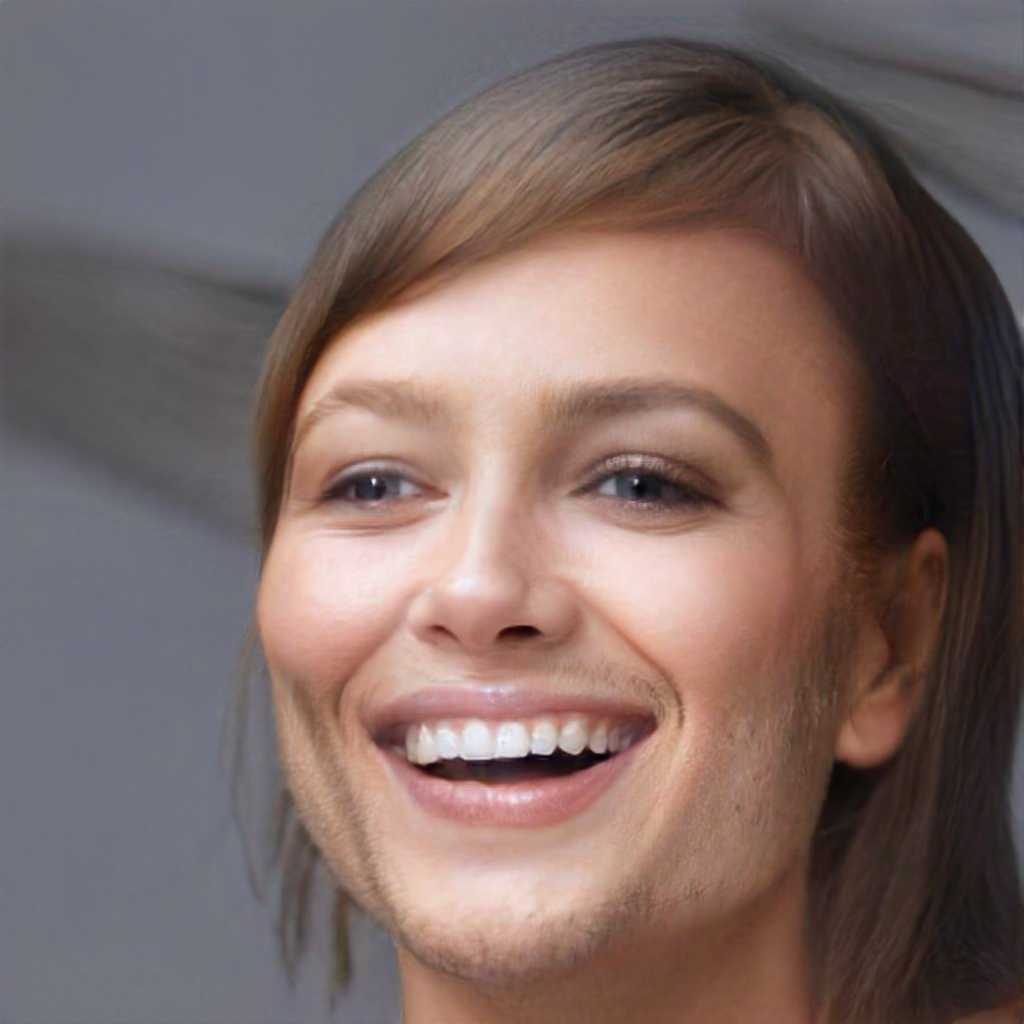}\linebreak[0]%
	\includegraphics[width=\sizea]{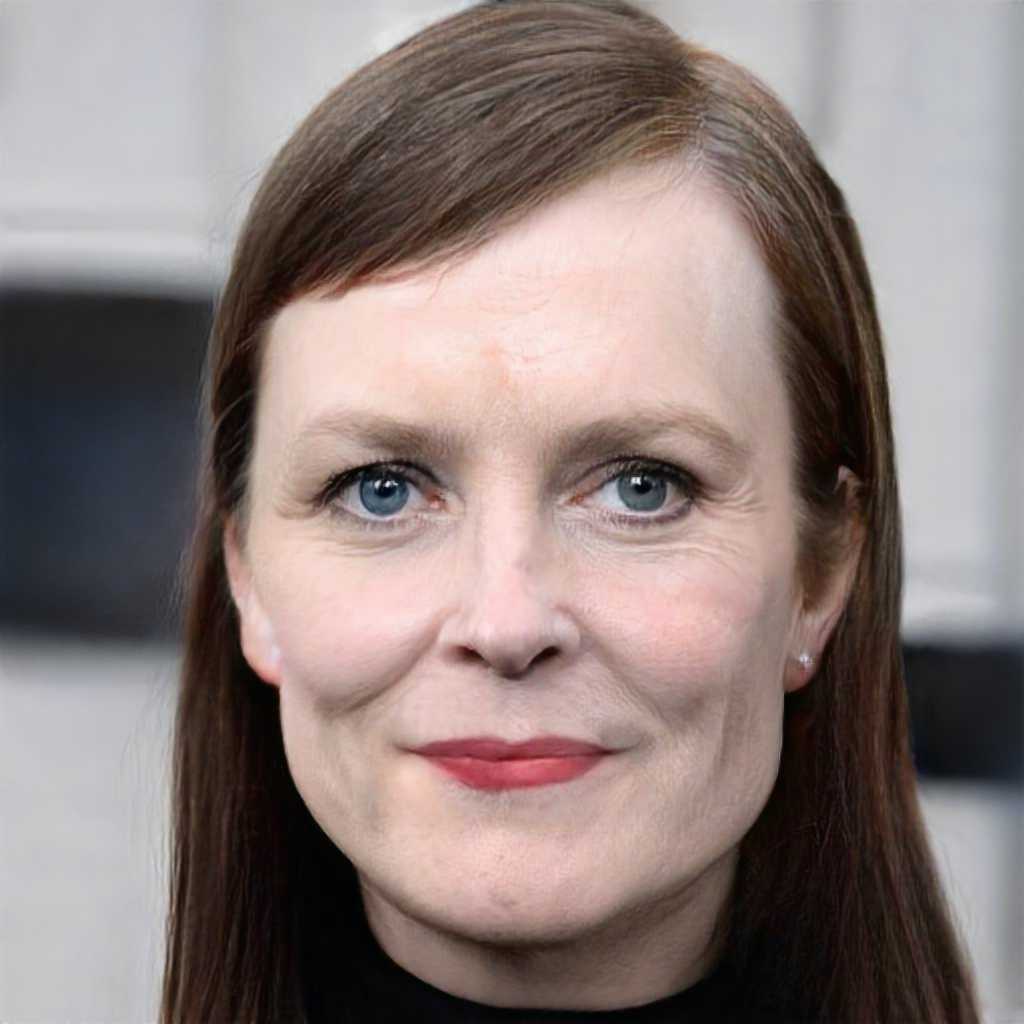}\linebreak[0]%
	\includegraphics[width=\sizea]{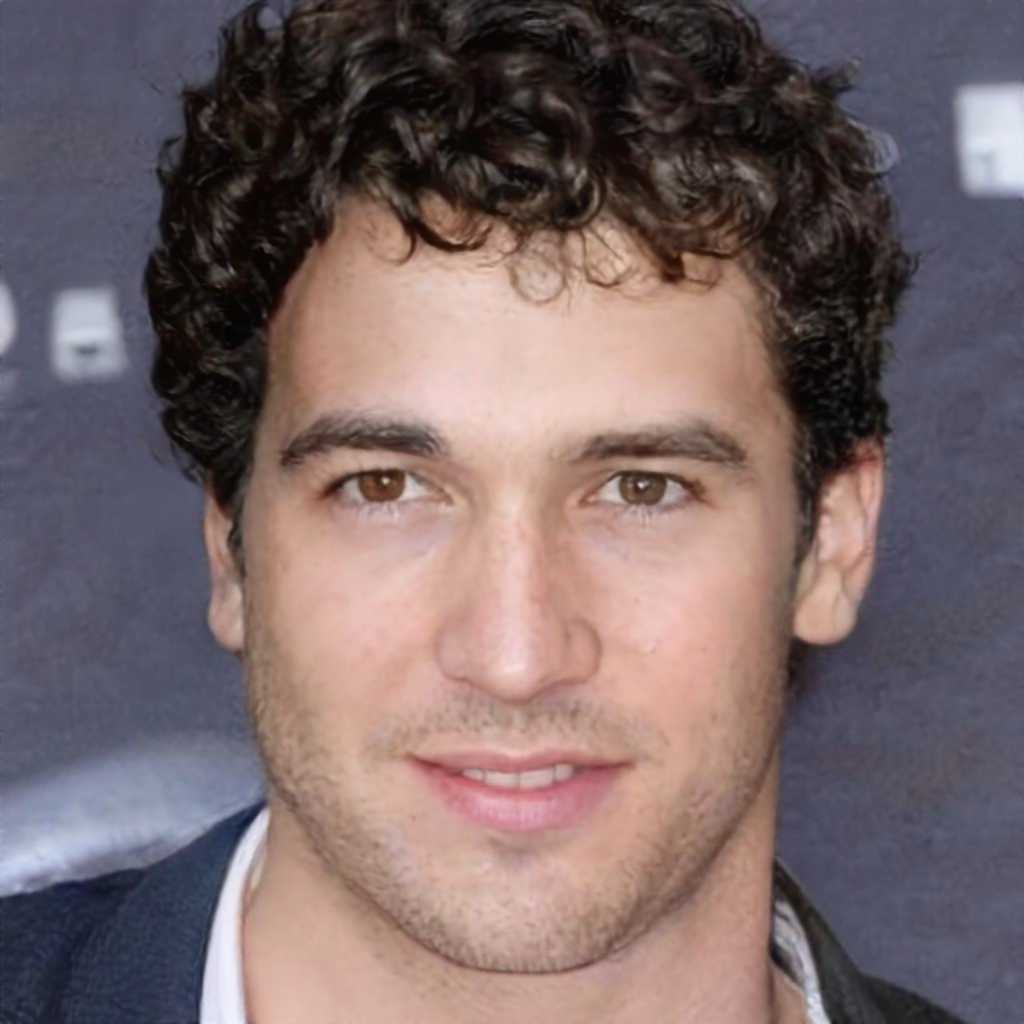}\linebreak[0]%
	\includegraphics[width=\sizea]{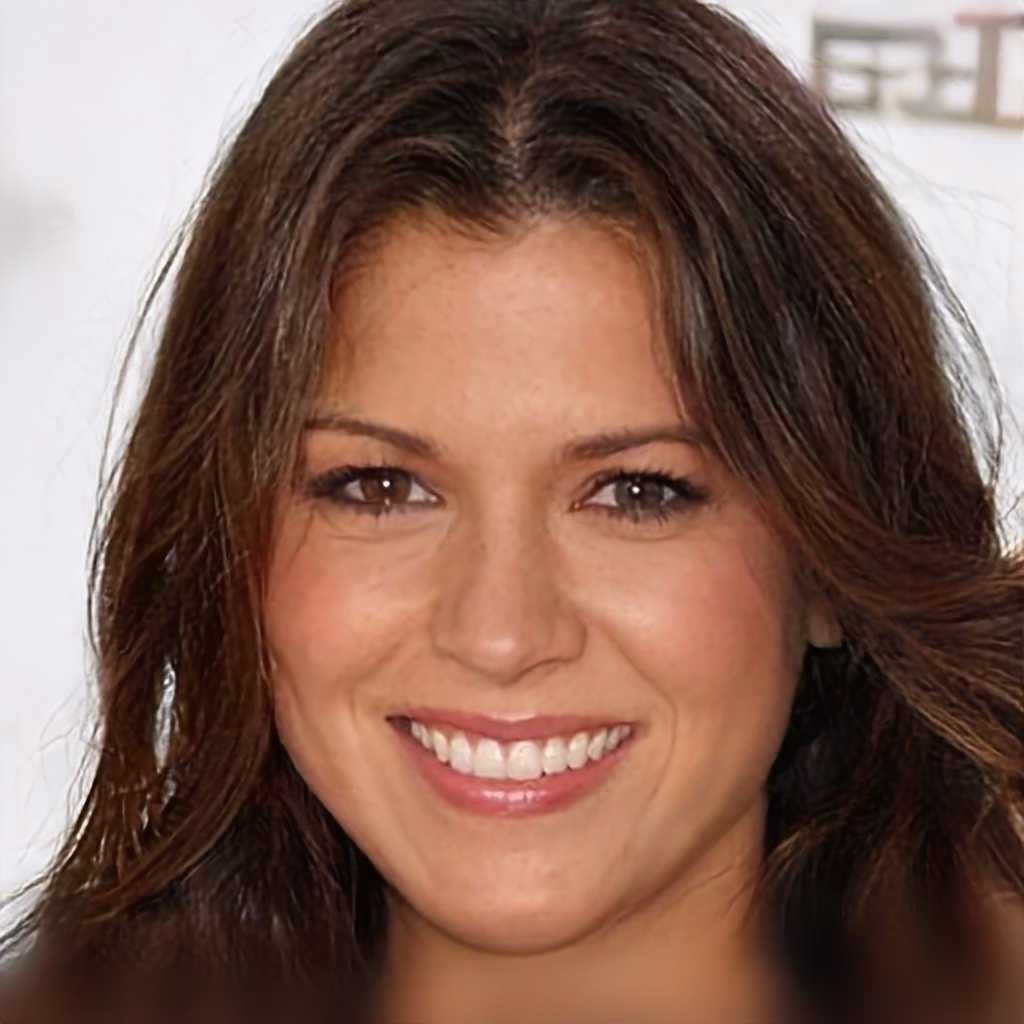}\linebreak[0]%
	\includegraphics[width=\sizea]{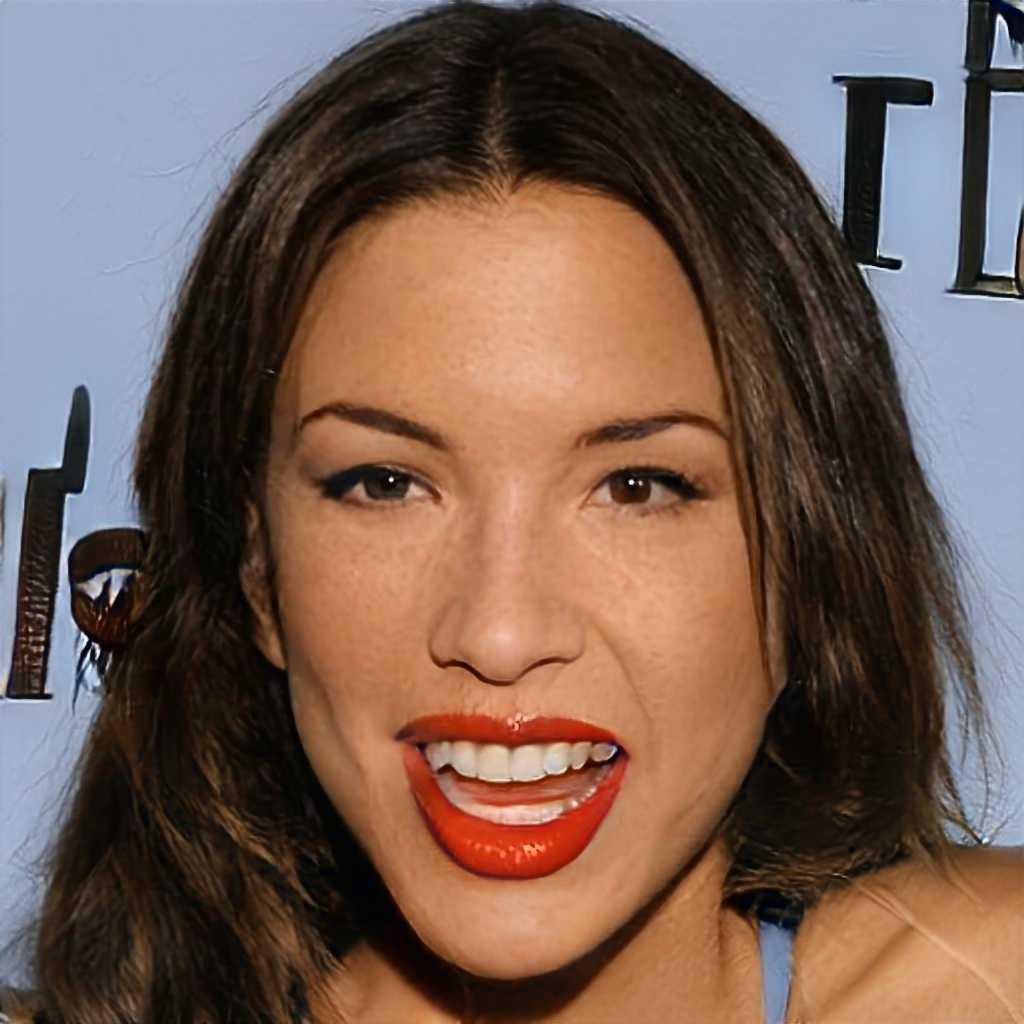}\linebreak[0]%
	\caption{Uncurated unconditional results on the 1024$\times$1024 MM-CelebA-HQ dataset.}
	\label{fig:celeba_uncond}
\end{figure*}
\begin{figure*}[!t]
	\centering
	\newcommand{\sizea}{0.125\linewidth}
	\setlength{\lineskip}{0pt}
	\includegraphics[width=\sizea]{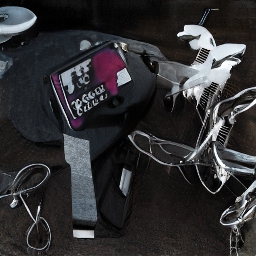}\linebreak[0]%
	\includegraphics[width=\sizea]{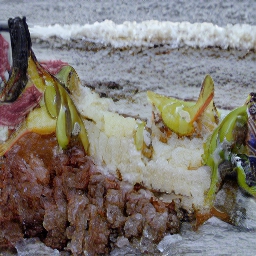}\linebreak[0]%
	\includegraphics[width=\sizea]{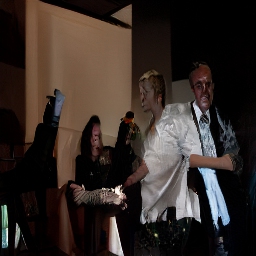}\linebreak[0]%
	\includegraphics[width=\sizea]{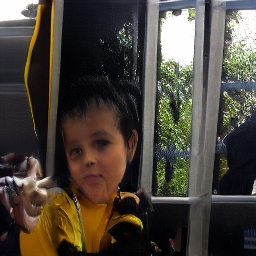}\linebreak[0]%
	\includegraphics[width=\sizea]{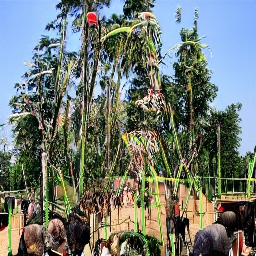}\linebreak[0]%
	\includegraphics[width=\sizea]{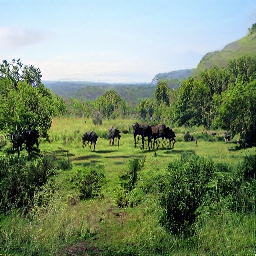}\linebreak[0]%
	\includegraphics[width=\sizea]{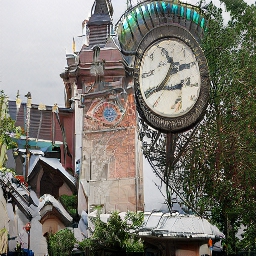}\linebreak[0]%
	\includegraphics[width=\sizea]{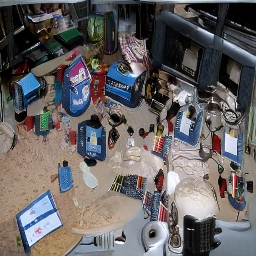}\linebreak[0]%
	\includegraphics[width=\sizea]{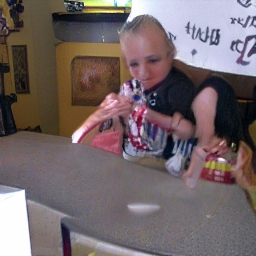}\linebreak[0]%
	\includegraphics[width=\sizea]{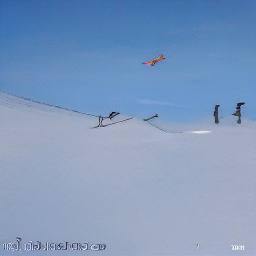}\linebreak[0]%
	\includegraphics[width=\sizea]{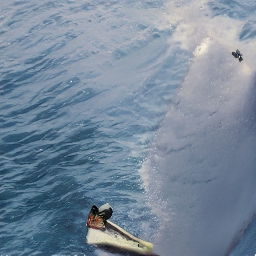}\linebreak[0]%
	\includegraphics[width=\sizea]{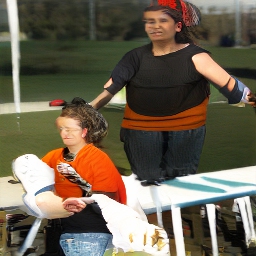}\linebreak[0]%
	\includegraphics[width=\sizea]{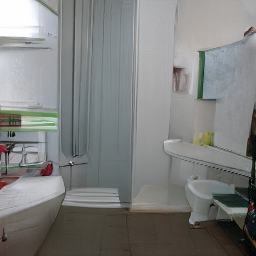}\linebreak[0]%
	\includegraphics[width=\sizea]{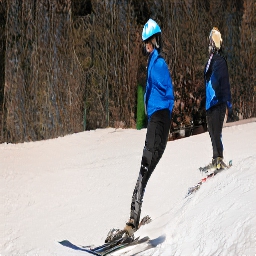}\linebreak[0]%
	\includegraphics[width=\sizea]{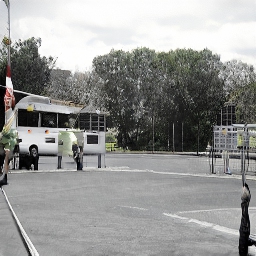}\linebreak[0]%
	\includegraphics[width=\sizea]{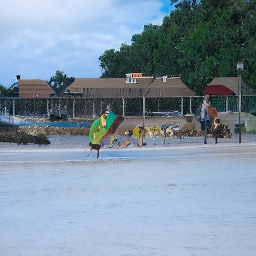}\linebreak[0]%
	\includegraphics[width=\sizea]{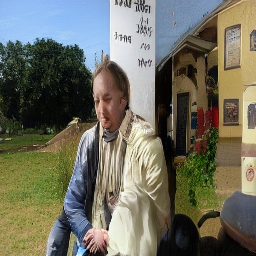}\linebreak[0]%
	\includegraphics[width=\sizea]{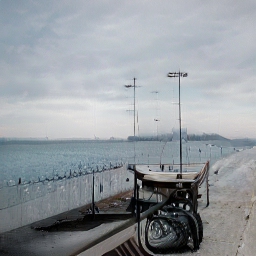}\linebreak[0]%
	\includegraphics[width=\sizea]{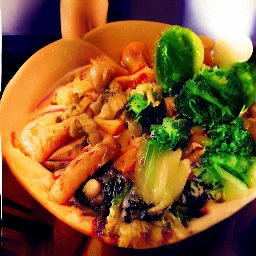}\linebreak[0]%
	\includegraphics[width=\sizea]{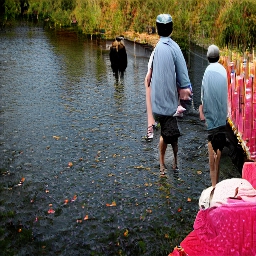}\linebreak[0]%
	\includegraphics[width=\sizea]{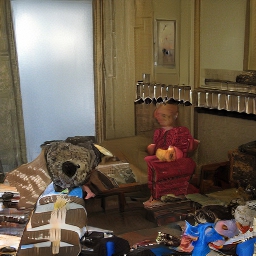}\linebreak[0]%
	\includegraphics[width=\sizea]{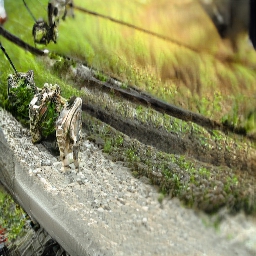}\linebreak[0]%
	\includegraphics[width=\sizea]{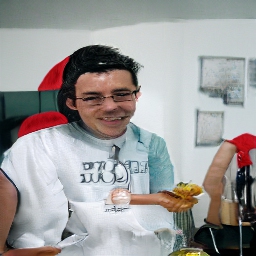}\linebreak[0]%
	\includegraphics[width=\sizea]{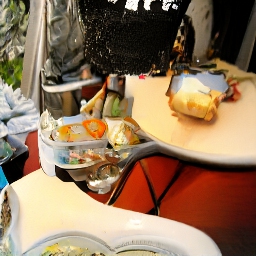}\linebreak[0]%
	\includegraphics[width=\sizea]{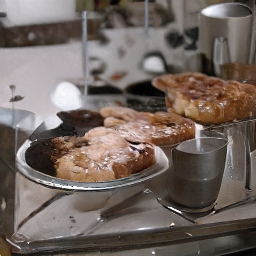}\linebreak[0]%
	\includegraphics[width=\sizea]{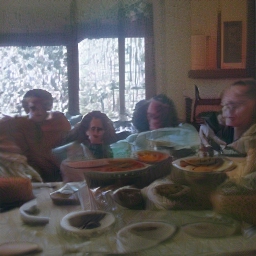}\linebreak[0]%
	\includegraphics[width=\sizea]{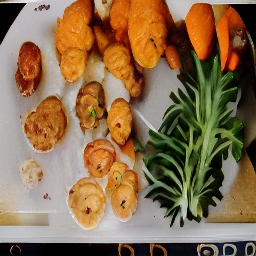}\linebreak[0]%
	\includegraphics[width=\sizea]{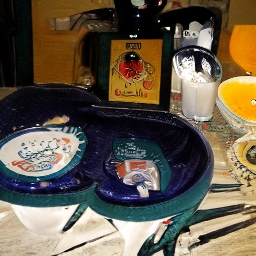}\linebreak[0]%
	\includegraphics[width=\sizea]{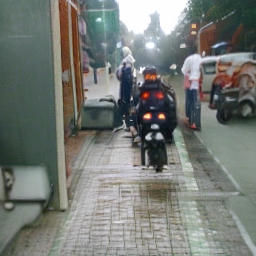}\linebreak[0]%
	\includegraphics[width=\sizea]{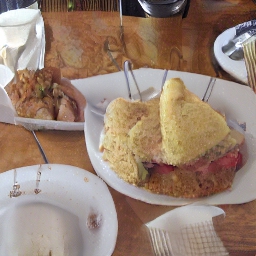}\linebreak[0]%
	\includegraphics[width=\sizea]{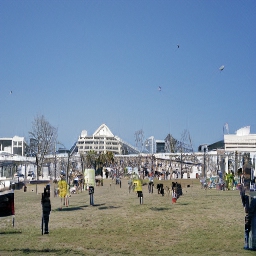}\linebreak[0]%
	\includegraphics[width=\sizea]{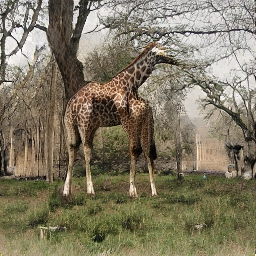}\linebreak[0]%
	\caption{Uncurated unconditional results on the 256$\times$256 MS-COCO dataset.}
	\label{fig:coco_uncond}
\end{figure*}

\begin{figure*}[!t]
	\centering
	\newcommand{\sizea}{0.125\linewidth}
	\setlength{\lineskip}{0pt}
	\includegraphics[width=\sizea]{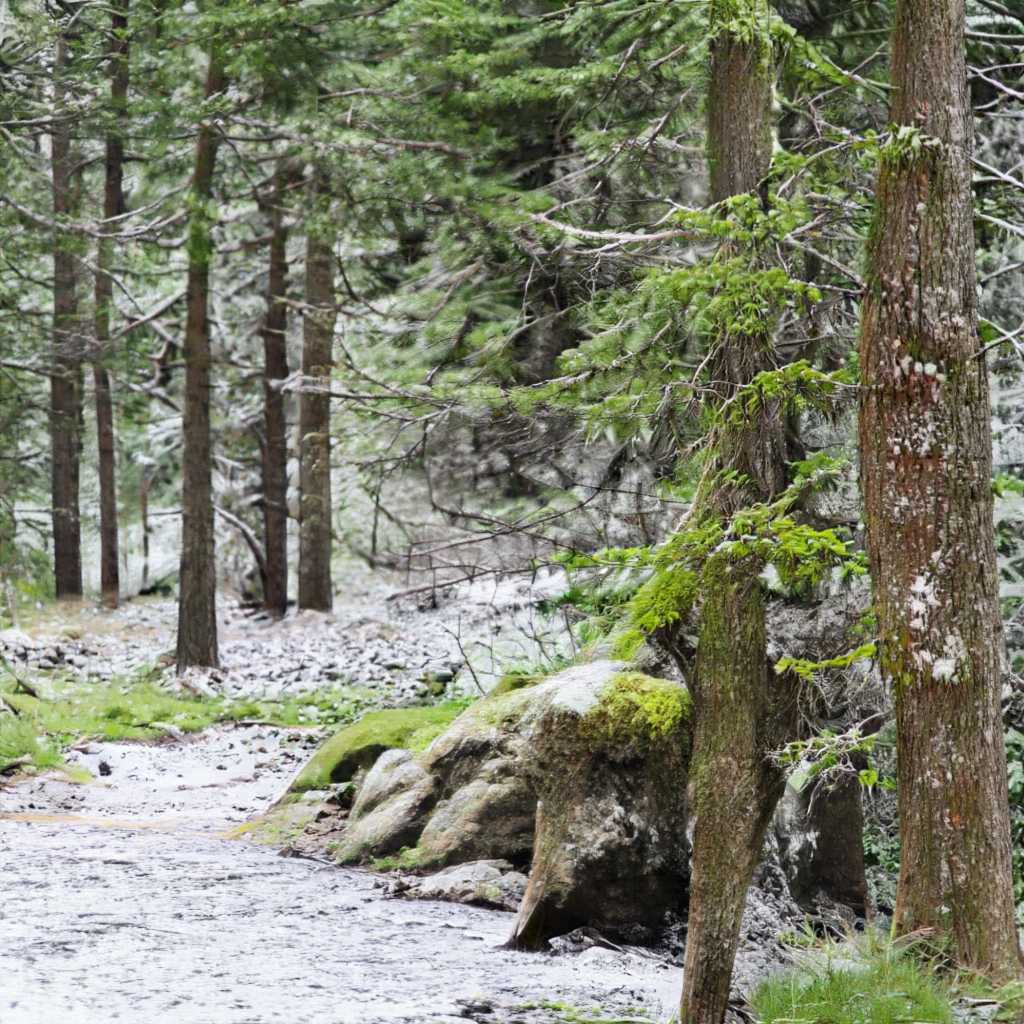}\linebreak[0]%
	\includegraphics[width=\sizea]{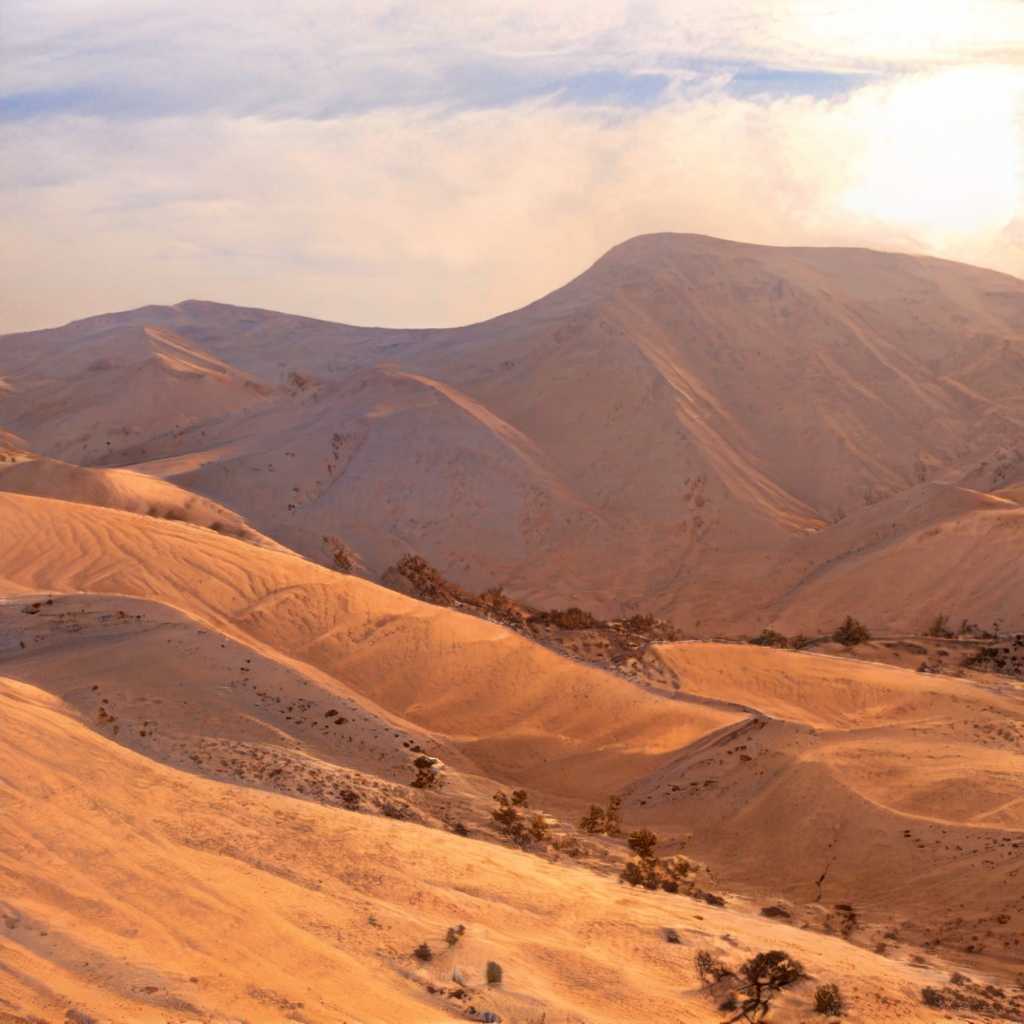}\linebreak[0]%
	\includegraphics[width=\sizea]{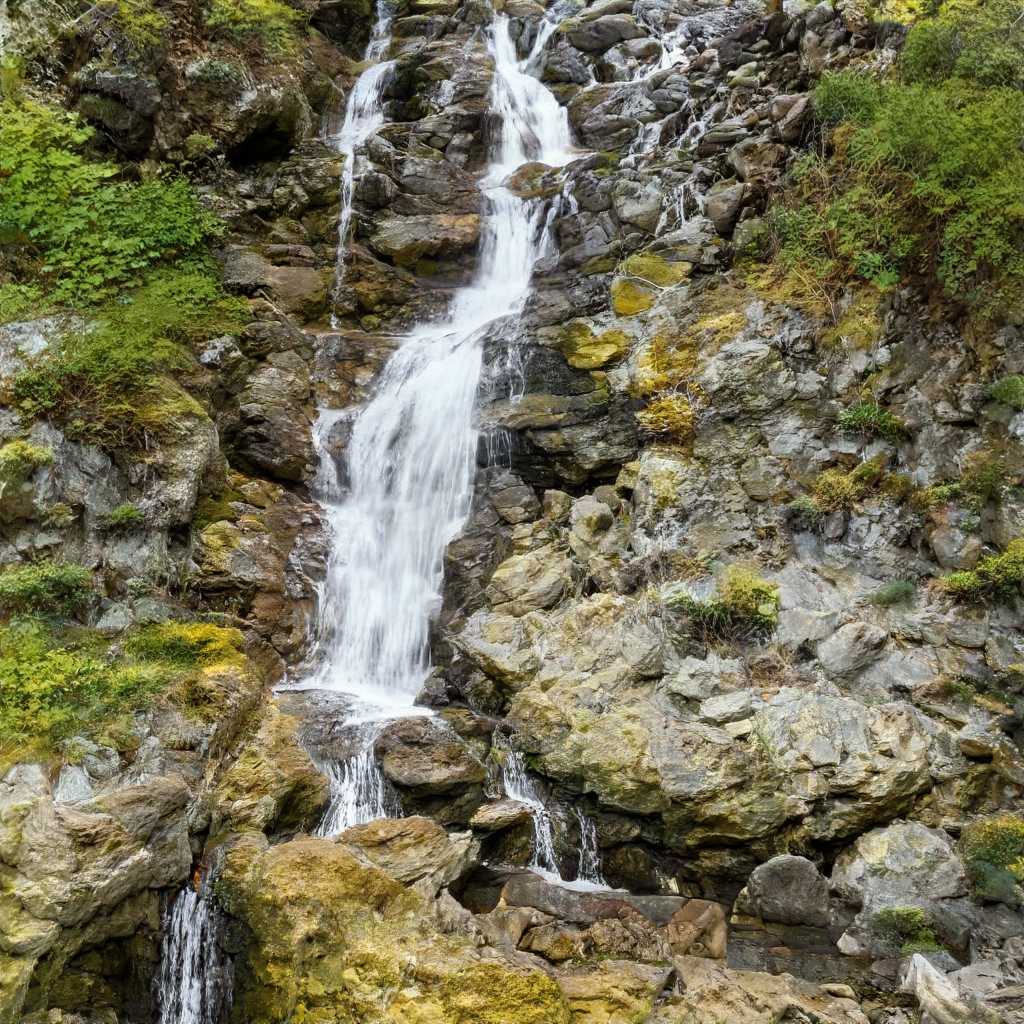}\linebreak[0]%
	\includegraphics[width=\sizea]{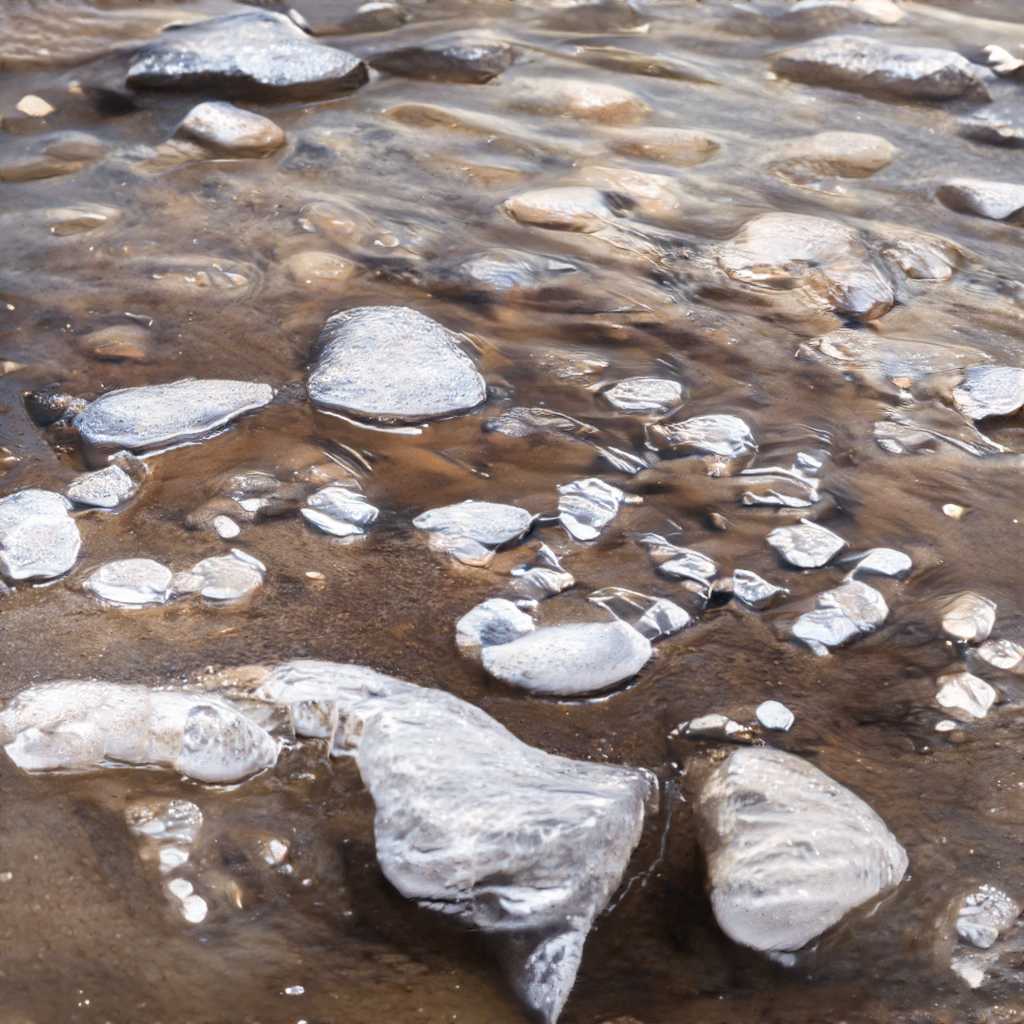}\linebreak[0]%
	\includegraphics[width=\sizea]{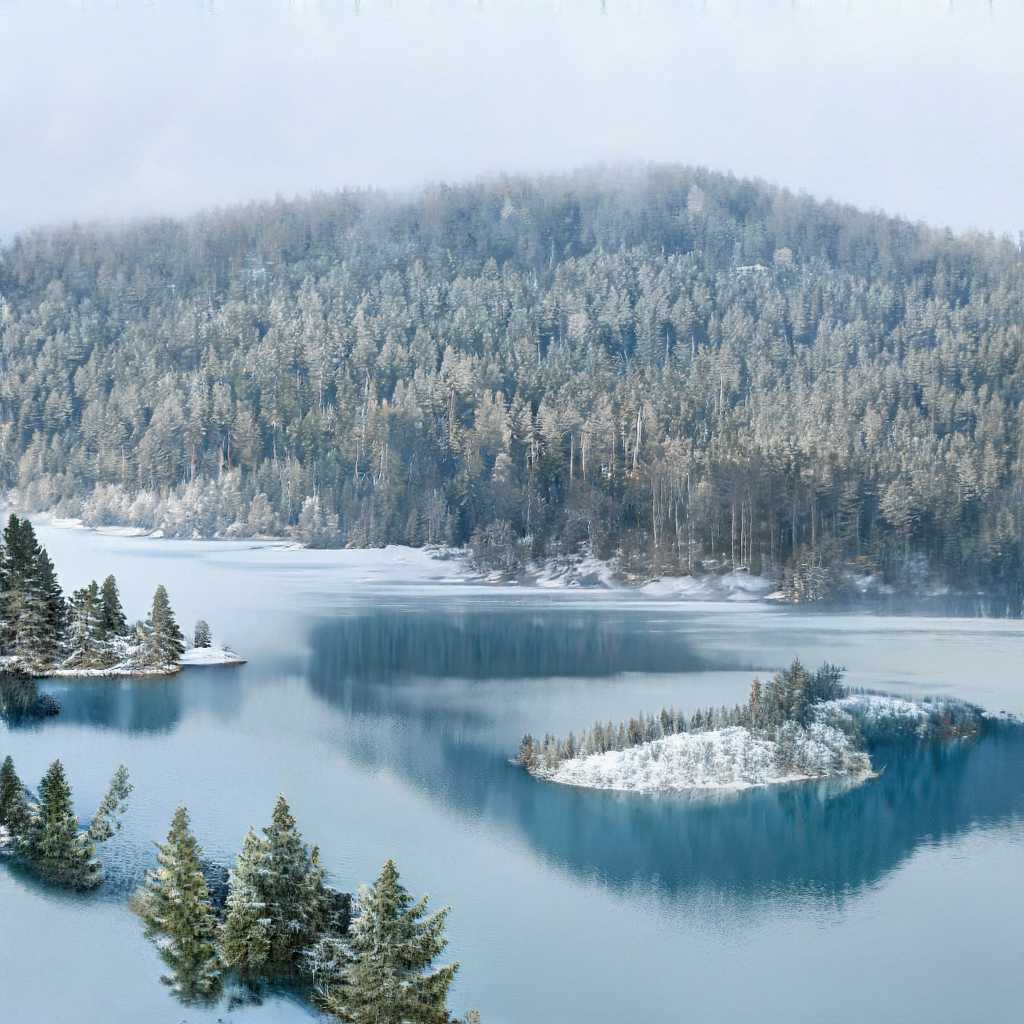}\linebreak[0]%
	\includegraphics[width=\sizea]{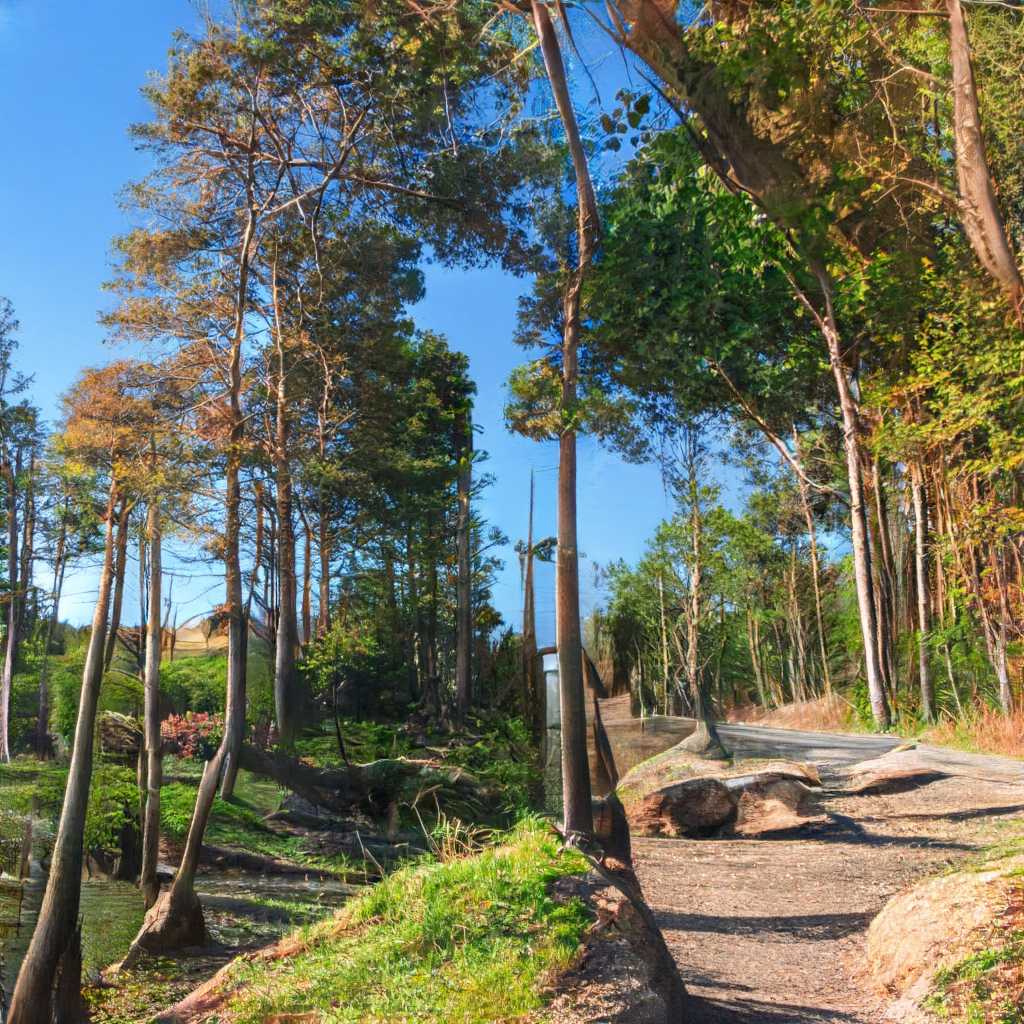}\linebreak[0]%
	\includegraphics[width=\sizea]{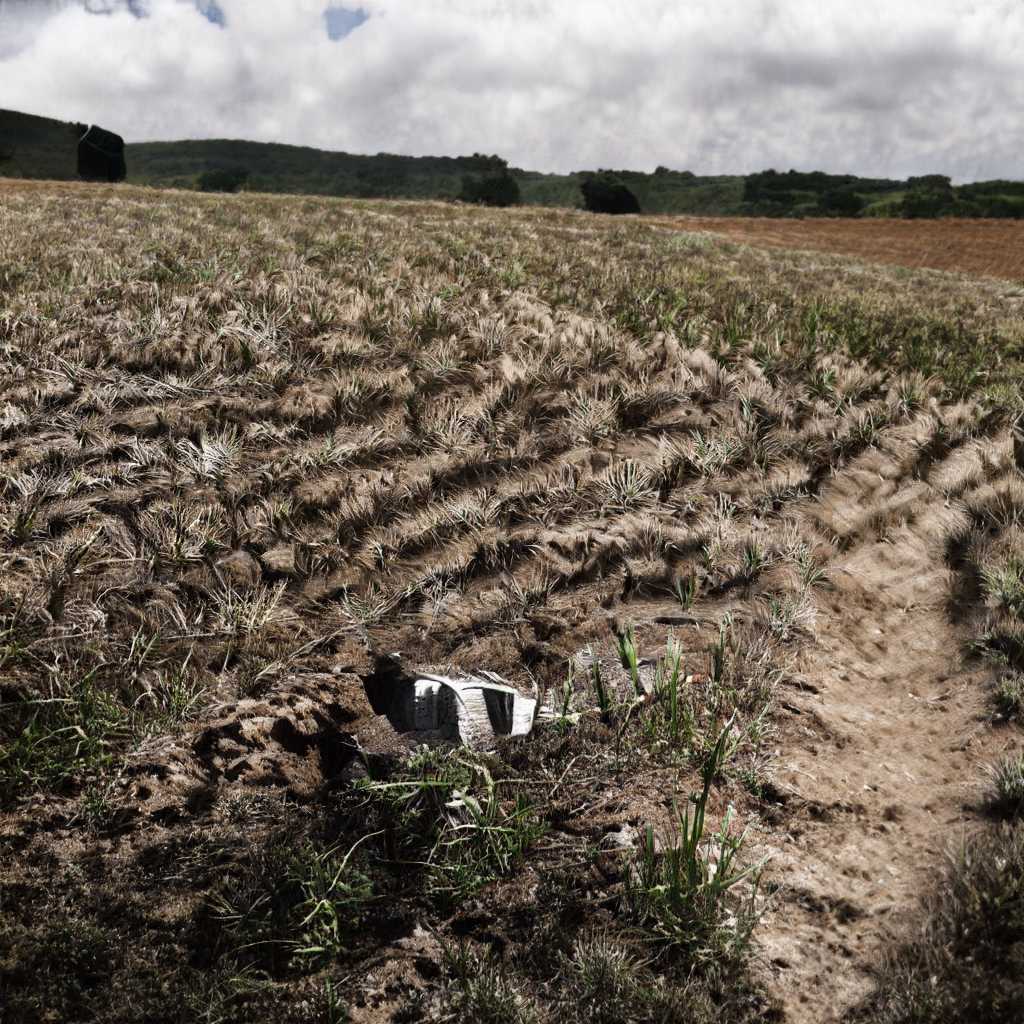}\linebreak[0]%
	\includegraphics[width=\sizea]{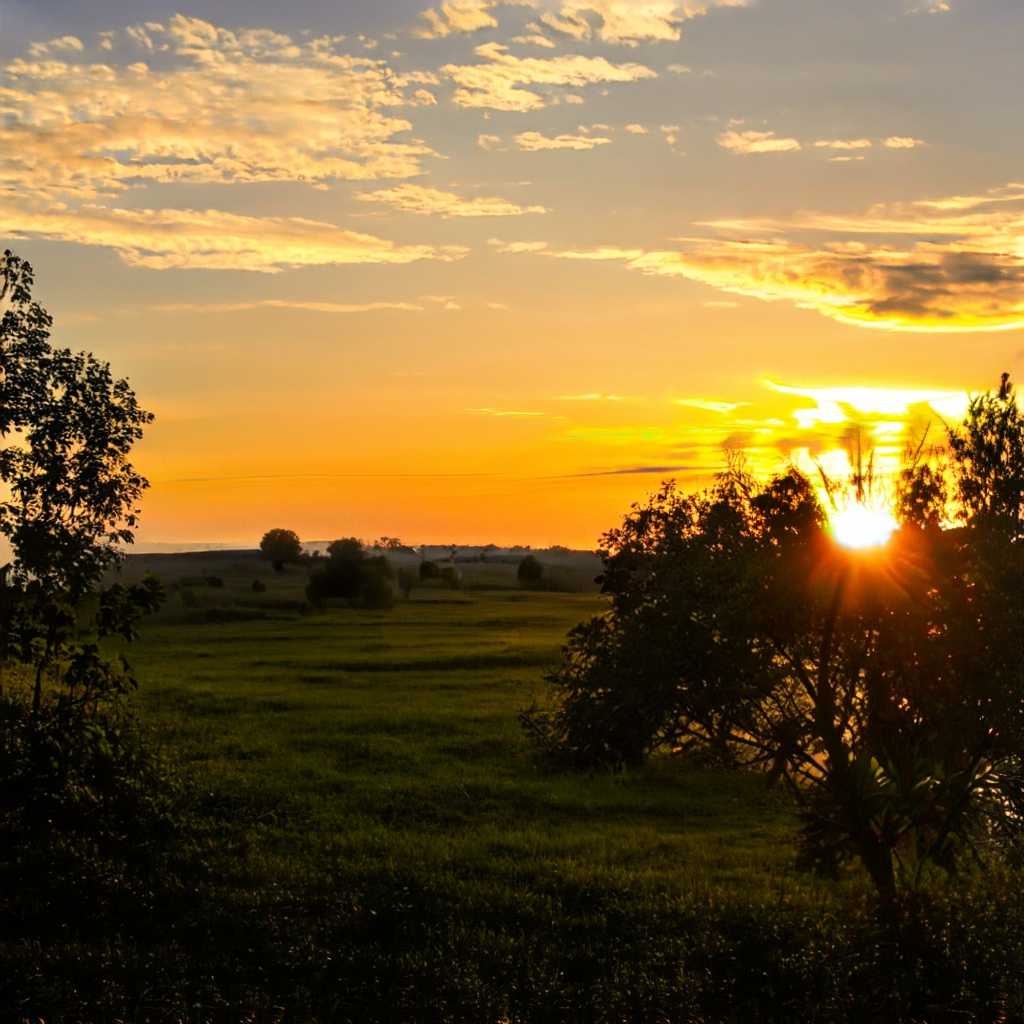}\linebreak[0]%
	\includegraphics[width=\sizea]{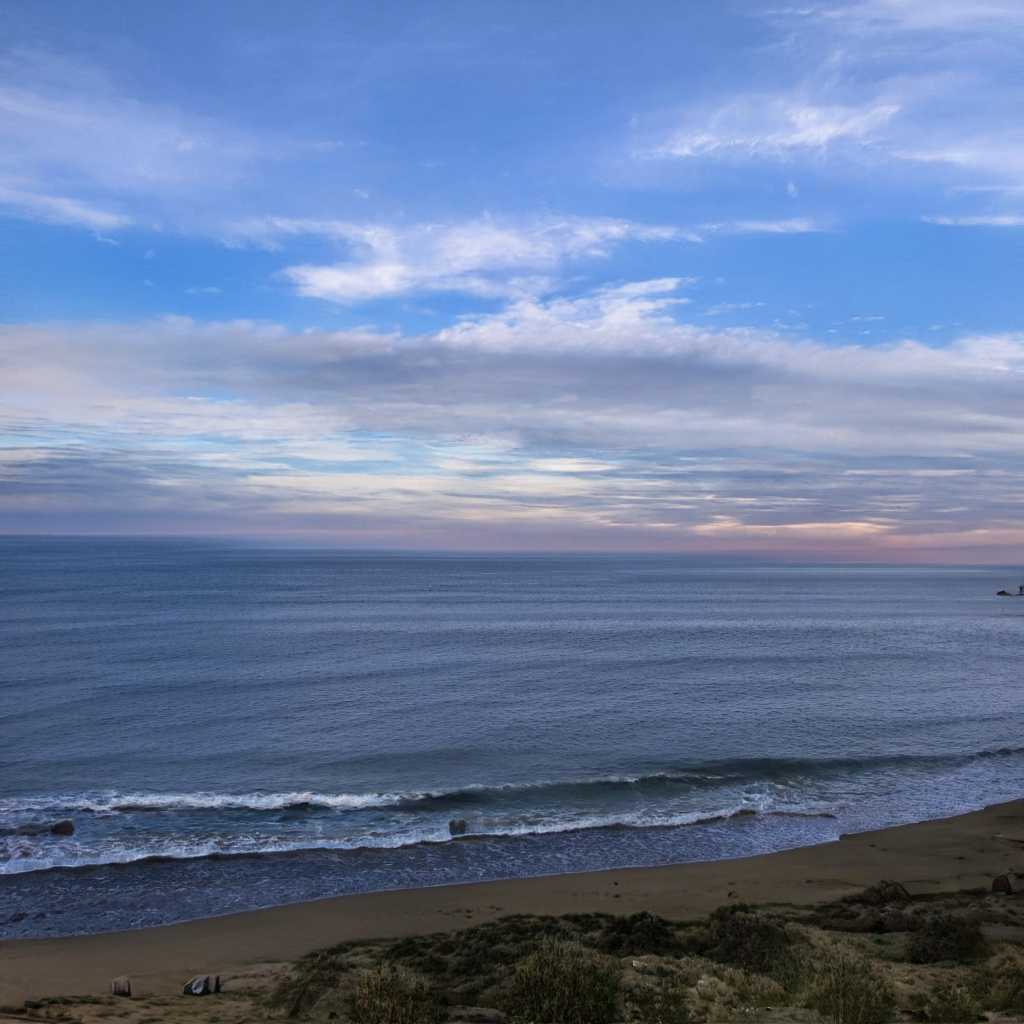}\linebreak[0]%
	\includegraphics[width=\sizea]{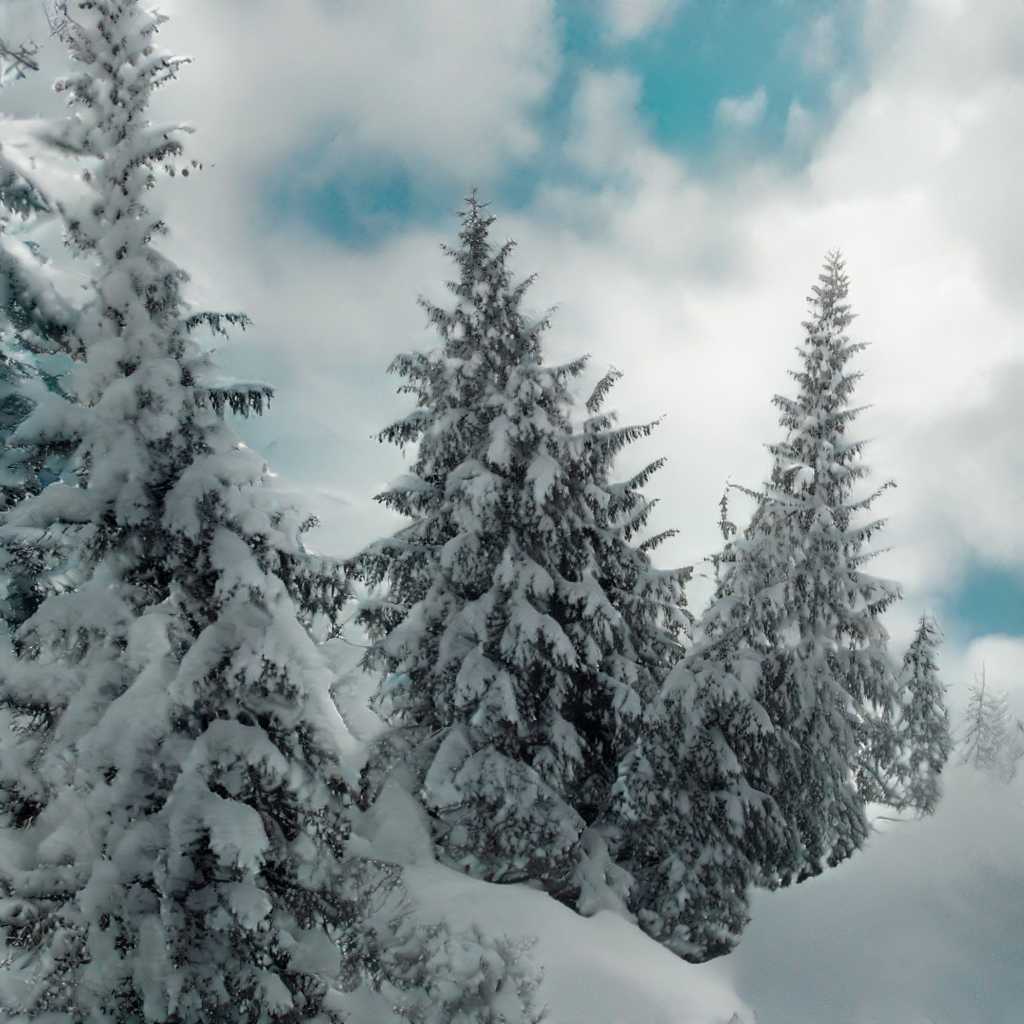}\linebreak[0]%
	\includegraphics[width=\sizea]{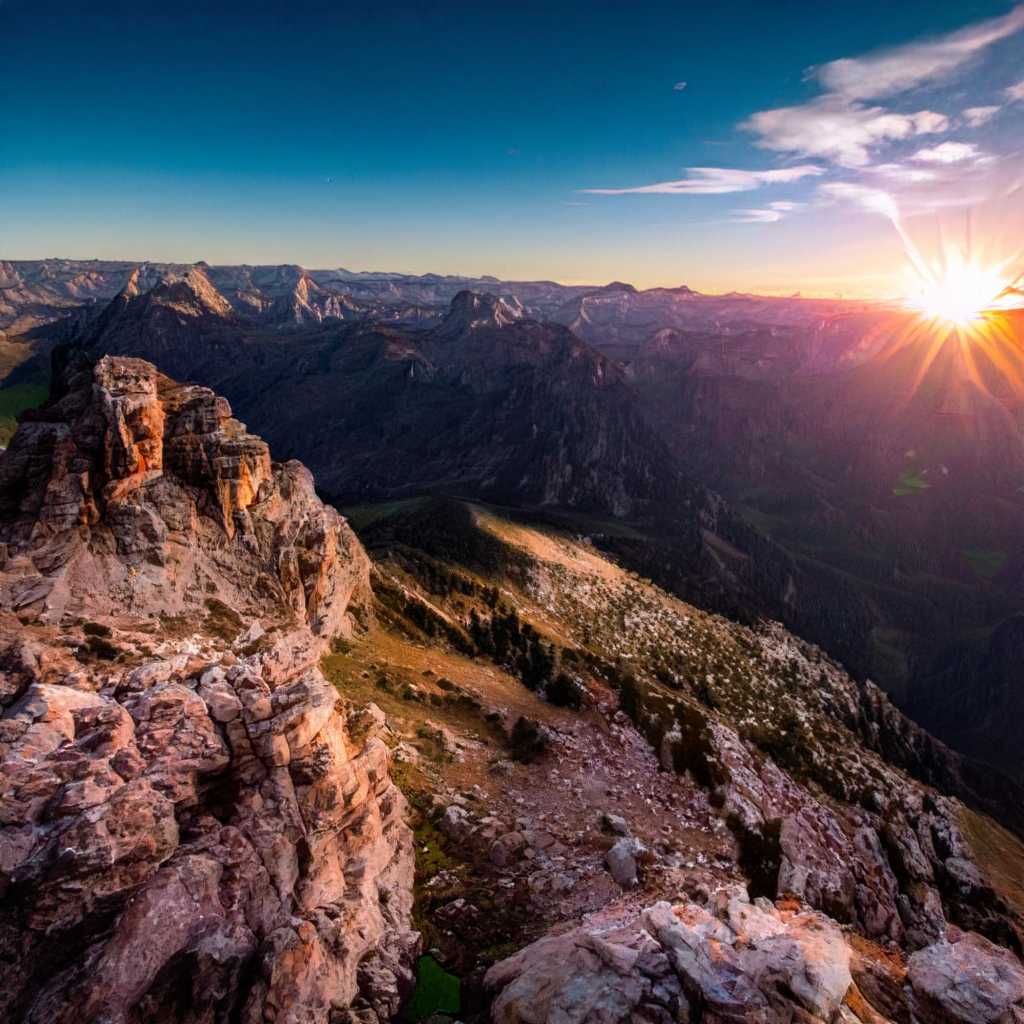}\linebreak[0]%
	\includegraphics[width=\sizea]{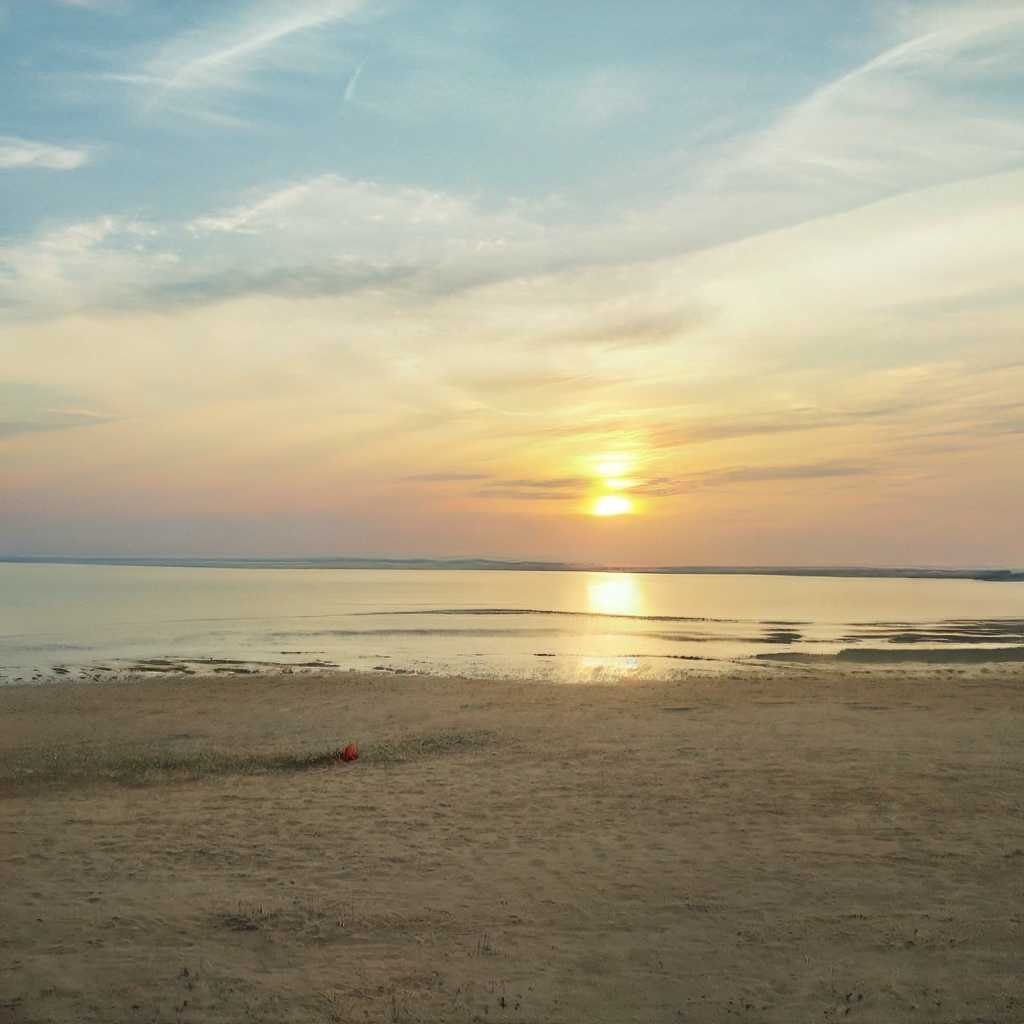}\linebreak[0]%
	\includegraphics[width=\sizea]{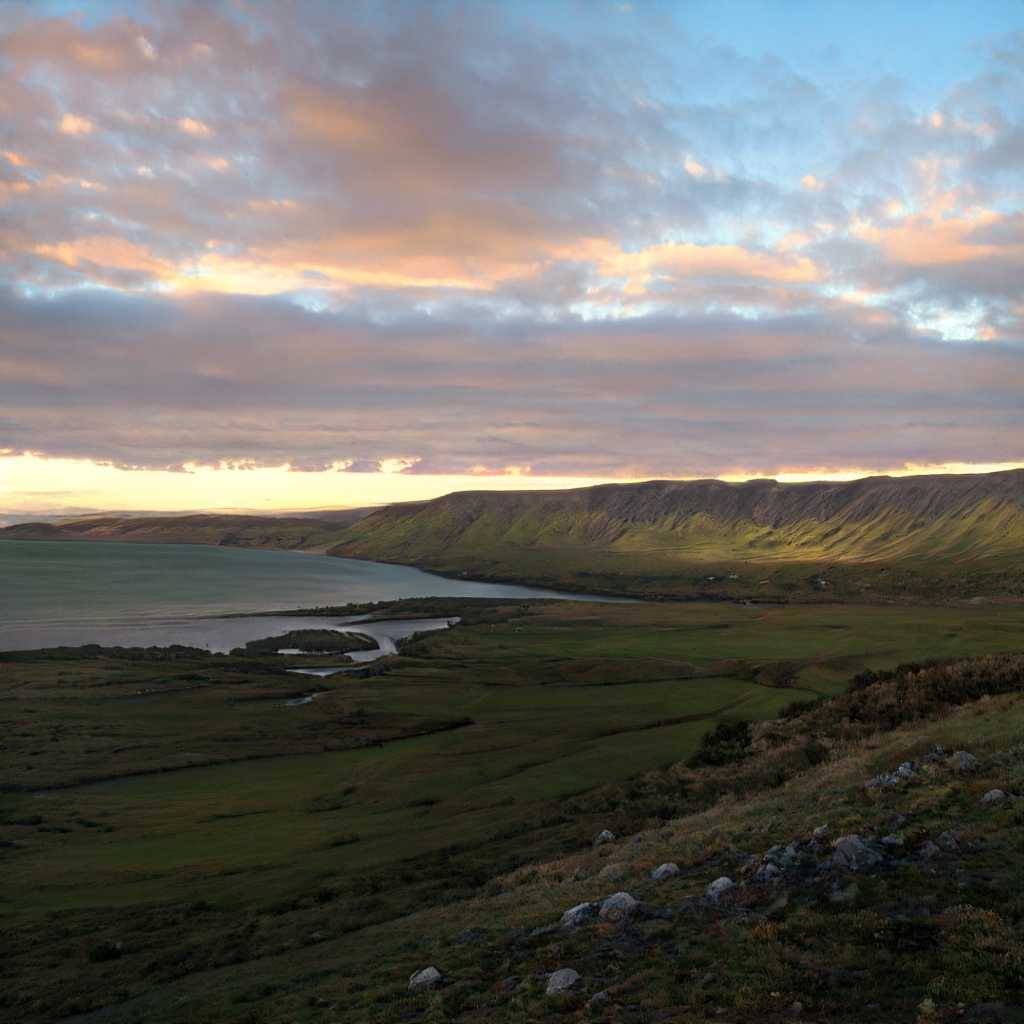}\linebreak[0]%
	\includegraphics[width=\sizea]{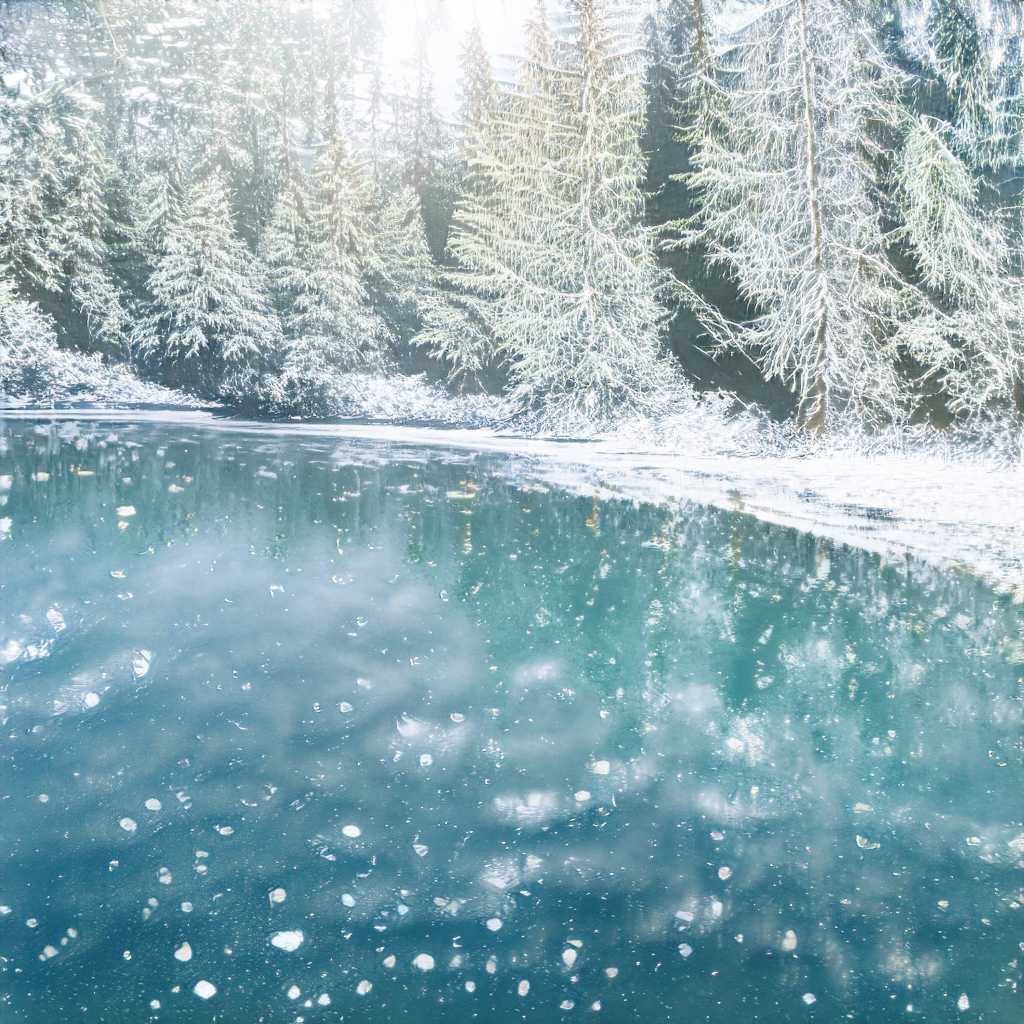}\linebreak[0]%
	\includegraphics[width=\sizea]{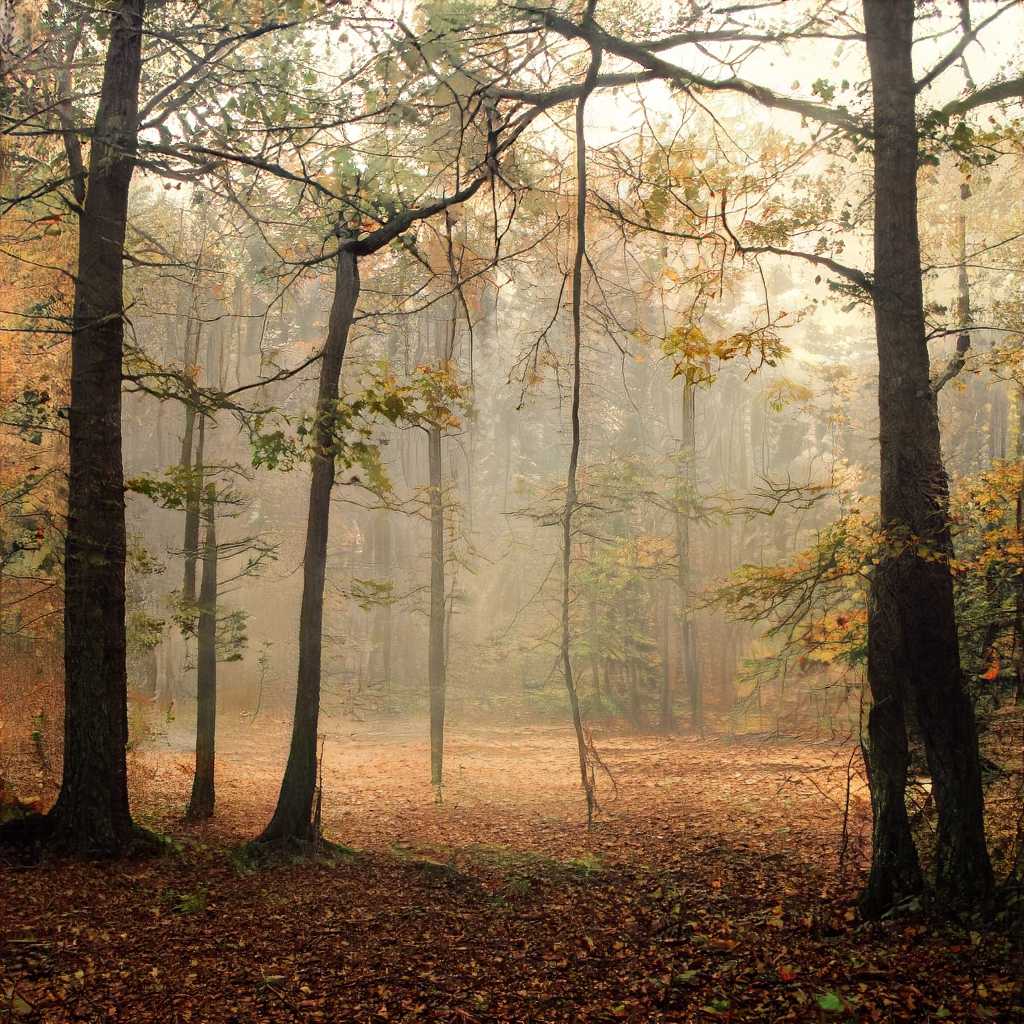}\linebreak[0]%
	\includegraphics[width=\sizea]{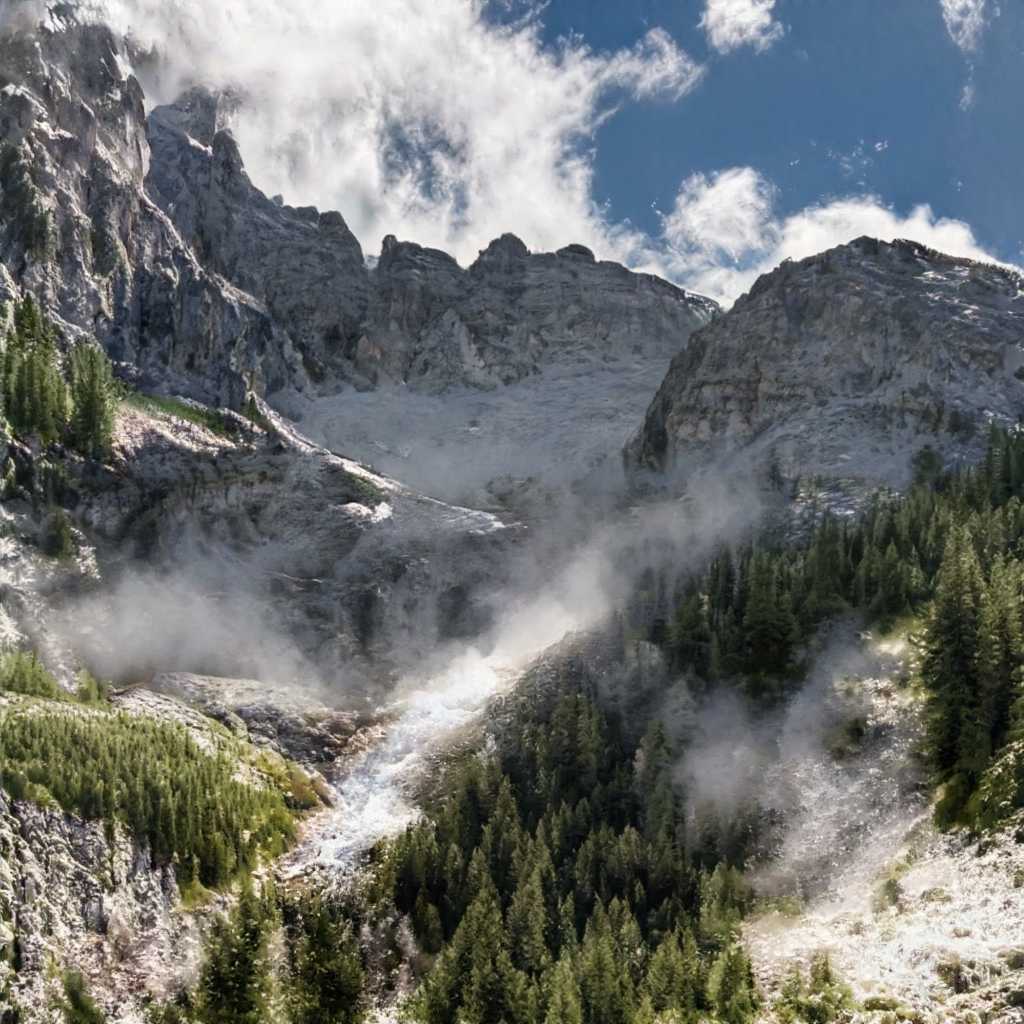}\linebreak[0]%
	\includegraphics[width=\sizea]{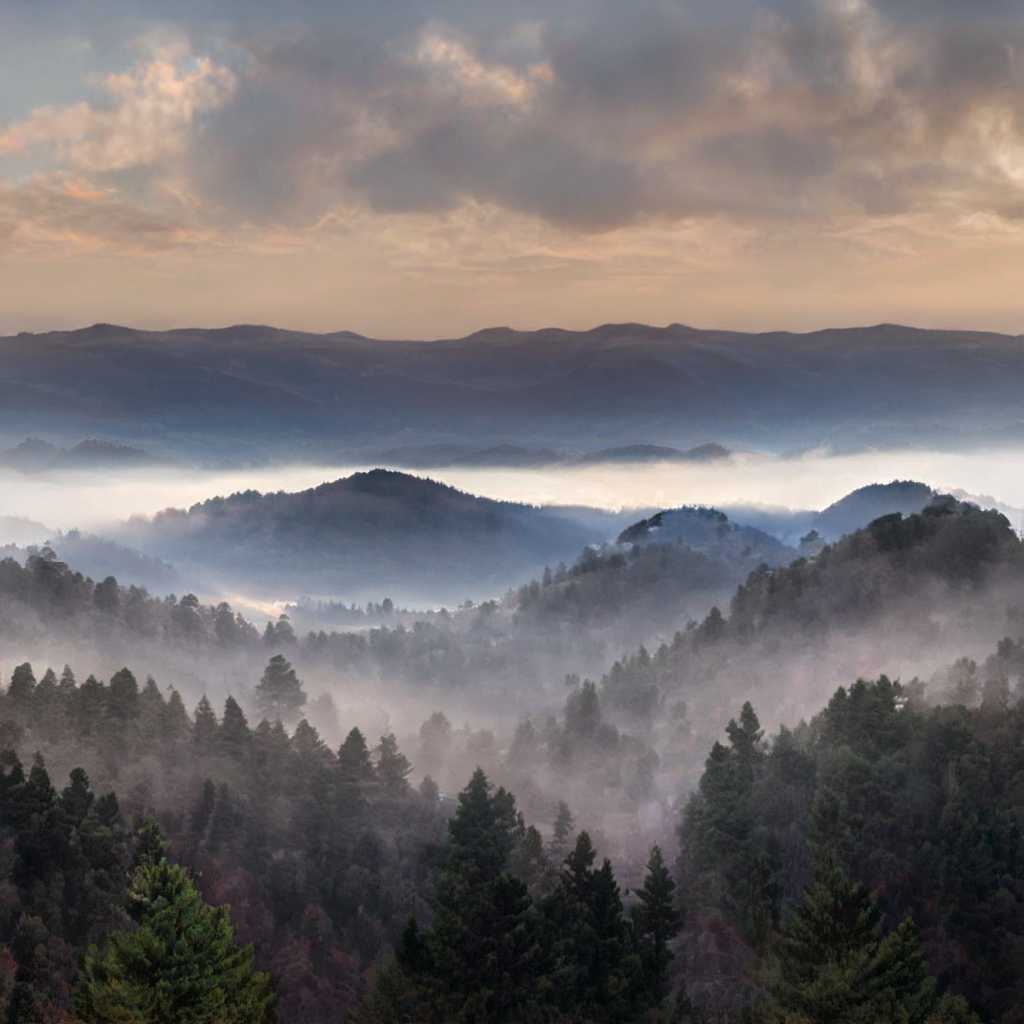}\linebreak[0]%
	\includegraphics[width=\sizea]{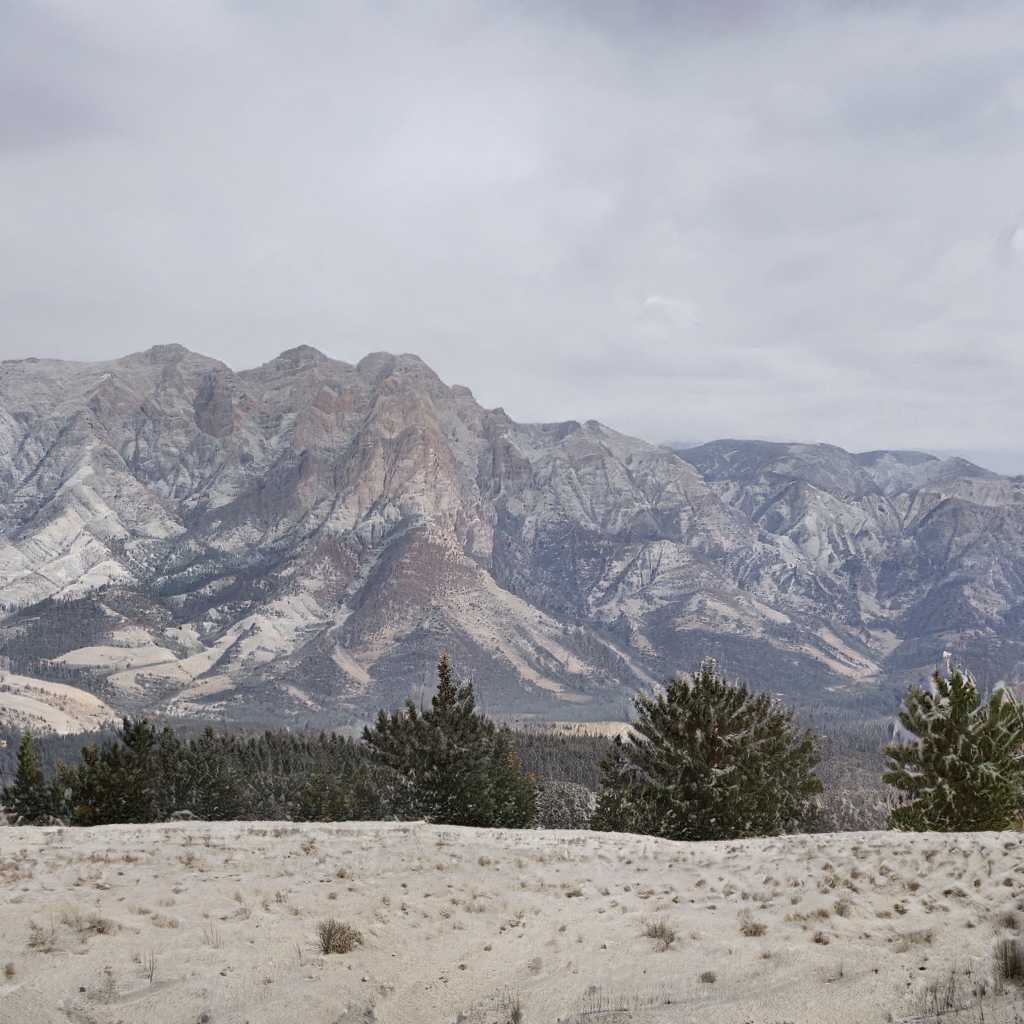}\linebreak[0]%
	\includegraphics[width=\sizea]{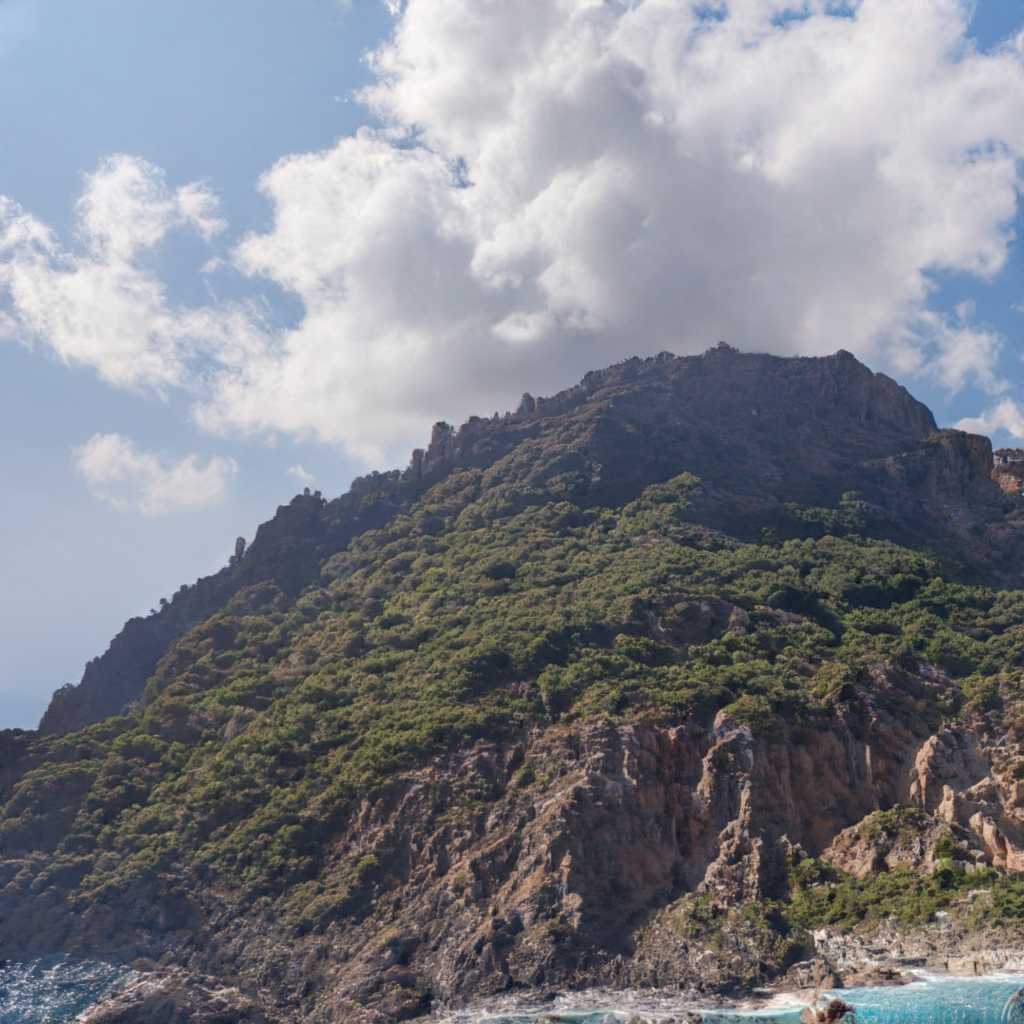}\linebreak[0]%
	\includegraphics[width=\sizea]{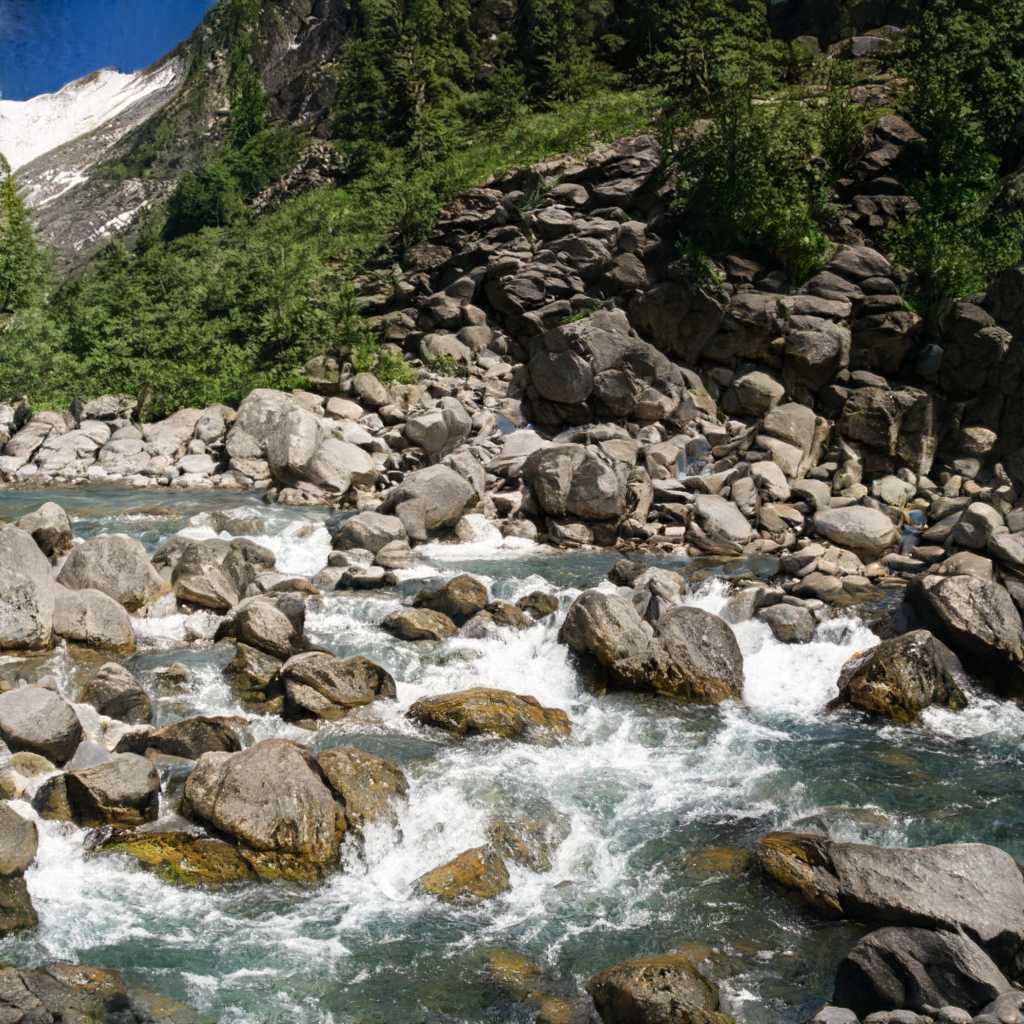}\linebreak[0]%
	\includegraphics[width=\sizea]{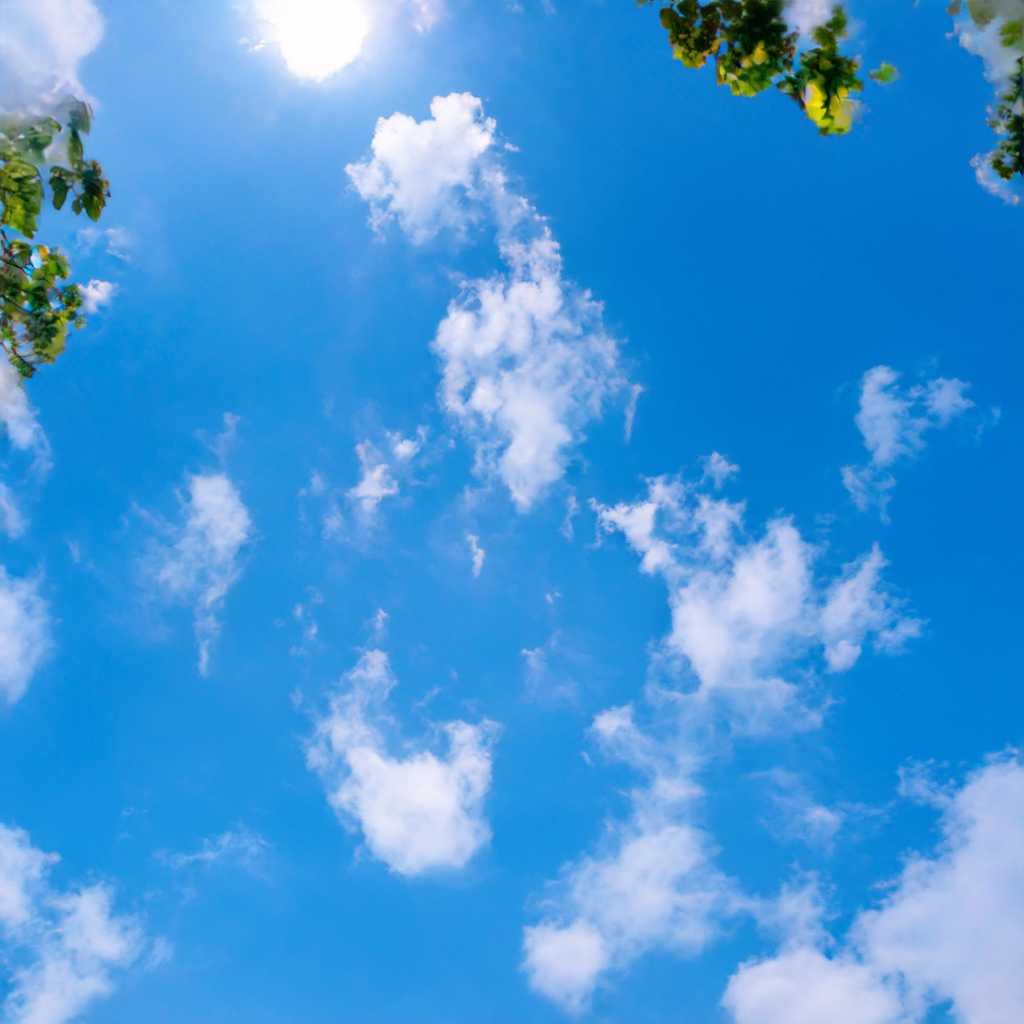}\linebreak[0]%
	\includegraphics[width=\sizea]{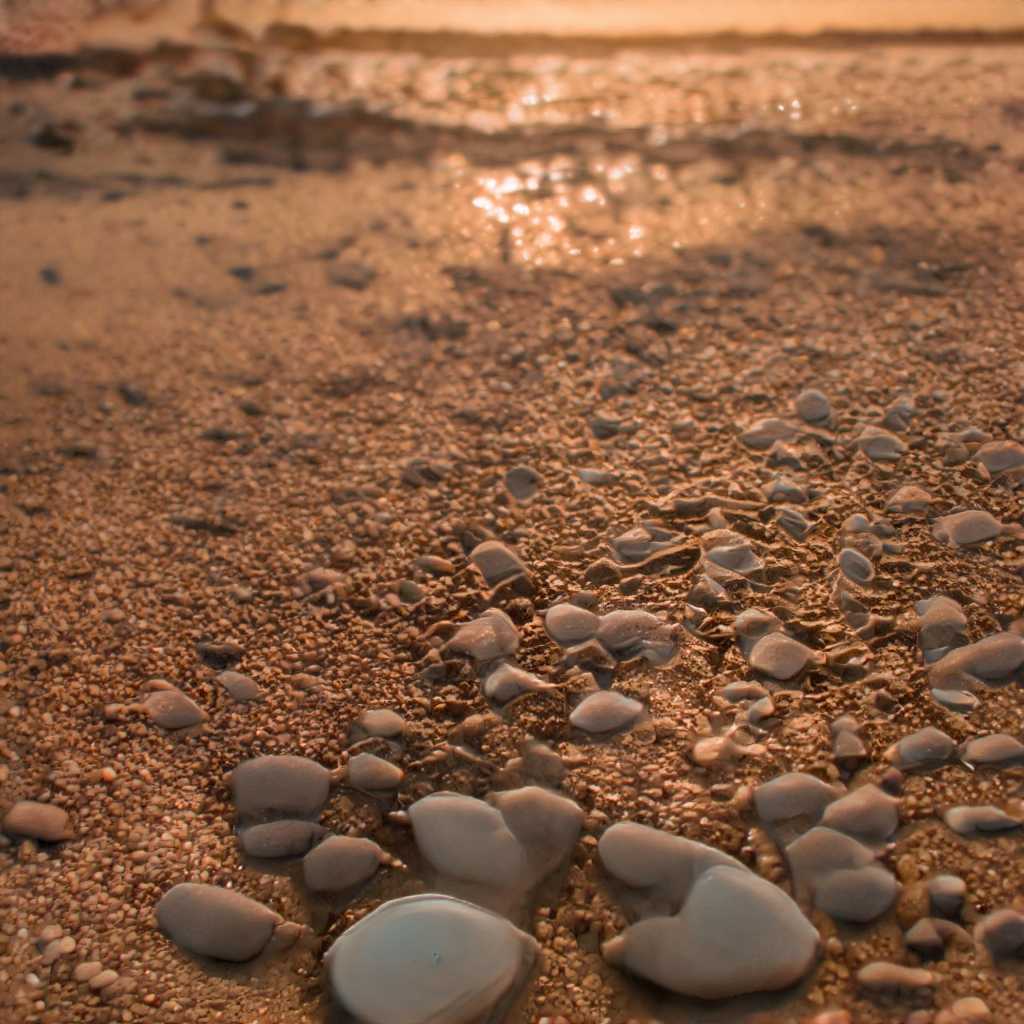}\linebreak[0]%
	\includegraphics[width=\sizea]{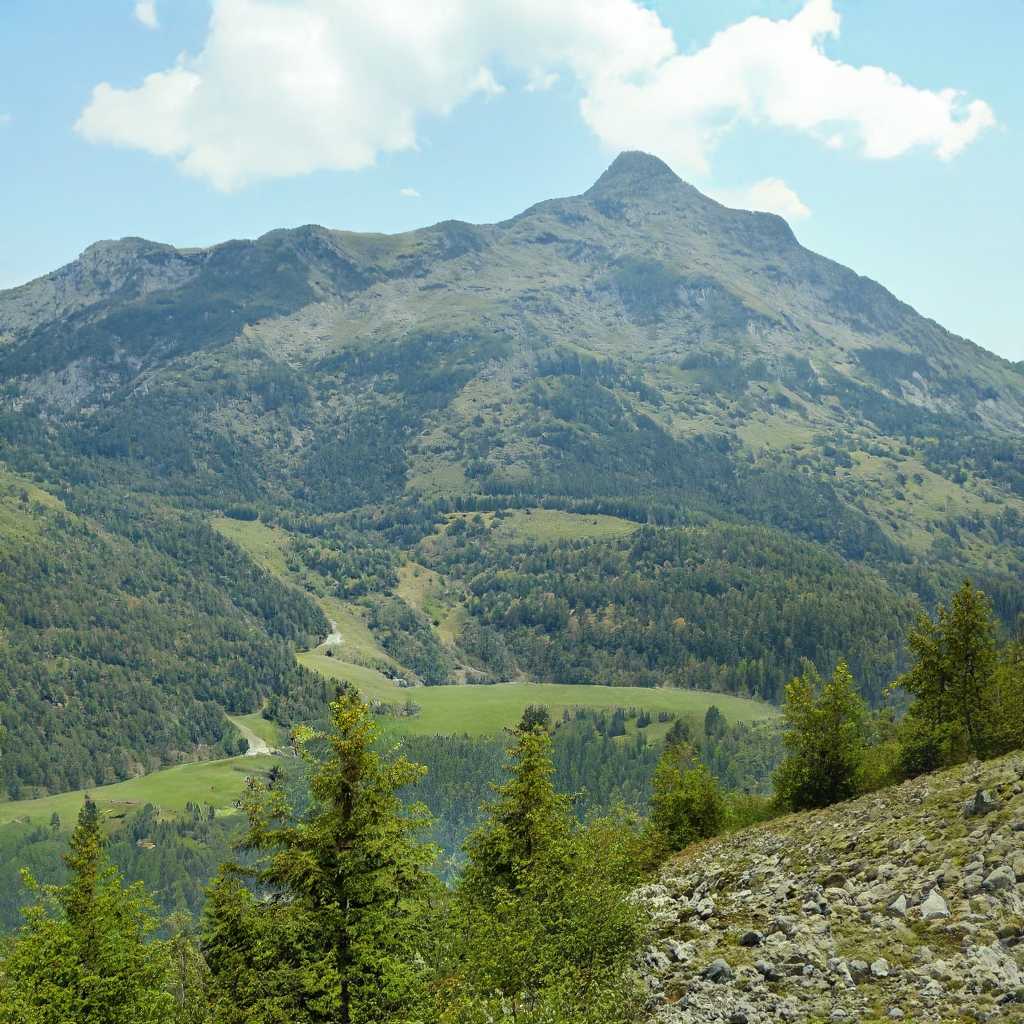}\linebreak[0]%
	\includegraphics[width=\sizea]{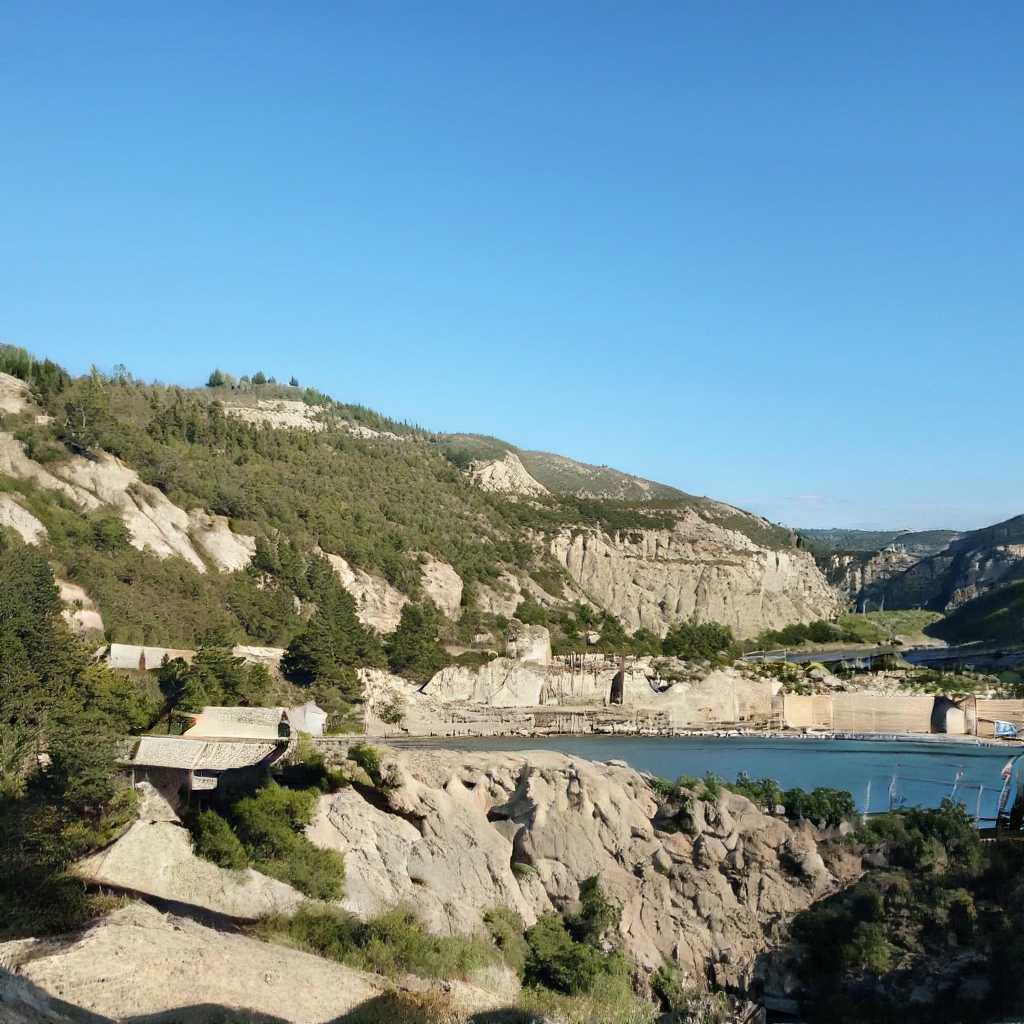}\linebreak[0]%
	\includegraphics[width=\sizea]{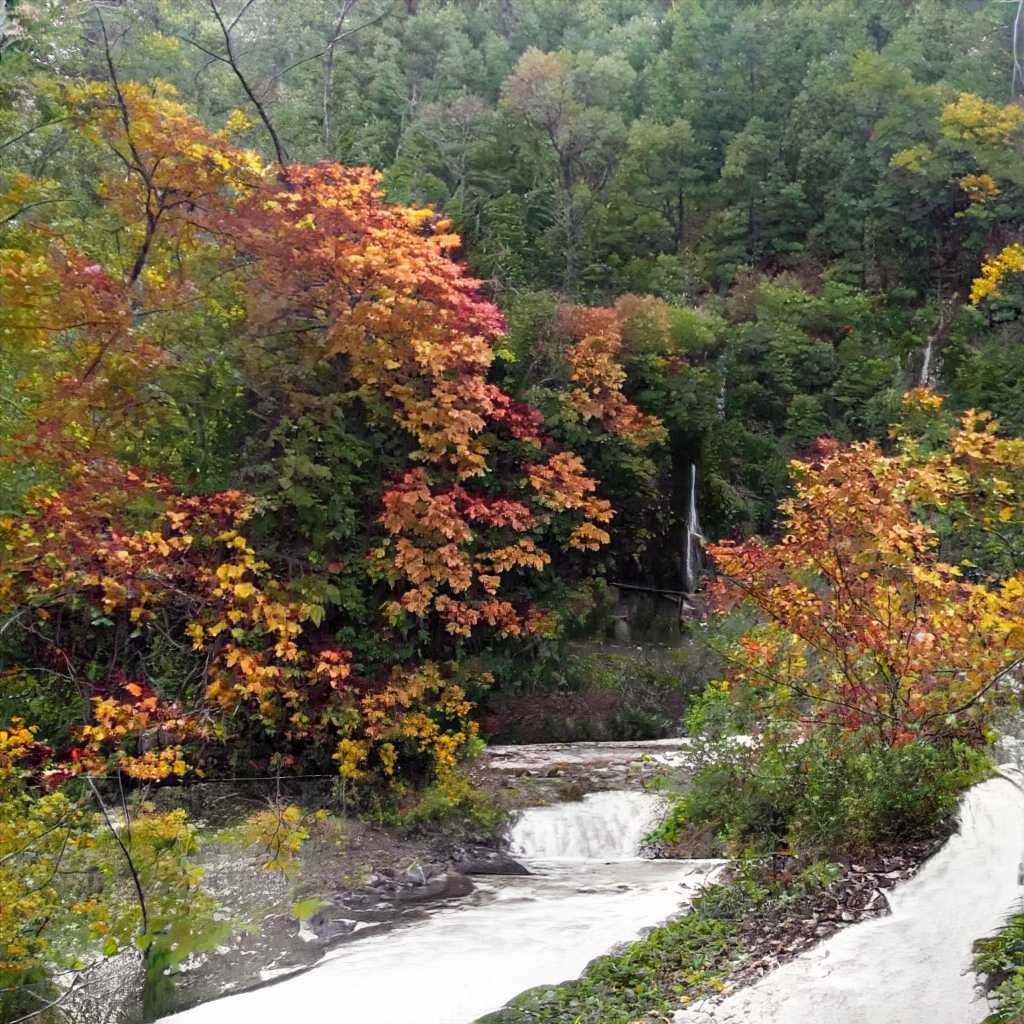}\linebreak[0]%
	\includegraphics[width=\sizea]{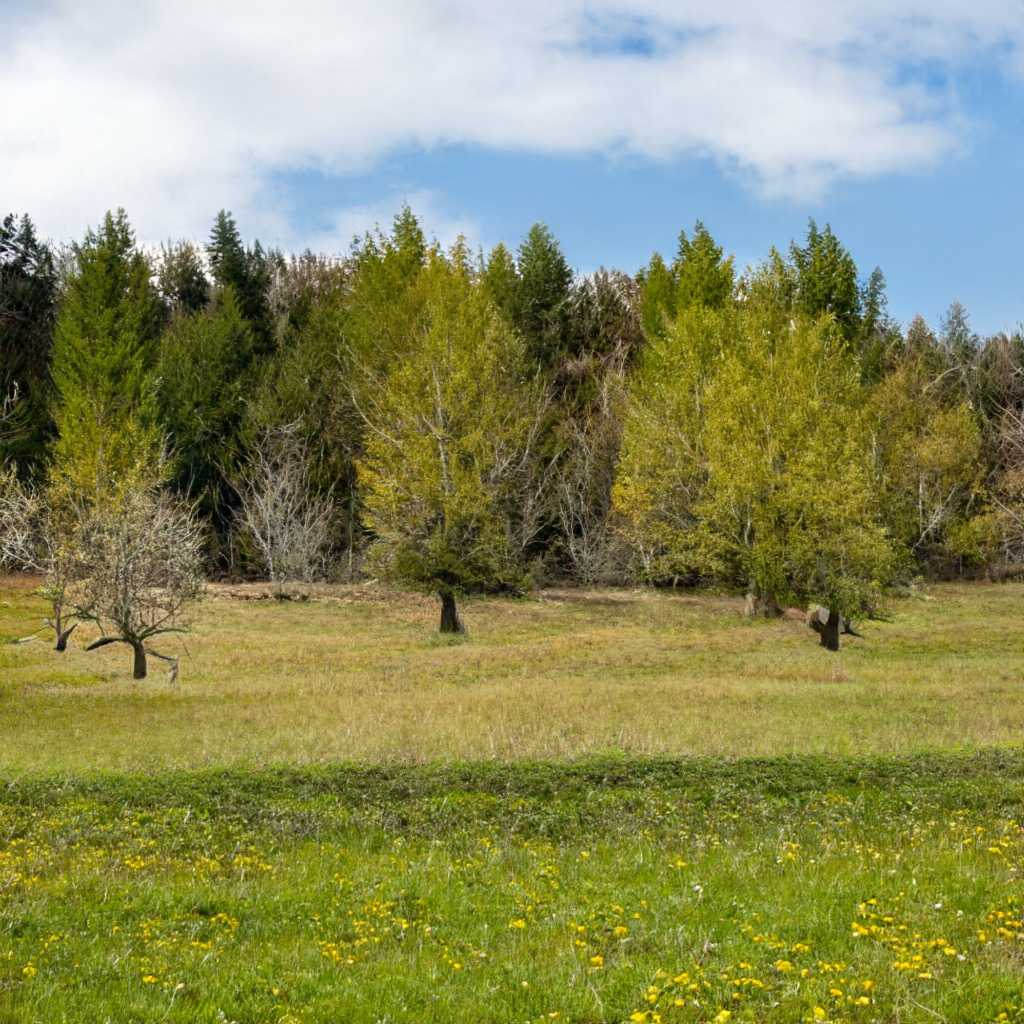}\linebreak[0]%
	\includegraphics[width=\sizea]{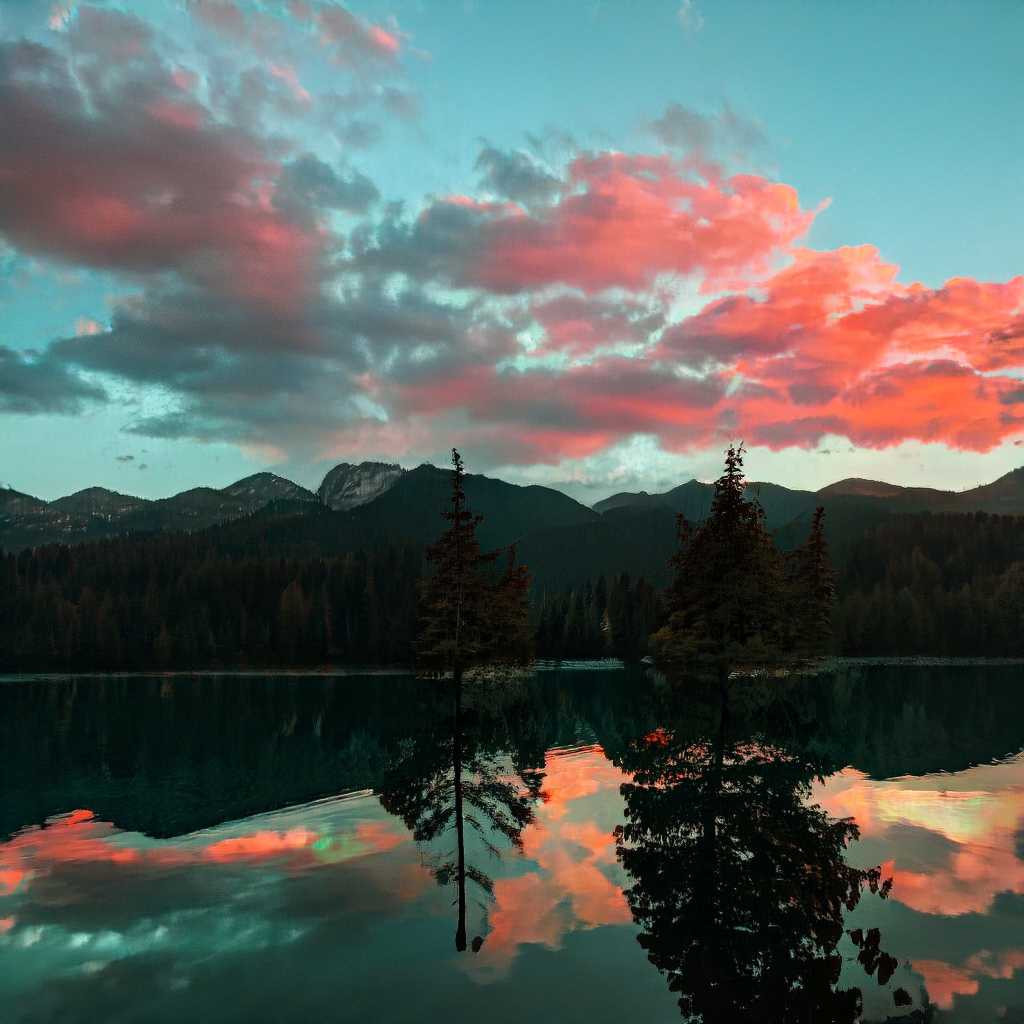}\linebreak[0]%
	\includegraphics[width=\sizea]{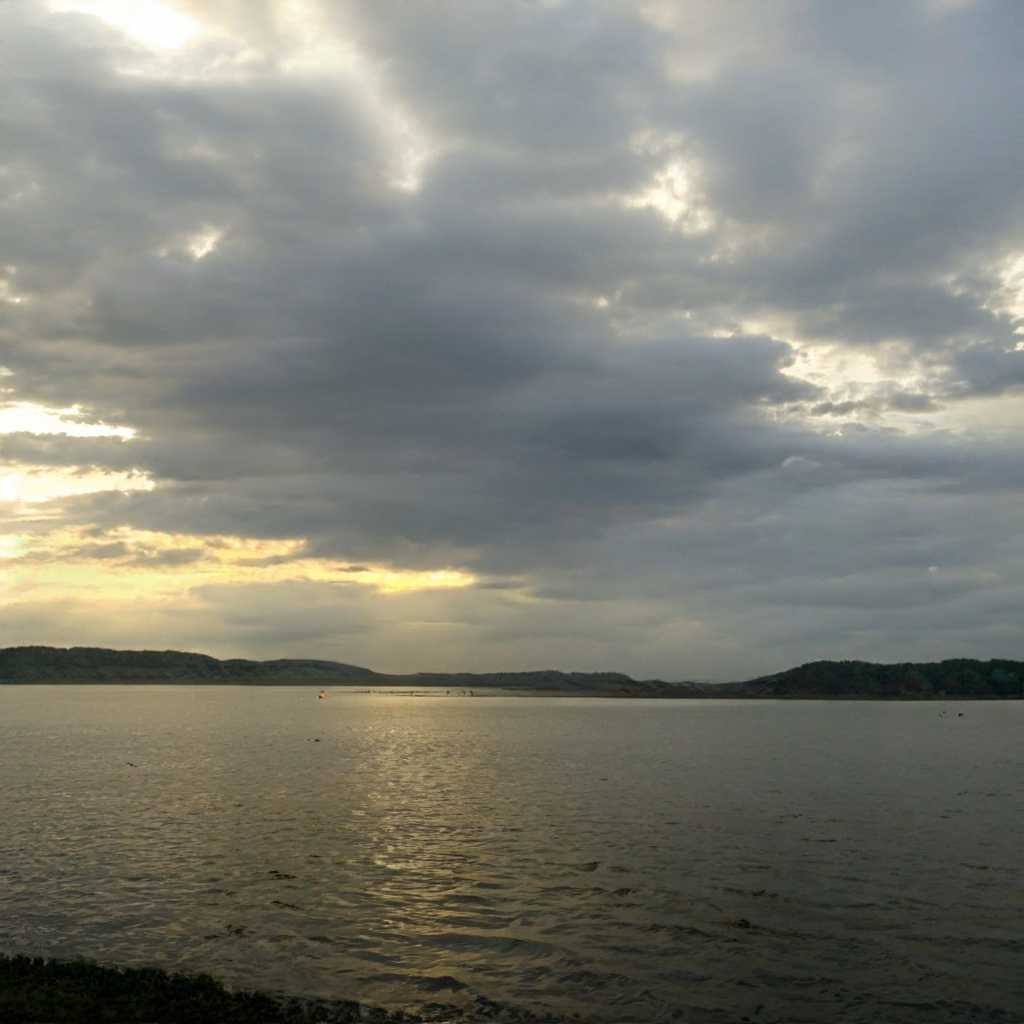}\linebreak[0]%
	\includegraphics[width=\sizea]{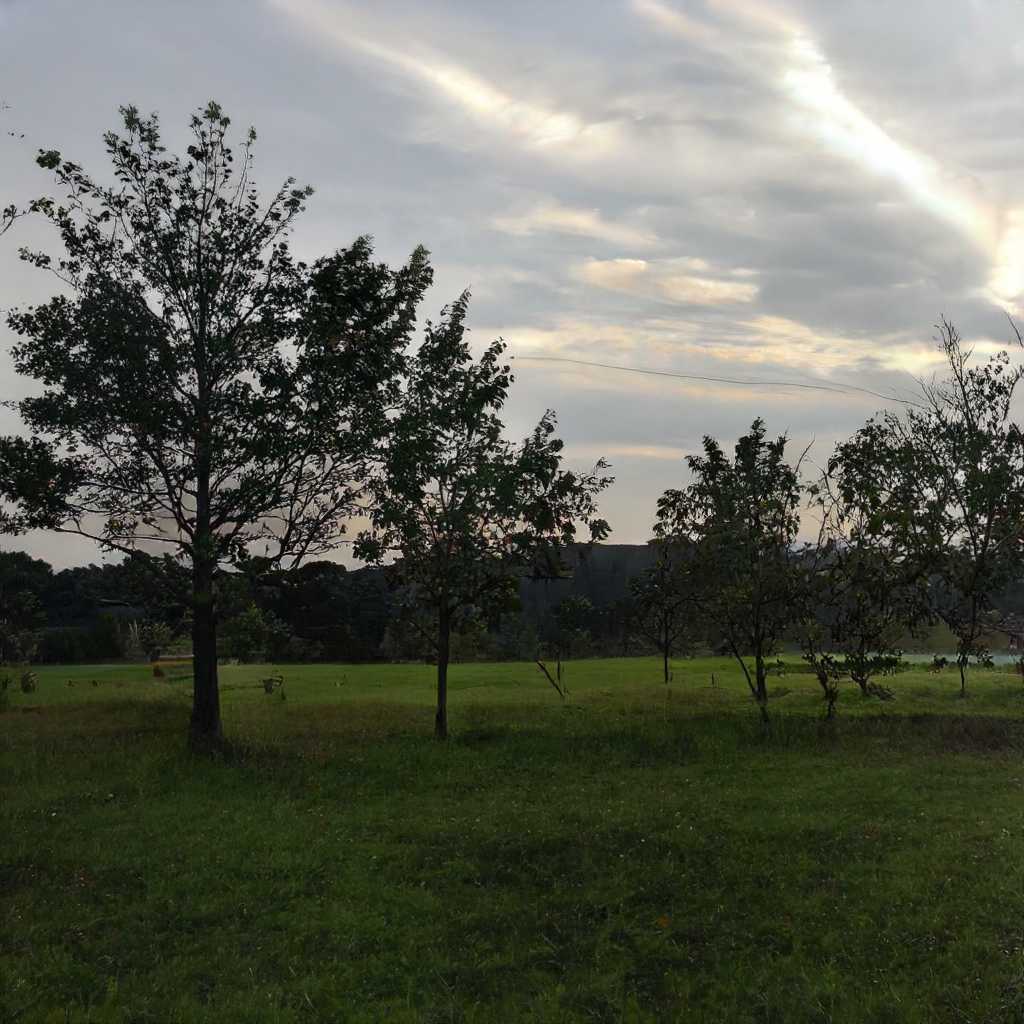}\linebreak[0]%
	\includegraphics[width=\sizea]{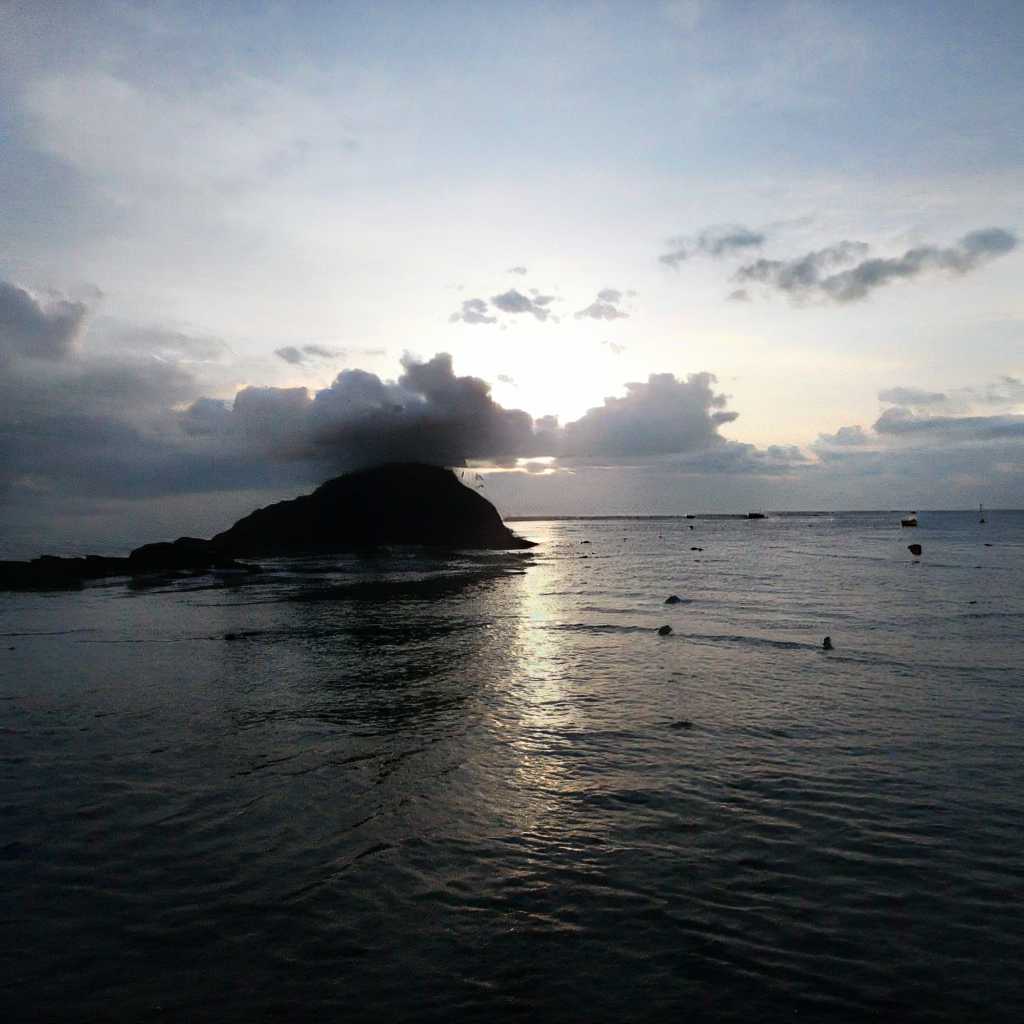}\linebreak[0]%
	\includegraphics[width=\sizea]{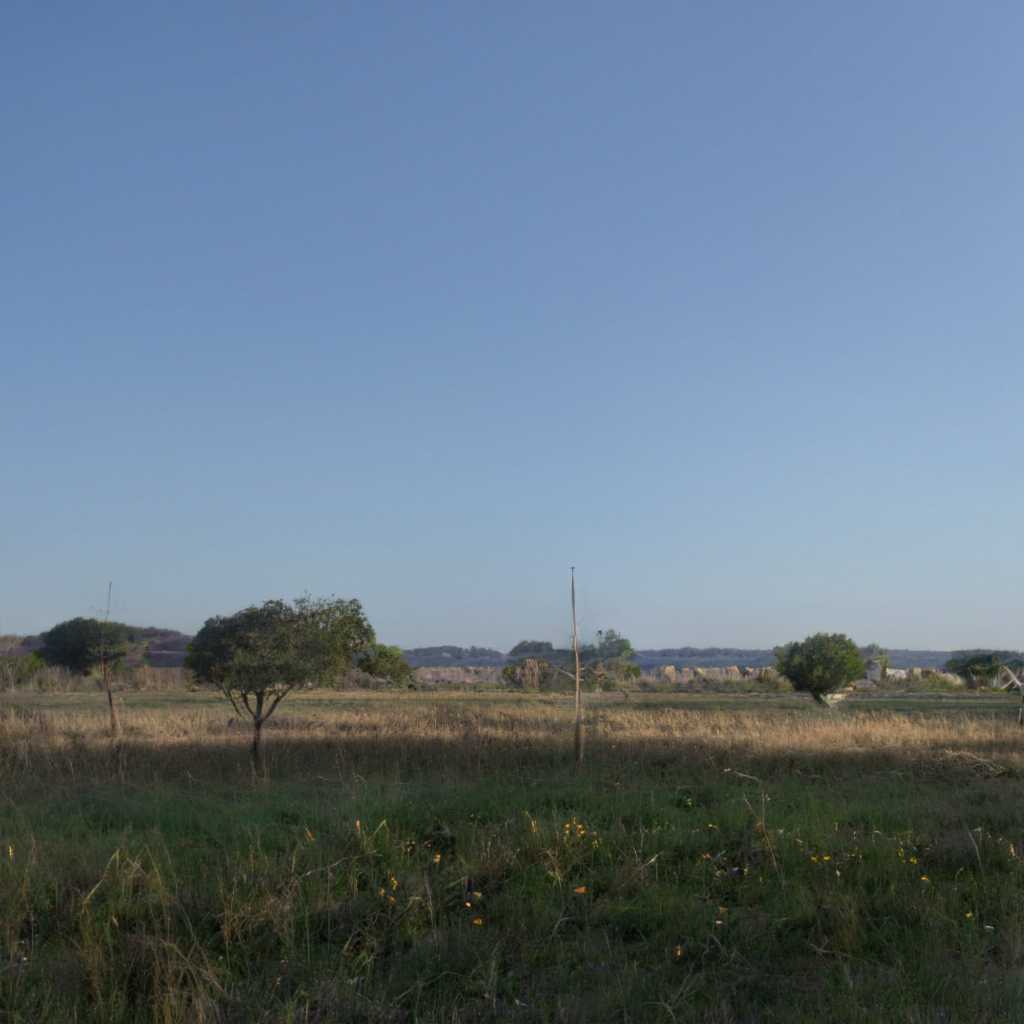}\linebreak[0]%
	\includegraphics[width=\sizea]{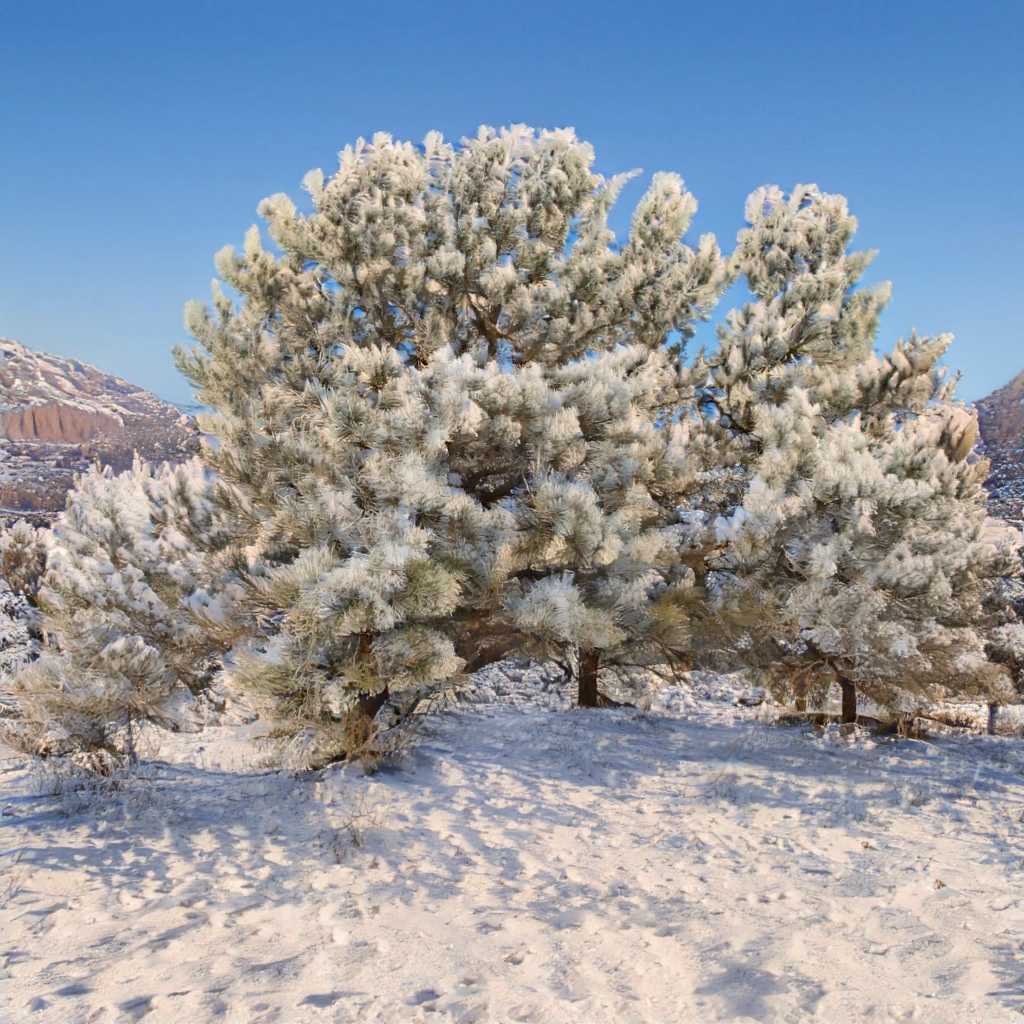}\linebreak[0]%
	\includegraphics[width=0.25\linewidth]{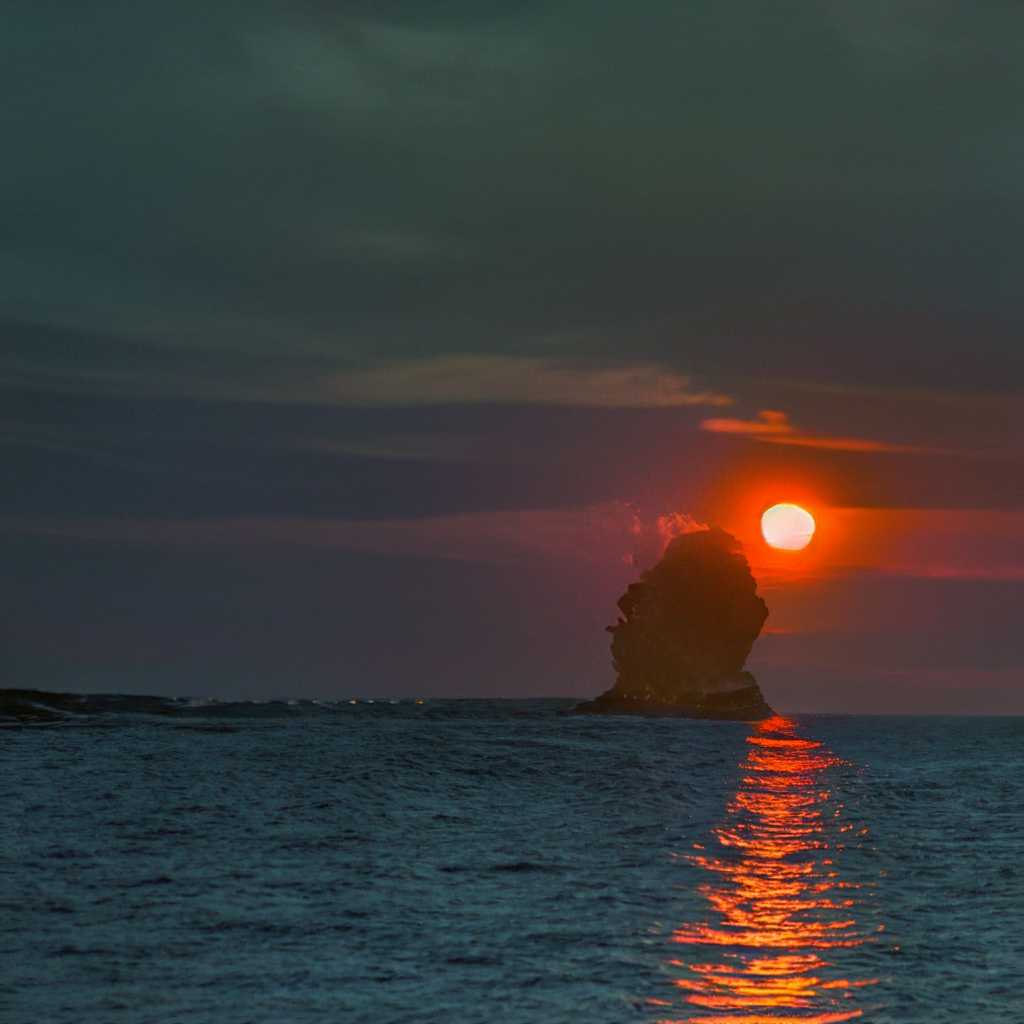}\linebreak[0]%
	\includegraphics[width=0.25\linewidth]{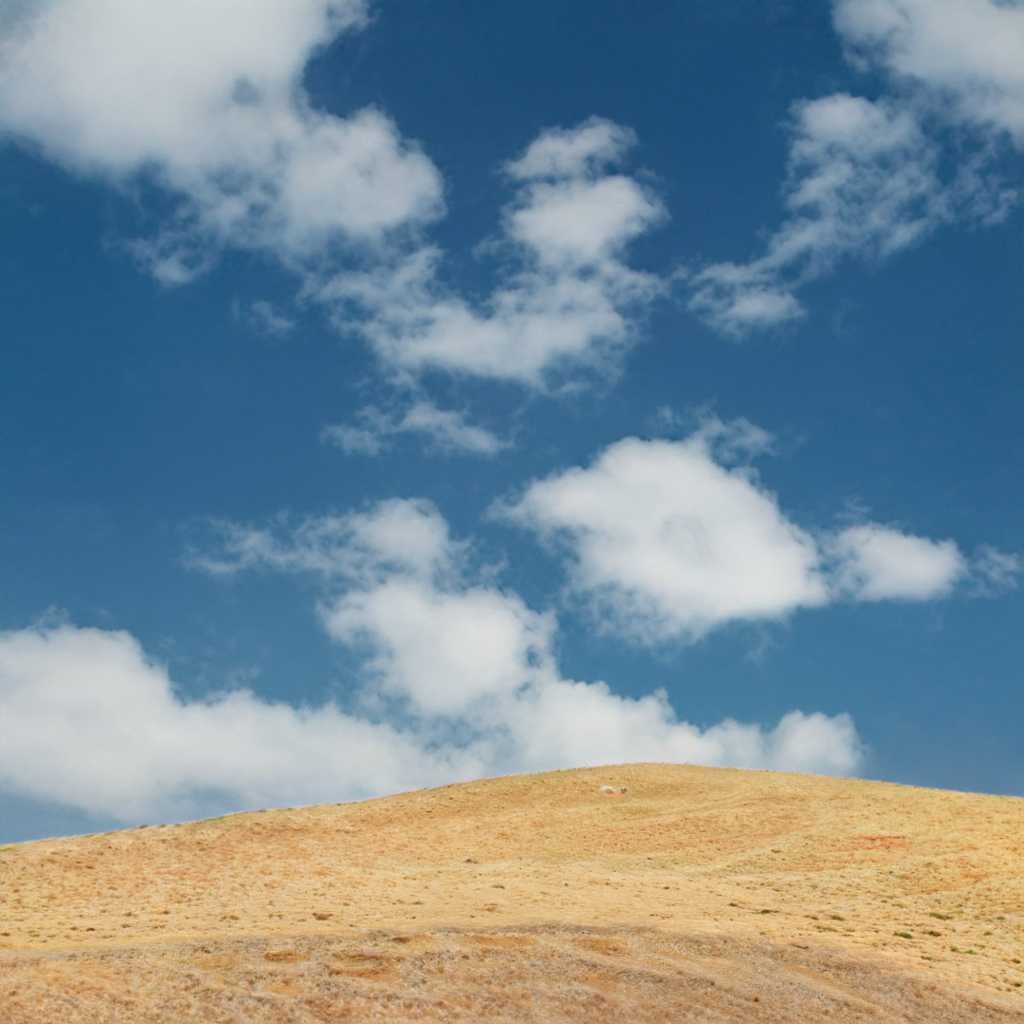}\linebreak[0]%
	\includegraphics[width=0.25\linewidth]{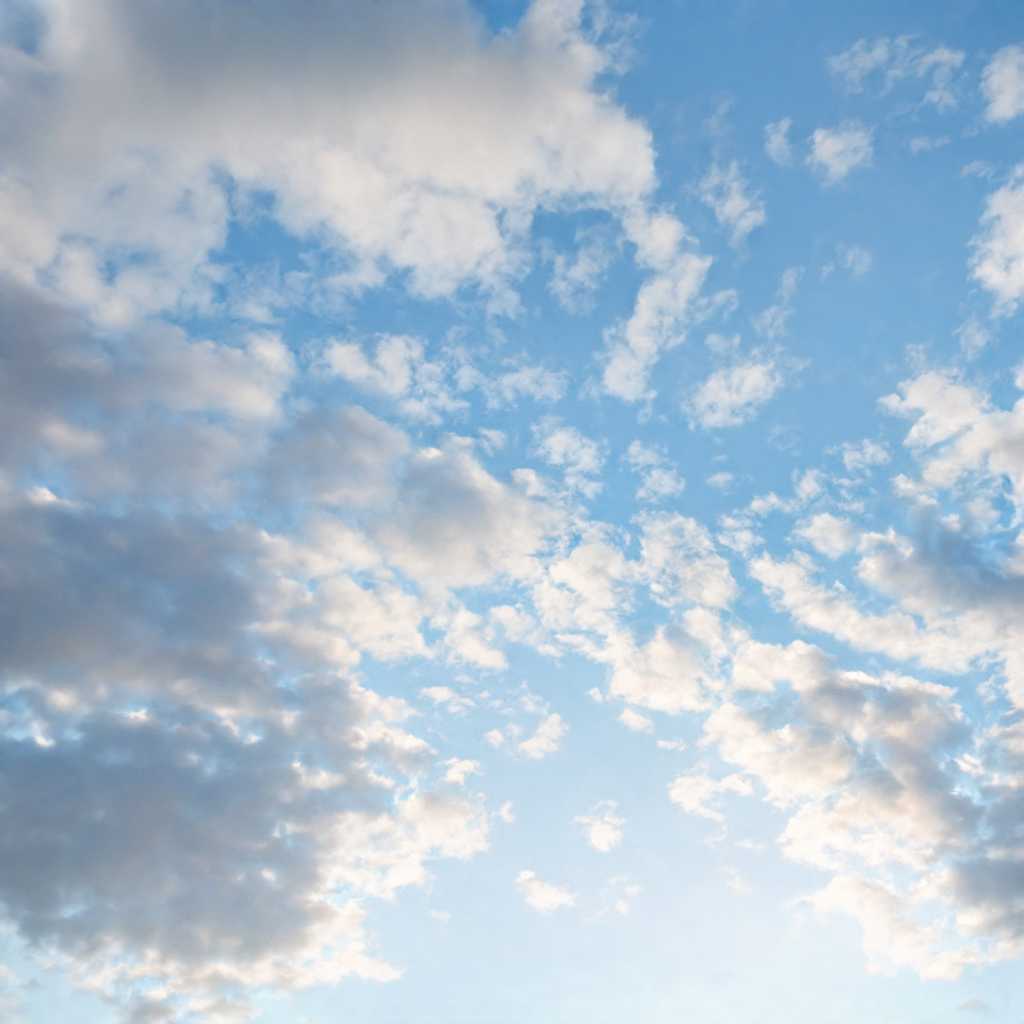}\linebreak[0]%
	\includegraphics[width=0.25\linewidth]{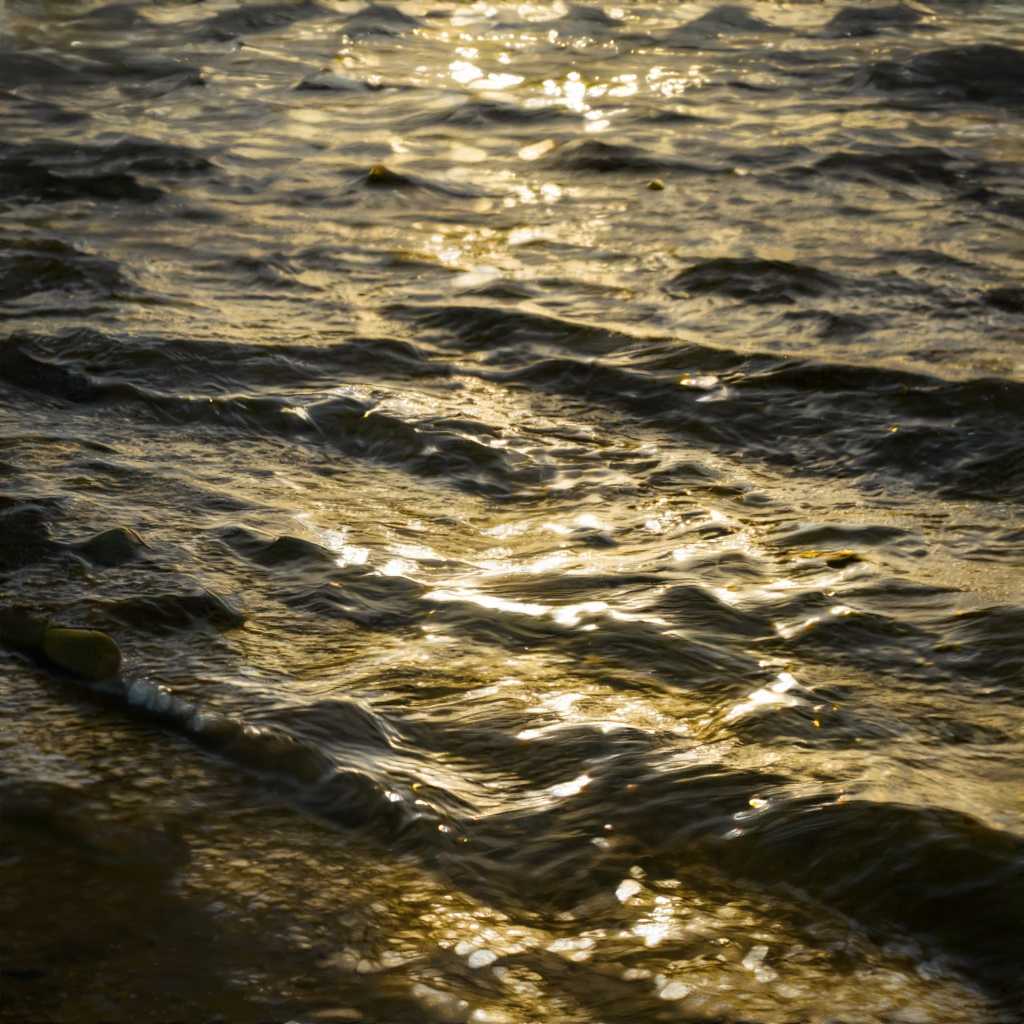}\linebreak[0]%
	\includegraphics[width=0.25\linewidth]{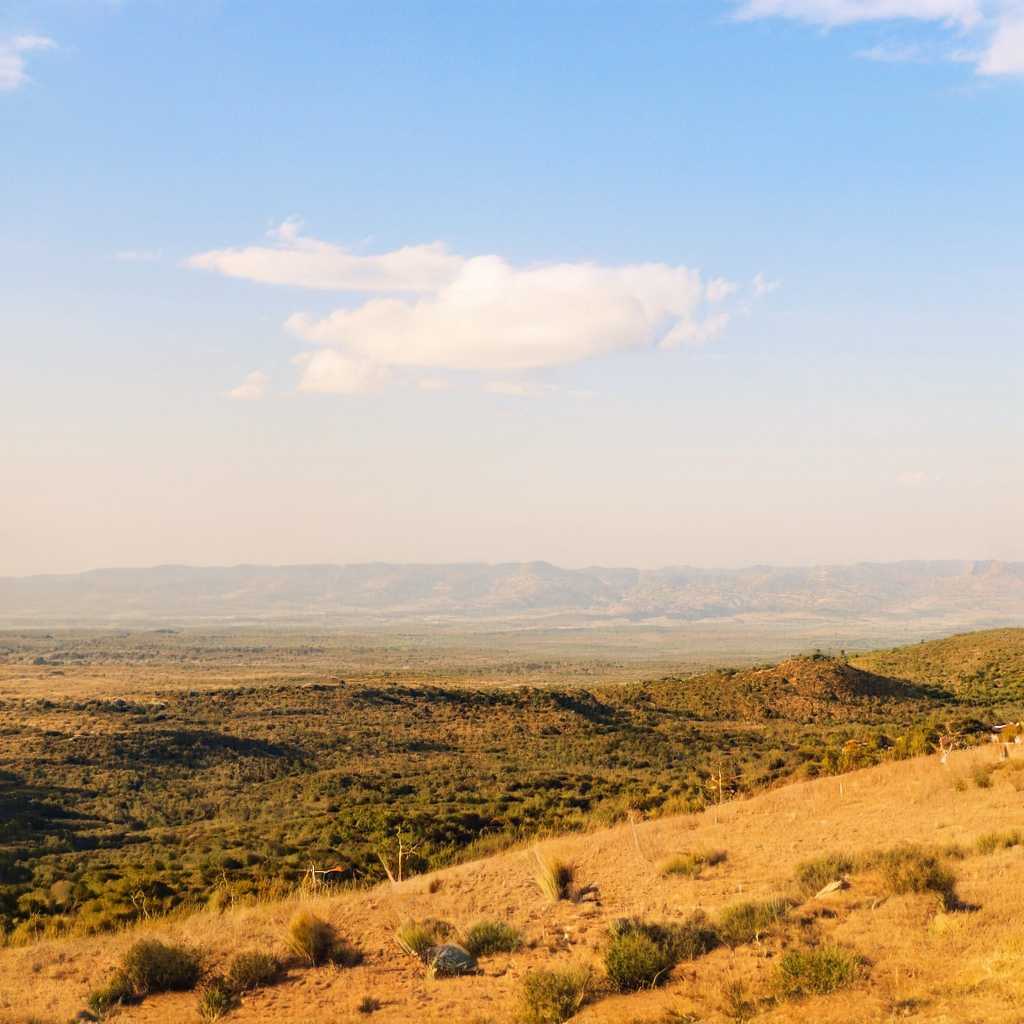}\linebreak[0]%
	\includegraphics[width=0.25\linewidth]{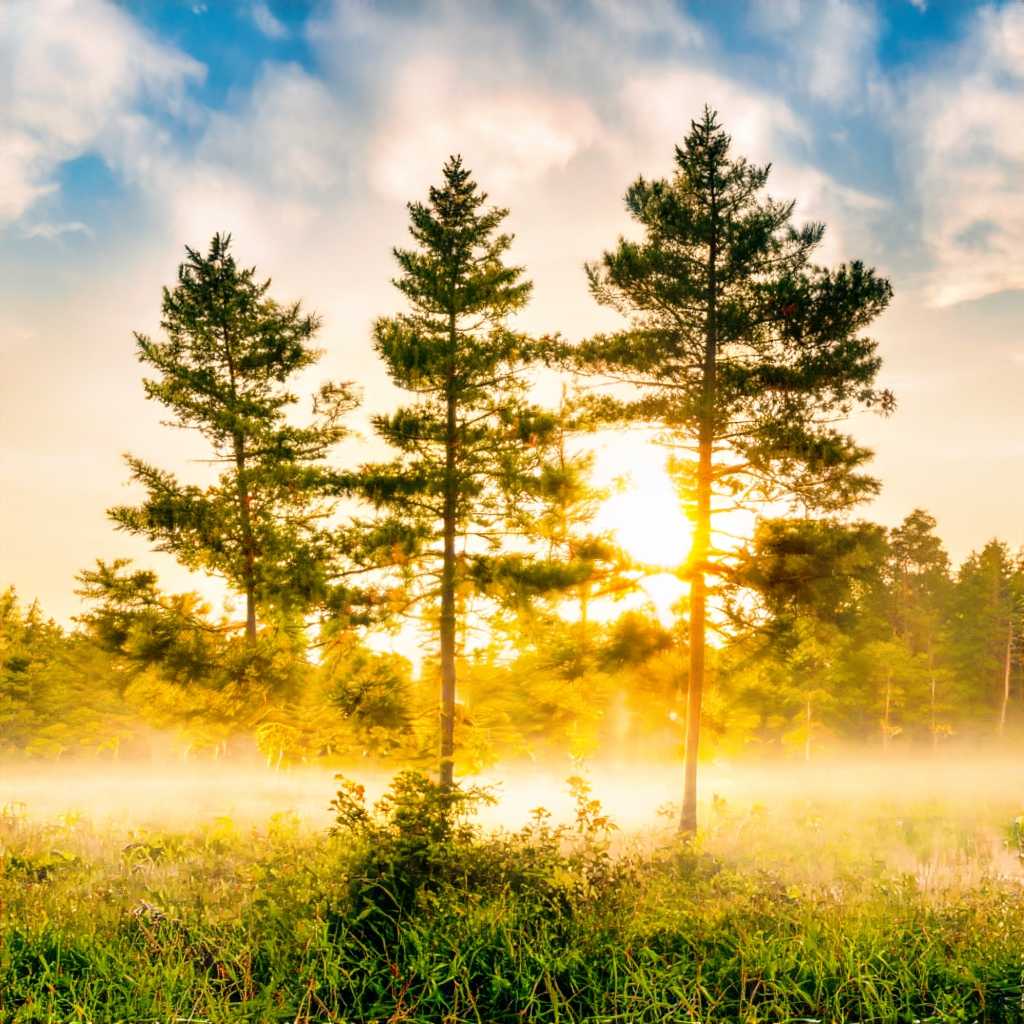}\linebreak[0]%
	\includegraphics[width=0.25\linewidth]{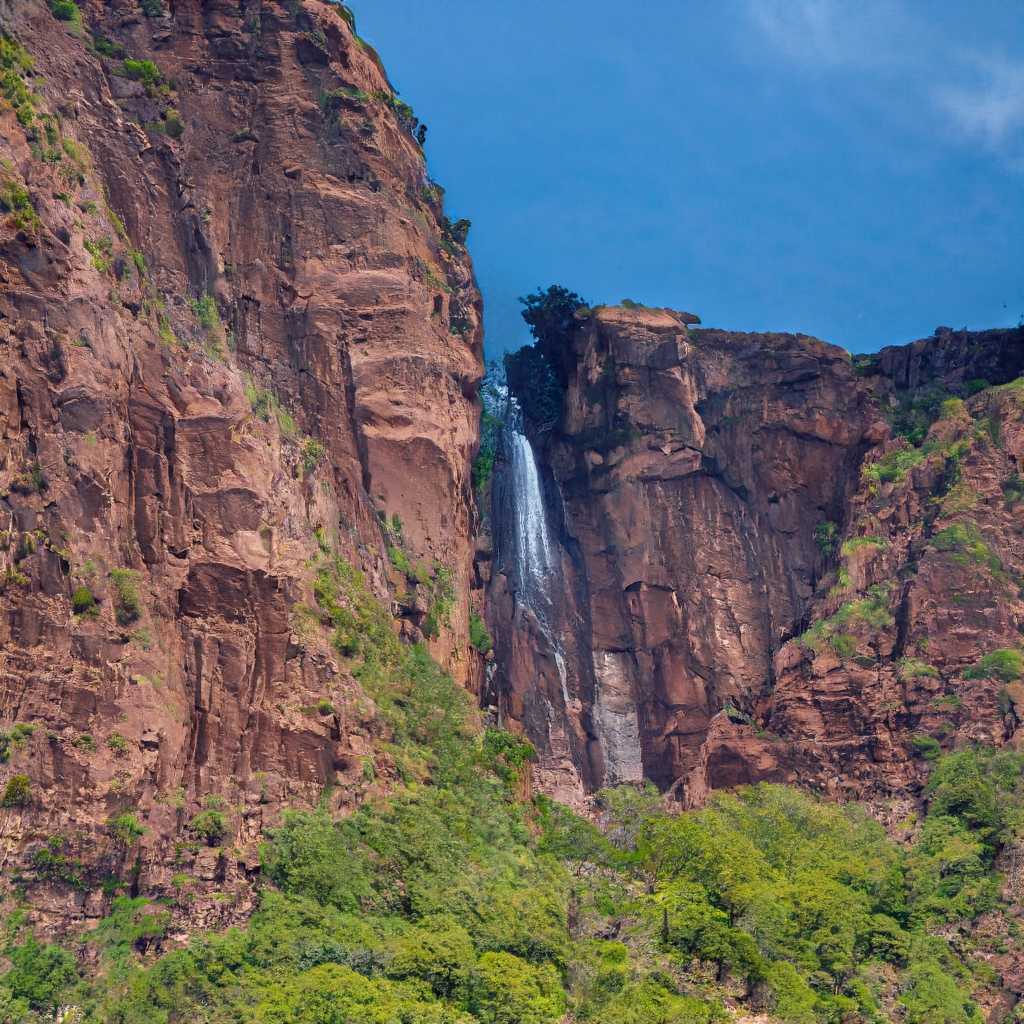}\linebreak[0]%
	\includegraphics[width=0.25\linewidth]{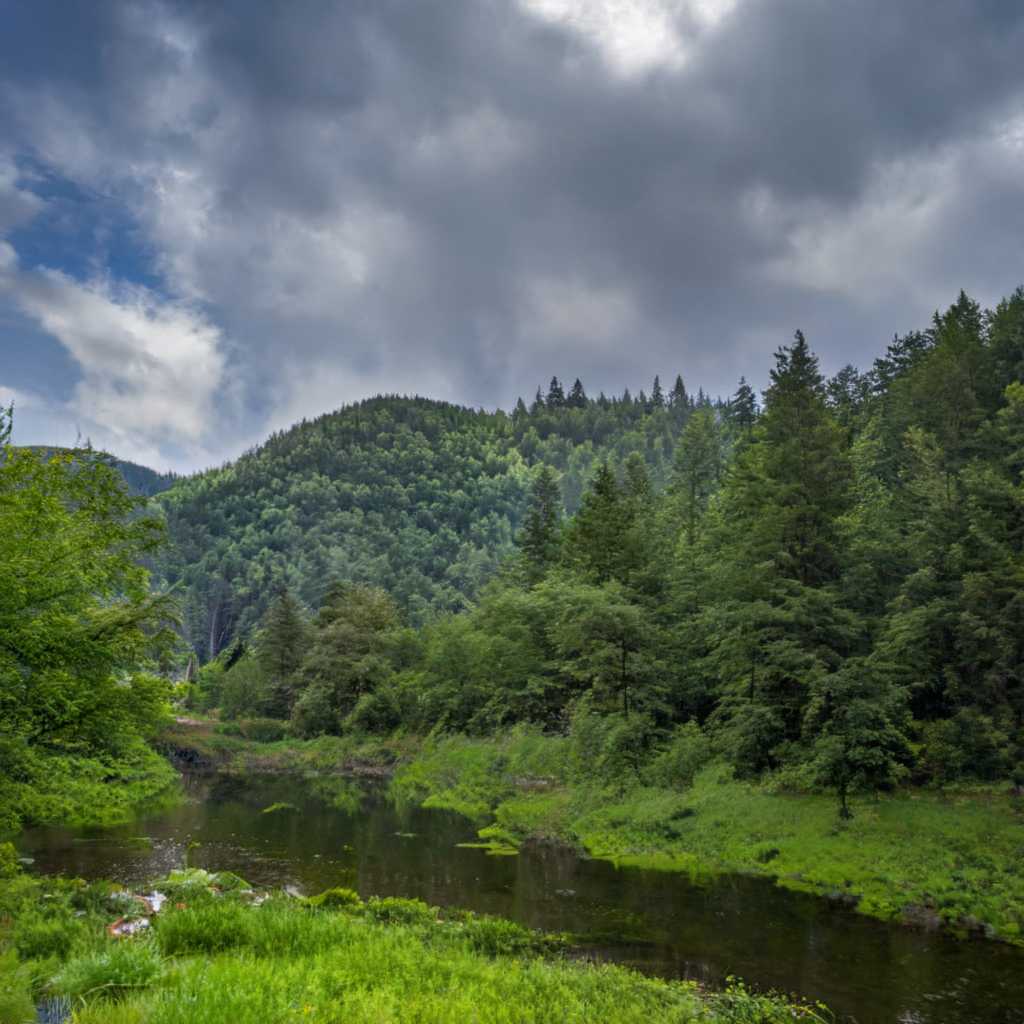}\linebreak[0]%
	\includegraphics[width=0.25\linewidth]{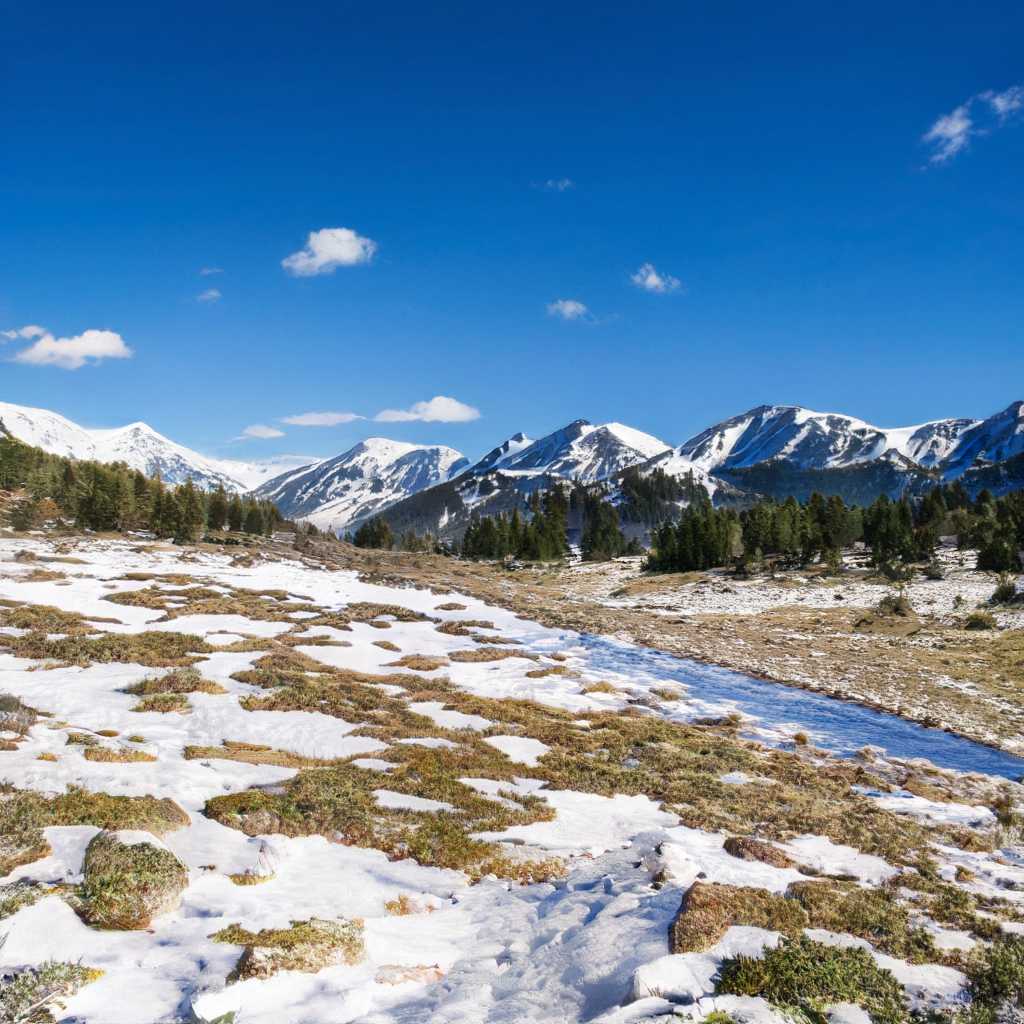}\linebreak[0]%
	\includegraphics[width=0.25\linewidth]{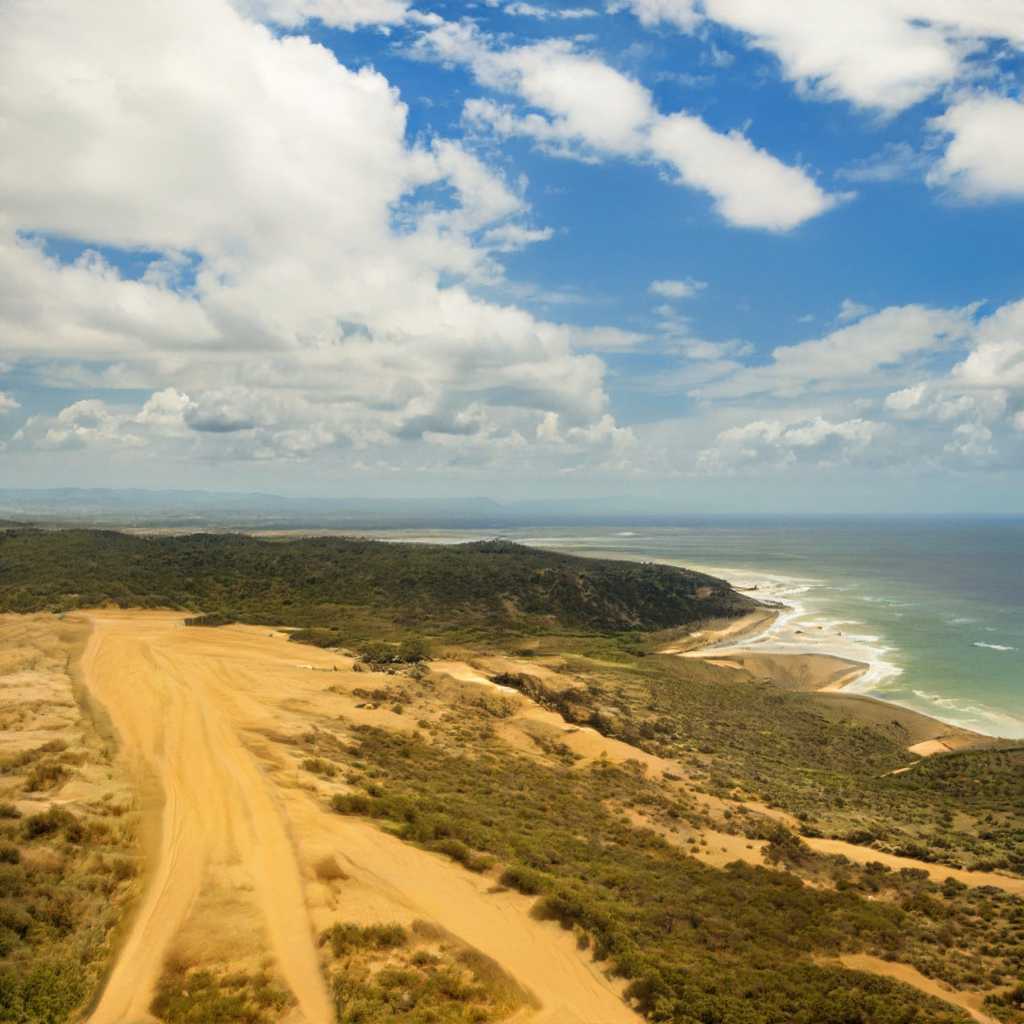}\linebreak[0]%
	\includegraphics[width=0.25\linewidth]{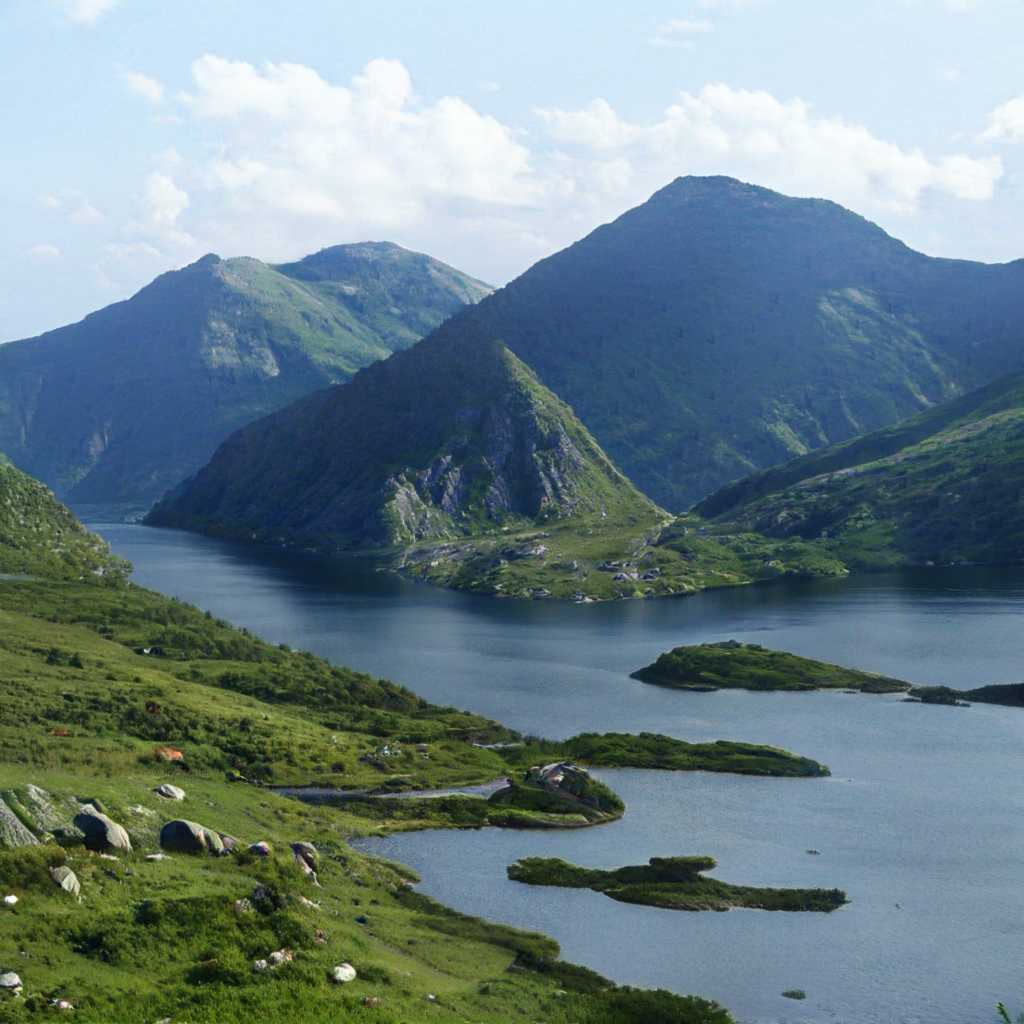}\linebreak[0]%
	\includegraphics[width=0.25\linewidth]{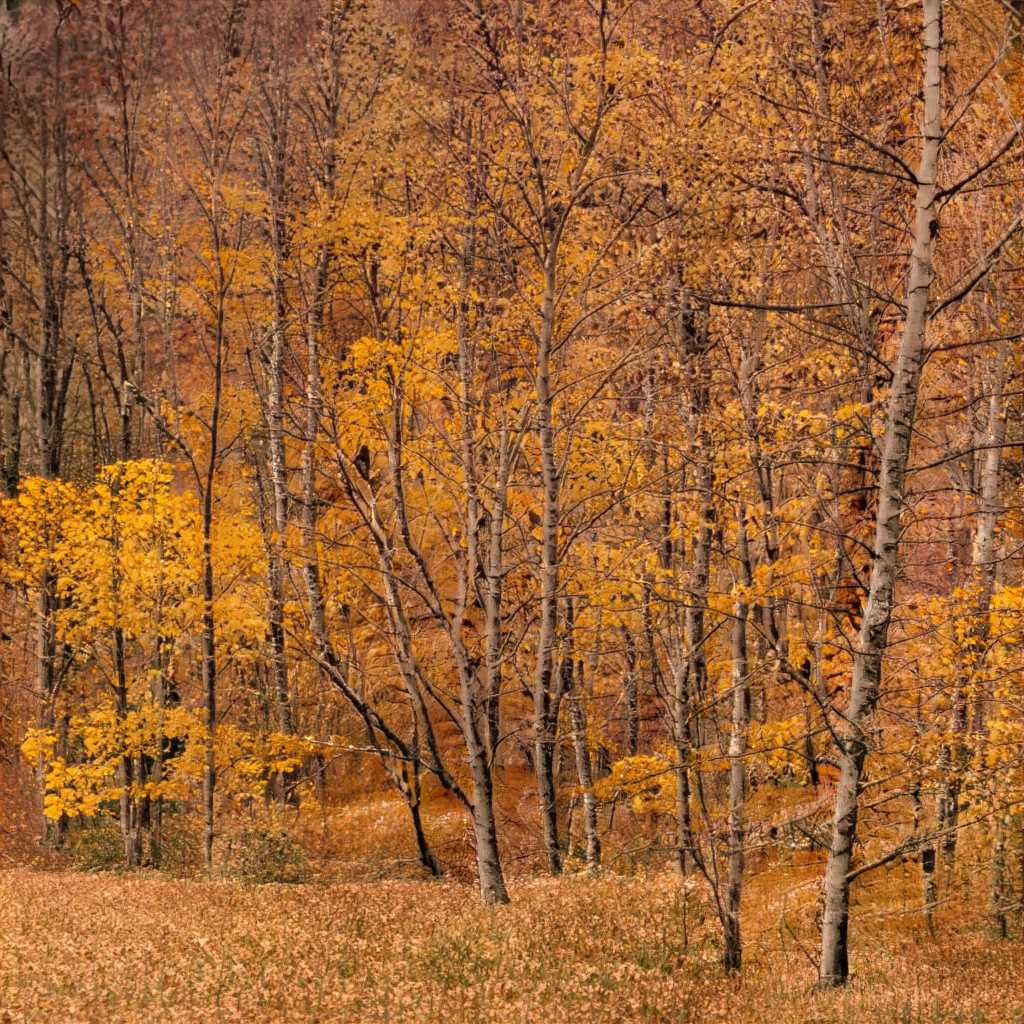}\linebreak[0]%
	\caption{Uncurated unconditional results on the 1024$\times$1024 landscape dataset. }
	\label{fig:getty_uncond}
\end{figure*}

	\end{appendix}

\end{document}